%% file: root_ArXiv.tex
\documentclass[letterpaper, 10 pt, journal]{IEEEtran}  

\IEEEoverridecommandlockouts                              

\makeatletter
\let\NAT@parse\undefined
\makeatother
\usepackage{pgfplots}
\usepackage{filecontents}
\usepackage{graphicx}
\usepackage{subfigure}
\usepackage{amsmath}
\usepackage{placeins}
\usepackage{float}
\usepackage[]{algorithm2e}
\usepackage{svg}
\usepackage{tikz}
\usepackage{lipsum}
\usepackage[T1]{fontenc}

\graphicspath{{images/}}

\newcommand{\executeiffilenewer}[3]{%
	\ifnum\pdfstrcmp{\pdffilemoddate{#1}}%
	{\pdffilemoddate{#2}}>0%
	{\immediate\write18{#3}}\fi%
}

\definecolor{darkgreen}{RGB}{0,90,0}

\title{\LARGE \bf
Local Histogram Matching for Efficient Optical Flow Computation Applied to Velocity Estimation on Pocket Drones
}

\author{Kimberly McGuire$^{1}$, Guido de Croon$^{1}$, Christophe de Wagter$^{1}$, Bart Remes$^{1}$, \\Karl Tuyls$^{1,2}$, and Hilbert  Kappen$^{3}$
\thanks{$^{1}$Delft University of Technology, The Netherlands  
  		\newline
  {\tt\small k.n.mcguire@tudelft.nl}, \newline {\tt\small g.c.h.e.decroon@tudelft.nl}}
\thanks{$^{2}$ University of Liverpool, United Kingdom }
     \thanks{$^{3}$ Radboud University of Nijmegen, The Netherlands}
}

\begin{document}

	\newlength\figureheight 
		\newlength\figurewidth 
\maketitle
\thispagestyle{empty}
\pagestyle{empty}

\begin{abstract}

Autonomous flight of pocket drones is challenging due to the severe limitations on on-board energy, sensing, and processing power. However, tiny drones have great potential as their small size allows maneuvering through narrow spaces while their small weight provides significant safety advantages.  This paper presents a computationally efficient algorithm for determining optical flow, which can be run on an STM32F4 microprocessor (168 MHz) of a 4 gram stereo-camera. The optical flow algorithm is based on edge histograms. We propose a matching scheme to determine local optical flow. Moreover, the method allows for sub-pixel flow determination based on time horizon adaptation. We demonstrate velocity measurements in flight and use it within a velocity control-loop on a pocket drone.

\end{abstract}

\section{INTRODUCTION}

\begin{figure}[t]
	\centering
	\footnotesize 
	\def\svgwidth{\columnwidth}
	\executeiffilenewer{images/dronestabilizationwithflow.svg}{images/dronestabilizationwithflow.pdf}%
	{inkscape -z -D --file=images/dronestabilizationwithflow.svg %
		--export-pdf=images/dronestabilizationwithflow.pdf --export-latex}%
	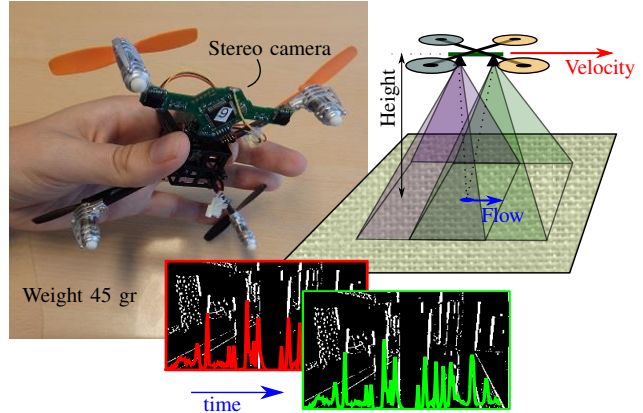%

	\caption{Pocket drone with velocity estimation using a downward looking stereo vision system. A novel efficient optical flow algorithm runs on-board an STM32F4 processor running at only 168 MHz and with only 192 kB of memory. The so-determining optical flow and height, with the
stereo-camera, provide the velocity estimates necessary for the pocket drone's low level control are obtained.} 
	\label{fig:dronestabilization}
\end{figure}

Pocket drones are Micro Air Vehicles (MAVs) small enough to fit in one's pocket and therefore small enough to maneuver in narrow spaces (Fig.~\ref{fig:dronestabilization}). The pocket drones' light weight and limited velocity make them inherently safe for humans. Their agility makes them ideal for search-and-rescue exploration in disaster areas (e.g. in partially collapsed buildings) or other indoor observation tasks. However, autonomous flight of pocket drones is challenging due to the severe limitations in on-board energy, sensing, and processing capabilities.

To deal with these limitations it is important to find efficient algorithms to enable low-level control on these aircraft. Examples of low-level control tasks are stabilization, velocity control and obstacle avoidance. To achieve these tasks, a pocket drone should be able to determine its own velocity, even in GPS-deprived environments. This can be done by measuring the optical flow detected with a bottom mounted camera \cite{bristeau2011navigation}. Flying insects like honeybees use optical flow as well for these low-level tasks \cite{srinivasan2011honeybees}. They serve as inspiration as they have limited processing capacity but can still achieve these tasks with ease.

Determining optical flow from sequences of images can be done in a dense manner with, e.g., Horn-Schunck \cite{horn1981determining}, or with more recent methods like Farneb{\"a}ck \cite{farneback2003two}. In robotics, computational efficiency is important and hence sparse optical flow is often determined with the help of a feature detector such as Shi-Tomasi \cite{shi1994good} or FAST \cite{rosten2005fusing}, followed by Lucas-Kanade feature tracking \cite{bouguet2001pyramidal}. Still, such a setup does not fit the processing limitations of a pocket drone's hardware, even if one is using small images. 

Optical flow based stabilization and velocity control is done with larger MAVs with a diameter of 50 cm and up \cite{romero2009real}\cite{grabe2015nonlinear}. As these aircraft can carry small commercial computers, they can calculate optical flow with more computationally heavy algorithms. A MAV's size is highly correlated on what it can carry and a pocket drone, which fits in the palm of your hand, cannot transport these types resources and therefore has to rely on off-board computing.

A few researchers have achieved optical flow based control fully on-board a tiny MAV. Dunkley et al. have flown a 25 gram helicopter with visual-inertial SLAM for stabilization, for which they use an external laptop to calculate its position by visual odometry \cite{dunkley2014visual}. Briod et al. produced on-board processing results, however they use multiple optical flow sensors which can only detect one direction of movement \cite{briod2013optic}. If more sensing capabilities are needed, the use of single-purpose sensors is not ideal. A combination of computer vision and a camera will result in a single, versatile, sensor, able to detect multiple variables and therefore saves weight on a tiny MAV. By limiting the weight it needs to carry, will increase its flight time significantly.

Closest to our work is the study by Moore et al., in which multiple optical flow cameras are used for obstacle avoidance \cite{moore2014autonomous}. Their vision algorithms heavily compress the images, apply a Sobel filter and do Sum of Absolute Difference (SAD) block matching on a low-power 32-bit Atmel micro controller (AT32UC3B1256).

This paper introduces a novel optical flow algorithm, computationally efficient enough to be run on-board a pocket drone. It is inspired by the optical flow method of Lee et al. \cite{lee2004see}, where image gradients are summed for each image column and row to obtain a horizontal and vertical edge histogram. The histograms are matched over time to estimate a global divergence and translational flow. In \cite{lee2004see} the algorithm is executed off-board with a set of images, however it shows great potential. In this paper, we extend the method to calculate local optical flow as well. This can be fitted to a linear model to determine both translational flow and divergence.  The later will be unused in the rest of this paper as we are focused on horizontal stabilization and velocity control. However, it will become useful for autonomous obstacle avoidance and landing tasks. Moreover, we introduce an adaptive time horizon rule to detect sub-pixel flow in case of slow movements. Equipped with a downward facing stereo camera, the pocket drone can determine its own speed and do basic stabilization and velocity control. 

The remainder of this paper is structured as follows. In Section \ref{sec:optflow}, the algorithm is explained with off-board results. Section \ref{sec:speedcontrol} will contain velocity control results of two MAVs, an AR.Drone 2.0 and a pocket drone, with both using the same 4 gr stereo-camera containing the optical flow algorithm on-board. Section \ref{sec:speedcontrol} will conclude these results and give remarks for future research possibilities.

\section{OPTICAL FLOW WITH EDGE FEATURE HISTOGRAMS}\label{sec:optflow}

This section explains the algorithm for the optical flow detection using edge-feature histograms. The gradient of the image is compressed into these histograms for  the horizontal and vertical direction. This reduces the 2D image search problem to 1D signal matching,  increasing its computational efficiency. Therefore, this algorithm is efficient enough to be run on-board a 4 gram stereo-camera module, which can used by an MAV to determine its own velocity.

\subsection{Edge Features Histograms}
\begin{figure}[t]
	\centering
\footnotesize

		\subfigure[Vision Loop]{
				\def\svgwidth{\columnwidth}
	\executeiffilenewer{images/edge_histogram2.svg}{images/edge_histogram2.pdf}%
	{inkscape -z -D --file=images/edge_histogram2.svg %
		--export-pdf=images/edge_histogram2.pdf --export-latex}%
	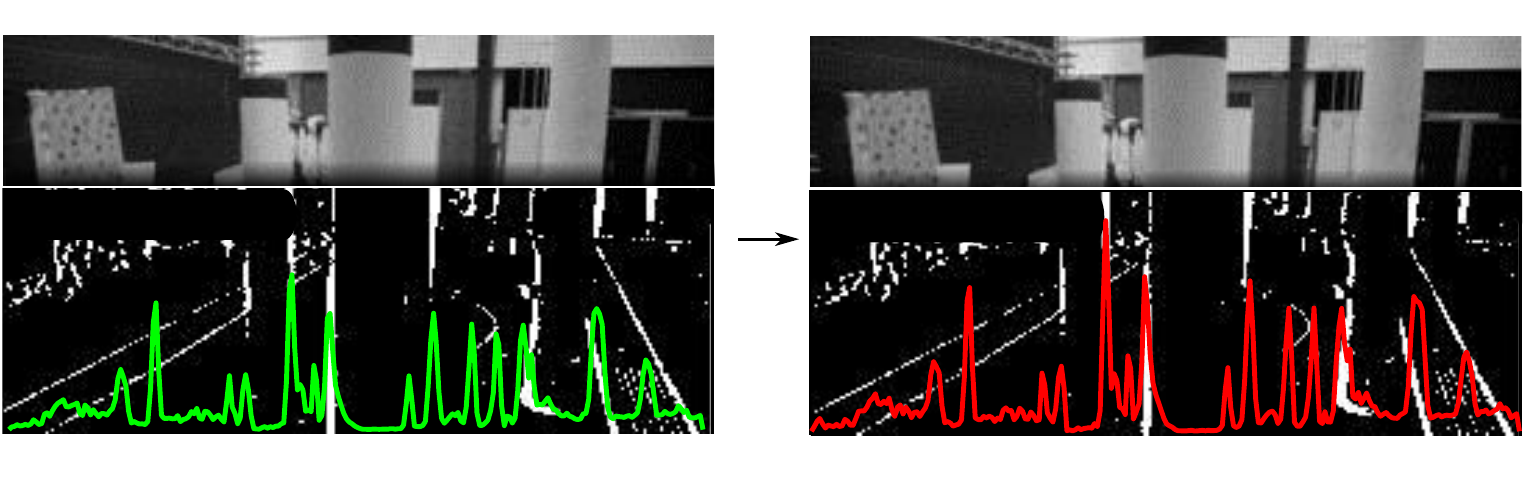%
}
			\subfigure[Local histogram matching]{	\setlength\figureheight{3.5cm} 
				\setlength\figurewidth{0.8\linewidth}\input{images/edge_hist_match.tikz}.}

\caption{(a) The vision loop with for creating the edge feature histograms and (b) the matching of the two histograms (previous frame (green) and current frame (red)) with SAD. This results in a pixel displacement (blue) which can be fitted to a linear model (dashed black line) for a global estimation.}
\label{fig:edgehistograms}
\end{figure}
The generated edge feature histograms are created by first calculating the gradient of the image  on the vertical and horizontal axis using a Sobel filter (Fig.~\ref{fig:edgehistograms}(a)). From these gradient intensity images, the histogram can be computed for each of the image's dimensions by summing up the intensities. The result is an edge feature histogram of the image gradients in the horizontal and vertical directions.

From two sequential frames, these edge histograms can be calculated and matched locally with the Sum of Absolute Differences (SAD). In Fig.~\ref{fig:edgehistograms}(b), this is done for a window size of 18 pixels and a maximum search distance of 10 pixels in both ways.
The displacement can be fitted to a linear model with least-square line fitting. 
This model has two parameters: a constant term for translational flow and a slope for divergence. Translational flow stands for the translational motion between the sequential images, which is measured if the camera is moved sideways. The slope/divergence is detected when a camera moves to and from a scene. In case of the displacement shown in  Fig.~\ref{fig:edgehistograms}(b) both types of flows are observed, however only translation flow will be considered in the remainder of this paper.

\subsection{Time Horizon Adaptation for Sub-Pixel Flow}

The previous section explained the matching of the edge feature histograms which gives translational flow. Due to a image sensor's resolution, existing variations within pixel boundaries can not be measured, so only integer flows can be considered. However, this will cause complication if the camera is moving slowly or is well above the ground. If these types of movements result in sub-pixel flow, this cannot be observed with the current state of the edge flow algorithm. This sub-pixel flow is important for to ensure velocity control on an MAV. 

To ensure the detection of sub-pixel flow, another factor is added to the algorithm. Instead of the immediate previous frame, the current frame is also compared with a certain time horizon $n$ before that. The longer the time horizon, the more resolution the sub-pixel flow detection will have. However, for higher velocities it will become necessary to compare the current edge histogram to the closest time horizon as possible. Therefore, this time horizon comparison must be adaptive. 

Which time horizon to use for the edge histogram matching, is determined by the translational flow calculated in the previous time step $p_{t-1}$:

\begin{eqnarray}\label{eq:previousframenumber}
n~=~\min\left(\dfrac{1}{|p_{t-1}|},N\right)
\end{eqnarray}

where $n$ is the number of the previous stored edge histogram that the current frame is compared to. The second term, $N$,  stands for the maximum number of edge histograms allowed to be stored in the memory. It needs to be limited due to the strict memory requirements and in our experiments is set to 10. Once the current histogram and time horizon histogram are compared, the resulting flow must be divided by $n$ to obtain the flow per frame.

\begin{figure}[t]
	\centering
	\small
		\def\svgwidth{0.7\columnwidth}
	\executeiffilenewer{images/velocity_estimation.svg}{images/velocity_estimation.pdf}%
	{inkscape -z -D --file=images/velocity_estimation.svg %
		--export-pdf=images/velocity_estimation.pdf --export-latex}%
	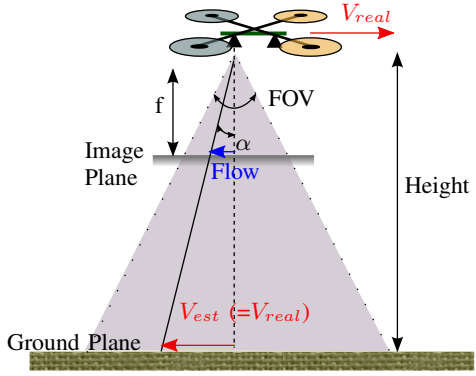%

	\caption{Velocity estimation by measuring optical flow with one camera and height with both cameras of the stereo-camera.}
	\label{fig:velocity_estimation}
\end{figure}


\subsection{Velocity Estimation on Set of Images}
\begin{figure}
	
	\small
	\centering

\begin{minipage}{0.8\linewidth}
			\includegraphics[width=0.49\linewidth]{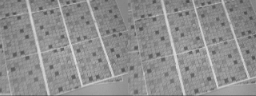}\hfill\includegraphics[width=0.49\linewidth]{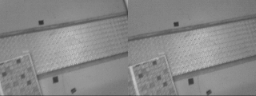}
			
			\vspace{1 mm}
			
			\includegraphics[width=0.49\linewidth]{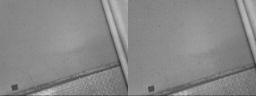}\hfill\includegraphics[width=0.49\linewidth]{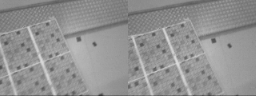}
			
			\vspace{1 mm}
			
			\includegraphics[width=0.49\linewidth]{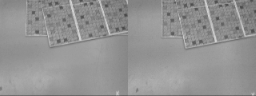}\hfill\includegraphics[width=0.49\linewidth]{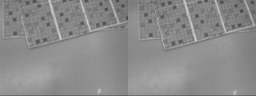}
		\end{minipage}
		
		\caption{
	Several screen shots of the set of images used for off-line estimation of the velocity. Here the diversity in amount of texture can be seen.}
		\label{fig:trajectory}
	\end{figure}

	\begin{figure}

		\footnotesize
		\flushright
		\subfigure[Amount of corners per time frame of the data set sample.]{\setlength\figureheight{1.5cm} 
			\setlength\figurewidth{0.85\linewidth}\input{images/dataset_cornerdetection.tikz}}

		\subfigure[Smoothed velocity estimates of EdgeFlow and Lucas-Kanade.]{	\setlength\figureheight{3cm} 
		\setlength\figurewidth{0.85\linewidth}\input{images/flow_total.tikz}}

	\subfigure[
	Comparison values for EdgeFlow and Lucas-Kanade.]{
\hspace{1.3cm}
			\begin{tabular}{|c|cccc|}
				\hline  & \multicolumn{2}{c}{Lucas-Kanade} & \multicolumn{2}{c|}{EdgeFlow} \\

				& hor. & ver. & hor. & ver. \\
				\hline
				\hline
				MSE  & 0.0696 & 0.0576 & 0.0476 &  0.0320 \\
				NMXM &  0.3030  & 0.3041 & 0.6716 & 0.6604 \\
				\hline
				Comp. Time &  \multicolumn{2}{c}{0.1337 s} & \multicolumn{2}{c|}{0.0234 s} \\
					\hline
					\multicolumn{1}{c}{}&\multicolumn{1}{c}{}&\multicolumn{1}{c}{}&\multicolumn{1}{c}{}&\multicolumn{1}{c}{}\\
			\end{tabular}}

		\caption{Off-line results of the optical flow measurements: (a) the measure of feature-richness of the image data-set by Shi-Tomasi corner detection and (b) a comparison of Lucas-Kanade and EdgeFlow with horizontal velocity estimation. In (c), the MSE and NMXM values are shown for the entire data set of 440 images, compared to the OptiTrack's measured velocities.}
		\label{fig:edgeflowreal}
	\end{figure}
The previous sections explained the calculation of the translational flow, for convenience now dubbed as \emph{EdgeFlow}. As seen in Fig.~\ref{fig:velocity_estimation}, the velocity estimation $V_{est}$ can be calculated with the height of the drone and the angle from the center axis of the camera: 
\begin{eqnarray}
V_{est}=h*tan(p_t*FOV/w)/\Delta t
\end{eqnarray}

where $p_t$ is the flow vector, $h$ is the height of the drone relative to the ground, and $w$ stands for the pixels size of the image (in x or y direction). FOV stands for the field-of-view of the image sensor. A MAV can monitor its height by means of a sonar, barometer or GPS. In our case we do it differently, as we match the left and right edge histogram from the stereo-camera with global SAD matching. This implies that only one sensor is used for both velocity and height estimation. 

For off-board velocity estimation, a dataset of stereo-camera images is produced and synchronized with ground truth velocity data. The ground truth is measured by a motion tracking system with reflective markers (OptiTrack, 24 infrared-cameras). 
  This dataset excites both the horizontal and vertical flow directions, which is equivalent to the x- and y-axis of the image plane, and contains areas of varying amounts of textures (Fig. \ref{fig:trajectory}). As an indication of the texture-richness of the surface, the number of features, as detected by the Shi-Tomasi corner detection, is plotted in Fig. \ref{fig:edgeflowreal}(a).

For estimating the velocity, the scripts run in Matlab R2014b on a Dell Latitude E7450 with an Intel(R) Core(TM) i7-5600U CPU @ 2.60GHz processor.  In Fig.~\ref{fig:edgeflowreal}(b), the results of a single pyramid-layer implementation of the Lucas-Kanade algorithm with Shi-Tomasi corner detection can be seen (from \cite{bouguet2001pyramidal}). The mean of the detected horizontal velocity vectors is shown per time frame and plotted against the measured velocity by the OptiTrack system, as well as the velocity measured by EdgeFlow. For Lucas-Kanade, the altitude data of the OptiTrack is used. For EdgeFlow, the height is determined by the stereo images alone by histogram matching.

In Fig.~\ref{fig:edgeflowreal}(c),  comparison values are shown of the EdgeFlow and Lucas-Kanade algorithm of the entire data set. The mean squared error (MSE) is lower for EdgeFlow than for Lucas-Kanade, where a lower value stands for a higher similarity between the compared velocity estimation and the OptiTrack data. The normalized maximum cross-correlation magnitude (NMXM) is used as a quality measure as well.  Here a higher value, between a range of 0 and 1, stands for a better shape correlation with the ground truth. The plot of Fig.~\ref{fig:edgeflowreal}(b) and the values in Fig.~\ref{fig:edgeflowreal}(c)  shows a better tracking of the velocity by EdgeFlow when compared. We think that the main reason for this is that it utilizes information present in lines, which are ignored in the corner detection stage of Lucas-Kanade. In terms of computational speed, the EdgeFlow algorithm has an average processing of 0.0234 sec for both velocity and height estimation, over 5 times faster than Lucas-Kanade. Although this algorithm is run off-board on a laptop computer, it is an indication of the computational efficiency of the algorithm. This is valuable as EdgeFlow needs to run embedded on the 4 gr stereo-board, which is done in the upcoming sections of this paper. 

\section{VELOCITY ESTIMATION AND CONTROL}\label{sec:speedcontrol}

\begin{figure}[t]
	\centering
		
		\scriptsize
	\subfigure[]{
	\includegraphics[width=0.4\linewidth]{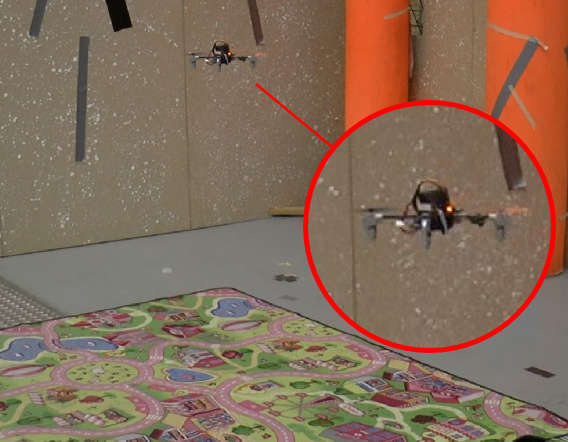}}\hfill 		\subfigure[]{				\def\svgwidth{.53\columnwidth}
	\executeiffilenewer{images/controlscheme.svg}{images/controlscheme.pdf}%
	{inkscape -z -D --file=images/controlscheme.svg %
		--export-pdf=images/controlscheme.pdf --export-latex}%
	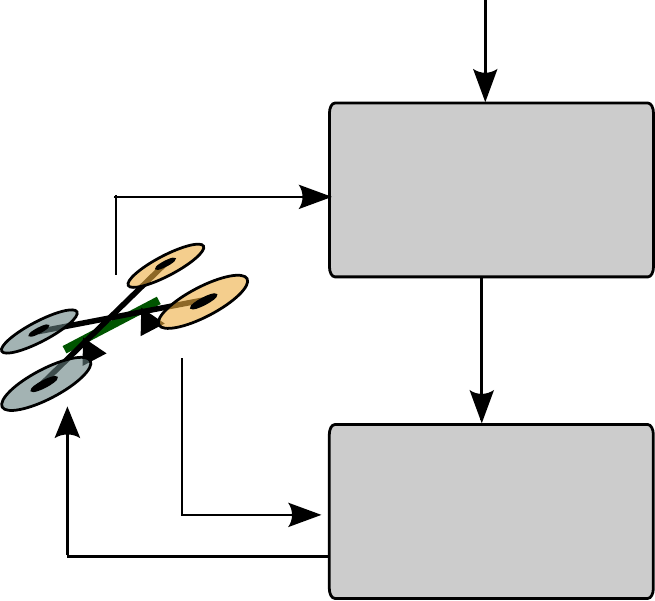%
}
	
\caption{ (a) A screen-shot of the video of the flight and (b) the control scheme of the velocity control. }
\label{fig:testsetup}
\end{figure}

The last subsection showed results with a data set of stereo images and OptiTrack data. In this section, the velocity estimated by EdgeFlow is run on-board the stereo-camera. Two platforms, an AR.Drone 2.0 and a pocket drone, will utilize the downward facing camera for velocity estimation and control.  Fig.~\ref{fig:testsetup}(a) gives a screen-shot of the video of the experiments\footnote{YouTube playlist:\\ https://www.youtube.com/playlist?list=PL\_KSX9GOn2P9TPb5nmFg-yH-UKC9eXbEE}, where it can be seen that the pocket drone is flying over a feature-rich mat. 
\begin{figure}[t]
	\centering

		\includegraphics[width=0.5\linewidth]{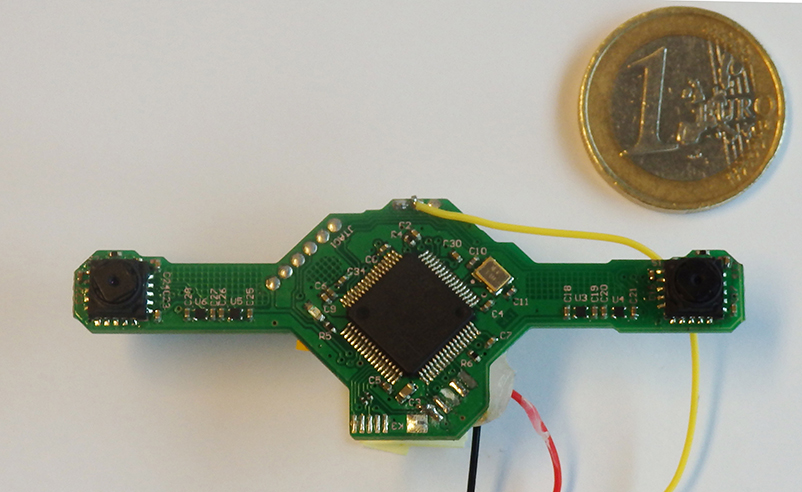}

	\caption{4 gram stereo-camera with a STM32F4 microprocessor with only 168 MHz speed and 192 kB of memory. The two cameras are located 6 cm apart from each other.}
	\label{fig:stereocamera}
\end{figure}



\begin{figure}[t]
	\begin{minipage}{\linewidth}
		
%
%
%
%
		
		\small
		\subfigure[Horizontal velocity estimate (MSE: 0.0224, NMXM: 0.6411)]{
			\setlength\figureheight{2.8cm} 
			\setlength\figurewidth{0.8\linewidth}\input{images/ardrone_velocity_estimate_control_hor.tikz}}
		\subfigure[Vertical velocity estimate (MSE: 0.0112, NMXM: 0.6068)]{
			\setlength\figureheight{2.5cm} 
			\setlength\figurewidth{0.84\linewidth}\input{images/ardrone_velocity_estimate_control_ver.tikz}}
		\caption{The velocity estimate of the AR.Drone 2.0 and stereo-board assembly during a velocity control task  with ground-truth as measured by OptiTrack. MSE and NMXM values are calculated for the entire flight.}
		\label{fig:onboardvelocity}
		
	\end{minipage}
\end{figure}
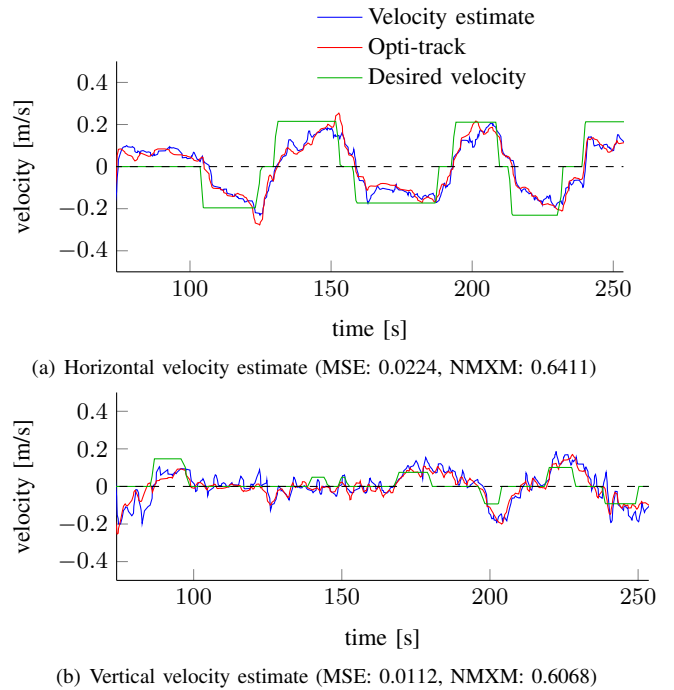

	\begin{figure}[t]	
		\centering
		\small
		\subfigure[Horizontal velocity estimate (MSE: 0.0064 m, NMXM: 0.5739)]{
			\setlength\figureheight{2.5cm} 
			\setlength\figurewidth{0.8\linewidth}\input{images/pocket_velocity_estimate_optitrack_hor.tikz}}
		\subfigure[Vertical velocity estimate (MSE: 0.0072 m, NMXM: 0.6066)]{
			\setlength\figureheight{2.5cm} 
			\setlength\figurewidth{0.84\linewidth}\input{images/pocket_velocity_estimate_optitrack_ver.tikz}}
		\caption{Velocity estimates calculated by the pocket drone and stereo-board assembly, during an OptiTrack guided flight. MSE and NMXM values are calculated for the entire flight.}
		\label{fig:onboardvelocitypocket}
		
	\end{figure}
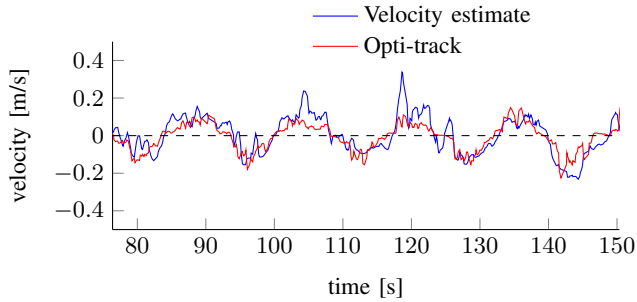
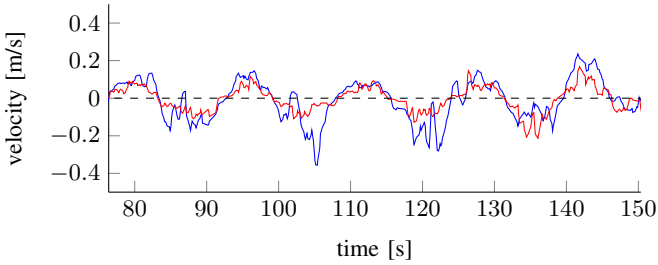

\subsection{Hardware and Software Specifics}
The AR.Drone 2.0\footnote{http://wiki.paparazziuav.org/wiki/AR\_Drone\_2} is a commercial drone with a weight of 380 grams and about 0.5 meter (with propellers considered) in diameter. The pocket drone\footnote{http://wiki.paparazziuav.org/wiki/Lisa/S/Tutorial/Nano\_Quadcopter} is 10 cm in diameter and has a total weight of 40 grams (including battery). 
It contains a Lisa S autopilot \cite{remes2014lisa}, which is mounted on a LadyBird quadcopter frame. The drone's movement is tracked by a motion tracking system, OptiTrack, which tracks passive reflective markers with its 24 infrared cameras. The registered motion will be used as ground truth to the experiments. 

The stereo-camera, introduced in \cite{de2014autonomous},  is attached to the bottom of both drones, facing downward to the ground plane (Fig.~\ref{fig:stereocamera}). It has two small cameras with two 1/6 inch image sensors, which are 6 cm apart. They have a horizontal FOV of 57.4$^o$ and vertical FOV of 44.5$^o$. The processor type is a STM32F4 with a speed of 168 MHz and 192 kB of memory. The processed stereo-camera images are grayscale and have $128 \times 96$ pixels. The maximum frame rate of the stereo-camera is 29 Hz, which is brought down to 25 Hz by the computation of EdgeFlow, with its average processing time of 0.0126 seconds. This is together with the height estimation using the same principle, all implemented on-board the stereo-camera.

The auto-pilot framework used for both MAV is  Paparazzi\footnote{http://wiki.paparazziuav.org/}. The AR.Drone 2.0's Wi-Fi and the pocket drone's Bluetooth module is used for communication with the Paparazzi ground, station to receive telemetry and send flight commands. Fig.~\ref{fig:testsetup}(b) shows the standard control scheme for the velocity control as implemented in paparazzi, which will receive a desired velocity references from the ground station for the guidance controller. This layer will send angle set-points to the attitude controller. The MAV's height should be kept constant by the altitude controller and measurements from the sonar (AR.drone) and barometer (pocket drone). Note that for these experiments, the height measured by the stereo-camera is only used for determining the velocity on-board and not for the control of the MAV's altitude.

\subsection{On-Board Velocity Control of a AR.Drone 2.0}

In this section, an AR.Drone 2.0 is used for velocity control with EdgeFlow, using the stereo-board instead of its standard bottom camera. Its difference with the desired velocity serves as the error signal for the guidance controller.
During the flight, several velocity references were sent to the AR.Drone, making it fly into specific direction. In Fig.~\ref{fig:onboardvelocity}, the stereo-camera's estimated velocity is plotted against its velocity measured by the OptiTrack for both horizontal and vertical direction of the image plane. This is equivalent to respectively sideways and forward direction in the AR.Drone's body fixed coordinate system. 

The AR.Drone was is able to determine its velocity with EdgeFlow computed on-board the stereo-camera, as the MSE and NMXM quality measures indicate a close correlation with the ground truth. This results in the AR.Drone's ability to correctly respond to the speed references given to the guidance controller


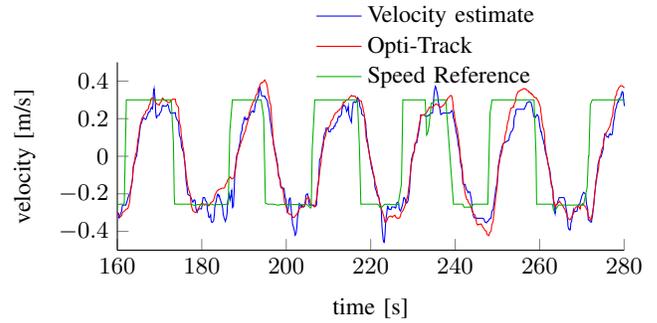
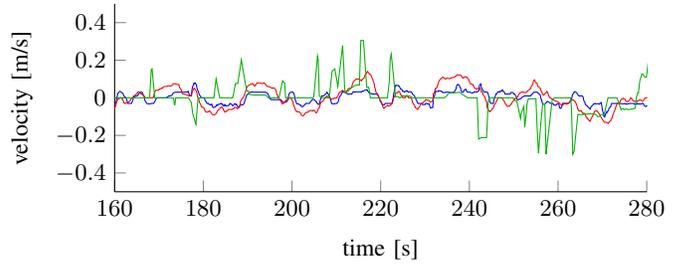
\begin{figure}[t]
	
	\centering
	\small

	\centering
	\small
	\subfigure[Horizontal velocity estimate (MSE: 0.0041 m, NMXM: 0.9631)]{
		\setlength\figureheight{2.5cm} 
		\setlength\figurewidth{0.8\linewidth}\input{images/pocket_velocity_estimate_control_ref_hor.tikz}}
	\subfigure[Vertical velocity estimate (MSE: 0.0025 m, NMXM: 0.7494)]{
		\setlength\figureheight{2.5cm} 
		\setlength\figurewidth{0.84\linewidth}\input{images/pocket_velocity_estimate_control_ref_ver.tikz}}
	\caption{Velocity estimates calculated by the pocket drone and stereo-board assembly, now using estimated velocity in the control. MSE and NMXM values are calculated for the entire flight which lasted for 370 seconds, where several external speed references were given for guidance.}
	\label{fig:onboardvelocitypocketcontrolref}

\end{figure}

.

\subsection{On-board Velocity Estimation of a Pocket Drone}

In the last subsection, we presented velocity control of an AR.Drone 2.0 to show the potential of using the stereo-camera for efficient velocity control. However, this needs to be shown on the pocket drone as well, which is smaller and hence has faster dynamics. Here the pocket drone is flown based on OptiTrack position measurement to present its on-board velocity estimation without using it in the control loop. During this flight, the velocity estimate calculated by the stereo-board is logged and plotted against its ground truth (Fig.~\ref{fig:onboardvelocitypocket}). 

The estimated velocity by the pocket drone is noisier than with the AR.Drone, which can be due of multiple reasons, from which the first is that the stereo-board is subjected to more vibrations on the pocket drone than the AR.Drone. This is because the camera is much closer to the rotors of the MAV and mounted directly on the frame. Another thing would be the control of the pocket drone, since it responds much faster as the AR.Drone. Additional filtering and de-rotation are essential to achieve the full on-board velocity control.

De-rotation is compensating for the camera rotations, where EdgeFlow will detect a flow not equivalent to translational velocity. Since the pocket drone has faster dynamics than the AR.Drone, the stereo-camera is subjected to faster rotations. De-rotation must be applied in order for the pocket drone to use optical flow for controls. In the experiments of the next subsection, the stereo-camera will receive rate measurement from the gyroscope. Hre it can estimate the resulting pixel shift in between frames due to rotation. The starting position of the histogram window search in the other image is offset with that pixel shift (an addition to section \ref{sec:optflow} A). 

%
%
%
%

\subsection{On-board Velocity Control of a Pocket drone}


Now the velocity estimate is used in the guidance control of the pocket drone and the OptiTrack measurements is only used for validation. The pocket drone's flight, during a guidance control task with externally given speed references, lasted for 370 seconds. Mostly horizontal (sideways) speed references where given,  however occasional horizontal speed references in the vertical direction were necessary to keep the pocket drone flying over the designated testing area. A portion of the velocity estimates during that same flight are displayed in Fig.~\ref{fig:onboardvelocitypocketcontrolref}. From the MSE and NMXM quality values for the horizontal speed, it can be determined that the EdgeFlow's estimated velocity correlates well with the ground truth. The pocket drone obeys the speed references given to the guidance controller.

 Noticeable in Fig.~\ref{fig:onboardvelocitypocketcontrolref}(b) is that the NMXM for vertical direction is lower than for the horizontal. As most of the speed references send to the guidance controller were for the horizontal direction, the correlation in shape is a lot more eminent, hence resulting in a higher NMXM value. Overall, it can be concluded that pocket drone can use the 4 gr stereo-board for its own velocity controlled guidance.

\section{CONCLUSION}\label{sec:conclusion}

In this paper we introduced a computationally efficient optical flow algorithm, which can run on a 4 gram stereo-camera with limited processing capabilities. The algorithm EdgeFlow uses a compressed representation of an image frame to match it with a previous time step. The adaptive time horizon enabled it to also detect sub-pixel flow, from which slower velocity could be estimated.

The stereo-camera is light enough to be carried by a 40 gram pocket drone. Together with the height and the optical flow calculated on-board, it can estimate its own velocity. The pocket drone uses that information within a guidance control loop, which enables it to compensate for drift and respond to external speed references. Our next focus is to use the same principle for a forward facing camera.

\addtolength{\textheight}{-12cm}   






\bibliographystyle{IEEEtran}
\bibliography{IEEEabrv,library_icra2016}

\end{document}

%% file: images/dronestabilizationwithflow.pdf_tex
\begingroup%
  \makeatletter%
  \providecommand\color[2][]{%
    \errmessage{(Inkscape) Color is used for the text in Inkscape, but the package 'color.sty' is not loaded}%
    \renewcommand\color[2][]{}%
  }%
  \providecommand\transparent[1]{%
    \errmessage{(Inkscape) Transparency is used (non-zero) for the text in Inkscape, but the package 'transparent.sty' is not loaded}%
    \renewcommand\transparent[1]{}%
  }%
  \providecommand\rotatebox[2]{#2}%
  \ifx\svgwidth\undefined%
    \setlength{\unitlength}{386.47575684bp}%
    \ifx\svgscale\undefined%
      \relax%
    \else%
      \setlength{\unitlength}{\unitlength * \real{\svgscale}}%
    \fi%
  \else%
    \setlength{\unitlength}{\svgwidth}%
  \fi%
  \global\let\svgwidth\undefined%
  \global\let\svgscale\undefined%
  \makeatother%
  \begin{picture}(1,0.61854725)%
    \put(0,0){\includegraphics[width=\unitlength]{dronestabilizationwithflow.pdf}}%
    \put(0.29076869,0.00435627){\color[rgb]{0,0,0}\makebox(0,0)[lb]{\smash{\color{blue}time}}}%
    \put(0.83320833,0.50953677){\color[rgb]{0,0,0}\makebox(0,0)[lb]{\smash{\color{red}Velocity}}}%
    \put(0.70776025,0.28734242){\color[rgb]{0,0.35294118,0}\makebox(0,0)[lb]{\smash{\color{blue}Flow}}}%
    \put(0.57805397,0.43055441){\color[rgb]{0,0,0}\rotatebox{90}{\makebox(0,0)[lb]{\smash{Height}}}}%
    \put(0.29836974,0.53886846){\color[rgb]{0,0,0}\makebox(0,0)[lb]{\smash{Stereo camera}}}%
    \put(0.01916435,0.16711081){\color[rgb]{0,0,0}\makebox(0,0)[lb]{\smash{Weight 45 gr}}}%
  \end{picture}%
\endgroup%

%% file: images/edge_histogram2.pdf_tex
\begingroup%
  \makeatletter%
  \providecommand\color[2][]{%
    \errmessage{(Inkscape) Color is used for the text in Inkscape, but the package 'color.sty' is not loaded}%
    \renewcommand\color[2][]{}%
  }%
  \providecommand\transparent[1]{%
    \errmessage{(Inkscape) Transparency is used (non-zero) for the text in Inkscape, but the package 'transparent.sty' is not loaded}%
    \renewcommand\transparent[1]{}%
  }%
  \providecommand\rotatebox[2]{#2}%
  \ifx\svgwidth\undefined%
    \setlength{\unitlength}{438.65bp}%
    \ifx\svgscale\undefined%
      \relax%
    \else%
      \setlength{\unitlength}{\unitlength * \real{\svgscale}}%
    \fi%
  \else%
    \setlength{\unitlength}{\svgwidth}%
  \fi%
  \global\let\svgwidth\undefined%
  \global\let\svgscale\undefined%
  \makeatother%
  \begin{picture}(1,0.32175781)%
    \put(0,0){\includegraphics[width=\unitlength]{edge_histogram2.pdf}}%
    \put(0.53946173,0.30848021){\color[rgb]{0,0,0}\makebox(0,0)[lb]{\smash{Image (t)}}}%
    \put(0.54116262,0.17147488){\color[rgb]{1,1,1}\makebox(0,0)[lb]{\smash{\color{white}Sobel}}}%
    \put(0.54116262,0.00383813){\color[rgb]{1,0,0}\makebox(0,0)[lb]{\smash{\color{red} Edge Histogram (t)}}}%
    \put(0.00900905,0.30848021){\color[rgb]{0,0,0}\makebox(0,0)[lb]{\smash{Image (t - 1)}}}%
    \put(0.01070994,0.17147488){\color[rgb]{1,1,1}\makebox(0,0)[lb]{\smash{\color{white}Sobel}}}%
    \put(0.01070994,0.00383813){\color[rgb]{1,0,0}\makebox(0,0)[lb]{\smash{\color{darkgreen} Edge Histogram (t-1)}}}%
  \end{picture}%
\endgroup%

%% file: images/edge_hist_match.tikz
%
%
\definecolor{mycolor1}{rgb}{0.00000,0.75000,0.75000}%
\begin{tikzpicture}

\begin{axis}[%
width=0.95092\figurewidth,
height=\figureheight,
at={(0\figurewidth,0\figureheight)},
scale only axis,
every outer x axis line/.append style={black},
every x tick label/.append style={font=\color{black}},
xmin=0.000,
xmax=300.000,
every outer y axis line/.append style={black},
every y tick label/.append style={font=\color{black}},
ymin=0.000,
ymax=10.000,
ytick={0, 5},
ylabel={Displacement [px]},
ylabel style={yshift=-8.5cm},
axis x line*=bottom,
axis y line*=right
]
\addplot [color=blue,solid,forget plot]
  table[row sep=crcr]{%
1.000	0.000\\
2.000	0.000\\
3.000	0.000\\
4.000	0.000\\
5.000	0.000\\
6.000	0.000\\
7.000	0.000\\
8.000	0.000\\
9.000	0.000\\
10.000	0.000\\
11.000	0.000\\
12.000	0.000\\
13.000	0.000\\
14.000	0.000\\
15.000	0.000\\
16.000	0.000\\
17.000	0.000\\
18.000	0.000\\
19.000	0.000\\
20.000	3.000\\
21.000	3.000\\
22.000	3.000\\
23.000	3.000\\
24.000	3.000\\
25.000	3.000\\
26.000	3.000\\
27.000	3.000\\
28.000	3.000\\
29.000	3.000\\
30.000	3.000\\
31.000	3.000\\
32.000	3.000\\
33.000	3.000\\
34.000	3.000\\
35.000	3.000\\
36.000	3.000\\
37.000	3.000\\
38.000	3.000\\
39.000	3.000\\
40.000	3.000\\
41.000	3.000\\
42.000	3.000\\
43.000	3.000\\
44.000	3.000\\
45.000	3.000\\
46.000	3.000\\
47.000	3.000\\
48.000	3.000\\
49.000	3.000\\
50.000	3.000\\
51.000	3.000\\
52.000	3.000\\
53.000	3.000\\
54.000	3.000\\
55.000	3.000\\
56.000	3.000\\
57.000	3.000\\
58.000	3.000\\
59.000	3.000\\
60.000	3.000\\
61.000	3.000\\
62.000	3.000\\
63.000	3.000\\
64.000	3.000\\
65.000	3.000\\
66.000	3.000\\
67.000	3.000\\
68.000	3.000\\
69.000	3.000\\
70.000	3.000\\
71.000	3.000\\
72.000	3.000\\
73.000	2.000\\
74.000	3.000\\
75.000	2.000\\
76.000	3.000\\
77.000	3.000\\
78.000	3.000\\
79.000	3.000\\
80.000	3.000\\
81.000	3.000\\
82.000	3.000\\
83.000	3.000\\
84.000	3.000\\
85.000	3.000\\
86.000	3.000\\
87.000	3.000\\
88.000	3.000\\
89.000	3.000\\
90.000	3.000\\
91.000	3.000\\
92.000	3.000\\
93.000	3.000\\
94.000	3.000\\
95.000	3.000\\
96.000	2.000\\
97.000	2.000\\
98.000	2.000\\
99.000	2.000\\
100.000	2.000\\
101.000	2.000\\
102.000	2.000\\
103.000	2.000\\
104.000	2.000\\
105.000	2.000\\
106.000	2.000\\
107.000	2.000\\
108.000	2.000\\
109.000	2.000\\
110.000	2.000\\
111.000	2.000\\
112.000	2.000\\
113.000	2.000\\
114.000	2.000\\
115.000	2.000\\
116.000	2.000\\
117.000	2.000\\
118.000	2.000\\
119.000	2.000\\
120.000	2.000\\
121.000	3.000\\
122.000	3.000\\
123.000	3.000\\
124.000	3.000\\
125.000	3.000\\
126.000	3.000\\
127.000	3.000\\
128.000	3.000\\
129.000	3.000\\
130.000	3.000\\
131.000	3.000\\
132.000	3.000\\
133.000	3.000\\
134.000	3.000\\
135.000	3.000\\
136.000	2.000\\
137.000	3.000\\
138.000	3.000\\
139.000	3.000\\
140.000	3.000\\
141.000	3.000\\
142.000	3.000\\
143.000	3.000\\
144.000	3.000\\
145.000	3.000\\
146.000	2.000\\
147.000	2.000\\
148.000	2.000\\
149.000	2.000\\
150.000	2.000\\
151.000	2.000\\
152.000	2.000\\
153.000	2.000\\
154.000	2.000\\
155.000	2.000\\
156.000	2.000\\
157.000	2.000\\
158.000	2.000\\
159.000	2.000\\
160.000	2.000\\
161.000	2.000\\
162.000	2.000\\
163.000	2.000\\
164.000	2.000\\
165.000	2.000\\
166.000	2.000\\
167.000	2.000\\
168.000	2.000\\
169.000	2.000\\
170.000	2.000\\
171.000	2.000\\
172.000	2.000\\
173.000	2.000\\
174.000	2.000\\
175.000	2.000\\
176.000	2.000\\
177.000	2.000\\
178.000	2.000\\
179.000	2.000\\
180.000	2.000\\
181.000	2.000\\
182.000	2.000\\
183.000	2.000\\
184.000	2.000\\
185.000	2.000\\
186.000	2.000\\
187.000	2.000\\
188.000	2.000\\
189.000	2.000\\
190.000	2.000\\
191.000	2.000\\
192.000	2.000\\
193.000	2.000\\
194.000	2.000\\
195.000	2.000\\
196.000	2.000\\
197.000	2.000\\
198.000	2.000\\
199.000	2.000\\
200.000	2.000\\
201.000	2.000\\
202.000	2.000\\
203.000	2.000\\
204.000	2.000\\
205.000	2.000\\
206.000	2.000\\
207.000	2.000\\
208.000	2.000\\
209.000	2.000\\
210.000	2.000\\
211.000	2.000\\
212.000	2.000\\
213.000	2.000\\
214.000	2.000\\
215.000	2.000\\
216.000	2.000\\
217.000	2.000\\
218.000	2.000\\
219.000	2.000\\
220.000	2.000\\
221.000	2.000\\
222.000	2.000\\
223.000	2.000\\
224.000	2.000\\
225.000	2.000\\
226.000	2.000\\
227.000	2.000\\
228.000	2.000\\
229.000	2.000\\
230.000	2.000\\
231.000	1.000\\
232.000	1.000\\
233.000	1.000\\
234.000	1.000\\
235.000	1.000\\
236.000	1.000\\
237.000	0.000\\
238.000	0.000\\
239.000	0.000\\
240.000	0.000\\
241.000	0.000\\
242.000	0.000\\
243.000	0.000\\
244.000	0.000\\
245.000	0.000\\
246.000	0.000\\
247.000	0.000\\
248.000	0.000\\
249.000	0.000\\
250.000	0.000\\
251.000	0.000\\
252.000	0.000\\
253.000	0.000\\
254.000	0.000\\
255.000	0.000\\
256.000	0.000\\
};
\addplot [color=black,dashed,forget plot]
  table[row sep=crcr]{%
1.000	3.284\\
2.000	3.277\\
3.000	3.270\\
4.000	3.264\\
5.000	3.257\\
6.000	3.250\\
7.000	3.243\\
8.000	3.237\\
9.000	3.230\\
10.000	3.223\\
11.000	3.216\\
12.000	3.209\\
13.000	3.203\\
14.000	3.196\\
15.000	3.189\\
16.000	3.182\\
17.000	3.176\\
18.000	3.169\\
19.000	3.162\\
20.000	3.155\\
21.000	3.149\\
22.000	3.142\\
23.000	3.135\\
24.000	3.128\\
25.000	3.121\\
26.000	3.115\\
27.000	3.108\\
28.000	3.101\\
29.000	3.094\\
30.000	3.088\\
31.000	3.081\\
32.000	3.074\\
33.000	3.067\\
34.000	3.060\\
35.000	3.054\\
36.000	3.047\\
37.000	3.040\\
38.000	3.033\\
39.000	3.027\\
40.000	3.020\\
41.000	3.013\\
42.000	3.006\\
43.000	3.000\\
44.000	2.993\\
45.000	2.986\\
46.000	2.979\\
47.000	2.972\\
48.000	2.966\\
49.000	2.959\\
50.000	2.952\\
51.000	2.945\\
52.000	2.939\\
53.000	2.932\\
54.000	2.925\\
55.000	2.918\\
56.000	2.912\\
57.000	2.905\\
58.000	2.898\\
59.000	2.891\\
60.000	2.884\\
61.000	2.878\\
62.000	2.871\\
63.000	2.864\\
64.000	2.857\\
65.000	2.851\\
66.000	2.844\\
67.000	2.837\\
68.000	2.830\\
69.000	2.823\\
70.000	2.817\\
71.000	2.810\\
72.000	2.803\\
73.000	2.796\\
74.000	2.790\\
75.000	2.783\\
76.000	2.776\\
77.000	2.769\\
78.000	2.763\\
79.000	2.756\\
80.000	2.749\\
81.000	2.742\\
82.000	2.735\\
83.000	2.729\\
84.000	2.722\\
85.000	2.715\\
86.000	2.708\\
87.000	2.702\\
88.000	2.695\\
89.000	2.688\\
90.000	2.681\\
91.000	2.675\\
92.000	2.668\\
93.000	2.661\\
94.000	2.654\\
95.000	2.647\\
96.000	2.641\\
97.000	2.634\\
98.000	2.627\\
99.000	2.620\\
100.000	2.614\\
101.000	2.607\\
102.000	2.600\\
103.000	2.593\\
104.000	2.586\\
105.000	2.580\\
106.000	2.573\\
107.000	2.566\\
108.000	2.559\\
109.000	2.553\\
110.000	2.546\\
111.000	2.539\\
112.000	2.532\\
113.000	2.526\\
114.000	2.519\\
115.000	2.512\\
116.000	2.505\\
117.000	2.498\\
118.000	2.492\\
119.000	2.485\\
120.000	2.478\\
121.000	2.471\\
122.000	2.465\\
123.000	2.458\\
124.000	2.451\\
125.000	2.444\\
126.000	2.438\\
127.000	2.431\\
128.000	2.424\\
129.000	2.417\\
130.000	2.410\\
131.000	2.404\\
132.000	2.397\\
133.000	2.390\\
134.000	2.383\\
135.000	2.377\\
136.000	2.370\\
137.000	2.363\\
138.000	2.356\\
139.000	2.349\\
140.000	2.343\\
141.000	2.336\\
142.000	2.329\\
143.000	2.322\\
144.000	2.316\\
145.000	2.309\\
146.000	2.302\\
147.000	2.295\\
148.000	2.289\\
149.000	2.282\\
150.000	2.275\\
151.000	2.268\\
152.000	2.261\\
153.000	2.255\\
154.000	2.248\\
155.000	2.241\\
156.000	2.234\\
157.000	2.228\\
158.000	2.221\\
159.000	2.214\\
160.000	2.207\\
161.000	2.201\\
162.000	2.194\\
163.000	2.187\\
164.000	2.180\\
165.000	2.173\\
166.000	2.167\\
167.000	2.160\\
168.000	2.153\\
169.000	2.146\\
170.000	2.140\\
171.000	2.133\\
172.000	2.126\\
173.000	2.119\\
174.000	2.112\\
175.000	2.106\\
176.000	2.099\\
177.000	2.092\\
178.000	2.085\\
179.000	2.079\\
180.000	2.072\\
181.000	2.065\\
182.000	2.058\\
183.000	2.052\\
184.000	2.045\\
185.000	2.038\\
186.000	2.031\\
187.000	2.024\\
188.000	2.018\\
189.000	2.011\\
190.000	2.004\\
191.000	1.997\\
192.000	1.991\\
193.000	1.984\\
194.000	1.977\\
195.000	1.970\\
196.000	1.964\\
197.000	1.957\\
198.000	1.950\\
199.000	1.943\\
200.000	1.936\\
201.000	1.930\\
202.000	1.923\\
203.000	1.916\\
204.000	1.909\\
205.000	1.903\\
206.000	1.896\\
207.000	1.889\\
208.000	1.882\\
209.000	1.875\\
210.000	1.869\\
211.000	1.862\\
212.000	1.855\\
213.000	1.848\\
214.000	1.842\\
215.000	1.835\\
216.000	1.828\\
217.000	1.821\\
218.000	1.815\\
219.000	1.808\\
220.000	1.801\\
221.000	1.794\\
222.000	1.787\\
223.000	1.781\\
224.000	1.774\\
225.000	1.767\\
226.000	1.760\\
227.000	1.754\\
228.000	1.747\\
229.000	1.740\\
230.000	1.733\\
231.000	1.726\\
232.000	1.720\\
233.000	1.713\\
234.000	1.706\\
235.000	1.699\\
236.000	1.693\\
237.000	1.686\\
238.000	1.679\\
239.000	1.672\\
240.000	1.666\\
241.000	1.659\\
242.000	1.652\\
243.000	1.645\\
244.000	1.638\\
245.000	1.632\\
246.000	1.625\\
247.000	1.618\\
248.000	1.611\\
249.000	1.605\\
250.000	1.598\\
251.000	1.591\\
252.000	1.584\\
253.000	1.578\\
254.000	1.571\\
255.000	1.564\\
256.000	1.557\\
};
\end{axis}

\begin{axis}[%
width=0.95092\figurewidth,
height=\figureheight,
at={(0\figurewidth,0\figureheight)},
scale only axis,
separate axis lines,
every outer x axis line/.append style={black},
every x tick label/.append style={font=\color{black}},
xmin=0.000,
xmax=300.000,
xlabel={Position on x-axis of image [px]},
every outer y axis line/.append style={black},
every y tick label/.append style={font=\color{black}},
ymin=0.000,
ymax=5000.000,
ytick={   0, 5000},
ylabel={Amount of features},
ylabel style={yshift=-0.5cm},
legend style={legend cell align=left,align=left,fill=none,draw=none}
]
\addplot [color=black!50!green,solid]
  table[row sep=crcr]{%
1.000	0.000\\
2.000	73.000\\
3.000	82.000\\
4.000	131.000\\
5.000	96.000\\
6.000	109.000\\
7.000	141.000\\
8.000	112.000\\
9.000	148.000\\
10.000	264.000\\
11.000	203.000\\
12.000	138.000\\
13.000	164.000\\
14.000	418.000\\
15.000	452.000\\
16.000	359.000\\
17.000	507.000\\
18.000	628.000\\
19.000	709.000\\
20.000	755.000\\
21.000	802.000\\
22.000	616.000\\
23.000	610.000\\
24.000	648.000\\
25.000	674.000\\
26.000	714.000\\
27.000	436.000\\
28.000	380.000\\
29.000	659.000\\
30.000	578.000\\
31.000	395.000\\
32.000	533.000\\
33.000	431.000\\
34.000	345.000\\
35.000	433.000\\
36.000	429.000\\
37.000	390.000\\
38.000	493.000\\
39.000	566.000\\
40.000	796.000\\
41.000	1365.000\\
42.000	1616.000\\
43.000	1425.000\\
44.000	1172.000\\
45.000	483.000\\
46.000	177.000\\
47.000	217.000\\
48.000	175.000\\
49.000	150.000\\
50.000	154.000\\
51.000	136.000\\
52.000	218.000\\
53.000	1101.000\\
54.000	2872.000\\
55.000	3405.000\\
56.000	1589.000\\
57.000	382.000\\
58.000	303.000\\
59.000	302.000\\
60.000	308.000\\
61.000	290.000\\
62.000	292.000\\
63.000	373.000\\
64.000	217.000\\
65.000	270.000\\
66.000	292.000\\
67.000	344.000\\
68.000	485.000\\
69.000	426.000\\
70.000	567.000\\
71.000	308.000\\
72.000	279.000\\
73.000	491.000\\
74.000	479.000\\
75.000	392.000\\
76.000	270.000\\
77.000	342.000\\
78.000	388.000\\
79.000	280.000\\
80.000	207.000\\
81.000	764.000\\
82.000	1448.000\\
83.000	605.000\\
84.000	435.000\\
85.000	158.000\\
86.000	308.000\\
87.000	1078.000\\
88.000	1475.000\\
89.000	1066.000\\
90.000	373.000\\
91.000	25.000\\
92.000	17.000\\
93.000	12.000\\
94.000	55.000\\
95.000	79.000\\
96.000	36.000\\
97.000	80.000\\
98.000	55.000\\
99.000	92.000\\
100.000	94.000\\
101.000	156.000\\
102.000	181.000\\
103.000	1313.000\\
104.000	3719.000\\
105.000	4159.000\\
106.000	2187.000\\
107.000	1065.000\\
108.000	1207.000\\
109.000	1068.000\\
110.000	492.000\\
111.000	522.000\\
112.000	485.000\\
113.000	1724.000\\
114.000	1303.000\\
115.000	249.000\\
116.000	382.000\\
117.000	1165.000\\
118.000	2987.000\\
119.000	3119.000\\
120.000	2098.000\\
121.000	1191.000\\
122.000	894.000\\
123.000	655.000\\
124.000	435.000\\
125.000	289.000\\
126.000	205.000\\
127.000	157.000\\
128.000	98.000\\
129.000	62.000\\
130.000	28.000\\
131.000	11.000\\
132.000	9.000\\
133.000	7.000\\
134.000	8.000\\
135.000	16.000\\
136.000	10.000\\
137.000	8.000\\
138.000	10.000\\
139.000	16.000\\
140.000	9.000\\
141.000	22.000\\
142.000	24.000\\
143.000	12.000\\
144.000	22.000\\
145.000	25.000\\
146.000	115.000\\
147.000	691.000\\
148.000	1444.000\\
149.000	814.000\\
150.000	94.000\\
151.000	82.000\\
152.000	79.000\\
153.000	107.000\\
154.000	217.000\\
155.000	1157.000\\
156.000	2417.000\\
157.000	3127.000\\
158.000	2368.000\\
159.000	1024.000\\
160.000	295.000\\
161.000	172.000\\
162.000	253.000\\
163.000	345.000\\
164.000	424.000\\
165.000	373.000\\
166.000	458.000\\
167.000	238.000\\
168.000	184.000\\
169.000	597.000\\
170.000	1663.000\\
171.000	2835.000\\
172.000	2070.000\\
173.000	449.000\\
174.000	116.000\\
175.000	115.000\\
176.000	176.000\\
177.000	258.000\\
178.000	511.000\\
179.000	995.000\\
180.000	2569.000\\
181.000	2272.000\\
182.000	644.000\\
183.000	105.000\\
184.000	360.000\\
185.000	339.000\\
186.000	178.000\\
187.000	307.000\\
188.000	1055.000\\
189.000	2399.000\\
190.000	2806.000\\
191.000	2202.000\\
192.000	1383.000\\
193.000	2013.000\\
194.000	1157.000\\
195.000	657.000\\
196.000	677.000\\
197.000	661.000\\
198.000	761.000\\
199.000	853.000\\
200.000	688.000\\
201.000	554.000\\
202.000	528.000\\
203.000	432.000\\
204.000	368.000\\
205.000	368.000\\
206.000	440.000\\
207.000	360.000\\
208.000	335.000\\
209.000	285.000\\
210.000	271.000\\
211.000	286.000\\
212.000	424.000\\
213.000	436.000\\
214.000	814.000\\
215.000	2086.000\\
216.000	3121.000\\
217.000	3249.000\\
218.000	3116.000\\
219.000	2825.000\\
220.000	1539.000\\
221.000	612.000\\
222.000	335.000\\
223.000	291.000\\
224.000	363.000\\
225.000	349.000\\
226.000	343.000\\
227.000	327.000\\
228.000	366.000\\
229.000	375.000\\
230.000	329.000\\
231.000	389.000\\
232.000	453.000\\
233.000	804.000\\
234.000	1460.000\\
235.000	1870.000\\
236.000	1766.000\\
237.000	1488.000\\
238.000	925.000\\
239.000	388.000\\
240.000	391.000\\
241.000	399.000\\
242.000	489.000\\
243.000	433.000\\
244.000	408.000\\
245.000	433.000\\
246.000	479.000\\
247.000	641.000\\
248.000	563.000\\
249.000	460.000\\
250.000	519.000\\
251.000	333.000\\
252.000	316.000\\
253.000	312.000\\
254.000	345.000\\
255.000	379.000\\
256.000	0.000\\
};
\addlegendentry{Edge-histogram (t-1)};

\addplot [color=red,solid]
  table[row sep=crcr]{%
1.000	0.000\\
2.000	102.000\\
3.000	223.000\\
4.000	289.000\\
5.000	168.000\\
6.000	74.000\\
7.000	138.000\\
8.000	118.000\\
9.000	106.000\\
10.000	158.000\\
11.000	109.000\\
12.000	122.000\\
13.000	268.000\\
14.000	226.000\\
15.000	139.000\\
16.000	134.000\\
17.000	286.000\\
18.000	454.000\\
19.000	462.000\\
20.000	462.000\\
21.000	550.000\\
22.000	687.000\\
23.000	749.000\\
24.000	847.000\\
25.000	627.000\\
26.000	609.000\\
27.000	681.000\\
28.000	627.000\\
29.000	749.000\\
30.000	489.000\\
31.000	339.000\\
32.000	549.000\\
33.000	594.000\\
34.000	391.000\\
35.000	542.000\\
36.000	459.000\\
37.000	310.000\\
38.000	429.000\\
39.000	442.000\\
40.000	338.000\\
41.000	441.000\\
42.000	531.000\\
43.000	699.000\\
44.000	1279.000\\
45.000	1578.000\\
46.000	1464.000\\
47.000	1327.000\\
48.000	580.000\\
49.000	193.000\\
50.000	218.000\\
51.000	187.000\\
52.000	119.000\\
53.000	142.000\\
54.000	161.000\\
55.000	259.000\\
56.000	1198.000\\
57.000	2953.000\\
58.000	3259.000\\
59.000	1463.000\\
60.000	307.000\\
61.000	265.000\\
62.000	260.000\\
63.000	337.000\\
64.000	301.000\\
65.000	340.000\\
66.000	313.000\\
67.000	222.000\\
68.000	298.000\\
69.000	266.000\\
70.000	478.000\\
71.000	531.000\\
72.000	461.000\\
73.000	469.000\\
74.000	261.000\\
75.000	336.000\\
76.000	520.000\\
77.000	458.000\\
78.000	280.000\\
79.000	272.000\\
80.000	435.000\\
81.000	361.000\\
82.000	237.000\\
83.000	259.000\\
84.000	1318.000\\
85.000	1045.000\\
86.000	417.000\\
87.000	296.000\\
88.000	210.000\\
89.000	513.000\\
90.000	1218.000\\
91.000	1475.000\\
92.000	975.000\\
93.000	14.000\\
94.000	32.000\\
95.000	11.000\\
96.000	31.000\\
97.000	80.000\\
98.000	36.000\\
99.000	59.000\\
100.000	71.000\\
101.000	81.000\\
102.000	69.000\\
103.000	173.000\\
104.000	107.000\\
105.000	468.000\\
106.000	2761.000\\
107.000	4775.000\\
108.000	3097.000\\
109.000	972.000\\
110.000	1303.000\\
111.000	1144.000\\
112.000	541.000\\
113.000	510.000\\
114.000	377.000\\
115.000	1708.000\\
116.000	1417.000\\
117.000	281.000\\
118.000	439.000\\
119.000	618.000\\
120.000	1569.000\\
121.000	3504.000\\
122.000	2847.000\\
123.000	1745.000\\
124.000	1097.000\\
125.000	833.000\\
126.000	648.000\\
127.000	460.000\\
128.000	278.000\\
129.000	165.000\\
130.000	123.000\\
131.000	93.000\\
132.000	38.000\\
133.000	13.000\\
134.000	11.000\\
135.000	7.000\\
136.000	7.000\\
137.000	15.000\\
138.000	21.000\\
139.000	6.000\\
140.000	7.000\\
141.000	17.000\\
142.000	19.000\\
143.000	9.000\\
144.000	23.000\\
145.000	12.000\\
146.000	15.000\\
147.000	19.000\\
148.000	48.000\\
149.000	131.000\\
150.000	913.000\\
151.000	1442.000\\
152.000	542.000\\
153.000	127.000\\
154.000	86.000\\
155.000	119.000\\
156.000	261.000\\
157.000	1148.000\\
158.000	2482.000\\
159.000	3411.000\\
160.000	2305.000\\
161.000	743.000\\
162.000	195.000\\
163.000	188.000\\
164.000	289.000\\
165.000	373.000\\
166.000	435.000\\
167.000	455.000\\
168.000	377.000\\
169.000	178.000\\
170.000	276.000\\
171.000	981.000\\
172.000	2017.000\\
173.000	2790.000\\
174.000	1672.000\\
175.000	202.000\\
176.000	116.000\\
177.000	116.000\\
178.000	208.000\\
179.000	319.000\\
180.000	772.000\\
181.000	1641.000\\
182.000	2793.000\\
183.000	1483.000\\
184.000	227.000\\
185.000	176.000\\
186.000	443.000\\
187.000	210.000\\
188.000	209.000\\
189.000	457.000\\
190.000	1551.000\\
191.000	2357.000\\
192.000	2787.000\\
193.000	1997.000\\
194.000	1597.000\\
195.000	1862.000\\
196.000	783.000\\
197.000	733.000\\
198.000	882.000\\
199.000	632.000\\
200.000	748.000\\
201.000	863.000\\
202.000	696.000\\
203.000	570.000\\
204.000	557.000\\
205.000	418.000\\
206.000	340.000\\
207.000	386.000\\
208.000	424.000\\
209.000	357.000\\
210.000	308.000\\
211.000	295.000\\
212.000	294.000\\
213.000	369.000\\
214.000	402.000\\
215.000	442.000\\
216.000	1005.000\\
217.000	2231.000\\
218.000	3054.000\\
219.000	2952.000\\
220.000	2913.000\\
221.000	2761.000\\
222.000	1679.000\\
223.000	584.000\\
224.000	299.000\\
225.000	317.000\\
226.000	355.000\\
227.000	377.000\\
228.000	323.000\\
229.000	353.000\\
230.000	365.000\\
231.000	349.000\\
232.000	369.000\\
233.000	389.000\\
234.000	625.000\\
235.000	1169.000\\
236.000	1683.000\\
237.000	1792.000\\
238.000	1575.000\\
239.000	1233.000\\
240.000	625.000\\
241.000	370.000\\
242.000	444.000\\
243.000	443.000\\
244.000	480.000\\
245.000	391.000\\
246.000	422.000\\
247.000	472.000\\
248.000	502.000\\
249.000	624.000\\
250.000	505.000\\
251.000	532.000\\
252.000	463.000\\
253.000	305.000\\
254.000	354.000\\
255.000	396.000\\
256.000	0.000\\
};
\addlegendentry{Edge-histogram (t)};

\addplot [color=blue,solid]
  table[row sep=crcr]{%
1.000	0.000\\
2.000	0.000\\
3.000	0.000\\
4.000	0.000\\
5.000	0.000\\
6.000	0.000\\
7.000	0.000\\
8.000	0.000\\
9.000	0.000\\
10.000	0.000\\
11.000	0.000\\
12.000	0.000\\
13.000	0.000\\
14.000	0.000\\
15.000	0.000\\
16.000	0.000\\
17.000	0.000\\
18.000	0.000\\
19.000	0.000\\
20.000	3.000\\
21.000	3.000\\
22.000	3.000\\
23.000	3.000\\
24.000	3.000\\
25.000	3.000\\
26.000	3.000\\
27.000	3.000\\
28.000	3.000\\
29.000	3.000\\
30.000	3.000\\
31.000	3.000\\
32.000	3.000\\
33.000	3.000\\
34.000	3.000\\
35.000	3.000\\
36.000	3.000\\
37.000	3.000\\
38.000	3.000\\
39.000	3.000\\
40.000	3.000\\
41.000	3.000\\
42.000	3.000\\
43.000	3.000\\
44.000	3.000\\
45.000	3.000\\
46.000	3.000\\
47.000	3.000\\
48.000	3.000\\
49.000	3.000\\
50.000	3.000\\
51.000	3.000\\
52.000	3.000\\
53.000	3.000\\
54.000	3.000\\
55.000	3.000\\
56.000	3.000\\
57.000	3.000\\
58.000	3.000\\
59.000	3.000\\
60.000	3.000\\
61.000	3.000\\
62.000	3.000\\
63.000	3.000\\
64.000	3.000\\
65.000	3.000\\
66.000	3.000\\
67.000	3.000\\
68.000	3.000\\
69.000	3.000\\
70.000	3.000\\
71.000	3.000\\
72.000	3.000\\
73.000	2.000\\
74.000	3.000\\
75.000	2.000\\
76.000	3.000\\
77.000	3.000\\
78.000	3.000\\
79.000	3.000\\
80.000	3.000\\
81.000	3.000\\
82.000	3.000\\
83.000	3.000\\
84.000	3.000\\
85.000	3.000\\
86.000	3.000\\
87.000	3.000\\
88.000	3.000\\
89.000	3.000\\
90.000	3.000\\
91.000	3.000\\
92.000	3.000\\
93.000	3.000\\
94.000	3.000\\
95.000	3.000\\
96.000	2.000\\
97.000	2.000\\
98.000	2.000\\
99.000	2.000\\
100.000	2.000\\
101.000	2.000\\
102.000	2.000\\
103.000	2.000\\
104.000	2.000\\
105.000	2.000\\
106.000	2.000\\
107.000	2.000\\
108.000	2.000\\
109.000	2.000\\
110.000	2.000\\
111.000	2.000\\
112.000	2.000\\
113.000	2.000\\
114.000	2.000\\
115.000	2.000\\
116.000	2.000\\
117.000	2.000\\
118.000	2.000\\
119.000	2.000\\
120.000	2.000\\
121.000	3.000\\
122.000	3.000\\
123.000	3.000\\
124.000	3.000\\
125.000	3.000\\
126.000	3.000\\
127.000	3.000\\
128.000	3.000\\
129.000	3.000\\
130.000	3.000\\
131.000	3.000\\
132.000	3.000\\
133.000	3.000\\
134.000	3.000\\
135.000	3.000\\
136.000	2.000\\
137.000	3.000\\
138.000	3.000\\
139.000	3.000\\
140.000	3.000\\
141.000	3.000\\
142.000	3.000\\
143.000	3.000\\
144.000	3.000\\
145.000	3.000\\
146.000	2.000\\
147.000	2.000\\
148.000	2.000\\
149.000	2.000\\
150.000	2.000\\
151.000	2.000\\
152.000	2.000\\
153.000	2.000\\
154.000	2.000\\
155.000	2.000\\
156.000	2.000\\
157.000	2.000\\
158.000	2.000\\
159.000	2.000\\
160.000	2.000\\
161.000	2.000\\
162.000	2.000\\
163.000	2.000\\
164.000	2.000\\
165.000	2.000\\
166.000	2.000\\
167.000	2.000\\
168.000	2.000\\
169.000	2.000\\
170.000	2.000\\
171.000	2.000\\
172.000	2.000\\
173.000	2.000\\
174.000	2.000\\
175.000	2.000\\
176.000	2.000\\
177.000	2.000\\
178.000	2.000\\
179.000	2.000\\
180.000	2.000\\
181.000	2.000\\
182.000	2.000\\
183.000	2.000\\
184.000	2.000\\
185.000	2.000\\
186.000	2.000\\
187.000	2.000\\
188.000	2.000\\
189.000	2.000\\
190.000	2.000\\
191.000	2.000\\
192.000	2.000\\
193.000	2.000\\
194.000	2.000\\
195.000	2.000\\
196.000	2.000\\
197.000	2.000\\
198.000	2.000\\
199.000	2.000\\
200.000	2.000\\
201.000	2.000\\
202.000	2.000\\
203.000	2.000\\
204.000	2.000\\
205.000	2.000\\
206.000	2.000\\
207.000	2.000\\
208.000	2.000\\
209.000	2.000\\
210.000	2.000\\
211.000	2.000\\
212.000	2.000\\
213.000	2.000\\
214.000	2.000\\
215.000	2.000\\
216.000	2.000\\
217.000	2.000\\
218.000	2.000\\
219.000	2.000\\
220.000	2.000\\
221.000	2.000\\
222.000	2.000\\
223.000	2.000\\
224.000	2.000\\
225.000	2.000\\
226.000	2.000\\
227.000	2.000\\
228.000	2.000\\
229.000	2.000\\
230.000	2.000\\
231.000	1.000\\
232.000	1.000\\
233.000	1.000\\
234.000	1.000\\
235.000	1.000\\
236.000	1.000\\
237.000	0.000\\
238.000	0.000\\
239.000	0.000\\
240.000	0.000\\
241.000	0.000\\
242.000	0.000\\
243.000	0.000\\
244.000	0.000\\
245.000	0.000\\
246.000	0.000\\
247.000	0.000\\
248.000	0.000\\
249.000	0.000\\
250.000	0.000\\
251.000	0.000\\
252.000	0.000\\
253.000	0.000\\
254.000	0.000\\
255.000	0.000\\
256.000	0.000\\
};
\addlegendentry{Displacement};

\end{axis}
\end{tikzpicture}%

%% file: images/velocity_estimation.pdf_tex
\begingroup%
  \makeatletter%
  \providecommand\color[2][]{%
    \errmessage{(Inkscape) Color is used for the text in Inkscape, but the package 'color.sty' is not loaded}%
    \renewcommand\color[2][]{}%
  }%
  \providecommand\transparent[1]{%
    \errmessage{(Inkscape) Transparency is used (non-zero) for the text in Inkscape, but the package 'transparent.sty' is not loaded}%
    \renewcommand\transparent[1]{}%
  }%
  \providecommand\rotatebox[2]{#2}%
  \ifx\svgwidth\undefined%
    \setlength{\unitlength}{284.43027344bp}%
    \ifx\svgscale\undefined%
      \relax%
    \else%
      \setlength{\unitlength}{\unitlength * \real{\svgscale}}%
    \fi%
  \else%
    \setlength{\unitlength}{\svgwidth}%
  \fi%
  \global\let\svgwidth\undefined%
  \global\let\svgscale\undefined%
  \makeatother%
  \begin{picture}(1,0.77769182)%
    \put(0,0){\includegraphics[width=\unitlength]{velocity_estimation.pdf}}%
    \put(0.55904029,0.56916633){\color[rgb]{0,0,0}\makebox(0,0)[lb]{\smash{FOV}}}%
    \put(0.43612591,0.41123429){\color[rgb]{0,0,1}\makebox(0,0)[lb]{\smash{Flow}}}%
    \put(0.48827863,0.47462358){\color[rgb]{0,0,0}\makebox(0,0)[lb]{\smash{$\alpha$}}}%
    \put(0.85121419,0.38266429){\color[rgb]{0,0,0}\makebox(0,0)[lb]{\smash{Height}}}%
    \put(0.31596177,0.55377166){\color[rgb]{0,0,0}\makebox(0,0)[lb]{\smash{f}}}%
    \put(-0.00149696,0.04554214){\color[rgb]{0,0,0}\makebox(0,0)[lb]{\smash{Ground Plane}}}%
    \put(0.71373292,0.75429801){\color[rgb]{0,0,0}\makebox(0,0)[lb]{\smash{\color{red}$V_{real}$}}}%
    \put(0.16477008,0.49018935){\color[rgb]{0,0,0}\makebox(0,0)[lt]{\begin{minipage}{0.18893952\unitlength}\raggedright Image Plane\end{minipage}}}%
    \put(0.36708928,0.11256895){\color[rgb]{0,0,0}\makebox(0,0)[lb]{\smash{\color{red}$V_{est}$   (=$V_{real}$)}}}%
  \end{picture}%
\endgroup%

%% file: images/dataset_cornerdetection.tikz
%
%
\begin{tikzpicture}

\begin{axis}[%
width=0.95092\figurewidth,
height=\figureheight,
at={(0\figurewidth,0\figureheight)},
scale only axis,
separate axis lines,
every outer x axis line/.append style={black},
every x tick label/.append style={font=\color{black}},
xmin=135.000,
xmax=160.000,
xlabel={time [s]},
every outer y axis line/.append style={black},
every y tick label/.append style={font=\color{black}},
ymin=0.000,
ymax=120.000,
ylabel={\# Corners},
title={Amount of Corners Detected}
]
\addplot [color=blue,solid,forget plot]
  table[row sep=crcr]{%
136.786	54.000\\
137.037	46.000\\
137.287	111.000\\
137.536	10.000\\
137.786	32.000\\
138.036	77.000\\
138.287	46.000\\
138.538	39.000\\
138.786	45.000\\
139.037	74.000\\
139.286	12.000\\
139.537	35.000\\
139.785	16.000\\
140.036	23.000\\
140.286	22.000\\
140.536	20.000\\
140.786	21.000\\
141.036	20.000\\
141.286	24.000\\
141.536	16.000\\
141.870	14.000\\
142.036	15.000\\
142.286	4.000\\
142.537	7.000\\
142.786	5.000\\
143.036	8.000\\
143.286	16.000\\
143.537	22.000\\
143.786	11.000\\
144.036	7.000\\
144.286	5.000\\
144.536	9.000\\
144.786	32.000\\
145.036	6.000\\
145.286	18.000\\
145.536	23.000\\
145.786	13.000\\
146.038	8.000\\
146.287	28.000\\
146.536	5.000\\
146.787	26.000\\
147.036	17.000\\
147.287	47.000\\
147.536	31.000\\
147.787	51.000\\
148.036	50.000\\
148.286	51.000\\
148.537	1.000\\
148.787	9.000\\
149.037	33.000\\
149.286	11.000\\
149.536	21.000\\
149.786	71.000\\
150.036	37.000\\
150.286	15.000\\
150.536	45.000\\
150.786	72.000\\
151.036	16.000\\
151.286	40.000\\
151.536	18.000\\
151.787	79.000\\
152.036	70.000\\
152.286	63.000\\
152.536	68.000\\
152.786	36.000\\
153.036	52.000\\
153.286	5.000\\
153.536	30.000\\
153.789	28.000\\
154.036	5.000\\
154.287	19.000\\
154.536	7.000\\
154.786	10.000\\
155.036	11.000\\
155.287	7.000\\
155.536	27.000\\
155.786	22.000\\
156.036	16.000\\
156.286	24.000\\
156.536	7.000\\
156.786	18.000\\
157.036	35.000\\
157.286	57.000\\
157.536	19.000\\
157.786	8.000\\
158.036	13.000\\
158.286	25.000\\
158.536	11.000\\
158.786	26.000\\
159.036	35.000\\
159.286	76.000\\
};
\end{axis}
\end{tikzpicture}%

%% file: images/flow_total.tikz
%
%
\begin{tikzpicture}

\begin{axis}[%
width=0.95092\figurewidth,
height=\figureheight,
at={(0\figurewidth,0\figureheight)},
scale only axis,
every outer x axis line/.append style={black},
every x tick label/.append style={font=\color{black}},
xmin=136.786,
xmax=159.286,
xlabel={Time [s]},
every outer y axis line/.append style={black},
every y tick label/.append style={font=\color{black}},
ymin=-0.600,
ymax=0.600,
ylabel={Velocity [m/s]},
axis x line*=bottom,
axis y line*=left,
legend style={at={(0.5,0.97)},anchor=north,legend cell align=left,align=left,fill=none,draw=none}
]
\addplot [color=blue,solid]
  table[row sep=crcr]{%
136.786	0.149\\
136.912	0.198\\
137.037	0.210\\
137.162	0.278\\
137.287	0.310\\
137.430	0.312\\
137.536	0.311\\
137.662	0.315\\
137.786	0.282\\
137.912	0.271\\
138.036	0.318\\
138.181	0.318\\
138.287	0.334\\
138.411	0.283\\
138.538	0.231\\
138.662	0.159\\
138.786	0.143\\
138.912	0.128\\
139.037	0.147\\
139.162	0.171\\
139.286	0.212\\
139.411	0.243\\
139.537	0.249\\
139.661	0.286\\
139.785	0.299\\
139.911	0.286\\
140.036	0.309\\
140.162	0.353\\
140.286	0.336\\
140.411	0.329\\
140.536	0.357\\
140.747	0.331\\
140.786	0.251\\
140.911	0.240\\
141.036	0.229\\
141.161	0.189\\
141.286	0.153\\
141.411	0.127\\
141.536	0.089\\
141.869	0.085\\
141.870	0.095\\
141.912	0.114\\
142.036	0.206\\
142.162	0.248\\
142.286	0.236\\
142.411	0.204\\
142.537	0.152\\
142.747	0.031\\
142.786	-0.001\\
142.911	-0.030\\
143.036	-0.022\\
143.162	-0.024\\
143.286	-0.006\\
143.411	-0.034\\
143.537	-0.043\\
143.661	-0.076\\
143.786	-0.046\\
143.911	-0.024\\
144.036	0.011\\
144.162	-0.010\\
144.286	-0.036\\
144.411	-0.111\\
144.536	-0.148\\
144.662	-0.199\\
144.786	-0.220\\
144.911	-0.160\\
145.036	-0.091\\
145.162	-0.069\\
145.286	-0.097\\
145.411	-0.064\\
145.536	-0.147\\
145.662	-0.213\\
145.786	-0.199\\
145.912	-0.109\\
146.038	-0.076\\
146.162	-0.029\\
146.287	-0.019\\
146.411	-0.074\\
146.536	-0.141\\
146.662	-0.188\\
146.787	-0.237\\
146.912	-0.255\\
147.036	-0.251\\
147.162	-0.243\\
147.287	-0.238\\
147.411	-0.187\\
147.536	-0.163\\
147.662	-0.203\\
147.787	-0.165\\
147.912	-0.163\\
148.036	-0.237\\
148.196	-0.283\\
148.286	-0.262\\
148.412	-0.340\\
148.537	-0.352\\
148.662	-0.307\\
148.787	-0.261\\
148.912	-0.211\\
149.037	-0.215\\
149.162	-0.251\\
149.286	-0.289\\
149.411	-0.327\\
149.536	-0.382\\
149.661	-0.352\\
149.786	-0.294\\
149.914	-0.195\\
150.036	-0.123\\
150.162	-0.104\\
150.286	-0.105\\
150.411	-0.099\\
150.536	-0.142\\
150.746	-0.153\\
150.786	-0.138\\
150.911	-0.165\\
151.036	-0.198\\
151.162	-0.237\\
151.286	-0.318\\
151.411	-0.380\\
151.536	-0.365\\
151.661	-0.355\\
151.787	-0.337\\
151.911	-0.296\\
152.036	-0.251\\
152.162	-0.265\\
152.286	-0.275\\
152.412	-0.288\\
152.536	-0.265\\
152.662	-0.283\\
152.786	-0.256\\
152.911	-0.250\\
153.036	-0.207\\
153.238	-0.153\\
153.286	-0.056\\
153.411	0.038\\
153.536	0.118\\
153.662	0.175\\
153.789	0.204\\
153.911	0.221\\
154.036	0.200\\
154.162	0.229\\
154.287	0.261\\
154.411	0.296\\
154.536	0.333\\
154.662	0.383\\
154.786	0.390\\
154.911	0.385\\
155.036	0.349\\
155.162	0.322\\
155.287	0.221\\
155.413	0.195\\
155.536	0.183\\
155.662	0.197\\
155.786	0.186\\
155.911	0.202\\
156.036	0.125\\
156.163	0.053\\
156.286	-0.030\\
156.412	-0.038\\
156.536	-0.024\\
156.745	0.048\\
156.786	0.084\\
156.911	0.124\\
157.036	0.181\\
157.162	0.208\\
157.286	0.200\\
157.428	0.246\\
157.536	0.265\\
157.662	0.165\\
157.786	0.149\\
157.912	0.136\\
158.036	0.091\\
158.161	0.009\\
158.286	-0.035\\
158.411	-0.127\\
158.536	-0.239\\
158.745	-0.211\\
158.786	-0.090\\
158.912	0.084\\
159.036	0.214\\
159.162	0.430\\
159.286	0.306\\
};
\addlegendentry{EdgeFlow};

\addplot [color=red,solid]
  table[row sep=crcr]{%
136.786	0.128\\
136.912	0.135\\
137.037	0.148\\
137.162	0.007\\
137.287	-0.008\\
137.430	0.002\\
137.536	-0.001\\
137.662	0.007\\
137.786	0.149\\
137.912	0.148\\
138.036	0.098\\
138.181	0.081\\
138.287	-0.114\\
138.411	-0.127\\
138.538	-0.129\\
138.662	-0.108\\
138.786	-0.101\\
138.912	0.074\\
139.037	0.074\\
139.162	0.078\\
139.286	0.100\\
139.411	0.117\\
139.537	0.108\\
139.661	0.073\\
139.785	0.088\\
139.911	0.096\\
140.036	0.087\\
140.162	0.113\\
140.286	0.155\\
140.411	0.152\\
140.536	0.093\\
140.747	0.082\\
140.786	0.040\\
140.911	0.055\\
141.036	0.033\\
141.161	0.083\\
141.286	0.083\\
141.411	0.094\\
141.536	0.022\\
141.869	0.027\\
141.870	0.013\\
141.912	0.016\\
142.036	-0.018\\
142.162	0.016\\
142.286	0.016\\
142.411	0.005\\
142.537	-0.011\\
142.747	0.015\\
142.786	-0.008\\
142.911	-0.030\\
143.036	-0.022\\
143.162	-0.032\\
143.286	-0.033\\
143.411	-0.031\\
143.537	-0.028\\
143.661	-0.057\\
143.786	-0.037\\
143.911	-0.011\\
144.036	0.033\\
144.162	0.039\\
144.286	0.066\\
144.411	0.032\\
144.536	-0.003\\
144.662	-0.052\\
144.786	-0.050\\
144.911	-0.098\\
145.036	-0.074\\
145.162	-0.074\\
145.286	-0.041\\
145.411	-0.047\\
145.536	-0.002\\
145.662	-0.019\\
145.786	-0.007\\
145.912	-0.010\\
146.038	0.002\\
146.162	-0.015\\
146.287	-0.029\\
146.411	-0.020\\
146.536	-0.042\\
146.662	-0.068\\
146.787	-0.077\\
146.912	-0.100\\
147.036	-0.072\\
147.162	-0.010\\
147.287	0.009\\
147.411	0.025\\
147.536	0.069\\
147.662	0.079\\
147.787	0.062\\
147.912	0.031\\
148.036	0.026\\
148.196	0.020\\
148.286	-0.057\\
148.412	-0.111\\
148.537	-0.105\\
148.662	-0.108\\
148.787	-0.104\\
148.912	-0.097\\
149.037	-0.034\\
149.162	-0.017\\
149.286	-0.067\\
149.411	-0.087\\
149.536	-0.091\\
149.661	-0.146\\
149.786	-0.154\\
149.914	-0.085\\
150.036	-0.070\\
150.162	-0.072\\
150.286	-0.072\\
150.411	-0.066\\
150.536	-0.092\\
150.746	-0.088\\
150.786	-0.080\\
150.911	-0.093\\
151.036	-0.119\\
151.162	-0.079\\
151.286	-0.068\\
151.411	-0.053\\
151.536	-0.027\\
151.661	-0.010\\
151.787	-0.048\\
151.911	-0.074\\
152.036	-0.106\\
152.162	-0.117\\
152.286	-0.112\\
152.412	-0.116\\
152.536	-0.102\\
152.662	-0.099\\
152.786	-0.116\\
152.911	-0.020\\
153.036	-0.006\\
153.238	0.015\\
153.286	0.071\\
153.411	0.152\\
153.536	0.095\\
153.662	0.132\\
153.789	0.138\\
153.911	0.140\\
154.036	0.105\\
154.162	0.098\\
154.287	0.107\\
154.411	0.146\\
154.536	0.112\\
154.662	0.125\\
154.786	0.146\\
154.911	0.140\\
155.036	0.106\\
155.162	0.132\\
155.287	0.099\\
155.413	0.085\\
155.536	0.076\\
155.662	0.100\\
155.786	0.110\\
155.911	0.141\\
156.036	0.116\\
156.163	0.087\\
156.286	0.037\\
156.412	0.034\\
156.536	0.018\\
156.745	0.039\\
156.786	0.051\\
156.911	0.065\\
157.036	0.135\\
157.162	0.142\\
157.286	0.140\\
157.428	0.168\\
157.536	0.186\\
157.662	0.072\\
157.786	0.053\\
157.912	0.042\\
158.036	0.019\\
158.161	-0.021\\
158.286	-0.011\\
158.411	-0.034\\
158.536	-0.048\\
158.745	-0.037\\
158.786	0.013\\
158.912	0.030\\
159.036	0.100\\
159.162	0.145\\
159.286	0.203\\
};
\addlegendentry{Lucas-Kanade};

\addplot [color=black!50!green,solid]
  table[row sep=crcr]{%
136.786	0.092\\
136.912	0.122\\
137.037	0.158\\
137.162	0.184\\
137.287	0.219\\
137.430	0.251\\
137.536	0.239\\
137.662	0.228\\
137.786	0.225\\
137.912	0.198\\
138.036	0.179\\
138.181	0.190\\
138.287	0.189\\
138.411	0.186\\
138.538	0.186\\
138.662	0.178\\
138.786	0.166\\
138.912	0.159\\
139.037	0.164\\
139.162	0.175\\
139.286	0.189\\
139.411	0.213\\
139.537	0.235\\
139.661	0.253\\
139.785	0.268\\
139.911	0.274\\
140.036	0.270\\
140.162	0.261\\
140.286	0.262\\
140.411	0.211\\
140.536	0.195\\
140.747	0.193\\
140.786	0.194\\
140.911	0.172\\
141.036	0.207\\
141.161	0.213\\
141.286	0.275\\
141.411	0.233\\
141.536	0.207\\
141.869	0.213\\
141.870	0.206\\
141.912	0.148\\
142.036	0.196\\
142.162	0.227\\
142.286	0.244\\
142.411	0.218\\
142.537	0.201\\
142.747	0.173\\
142.786	0.147\\
142.911	0.097\\
143.036	0.103\\
143.162	0.080\\
143.286	0.058\\
143.411	0.031\\
143.537	0.005\\
143.661	-0.033\\
143.786	-0.056\\
143.911	-0.074\\
144.036	-0.083\\
144.162	-0.097\\
144.286	-0.101\\
144.411	-0.114\\
144.536	-0.127\\
144.662	-0.137\\
144.786	-0.146\\
144.911	-0.147\\
145.036	-0.152\\
145.162	-0.143\\
145.286	-0.148\\
145.411	-0.148\\
145.536	-0.156\\
145.662	-0.158\\
145.786	-0.170\\
145.912	-0.168\\
146.038	-0.175\\
146.162	-0.187\\
146.287	-0.187\\
146.411	-0.207\\
146.536	-0.225\\
146.662	-0.233\\
146.787	-0.231\\
146.912	-0.241\\
147.036	-0.226\\
147.162	-0.222\\
147.287	-0.222\\
147.411	-0.206\\
147.536	-0.194\\
147.662	-0.189\\
147.787	-0.194\\
147.912	-0.183\\
148.036	-0.207\\
148.196	-0.233\\
148.286	-0.260\\
148.412	-0.274\\
148.537	-0.293\\
148.662	-0.292\\
148.787	-0.279\\
148.912	-0.258\\
149.037	-0.242\\
149.162	-0.232\\
149.286	-0.224\\
149.411	-0.227\\
149.536	-0.231\\
149.661	-0.225\\
149.786	-0.229\\
149.914	-0.243\\
150.036	-0.232\\
150.162	-0.232\\
150.286	-0.263\\
150.411	-0.225\\
150.536	-0.193\\
150.746	-0.188\\
150.786	-0.160\\
150.911	-0.116\\
151.036	-0.139\\
151.162	-0.143\\
151.286	-0.148\\
151.411	-0.168\\
151.536	-0.174\\
151.661	-0.174\\
151.787	-0.170\\
151.911	-0.162\\
152.036	-0.161\\
152.162	-0.174\\
152.286	-0.198\\
152.412	-0.220\\
152.536	-0.232\\
152.662	-0.233\\
152.786	-0.237\\
152.911	-0.193\\
153.036	-0.153\\
153.238	-0.120\\
153.286	-0.086\\
153.411	-0.033\\
153.536	-0.016\\
153.662	0.007\\
153.789	0.037\\
153.911	0.069\\
154.036	0.104\\
154.162	0.135\\
154.287	0.165\\
154.411	0.195\\
154.536	0.225\\
154.662	0.245\\
154.786	0.263\\
154.911	0.271\\
155.036	0.270\\
155.162	0.250\\
155.287	0.223\\
155.413	0.193\\
155.536	0.177\\
155.662	0.153\\
155.786	0.143\\
155.911	0.143\\
156.036	0.148\\
156.163	0.142\\
156.286	0.163\\
156.412	0.148\\
156.536	0.145\\
156.745	0.145\\
156.786	0.152\\
156.911	0.129\\
157.036	0.138\\
157.162	0.135\\
157.286	0.128\\
157.428	0.111\\
157.536	0.107\\
157.662	0.096\\
157.786	0.084\\
157.912	0.076\\
158.036	0.074\\
158.161	0.074\\
158.286	0.084\\
158.411	0.077\\
158.536	0.089\\
158.745	0.107\\
158.786	0.114\\
158.912	0.107\\
159.036	0.103\\
159.162	0.071\\
159.286	0.004\\
};
\addlegendentry{Optitrack};

\addplot [color=black,dashed,forget plot]
  table[row sep=crcr]{%
134.536	0.000\\
159.286	0.000\\
};
\end{axis}
\end{tikzpicture}%

%% file: images/controlscheme.pdf_tex
\begingroup%
  \makeatletter%
  \providecommand\color[2][]{%
    \errmessage{(Inkscape) Color is used for the text in Inkscape, but the package 'color.sty' is not loaded}%
    \renewcommand\color[2][]{}%
  }%
  \providecommand\transparent[1]{%
    \errmessage{(Inkscape) Transparency is used (non-zero) for the text in Inkscape, but the package 'transparent.sty' is not loaded}%
    \renewcommand\transparent[1]{}%
  }%
  \providecommand\rotatebox[2]{#2}%
  \ifx\svgwidth\undefined%
    \setlength{\unitlength}{188.61258269bp}%
    \ifx\svgscale\undefined%
      \relax%
    \else%
      \setlength{\unitlength}{\unitlength * \real{\svgscale}}%
    \fi%
  \else%
    \setlength{\unitlength}{\svgwidth}%
  \fi%
  \global\let\svgwidth\undefined%
  \global\let\svgscale\undefined%
  \makeatother%
  \begin{picture}(1,0.91646529)%
    \put(0,0){\includegraphics[width=\unitlength]{controlscheme.pdf}}%
    \put(0.56415634,0.69244514){\color[rgb]{0,0,0}\makebox(0,0)[lt]{\begin{minipage}{0.37849485\unitlength}\centering Guidance Controller\end{minipage}}}%
    \put(0.35236984,0.86868332){\color[rgb]{0,0,0}\makebox(0,0)[lb]{\smash{Desired Velocity}}}%
    \put(0.046628,0.73498988){\color[rgb]{0,0,0}\makebox(0,0)[lt]{\begin{minipage}{0.51564639\unitlength}\raggedright Measured Velocity\end{minipage}}}%
    \put(0.56661228,0.21975358){\color[rgb]{0,0,0}\makebox(0,0)[lt]{\begin{minipage}{0.37849485\unitlength}\centering Atitude and Altitude Controller\end{minipage}}}%
    \put(0.04309233,0.00051776){\color[rgb]{0,0,0}\makebox(0,0)[lb]{\smash{Motor Commands}}}%
    \put(0.31443826,0.36484034){\color[rgb]{0,0,0}\makebox(0,0)[lt]{\begin{minipage}{0.17255217\unitlength}\raggedright Angle \& Height\end{minipage}}}%
    \put(0.77008906,0.45010769){\color[rgb]{0,0,0}\makebox(0,0)[lt]{\begin{minipage}{0.22154928\unitlength}\raggedright Angle setpoint\end{minipage}}}%
  \end{picture}%
\endgroup%

%% file: images/ardrone_velocity_estimate_control_hor.tikz
%
%
\begin{tikzpicture}

\begin{axis}[%
width=0.95092\figurewidth,
height=\figureheight,
at={(0\figurewidth,0\figureheight)},
scale only axis,
unbounded coords=jump,
every outer x axis line/.append style={black},
every x tick label/.append style={font=\color{black}},
xmin=73.999,
xmax=253.600,
xlabel={time [s]},
every outer y axis line/.append style={black},
every y tick label/.append style={font=\color{black}},
ymin=-0.500,
ymax=0.500,
ylabel={velocity [m/s]},
axis x line*=bottom,
axis y line*=left,
legend style={at={(0.378224,0.822597)},anchor=south west,legend cell align=left,align=left,fill=none,draw=none}
]
\addplot [color=blue,solid]
  table[row sep=crcr]{%
1.016	-0.129\\
1.277	-0.090\\
2.821	-0.085\\
3.840	-0.076\\
4.607	-0.082\\
4.861	-0.062\\
5.628	-0.062\\
6.655	-0.057\\
7.673	-0.072\\
8.725	-0.052\\
9.491	-0.041\\
9.750	-0.047\\
10.509	-0.045\\
11.537	-0.030\\
12.803	-0.052\\
14.336	-0.058\\
14.594	-0.059\\
15.355	-0.044\\
16.423	-0.061\\
16.535	-0.065\\
16.762	-0.059\\
16.873	-0.060\\
17.207	-0.075\\
17.319	-0.062\\
18.108	-0.042\\
18.571	-0.052\\
18.683	-0.065\\
19.239	-0.066\\
19.458	-0.071\\
19.913	-0.054\\
20.023	-0.043\\
20.248	-0.027\\
20.359	-0.037\\
20.927	-0.047\\
21.037	-0.040\\
21.150	-0.012\\
21.376	-0.001\\
21.715	0.006\\
21.828	-0.013\\
21.939	-0.010\\
22.507	-0.007\\
22.619	-0.001\\
22.730	0.033\\
23.067	0.049\\
23.178	0.030\\
23.292	0.031\\
23.627	0.040\\
23.739	0.042\\
24.183	0.032\\
24.294	0.028\\
24.632	0.037\\
24.968	0.019\\
25.079	-0.007\\
25.531	-0.010\\
25.643	-0.011\\
26.541	-0.017\\
27.330	-0.016\\
27.441	-0.017\\
27.891	-0.025\\
28.119	-0.022\\
28.232	-0.025\\
28.343	-0.010\\
28.688	-0.010\\
28.797	0.001\\
28.909	0.010\\
29.246	-0.002\\
29.358	0.001\\
29.809	0.011\\
29.922	-0.000\\
30.256	-0.005\\
30.479	-0.005\\
30.821	0.002\\
30.932	-0.016\\
31.044	-0.030\\
31.382	-0.024\\
31.604	-0.045\\
31.830	-0.060\\
31.942	-0.066\\
32.610	-0.072\\
32.833	-0.079\\
33.054	-0.082\\
33.167	-0.066\\
34.295	-0.061\\
34.406	-0.057\\
34.517	-0.034\\
34.737	-0.024\\
35.073	-0.014\\
35.184	-0.005\\
35.636	-0.015\\
36.198	-0.006\\
36.538	-0.004\\
36.757	-0.000\\
37.088	-0.005\\
37.989	-0.009\\
38.101	-0.010\\
38.326	-0.005\\
38.438	-0.002\\
38.676	0.011\\
39.013	0.002\\
39.572	-0.014\\
39.907	-0.030\\
40.132	-0.038\\
40.580	-0.042\\
40.910	-0.030\\
41.020	-0.033\\
41.132	-0.046\\
41.357	-0.056\\
41.466	-0.061\\
41.581	-0.063\\
42.480	-0.064\\
42.592	-0.065\\
43.039	-0.064\\
43.154	-0.079\\
43.265	-0.082\\
43.596	-0.068\\
43.708	-0.062\\
43.820	-0.071\\
43.935	-0.067\\
44.344	-0.067\\
44.654	-0.058\\
44.859	-0.058\\
44.939	-0.066\\
45.471	-0.060\\
46.164	-0.069\\
46.835	-0.049\\
47.927	-0.023\\
48.869	-0.008\\
49.432	0.029\\
49.653	0.048\\
50.322	0.062\\
50.545	0.073\\
50.768	0.070\\
50.879	0.056\\
51.104	0.033\\
51.555	0.024\\
51.667	0.015\\
51.888	-0.013\\
52.004	-0.030\\
52.451	-0.048\\
52.562	-0.052\\
53.119	-0.059\\
53.455	-0.054\\
53.569	-0.058\\
53.681	-0.067\\
54.014	-0.076\\
54.123	-0.074\\
54.567	-0.071\\
54.678	-0.048\\
54.900	-0.037\\
55.014	-0.023\\
55.346	-0.017\\
55.456	0.002\\
55.679	-0.001\\
56.356	0.004\\
56.692	0.004\\
56.804	-0.005\\
57.362	-0.053\\
57.922	-0.075\\
58.034	-0.087\\
58.156	-0.094\\
58.587	-0.114\\
58.699	-0.115\\
59.051	-0.125\\
59.278	-0.138\\
59.389	-0.146\\
59.612	-0.131\\
59.723	-0.128\\
60.966	-0.144\\
61.193	-0.139\\
61.308	-0.120\\
61.988	-0.102\\
62.095	-0.081\\
62.541	-0.057\\
63.551	-0.033\\
63.775	-0.007\\
64.335	0.018\\
64.447	0.045\\
64.559	0.054\\
64.919	0.055\\
65.024	0.058\\
65.130	0.063\\
65.349	0.073\\
66.009	0.082\\
66.120	0.079\\
67.586	0.077\\
67.706	0.074\\
68.148	0.073\\
68.487	0.084\\
68.949	0.057\\
69.060	0.049\\
69.506	0.003\\
69.617	-0.041\\
70.173	-0.063\\
71.407	-0.098\\
72.086	-0.135\\
72.412	-0.155\\
72.519	-0.189\\
73.521	-0.172\\
73.633	-0.163\\
73.999	-0.120\\
74.082	-0.087\\
74.193	-0.054\\
74.526	-0.009\\
74.641	0.034\\
75.320	0.063\\
75.880	0.084\\
76.320	0.092\\
76.882	0.095\\
76.997	0.094\\
77.553	0.100\\
78.001	0.094\\
79.137	0.092\\
79.695	0.095\\
79.810	0.092\\
80.146	0.090\\
80.376	0.086\\
80.481	0.072\\
80.950	0.067\\
81.403	0.067\\
81.518	0.074\\
82.061	0.070\\
82.273	0.061\\
82.494	0.070\\
82.608	0.074\\
83.400	0.076\\
84.622	0.068\\
85.299	0.070\\
85.978	0.067\\
86.088	0.055\\
86.758	0.050\\
87.319	0.060\\
87.432	0.053\\
88.434	0.058\\
88.545	0.060\\
88.877	0.074\\
91.239	0.075\\
91.348	0.077\\
92.589	0.092\\
92.701	0.094\\
92.813	0.087\\
93.032	0.092\\
93.596	0.088\\
94.053	0.090\\
94.496	0.088\\
95.722	0.083\\
97.171	0.061\\
97.283	0.042\\
97.508	0.042\\
98.744	0.044\\
98.968	0.038\\
99.210	0.038\\
100.226	0.040\\
100.562	0.050\\
100.672	0.048\\
101.121	0.063\\
101.233	0.068\\
101.458	0.067\\
102.241	0.044\\
102.582	0.034\\
103.030	0.034\\
103.142	0.031\\
103.587	0.024\\
103.923	0.037\\
104.373	0.049\\
104.597	0.032\\
104.709	0.018\\
104.931	0.025\\
105.042	0.021\\
106.734	0.006\\
106.846	-0.007\\
107.301	-0.021\\
107.642	-0.046\\
107.756	-0.079\\
107.979	-0.087\\
108.321	-0.092\\
108.548	-0.093\\
109.341	-0.096\\
109.455	-0.096\\
110.125	-0.098\\
110.795	-0.100\\
110.904	-0.110\\
111.803	-0.111\\
111.913	-0.115\\
112.024	-0.107\\
112.916	-0.107\\
113.025	-0.121\\
113.359	-0.128\\
115.040	-0.133\\
115.152	-0.142\\
115.492	-0.141\\
115.824	-0.138\\
115.936	-0.132\\
116.278	-0.139\\
116.837	-0.146\\
116.947	-0.144\\
117.063	-0.146\\
117.291	-0.152\\
117.740	-0.152\\
118.746	-0.153\\
118.971	-0.157\\
119.083	-0.160\\
119.225	-0.162\\
119.321	-0.168\\
119.885	-0.167\\
119.999	-0.165\\
120.332	-0.163\\
120.667	-0.162\\
121.336	-0.163\\
122.006	-0.163\\
122.235	-0.170\\
122.343	-0.172\\
122.565	-0.182\\
122.677	-0.191\\
123.016	-0.206\\
123.137	-0.219\\
124.581	-0.222\\
124.694	-0.229\\
124.806	-0.231\\
124.918	-0.224\\
125.697	-0.224\\
125.817	-0.210\\
125.920	-0.197\\
126.263	-0.179\\
126.375	-0.156\\
127.156	-0.144\\
127.267	-0.129\\
127.489	-0.120\\
127.823	-0.114\\
127.935	-0.100\\
128.274	-0.087\\
128.384	-0.076\\
128.608	-0.070\\
128.942	-0.069\\
129.061	-0.066\\
129.409	-0.069\\
129.523	-0.059\\
129.748	-0.041\\
129.860	-0.023\\
130.304	-0.006\\
131.199	0.014\\
131.755	0.022\\
132.202	0.029\\
132.425	0.033\\
132.879	0.046\\
133.442	0.050\\
133.577	0.052\\
133.786	0.051\\
133.915	0.052\\
134.014	0.053\\
134.341	0.067\\
134.455	0.078\\
134.679	0.094\\
134.794	0.109\\
135.020	0.127\\
135.580	0.128\\
135.918	0.122\\
136.592	0.114\\
136.703	0.105\\
137.369	0.093\\
137.927	0.087\\
138.041	0.079\\
138.487	0.076\\
139.963	0.083\\
140.294	0.093\\
140.408	0.106\\
141.310	0.118\\
141.634	0.130\\
141.747	0.144\\
141.864	0.158\\
142.086	0.169\\
142.416	0.169\\
142.528	0.157\\
142.761	0.151\\
142.977	0.145\\
143.200	0.140\\
143.311	0.140\\
143.422	0.141\\
143.646	0.144\\
143.981	0.151\\
144.654	0.155\\
145.103	0.161\\
146.003	0.162\\
146.118	0.163\\
146.568	0.172\\
147.693	0.178\\
148.032	0.182\\
148.250	0.181\\
148.599	0.174\\
148.702	0.180\\
148.805	0.184\\
148.919	0.182\\
149.032	0.181\\
149.383	0.180\\
149.609	0.176\\
149.962	0.181\\
150.061	0.179\\
150.170	0.180\\
150.304	0.180\\
150.737	0.165\\
150.852	0.138\\
151.071	0.143\\
151.189	0.149\\
151.525	0.151\\
151.636	0.143\\
152.639	0.148\\
152.753	0.148\\
153.087	0.133\\
153.198	0.139\\
153.989	0.163\\
154.098	0.151\\
154.761	0.128\\
154.878	0.109\\
156.010	0.092\\
156.111	0.076\\
156.226	0.066\\
156.335	0.065\\
156.893	0.060\\
157.339	0.056\\
157.450	0.057\\
157.906	0.071\\
158.026	0.083\\
158.240	0.071\\
158.353	0.052\\
158.703	0.031\\
158.798	0.008\\
159.247	-0.020\\
159.364	-0.025\\
159.597	-0.034\\
159.708	-0.051\\
160.042	-0.066\\
161.387	-0.067\\
161.503	-0.072\\
161.941	-0.080\\
162.053	-0.090\\
162.279	-0.096\\
162.498	-0.121\\
162.612	-0.147\\
162.945	-0.170\\
163.054	-0.174\\
163.167	-0.167\\
163.390	-0.166\\
163.613	-0.160\\
164.066	-0.148\\
164.180	-0.138\\
165.280	-0.125\\
165.963	-0.108\\
166.308	-0.095\\
166.614	-0.095\\
167.195	-0.107\\
167.531	-0.108\\
168.320	-0.103\\
169.324	-0.106\\
170.002	-0.108\\
170.672	-0.109\\
170.899	-0.111\\
171.803	-0.113\\
171.915	-0.122\\
172.251	-0.118\\
172.367	-0.111\\
172.593	-0.109\\
172.711	-0.099\\
172.824	-0.088\\
172.928	-0.088\\
173.069	-0.092\\
173.152	-0.088\\
173.262	-0.101\\
174.047	-0.111\\
174.158	-0.111\\
175.061	-0.111\\
175.620	-0.114\\
176.071	-0.118\\
176.532	-0.120\\
176.966	-0.124\\
177.301	-0.136\\
177.413	-0.130\\
177.524	-0.121\\
177.860	-0.122\\
178.196	-0.129\\
178.311	-0.140\\
178.752	-0.145\\
178.865	-0.146\\
179.455	-0.142\\
179.563	-0.136\\
180.574	-0.129\\
180.676	-0.138\\
181.689	-0.151\\
182.135	-0.156\\
182.248	-0.155\\
182.472	-0.150\\
182.582	-0.142\\
183.371	-0.136\\
184.040	-0.143\\
184.374	-0.145\\
184.487	-0.138\\
184.819	-0.133\\
185.818	-0.129\\
186.155	-0.127\\
186.269	-0.134\\
186.718	-0.139\\
186.830	-0.160\\
187.164	-0.170\\
187.276	-0.169\\
187.388	-0.170\\
188.289	-0.160\\
188.400	-0.153\\
188.858	-0.138\\
189.080	-0.126\\
189.298	-0.120\\
189.646	-0.101\\
189.897	-0.083\\
189.989	-0.081\\
190.216	-0.079\\
190.322	-0.084\\
190.657	-0.099\\
190.766	-0.095\\
190.879	-0.091\\
191.218	-0.094\\
191.330	-0.089\\
192.117	-0.081\\
192.229	-0.067\\
192.344	-0.052\\
192.470	-0.036\\
192.574	-0.013\\
193.129	-0.005\\
193.244	0.010\\
193.354	0.028\\
193.464	0.044\\
194.249	0.058\\
194.919	0.064\\
195.814	0.080\\
195.928	0.098\\
196.043	0.115\\
196.266	0.125\\
197.496	0.125\\
198.275	0.124\\
198.835	0.114\\
198.946	0.126\\
199.058	0.138\\
199.506	0.117\\
199.630	0.116\\
199.965	0.121\\
200.185	0.092\\
200.631	0.093\\
201.093	0.107\\
201.193	0.128\\
201.415	0.128\\
201.755	0.132\\
202.091	0.153\\
202.214	0.152\\
202.318	0.145\\
202.657	0.170\\
202.892	0.171\\
203.780	0.165\\
203.892	0.165\\
204.021	0.163\\
204.112	0.163\\
204.227	0.164\\
204.339	0.163\\
204.450	0.171\\
204.787	0.178\\
204.899	0.181\\
205.010	0.185\\
205.580	0.190\\
205.686	0.192\\
205.796	0.195\\
206.128	0.199\\
206.572	0.207\\
207.130	0.203\\
207.241	0.199\\
207.352	0.203\\
207.467	0.195\\
208.139	0.190\\
208.365	0.187\\
208.594	0.174\\
209.031	0.166\\
209.720	0.154\\
209.832	0.151\\
210.170	0.143\\
210.291	0.124\\
210.393	0.112\\
210.505	0.088\\
210.958	0.066\\
211.955	0.047\\
212.068	0.048\\
212.180	0.051\\
212.515	0.046\\
212.627	0.041\\
213.296	0.039\\
213.527	0.032\\
214.084	0.029\\
214.417	0.029\\
214.532	0.028\\
215.092	-0.005\\
215.200	-0.038\\
215.421	-0.061\\
215.532	-0.071\\
215.982	-0.087\\
216.316	-0.092\\
216.877	-0.103\\
217.325	-0.114\\
217.548	-0.120\\
217.660	-0.113\\
218.330	-0.106\\
218.775	-0.106\\
219.136	-0.117\\
220.030	-0.126\\
220.140	-0.129\\
220.932	-0.127\\
221.064	-0.127\\
221.156	-0.133\\
221.845	-0.142\\
221.949	-0.154\\
222.161	-0.158\\
222.277	-0.155\\
222.390	-0.145\\
222.500	-0.146\\
222.953	-0.153\\
223.178	-0.158\\
223.290	-0.160\\
223.968	-0.165\\
224.304	-0.162\\
224.414	-0.163\\
224.636	-0.164\\
224.748	-0.173\\
224.859	-0.180\\
224.973	-0.169\\
225.299	-0.158\\
226.303	-0.149\\
226.538	-0.141\\
226.870	-0.144\\
226.981	-0.148\\
227.203	-0.151\\
227.318	-0.141\\
227.545	-0.137\\
227.766	-0.145\\
227.878	-0.163\\
228.214	-0.172\\
228.998	-0.187\\
229.111	-0.193\\
229.561	-0.189\\
230.141	-0.174\\
231.497	-0.163\\
231.722	-0.174\\
232.058	-0.179\\
232.173	-0.160\\
232.392	-0.146\\
233.072	-0.124\\
233.180	-0.103\\
234.068	-0.096\\
234.181	-0.100\\
234.296	-0.105\\
234.407	-0.082\\
234.629	-0.074\\
234.742	-0.080\\
235.188	-0.087\\
235.298	-0.085\\
235.629	-0.082\\
235.963	-0.073\\
236.075	-0.066\\
236.519	-0.065\\
236.630	-0.065\\
237.867	-0.064\\
237.983	-0.067\\
238.204	-0.069\\
238.422	-0.075\\
238.868	-0.080\\
239.094	-0.088\\
239.208	-0.099\\
239.323	-0.098\\
239.548	-0.085\\
239.658	-0.058\\
240.012	-0.024\\
240.235	0.014\\
240.457	0.044\\
240.568	0.074\\
240.793	0.102\\
240.910	0.130\\
241.245	0.141\\
241.357	0.141\\
241.472	0.140\\
241.804	0.140\\
242.168	0.142\\
242.256	0.137\\
242.925	0.125\\
243.036	0.108\\
243.266	0.098\\
244.281	0.102\\
244.623	0.105\\
245.091	0.103\\
245.184	0.097\\
245.629	0.086\\
245.851	0.083\\
245.962	0.084\\
246.419	0.099\\
246.530	0.101\\
246.641	0.103\\
246.752	0.104\\
247.657	0.106\\
247.766	0.105\\
247.990	0.098\\
248.103	0.092\\
248.552	0.093\\
248.681	0.087\\
249.114	0.088\\
249.224	0.080\\
249.562	0.076\\
249.675	0.075\\
250.136	0.077\\
250.249	0.088\\
250.481	0.101\\
251.024	0.109\\
251.135	0.115\\
251.245	0.126\\
251.357	0.139\\
251.472	0.147\\
251.584	0.152\\
251.924	0.150\\
252.151	0.142\\
252.489	0.134\\
253.600	0.120\\
254.609	0.105\\
254.721	0.112\\
255.169	0.103\\
255.277	0.105\\
255.618	0.106\\
256.289	0.111\\
256.402	0.112\\
256.630	0.123\\
257.087	0.142\\
257.528	0.163\\
258.646	0.144\\
259.209	0.154\\
260.003	0.154\\
260.114	0.156\\
262.133	0.141\\
262.245	0.127\\
262.583	0.103\\
262.806	0.084\\
263.481	0.072\\
263.605	0.090\\
264.269	0.075\\
264.935	0.064\\
265.384	0.048\\
265.497	0.066\\
266.160	0.092\\
266.271	0.114\\
266.495	0.108\\
266.952	0.088\\
267.055	0.059\\
267.503	0.042\\
267.612	0.046\\
268.288	-0.073\\
268.629	-0.109\\
268.965	-0.149\\
269.413	-0.186\\
269.875	-0.200\\
270.329	-0.186\\
270.550	-0.189\\
271.333	-0.190\\
271.669	-0.215\\
272.111	-0.110\\
272.340	-0.093\\
272.467	-0.071\\
272.567	-0.043\\
273.008	-0.027\\
273.568	-0.039\\
273.904	-0.028\\
274.017	-0.004\\
274.356	0.018\\
274.466	0.028\\
274.912	0.027\\
275.024	0.019\\
275.135	0.007\\
275.695	0.004\\
275.805	0.015\\
276.476	0.021\\
276.589	0.010\\
276.817	0.006\\
276.926	0.022\\
277.260	0.050\\
277.596	0.014\\
};
\addlegendentry{Velocity estimate};

\addplot [color=red,solid]
  table[row sep=crcr]{%
nan	-0.000\\
1.277	-0.000\\
2.821	-0.000\\
3.840	0.000\\
4.607	0.000\\
4.861	0.000\\
5.628	0.000\\
6.655	0.000\\
7.673	0.000\\
8.725	0.000\\
9.491	0.000\\
9.750	0.000\\
10.509	-0.000\\
11.537	-0.000\\
12.803	-0.000\\
14.336	0.000\\
14.594	0.000\\
15.355	0.000\\
16.423	0.000\\
16.535	0.000\\
16.762	0.000\\
16.873	-0.000\\
17.207	-0.000\\
17.319	-0.000\\
18.108	-0.000\\
18.571	-0.000\\
18.683	-0.000\\
19.239	0.000\\
19.458	0.000\\
19.913	0.000\\
20.023	0.000\\
20.248	0.000\\
20.359	0.000\\
20.927	0.000\\
21.037	0.000\\
21.150	-0.000\\
21.376	-0.000\\
21.715	-0.000\\
21.828	0.000\\
21.939	0.000\\
22.507	0.000\\
22.619	0.000\\
22.730	-0.000\\
23.067	-0.000\\
23.178	-0.000\\
23.292	-0.000\\
23.627	-0.000\\
23.739	-0.000\\
24.183	-0.000\\
24.294	0.000\\
24.632	0.000\\
24.968	0.000\\
25.079	0.000\\
25.531	0.000\\
25.643	0.000\\
26.541	0.000\\
27.330	-0.000\\
27.441	-0.000\\
27.891	-0.000\\
28.119	-0.000\\
28.232	-0.000\\
28.343	-0.000\\
28.688	-0.000\\
28.797	-0.000\\
28.909	-0.000\\
29.246	0.000\\
29.358	0.000\\
29.809	0.000\\
29.922	0.000\\
30.256	0.000\\
30.479	0.000\\
30.821	0.000\\
30.932	0.000\\
31.044	0.000\\
31.382	0.000\\
31.604	-0.000\\
31.830	-0.000\\
31.942	-0.000\\
32.610	-0.000\\
32.833	-0.000\\
33.054	0.000\\
33.167	0.000\\
34.295	0.000\\
34.406	0.000\\
34.517	0.000\\
34.737	0.000\\
35.073	0.000\\
35.184	0.000\\
35.636	-0.000\\
36.198	-0.000\\
36.538	-0.000\\
36.757	-0.000\\
37.088	-0.000\\
37.989	0.000\\
38.101	0.000\\
38.326	0.000\\
38.438	0.000\\
38.676	0.000\\
39.013	0.000\\
39.572	0.000\\
39.907	0.000\\
40.132	0.000\\
40.580	0.000\\
40.910	0.000\\
41.020	0.000\\
41.132	0.000\\
41.357	0.000\\
41.466	0.000\\
41.581	0.000\\
42.480	0.000\\
42.592	0.000\\
43.039	0.000\\
43.154	-0.000\\
43.265	-0.000\\
43.596	-0.000\\
43.708	-0.000\\
43.820	-0.000\\
43.935	-0.000\\
44.344	-0.000\\
44.654	-0.000\\
44.859	0.000\\
44.939	0.000\\
45.471	0.000\\
46.164	-0.000\\
46.835	-0.000\\
47.927	-0.000\\
48.869	-0.000\\
49.432	-0.000\\
49.653	-0.000\\
50.322	-0.000\\
50.545	-0.000\\
50.768	-0.000\\
50.879	-0.000\\
51.104	-0.000\\
51.555	-0.000\\
51.667	-0.000\\
51.888	-0.000\\
52.004	-0.000\\
52.451	-0.000\\
52.562	-0.000\\
53.119	0.000\\
53.455	0.000\\
53.569	0.000\\
53.681	0.000\\
54.014	0.000\\
54.123	0.000\\
54.567	-0.000\\
54.678	-0.000\\
54.900	-0.000\\
55.014	-0.000\\
55.346	-0.001\\
55.456	-0.001\\
55.679	-0.000\\
56.356	-0.001\\
56.692	-0.002\\
56.804	-0.006\\
57.362	-0.015\\
57.922	-0.032\\
58.034	-0.066\\
58.156	-0.122\\
58.587	-0.194\\
58.699	-0.268\\
59.051	-0.335\\
59.278	-0.386\\
59.389	-0.408\\
59.612	-0.398\\
59.723	-0.361\\
60.966	-0.310\\
61.193	-0.252\\
61.308	-0.190\\
61.988	-0.130\\
62.095	-0.078\\
62.541	-0.034\\
63.551	0.001\\
63.775	0.025\\
64.335	0.040\\
64.447	0.047\\
64.559	0.051\\
64.919	0.055\\
65.024	0.061\\
65.130	0.068\\
65.349	0.083\\
66.009	0.103\\
66.120	0.127\\
67.586	0.151\\
67.706	0.173\\
68.148	0.190\\
68.487	0.198\\
68.949	0.196\\
69.060	0.183\\
69.506	0.143\\
69.617	0.074\\
70.173	-0.008\\
71.407	-0.086\\
72.086	-0.147\\
72.412	-0.184\\
72.519	-0.183\\
73.521	-0.143\\
73.633	-0.085\\
73.999	-0.027\\
74.082	0.020\\
74.193	0.053\\
74.526	0.073\\
74.641	0.082\\
75.320	0.084\\
75.880	0.082\\
76.320	0.076\\
76.882	0.070\\
76.997	0.065\\
77.553	0.060\\
78.001	0.055\\
79.137	0.051\\
79.695	0.045\\
79.810	0.039\\
80.146	0.033\\
80.376	0.029\\
80.481	0.027\\
80.950	0.030\\
81.403	0.036\\
81.518	0.044\\
82.061	0.053\\
82.273	0.060\\
82.494	0.064\\
82.608	0.064\\
83.400	0.060\\
84.622	0.053\\
85.299	0.049\\
85.978	0.050\\
86.088	0.054\\
86.758	0.060\\
87.319	0.068\\
87.432	0.077\\
88.434	0.083\\
88.545	0.084\\
88.877	0.084\\
91.239	0.084\\
91.348	0.083\\
92.589	0.080\\
92.701	0.076\\
92.813	0.073\\
93.032	0.068\\
93.596	0.063\\
94.053	0.059\\
94.496	0.057\\
95.722	0.056\\
97.171	0.056\\
97.283	0.055\\
97.508	0.053\\
98.744	0.052\\
98.968	0.050\\
99.210	0.048\\
100.226	0.045\\
100.562	0.043\\
100.672	0.040\\
101.121	0.037\\
101.233	0.033\\
101.458	0.030\\
102.241	0.027\\
102.582	0.026\\
103.030	0.028\\
103.142	0.032\\
103.587	0.038\\
103.923	0.042\\
104.373	0.045\\
104.597	0.044\\
104.709	0.033\\
104.931	0.014\\
105.042	-0.010\\
106.734	-0.034\\
106.846	-0.056\\
107.301	-0.075\\
107.642	-0.088\\
107.756	-0.096\\
107.979	-0.100\\
108.321	-0.103\\
108.548	-0.103\\
109.341	-0.104\\
109.455	-0.104\\
110.125	-0.104\\
110.795	-0.105\\
110.904	-0.108\\
111.803	-0.113\\
111.913	-0.119\\
112.024	-0.124\\
112.916	-0.128\\
113.025	-0.134\\
113.359	-0.138\\
115.040	-0.141\\
115.152	-0.142\\
115.492	-0.142\\
115.824	-0.140\\
115.936	-0.136\\
116.278	-0.131\\
116.837	-0.128\\
116.947	-0.127\\
117.063	-0.128\\
117.291	-0.130\\
117.740	-0.133\\
118.746	-0.136\\
118.971	-0.139\\
119.083	-0.141\\
119.225	-0.143\\
119.321	-0.145\\
119.885	-0.147\\
119.999	-0.150\\
120.332	-0.157\\
120.667	-0.166\\
121.336	-0.177\\
122.006	-0.190\\
122.235	-0.204\\
122.343	-0.219\\
122.565	-0.232\\
122.677	-0.244\\
123.016	-0.257\\
123.137	-0.268\\
124.581	-0.275\\
124.694	-0.277\\
124.806	-0.274\\
124.918	-0.265\\
125.697	-0.250\\
125.817	-0.229\\
125.920	-0.203\\
126.263	-0.177\\
126.375	-0.150\\
127.156	-0.124\\
127.267	-0.100\\
127.489	-0.078\\
127.823	-0.061\\
127.935	-0.048\\
128.274	-0.041\\
128.384	-0.040\\
128.608	-0.043\\
128.942	-0.049\\
129.061	-0.058\\
129.409	-0.065\\
129.523	-0.067\\
129.748	-0.059\\
129.860	-0.042\\
130.304	-0.020\\
131.199	0.005\\
131.755	0.026\\
132.202	0.042\\
132.425	0.051\\
132.879	0.055\\
133.442	0.057\\
133.577	0.060\\
133.786	0.066\\
133.915	0.074\\
134.014	0.082\\
134.341	0.087\\
134.455	0.090\\
134.679	0.090\\
134.794	0.088\\
135.020	0.085\\
135.580	0.080\\
135.918	0.074\\
136.592	0.069\\
136.703	0.064\\
137.369	0.060\\
137.927	0.059\\
138.041	0.062\\
138.487	0.068\\
139.963	0.078\\
140.294	0.088\\
140.408	0.097\\
141.310	0.105\\
141.634	0.110\\
141.747	0.112\\
141.864	0.112\\
142.086	0.110\\
142.416	0.107\\
142.528	0.105\\
142.761	0.103\\
142.977	0.104\\
143.200	0.106\\
143.311	0.111\\
143.422	0.119\\
143.646	0.128\\
143.981	0.140\\
144.654	0.152\\
145.103	0.163\\
146.003	0.170\\
146.118	0.176\\
146.568	0.180\\
147.693	0.183\\
148.032	0.185\\
148.250	0.188\\
148.599	0.193\\
148.702	0.197\\
148.805	0.201\\
148.919	0.201\\
149.032	0.200\\
149.383	0.197\\
149.609	0.194\\
149.962	0.191\\
150.061	0.189\\
150.170	0.187\\
150.304	0.187\\
150.737	0.188\\
150.852	0.191\\
151.071	0.198\\
151.189	0.210\\
151.525	0.226\\
151.636	0.241\\
152.639	0.252\\
152.753	0.254\\
153.087	0.244\\
153.198	0.223\\
153.989	0.192\\
154.098	0.158\\
154.761	0.124\\
154.878	0.097\\
156.010	0.077\\
156.111	0.065\\
156.226	0.058\\
156.335	0.055\\
156.893	0.054\\
157.339	0.052\\
157.450	0.049\\
157.906	0.043\\
158.026	0.035\\
158.240	0.023\\
158.353	0.009\\
158.703	-0.008\\
158.798	-0.024\\
159.247	-0.037\\
159.364	-0.046\\
159.597	-0.049\\
159.708	-0.048\\
160.042	-0.047\\
161.387	-0.049\\
161.503	-0.055\\
161.941	-0.067\\
162.053	-0.081\\
162.279	-0.095\\
162.498	-0.105\\
162.612	-0.111\\
162.945	-0.112\\
163.054	-0.108\\
163.167	-0.102\\
163.390	-0.097\\
163.613	-0.094\\
164.066	-0.092\\
164.180	-0.092\\
165.280	-0.093\\
165.963	-0.093\\
166.308	-0.091\\
166.614	-0.088\\
167.195	-0.084\\
167.531	-0.081\\
168.320	-0.080\\
169.324	-0.081\\
170.002	-0.083\\
170.672	-0.086\\
170.899	-0.088\\
171.803	-0.089\\
171.915	-0.089\\
172.251	-0.089\\
172.367	-0.089\\
172.593	-0.089\\
172.711	-0.088\\
172.824	-0.087\\
172.928	-0.088\\
173.069	-0.090\\
173.152	-0.093\\
173.262	-0.097\\
174.047	-0.102\\
174.158	-0.106\\
175.061	-0.109\\
175.620	-0.110\\
176.071	-0.111\\
176.532	-0.112\\
176.966	-0.113\\
177.301	-0.114\\
177.413	-0.115\\
177.524	-0.114\\
177.860	-0.113\\
178.196	-0.111\\
178.311	-0.112\\
178.752	-0.116\\
178.865	-0.121\\
179.455	-0.129\\
179.563	-0.140\\
180.574	-0.150\\
180.676	-0.159\\
181.689	-0.166\\
182.135	-0.170\\
182.248	-0.171\\
182.472	-0.165\\
182.582	-0.157\\
183.371	-0.150\\
184.040	-0.146\\
184.374	-0.143\\
184.487	-0.144\\
184.819	-0.149\\
185.818	-0.154\\
186.155	-0.159\\
186.269	-0.164\\
186.718	-0.167\\
186.830	-0.168\\
187.164	-0.166\\
187.276	-0.159\\
187.388	-0.148\\
188.289	-0.132\\
188.400	-0.115\\
188.858	-0.100\\
189.080	-0.090\\
189.298	-0.084\\
189.646	-0.083\\
189.897	-0.085\\
189.989	-0.088\\
190.216	-0.090\\
190.322	-0.091\\
190.657	-0.089\\
190.766	-0.084\\
190.879	-0.075\\
191.218	-0.063\\
191.330	-0.051\\
192.117	-0.040\\
192.229	-0.030\\
192.344	-0.023\\
192.470	-0.019\\
192.574	-0.016\\
193.129	-0.012\\
193.244	-0.003\\
193.354	0.011\\
193.464	0.029\\
194.249	0.049\\
194.919	0.070\\
195.814	0.091\\
195.928	0.110\\
196.043	0.125\\
196.266	0.136\\
197.496	0.143\\
198.275	0.149\\
198.835	0.154\\
198.946	0.158\\
199.058	0.162\\
199.506	0.169\\
199.630	0.179\\
199.965	0.189\\
200.185	0.199\\
200.631	0.207\\
201.093	0.213\\
201.193	0.216\\
201.415	0.216\\
201.755	0.213\\
202.091	0.209\\
202.214	0.203\\
202.318	0.195\\
202.657	0.185\\
202.892	0.176\\
203.780	0.167\\
203.892	0.160\\
204.021	0.155\\
204.112	0.153\\
204.227	0.154\\
204.339	0.155\\
204.450	0.155\\
204.787	0.156\\
204.899	0.158\\
205.010	0.161\\
205.580	0.165\\
205.686	0.171\\
205.796	0.177\\
206.128	0.182\\
206.572	0.185\\
207.130	0.187\\
207.241	0.184\\
207.352	0.178\\
207.467	0.171\\
208.139	0.165\\
208.365	0.160\\
208.594	0.154\\
209.031	0.147\\
209.720	0.139\\
209.832	0.129\\
210.170	0.116\\
210.291	0.100\\
210.393	0.083\\
210.505	0.068\\
210.958	0.057\\
211.955	0.050\\
212.068	0.046\\
212.180	0.046\\
212.515	0.049\\
212.627	0.049\\
213.296	0.043\\
213.527	0.029\\
214.084	0.010\\
214.417	-0.013\\
214.532	-0.036\\
215.092	-0.057\\
215.200	-0.072\\
215.421	-0.083\\
215.532	-0.089\\
215.982	-0.095\\
216.316	-0.100\\
216.877	-0.103\\
217.325	-0.103\\
217.548	-0.101\\
217.660	-0.099\\
218.330	-0.098\\
218.775	-0.098\\
219.136	-0.101\\
220.030	-0.108\\
220.140	-0.117\\
220.932	-0.126\\
221.064	-0.134\\
221.156	-0.140\\
221.845	-0.142\\
221.949	-0.141\\
222.161	-0.139\\
222.277	-0.138\\
222.390	-0.138\\
222.500	-0.140\\
222.953	-0.145\\
223.178	-0.150\\
223.290	-0.156\\
223.968	-0.163\\
224.304	-0.169\\
224.414	-0.173\\
224.636	-0.177\\
224.748	-0.179\\
224.859	-0.181\\
224.973	-0.182\\
225.299	-0.181\\
226.303	-0.178\\
226.538	-0.177\\
226.870	-0.176\\
226.981	-0.178\\
227.203	-0.179\\
227.318	-0.180\\
227.545	-0.181\\
227.766	-0.181\\
227.878	-0.182\\
228.214	-0.183\\
228.998	-0.185\\
229.111	-0.189\\
229.561	-0.196\\
230.141	-0.203\\
231.497	-0.208\\
231.722	-0.211\\
232.058	-0.209\\
232.173	-0.200\\
232.392	-0.183\\
233.072	-0.161\\
233.180	-0.136\\
234.068	-0.113\\
234.181	-0.092\\
234.296	-0.077\\
234.407	-0.070\\
234.629	-0.068\\
234.742	-0.067\\
235.188	-0.066\\
235.298	-0.064\\
235.629	-0.061\\
235.963	-0.057\\
236.075	-0.053\\
236.519	-0.052\\
236.630	-0.054\\
237.867	-0.057\\
237.983	-0.060\\
238.204	-0.062\\
238.422	-0.062\\
238.868	-0.060\\
239.094	-0.055\\
239.208	-0.049\\
239.323	-0.037\\
239.548	-0.019\\
239.658	0.005\\
240.012	0.034\\
240.235	0.063\\
240.457	0.091\\
240.568	0.113\\
240.793	0.126\\
240.910	0.132\\
241.245	0.133\\
241.357	0.131\\
241.472	0.129\\
241.804	0.128\\
242.168	0.132\\
242.256	0.136\\
242.925	0.137\\
243.036	0.135\\
243.266	0.131\\
244.281	0.127\\
244.623	0.120\\
245.091	0.112\\
245.184	0.107\\
245.629	0.105\\
245.851	0.103\\
245.962	0.101\\
246.419	0.099\\
246.530	0.096\\
246.641	0.092\\
246.752	0.088\\
247.657	0.083\\
247.766	0.077\\
247.990	0.072\\
248.103	0.069\\
248.552	0.068\\
248.681	0.069\\
249.114	0.071\\
249.224	0.073\\
249.562	0.076\\
249.675	0.081\\
250.136	0.088\\
250.249	0.096\\
250.481	0.105\\
251.024	0.113\\
251.135	0.119\\
251.245	0.123\\
251.357	0.123\\
251.472	0.119\\
251.584	0.114\\
251.924	0.110\\
252.151	0.109\\
252.489	0.110\\
253.600	0.113\\
254.609	0.118\\
254.721	0.125\\
255.169	0.132\\
255.277	0.136\\
255.618	0.137\\
256.289	0.137\\
256.402	0.136\\
256.630	0.136\\
257.087	0.139\\
257.528	0.145\\
258.646	0.155\\
259.209	0.173\\
260.003	0.196\\
260.114	0.223\\
262.133	0.250\\
262.245	0.278\\
262.583	0.307\\
262.806	0.331\\
263.481	0.346\\
263.605	0.353\\
264.269	0.357\\
264.935	0.360\\
265.384	0.362\\
265.497	0.365\\
266.160	0.372\\
266.271	0.381\\
266.495	0.383\\
266.952	0.350\\
267.055	0.239\\
267.503	0.064\\
267.612	-0.127\\
268.288	-0.303\\
268.629	-0.477\\
268.965	-0.685\\
269.413	-0.893\\
269.875	-1.045\\
270.329	-1.153\\
270.550	-1.234\\
271.333	-1.270\\
271.669	-1.216\\
272.111	-1.046\\
272.340	-0.821\\
272.467	-0.609\\
272.567	-0.425\\
273.008	-0.264\\
273.568	-0.130\\
273.904	-0.040\\
274.017	-0.005\\
274.356	0.000\\
274.466	-0.000\\
274.912	-0.001\\
275.024	-0.001\\
275.135	-0.001\\
275.695	-0.001\\
275.805	-0.001\\
276.476	-0.001\\
276.589	-0.001\\
276.817	-0.001\\
276.926	-0.001\\
277.260	-0.001\\
nan	-0.001\\
};
\addlegendentry{Opti-track};

\addplot [color=black!30!green,solid]
  table[row sep=crcr]{%
1.016	0.000\\
1.277	0.000\\
2.821	0.000\\
3.840	0.000\\
4.607	0.000\\
4.861	0.000\\
5.628	0.000\\
6.655	0.000\\
7.673	0.000\\
8.725	0.000\\
9.491	0.000\\
9.750	0.000\\
10.509	0.000\\
11.537	0.000\\
12.803	0.000\\
14.336	0.000\\
14.594	0.000\\
15.355	0.000\\
16.423	0.000\\
16.535	0.000\\
16.762	0.000\\
16.873	0.000\\
17.207	0.000\\
17.319	0.000\\
18.108	0.000\\
18.571	0.000\\
18.683	0.000\\
19.239	0.000\\
19.458	0.000\\
19.913	0.000\\
20.023	0.000\\
20.248	0.000\\
20.359	0.000\\
20.927	0.000\\
21.037	0.000\\
21.150	0.000\\
21.376	0.000\\
21.715	0.000\\
21.828	0.000\\
21.939	0.000\\
22.507	0.000\\
22.619	0.000\\
22.730	0.000\\
23.067	0.000\\
23.178	0.000\\
23.292	0.000\\
23.627	0.000\\
23.739	0.000\\
24.183	0.000\\
24.294	0.000\\
24.632	0.000\\
24.968	0.000\\
25.079	0.000\\
25.531	0.000\\
25.643	0.000\\
26.541	0.000\\
27.330	0.000\\
27.441	0.000\\
27.891	0.000\\
28.119	0.000\\
28.232	0.000\\
28.343	0.000\\
28.688	0.000\\
28.797	0.000\\
28.909	0.000\\
29.246	0.000\\
29.358	0.000\\
29.809	0.000\\
29.922	0.000\\
30.256	0.000\\
30.479	0.000\\
30.821	0.000\\
30.932	0.000\\
31.044	0.000\\
31.382	0.000\\
31.604	0.000\\
31.830	0.000\\
31.942	0.000\\
32.610	0.000\\
32.833	0.000\\
33.054	0.000\\
33.167	0.000\\
34.295	0.000\\
34.406	0.000\\
34.517	0.000\\
34.737	0.000\\
35.073	0.000\\
35.184	0.000\\
35.636	0.000\\
36.198	0.000\\
36.538	0.000\\
36.757	0.000\\
37.088	0.000\\
37.989	0.000\\
38.101	0.000\\
38.326	0.000\\
38.438	0.000\\
38.676	0.000\\
39.013	0.000\\
39.572	0.000\\
39.907	0.000\\
40.132	0.000\\
40.580	0.000\\
40.910	0.000\\
41.020	0.000\\
41.132	0.000\\
41.357	0.000\\
41.466	0.000\\
41.581	0.000\\
42.480	0.000\\
42.592	0.000\\
43.039	0.000\\
43.154	0.000\\
43.265	0.000\\
43.596	0.000\\
43.708	0.000\\
43.820	0.000\\
43.935	0.000\\
44.344	0.000\\
44.654	0.000\\
44.859	0.000\\
44.939	0.000\\
45.471	0.000\\
46.164	0.000\\
46.835	0.000\\
47.927	0.000\\
48.869	0.000\\
49.432	0.000\\
49.653	0.000\\
50.322	0.000\\
50.545	0.000\\
50.768	0.000\\
50.879	0.000\\
51.104	0.000\\
51.555	0.000\\
51.667	0.000\\
51.888	0.000\\
52.004	0.000\\
52.451	0.000\\
52.562	0.000\\
53.119	0.000\\
53.455	0.000\\
53.569	0.000\\
53.681	0.000\\
54.014	0.000\\
54.123	0.000\\
54.567	0.000\\
54.678	0.000\\
54.900	0.000\\
55.014	0.000\\
55.346	0.000\\
55.456	0.000\\
55.679	0.000\\
56.356	0.000\\
56.692	0.000\\
56.804	0.000\\
57.362	0.000\\
57.922	0.000\\
58.034	0.000\\
58.156	0.000\\
58.587	0.000\\
58.699	0.000\\
59.051	0.000\\
59.278	0.000\\
59.389	0.000\\
59.612	0.000\\
59.723	0.000\\
60.966	0.000\\
61.193	0.000\\
61.308	0.000\\
61.988	0.000\\
62.095	0.000\\
62.541	0.000\\
63.551	0.000\\
63.775	0.000\\
64.335	0.000\\
64.447	0.000\\
64.559	0.000\\
64.919	0.000\\
65.024	0.000\\
65.130	0.000\\
65.349	0.000\\
66.009	0.000\\
66.120	0.000\\
67.586	0.000\\
67.706	0.000\\
68.148	0.000\\
68.487	0.000\\
68.949	0.000\\
69.060	0.000\\
69.506	0.000\\
69.617	0.000\\
70.173	0.000\\
71.407	0.000\\
72.086	0.000\\
72.412	0.000\\
72.519	0.000\\
73.521	0.000\\
73.633	0.000\\
73.999	0.000\\
74.082	0.000\\
74.193	0.000\\
74.526	0.000\\
74.641	0.000\\
75.320	0.000\\
75.880	0.000\\
76.320	0.000\\
76.882	0.000\\
76.997	0.000\\
77.553	0.000\\
78.001	0.000\\
79.137	0.000\\
79.695	0.000\\
79.810	0.000\\
80.146	0.000\\
80.376	0.000\\
80.481	0.000\\
80.950	0.000\\
81.403	0.000\\
81.518	0.000\\
82.061	0.000\\
82.273	0.000\\
82.494	0.000\\
82.608	0.000\\
83.400	0.000\\
84.622	0.000\\
85.299	0.000\\
85.978	0.000\\
86.088	0.000\\
86.758	0.000\\
87.319	0.000\\
87.432	0.000\\
88.434	0.000\\
88.545	0.000\\
88.877	0.000\\
91.239	0.000\\
91.348	0.000\\
92.589	0.000\\
92.701	0.000\\
92.813	0.000\\
93.032	0.000\\
93.596	0.000\\
94.053	0.000\\
94.496	0.000\\
95.722	0.000\\
97.171	0.000\\
97.283	0.000\\
97.508	0.000\\
98.744	0.000\\
98.968	0.000\\
99.210	0.000\\
100.226	0.000\\
100.562	0.000\\
100.672	0.000\\
101.121	0.000\\
101.233	0.000\\
101.458	0.000\\
102.241	0.000\\
102.582	0.000\\
103.030	0.000\\
103.142	0.000\\
103.587	0.000\\
103.923	-0.057\\
104.373	-0.135\\
104.597	-0.174\\
104.709	-0.194\\
104.931	-0.196\\
105.042	-0.196\\
106.734	-0.196\\
106.846	-0.196\\
107.301	-0.196\\
107.642	-0.196\\
107.756	-0.196\\
107.979	-0.196\\
108.321	-0.196\\
108.548	-0.196\\
109.341	-0.196\\
109.455	-0.196\\
110.125	-0.196\\
110.795	-0.196\\
110.904	-0.196\\
111.803	-0.196\\
111.913	-0.196\\
112.024	-0.196\\
112.916	-0.196\\
113.025	-0.196\\
113.359	-0.196\\
115.040	-0.196\\
115.152	-0.196\\
115.492	-0.196\\
115.824	-0.196\\
115.936	-0.196\\
116.278	-0.196\\
116.837	-0.196\\
116.947	-0.196\\
117.063	-0.196\\
117.291	-0.196\\
117.740	-0.196\\
118.746	-0.196\\
118.971	-0.196\\
119.083	-0.196\\
119.225	-0.196\\
119.321	-0.196\\
119.885	-0.196\\
119.999	-0.196\\
120.332	-0.196\\
120.667	-0.196\\
121.336	-0.196\\
122.006	-0.196\\
122.235	-0.196\\
122.343	-0.196\\
122.565	-0.196\\
122.677	-0.196\\
123.016	-0.196\\
123.137	-0.196\\
124.581	-0.080\\
124.694	-0.060\\
124.806	-0.041\\
124.918	-0.021\\
125.697	0.000\\
125.817	0.000\\
125.920	0.000\\
126.263	0.000\\
126.375	0.000\\
127.156	0.000\\
127.267	0.000\\
127.489	0.000\\
127.823	0.000\\
127.935	0.000\\
128.274	0.000\\
128.384	0.000\\
128.608	0.000\\
128.942	0.000\\
129.061	0.000\\
129.409	0.000\\
129.523	0.000\\
129.748	0.037\\
129.860	0.058\\
130.304	0.143\\
131.199	0.215\\
131.755	0.215\\
132.202	0.215\\
132.425	0.215\\
132.879	0.215\\
133.442	0.215\\
133.577	0.215\\
133.786	0.215\\
133.915	0.215\\
134.014	0.215\\
134.341	0.215\\
134.455	0.215\\
134.679	0.215\\
134.794	0.215\\
135.020	0.215\\
135.580	0.215\\
135.918	0.215\\
136.592	0.215\\
136.703	0.215\\
137.369	0.215\\
137.927	0.215\\
138.041	0.215\\
138.487	0.215\\
139.963	0.215\\
140.294	0.215\\
140.408	0.215\\
141.310	0.215\\
141.634	0.215\\
141.747	0.215\\
141.864	0.215\\
142.086	0.215\\
142.416	0.215\\
142.528	0.215\\
142.761	0.215\\
142.977	0.215\\
143.200	0.215\\
143.311	0.215\\
143.422	0.215\\
143.646	0.215\\
143.981	0.215\\
144.654	0.215\\
145.103	0.215\\
146.003	0.215\\
146.118	0.215\\
146.568	0.215\\
147.693	0.215\\
148.032	0.215\\
148.250	0.215\\
148.599	0.215\\
148.702	0.215\\
148.805	0.215\\
148.919	0.215\\
149.032	0.215\\
149.383	0.215\\
149.609	0.215\\
149.962	0.215\\
150.061	0.215\\
150.170	0.215\\
150.304	0.215\\
150.737	0.215\\
150.852	0.215\\
151.071	0.215\\
151.189	0.215\\
151.525	0.215\\
151.636	0.215\\
152.639	0.116\\
152.753	0.095\\
153.087	0.031\\
153.198	0.009\\
153.989	0.000\\
154.098	0.000\\
154.761	0.000\\
154.878	0.000\\
156.010	0.000\\
156.111	0.000\\
156.226	0.000\\
156.335	0.000\\
156.893	0.000\\
157.339	0.000\\
157.450	0.000\\
157.906	-0.024\\
158.026	-0.042\\
158.240	-0.075\\
158.353	-0.093\\
158.703	-0.147\\
158.798	-0.161\\
159.247	-0.173\\
159.364	-0.173\\
159.597	-0.173\\
159.708	-0.173\\
160.042	-0.173\\
161.387	-0.173\\
161.503	-0.173\\
161.941	-0.173\\
162.053	-0.173\\
162.279	-0.173\\
162.498	-0.173\\
162.612	-0.173\\
162.945	-0.173\\
163.054	-0.173\\
163.167	-0.173\\
163.390	-0.173\\
163.613	-0.173\\
164.066	-0.173\\
164.180	-0.173\\
165.280	-0.173\\
165.963	-0.173\\
166.308	-0.173\\
166.614	-0.173\\
167.195	-0.173\\
167.531	-0.173\\
168.320	-0.173\\
169.324	-0.173\\
170.002	-0.173\\
170.672	-0.173\\
170.899	-0.173\\
171.803	-0.173\\
171.915	-0.173\\
172.251	-0.173\\
172.367	-0.173\\
172.593	-0.173\\
172.711	-0.173\\
172.824	-0.173\\
172.928	-0.173\\
173.069	-0.173\\
173.152	-0.173\\
173.262	-0.173\\
174.047	-0.173\\
174.158	-0.173\\
175.061	-0.173\\
175.620	-0.173\\
176.071	-0.173\\
176.532	-0.173\\
176.966	-0.173\\
177.301	-0.173\\
177.413	-0.173\\
177.524	-0.173\\
177.860	-0.173\\
178.196	-0.173\\
178.311	-0.173\\
178.752	-0.173\\
178.865	-0.173\\
179.455	-0.173\\
179.563	-0.173\\
180.574	-0.173\\
180.676	-0.173\\
181.689	-0.173\\
182.135	-0.173\\
182.248	-0.173\\
182.472	-0.173\\
182.582	-0.173\\
183.371	-0.173\\
184.040	-0.173\\
184.374	-0.173\\
184.487	-0.173\\
184.819	-0.173\\
185.818	-0.173\\
186.155	-0.173\\
186.269	-0.173\\
186.718	-0.173\\
186.830	-0.173\\
187.164	-0.157\\
187.276	-0.140\\
187.388	-0.123\\
188.289	0.000\\
188.400	0.000\\
188.858	0.000\\
189.080	0.000\\
189.298	0.000\\
189.646	0.000\\
189.897	0.000\\
189.989	0.000\\
190.216	0.000\\
190.322	0.000\\
190.657	0.000\\
190.766	0.000\\
190.879	0.000\\
191.218	0.000\\
191.330	0.000\\
192.117	0.000\\
192.229	0.000\\
192.344	0.000\\
192.470	0.000\\
192.574	0.000\\
193.129	0.077\\
193.244	0.099\\
193.354	0.119\\
193.464	0.140\\
194.249	0.211\\
194.919	0.211\\
195.814	0.211\\
195.928	0.211\\
196.043	0.211\\
196.266	0.211\\
197.496	0.211\\
198.275	0.211\\
198.835	0.211\\
198.946	0.211\\
199.058	0.211\\
199.506	0.211\\
199.630	0.211\\
199.965	0.211\\
200.185	0.211\\
200.631	0.211\\
201.093	0.211\\
201.193	0.211\\
201.415	0.211\\
201.755	0.211\\
202.091	0.211\\
202.214	0.211\\
202.318	0.211\\
202.657	0.211\\
202.892	0.211\\
203.780	0.211\\
203.892	0.211\\
204.021	0.211\\
204.112	0.211\\
204.227	0.211\\
204.339	0.211\\
204.450	0.211\\
204.787	0.211\\
204.899	0.211\\
205.010	0.211\\
205.580	0.211\\
205.686	0.211\\
205.796	0.211\\
206.128	0.211\\
206.572	0.211\\
207.130	0.211\\
207.241	0.211\\
207.352	0.211\\
207.467	0.211\\
208.139	0.211\\
208.365	0.211\\
208.594	0.191\\
209.031	0.108\\
209.720	0.000\\
209.832	0.000\\
210.170	0.000\\
210.291	0.000\\
210.393	0.000\\
210.505	0.000\\
210.958	0.000\\
211.955	0.000\\
212.068	0.000\\
212.180	0.000\\
212.515	0.000\\
212.627	0.000\\
213.296	-0.062\\
213.527	-0.109\\
214.084	-0.224\\
214.417	-0.231\\
214.532	-0.231\\
215.092	-0.231\\
215.200	-0.231\\
215.421	-0.231\\
215.532	-0.231\\
215.982	-0.231\\
216.316	-0.231\\
216.877	-0.231\\
217.325	-0.231\\
217.548	-0.231\\
217.660	-0.231\\
218.330	-0.231\\
218.775	-0.231\\
219.136	-0.231\\
220.030	-0.231\\
220.140	-0.231\\
220.932	-0.231\\
221.064	-0.231\\
221.156	-0.231\\
221.845	-0.231\\
221.949	-0.231\\
222.161	-0.231\\
222.277	-0.231\\
222.390	-0.231\\
222.500	-0.231\\
222.953	-0.231\\
223.178	-0.231\\
223.290	-0.231\\
223.968	-0.231\\
224.304	-0.231\\
224.414	-0.231\\
224.636	-0.231\\
224.748	-0.231\\
224.859	-0.231\\
224.973	-0.231\\
225.299	-0.231\\
226.303	-0.231\\
226.538	-0.231\\
226.870	-0.231\\
226.981	-0.231\\
227.203	-0.231\\
227.318	-0.231\\
227.545	-0.231\\
227.766	-0.231\\
227.878	-0.231\\
228.214	-0.231\\
228.998	-0.231\\
229.111	-0.231\\
229.561	-0.231\\
230.141	-0.231\\
231.497	-0.141\\
231.722	-0.096\\
232.058	-0.027\\
232.173	-0.004\\
232.392	0.000\\
233.072	0.000\\
233.180	0.000\\
234.068	0.000\\
234.181	0.000\\
234.296	0.000\\
234.407	0.000\\
234.629	0.000\\
234.742	0.000\\
235.188	0.000\\
235.298	0.000\\
235.629	0.000\\
235.963	0.000\\
236.075	0.000\\
236.519	0.000\\
236.630	0.000\\
237.867	0.000\\
237.983	0.000\\
238.204	0.000\\
238.422	0.000\\
238.868	0.000\\
239.094	0.031\\
239.208	0.052\\
239.323	0.074\\
239.548	0.115\\
239.658	0.136\\
240.012	0.201\\
240.235	0.213\\
240.457	0.213\\
240.568	0.213\\
240.793	0.213\\
240.910	0.213\\
241.245	0.213\\
241.357	0.213\\
241.472	0.213\\
241.804	0.213\\
242.168	0.213\\
242.256	0.213\\
242.925	0.213\\
243.036	0.213\\
243.266	0.213\\
244.281	0.213\\
244.623	0.213\\
245.091	0.213\\
245.184	0.213\\
245.629	0.213\\
245.851	0.213\\
245.962	0.213\\
246.419	0.213\\
246.530	0.213\\
246.641	0.213\\
246.752	0.213\\
247.657	0.213\\
247.766	0.213\\
247.990	0.213\\
248.103	0.213\\
248.552	0.213\\
248.681	0.213\\
249.114	0.213\\
249.224	0.213\\
249.562	0.213\\
249.675	0.213\\
250.136	0.213\\
250.249	0.213\\
250.481	0.213\\
251.024	0.213\\
251.135	0.213\\
251.245	0.213\\
251.357	0.213\\
251.472	0.213\\
251.584	0.213\\
251.924	0.213\\
252.151	0.213\\
252.489	0.213\\
253.600	0.213\\
254.609	0.213\\
254.721	0.213\\
255.169	0.213\\
255.277	0.213\\
255.618	0.213\\
256.289	0.213\\
256.402	0.213\\
256.630	0.213\\
257.087	0.213\\
257.528	0.213\\
258.646	0.213\\
259.209	0.213\\
260.003	0.213\\
260.114	0.213\\
262.133	0.213\\
262.245	0.213\\
262.583	0.213\\
262.806	0.186\\
263.481	0.059\\
263.605	0.035\\
264.269	0.000\\
264.935	0.000\\
265.384	0.000\\
265.497	0.000\\
266.160	0.000\\
266.271	0.000\\
266.495	0.000\\
266.952	0.000\\
267.055	0.000\\
267.503	0.000\\
267.612	0.000\\
268.288	0.000\\
268.629	0.000\\
268.965	0.000\\
269.413	0.000\\
269.875	0.000\\
270.329	0.000\\
270.550	0.000\\
271.333	0.000\\
271.669	0.000\\
272.111	0.000\\
272.340	0.000\\
272.467	0.000\\
272.567	0.000\\
273.008	0.000\\
273.568	0.000\\
273.904	0.000\\
274.017	0.000\\
274.356	0.000\\
274.466	0.000\\
274.912	0.000\\
275.024	0.000\\
275.135	0.000\\
275.695	0.000\\
275.805	0.000\\
276.476	0.000\\
276.589	0.000\\
276.817	0.000\\
276.926	0.000\\
277.260	0.000\\
277.596	nan\\
};
\addlegendentry{Desired velocity};

\addplot [color=black,dashed,forget plot]
  table[row sep=crcr]{%
0.000	0.000\\
253.600	0.000\\
};
\end{axis}
\end{tikzpicture}%

%% file: images/ardrone_velocity_estimate_control_ver.tikz
%
%
\begin{tikzpicture}

\begin{axis}[%
width=0.95092\figurewidth,
height=\figureheight,
at={(0\figurewidth,0\figureheight)},
scale only axis,
unbounded coords=jump,
every outer x axis line/.append style={black},
every x tick label/.append style={font=\color{black}},
xmin=73.999,
xmax=253.600,
xlabel={time [s]},
every outer y axis line/.append style={black},
every y tick label/.append style={font=\color{black}},
ymin=-0.500,
ymax=0.500,
ylabel={velocity [m/s]},
axis x line*=bottom,
axis y line*=left
]
\addplot [color=blue,solid,forget plot]
  table[row sep=crcr]{%
1.016	-0.010\\
1.277	0.030\\
2.821	0.068\\
3.840	0.074\\
4.607	0.048\\
4.861	0.072\\
5.628	0.057\\
6.655	0.014\\
7.673	0.020\\
8.725	0.010\\
9.491	0.019\\
9.750	0.007\\
10.509	0.046\\
11.537	0.057\\
12.803	0.052\\
14.336	0.050\\
14.594	0.058\\
15.355	0.055\\
16.423	0.031\\
16.535	0.034\\
16.762	0.032\\
16.873	0.021\\
17.207	0.005\\
17.319	0.003\\
18.108	-0.052\\
18.571	-0.055\\
18.683	0.059\\
19.239	0.084\\
19.458	0.101\\
19.913	0.145\\
20.023	0.137\\
20.248	0.008\\
20.359	-0.018\\
20.927	-0.032\\
21.037	-0.058\\
21.150	-0.059\\
21.376	-0.075\\
21.715	-0.085\\
21.828	-0.087\\
21.939	-0.065\\
22.507	0.023\\
22.619	0.081\\
22.730	0.123\\
23.067	0.125\\
23.178	0.085\\
23.292	-0.018\\
23.627	-0.063\\
23.739	-0.104\\
24.183	-0.106\\
24.294	-0.072\\
24.632	-0.059\\
24.968	-0.040\\
25.079	-0.037\\
25.531	-0.050\\
25.643	-0.050\\
26.541	-0.057\\
27.330	-0.034\\
27.441	-0.019\\
27.891	0.018\\
28.119	0.063\\
28.232	0.082\\
28.343	0.053\\
28.688	0.046\\
28.797	0.029\\
28.909	0.010\\
29.246	0.009\\
29.358	0.008\\
29.809	0.001\\
29.922	-0.013\\
30.256	-0.035\\
30.479	-0.035\\
30.821	-0.023\\
30.932	-0.005\\
31.044	0.016\\
31.382	0.041\\
31.604	0.059\\
31.830	0.080\\
31.942	0.070\\
32.610	0.066\\
32.833	0.047\\
33.054	0.026\\
33.167	-0.024\\
34.295	-0.049\\
34.406	-0.061\\
34.517	-0.056\\
34.737	-0.052\\
35.073	-0.033\\
35.184	-0.001\\
35.636	0.001\\
36.198	0.002\\
36.538	0.009\\
36.757	0.001\\
37.088	-0.009\\
37.989	0.001\\
38.101	-0.028\\
38.326	-0.005\\
38.438	-0.008\\
38.676	-0.013\\
39.013	-0.009\\
39.572	0.031\\
39.907	-0.014\\
40.132	0.002\\
40.580	0.027\\
40.910	0.025\\
41.020	0.018\\
41.132	0.037\\
41.357	0.009\\
41.466	-0.040\\
41.581	-0.072\\
42.480	-0.081\\
42.592	-0.069\\
43.039	-0.036\\
43.154	-0.003\\
43.265	0.029\\
43.596	0.034\\
43.708	0.027\\
43.820	0.026\\
43.935	0.051\\
44.344	0.046\\
44.654	0.040\\
44.859	0.021\\
44.939	-0.002\\
45.471	-0.045\\
46.164	-0.037\\
46.835	-0.057\\
47.927	-0.030\\
48.869	0.011\\
49.432	0.055\\
49.653	0.043\\
50.322	0.084\\
50.545	0.075\\
50.768	0.046\\
50.879	0.006\\
51.104	-0.013\\
51.555	-0.014\\
51.667	-0.015\\
51.888	-0.016\\
52.004	-0.009\\
52.451	0.021\\
52.562	0.017\\
53.119	0.014\\
53.455	0.019\\
53.569	0.054\\
53.681	0.055\\
54.014	0.043\\
54.123	0.042\\
54.567	0.045\\
54.678	0.020\\
54.900	0.029\\
55.014	0.042\\
55.346	0.048\\
55.456	0.043\\
55.679	0.036\\
56.356	0.025\\
56.692	0.008\\
56.804	0.020\\
57.362	0.045\\
57.922	0.078\\
58.034	0.039\\
58.156	0.042\\
58.587	0.147\\
58.699	0.159\\
59.051	0.145\\
59.278	0.179\\
59.389	0.186\\
59.612	0.075\\
59.723	0.051\\
60.966	0.048\\
61.193	0.006\\
61.308	-0.033\\
61.988	-0.086\\
62.095	-0.112\\
62.541	-0.138\\
63.551	-0.123\\
63.775	-0.106\\
64.335	-0.087\\
64.447	-0.086\\
64.559	-0.082\\
64.919	-0.079\\
65.024	-0.075\\
65.130	-0.075\\
65.349	-0.084\\
66.009	-0.108\\
66.120	-0.119\\
67.586	-0.198\\
67.706	-0.267\\
68.148	-0.258\\
68.487	-0.240\\
68.949	-0.234\\
69.060	-0.162\\
69.506	-0.153\\
69.617	-0.123\\
70.173	-0.049\\
71.407	0.022\\
72.086	0.090\\
72.412	0.177\\
72.519	0.163\\
73.521	0.096\\
73.633	0.026\\
73.999	-0.052\\
74.082	-0.102\\
74.193	-0.151\\
74.526	-0.172\\
74.641	-0.200\\
75.320	-0.200\\
75.880	-0.166\\
76.320	-0.128\\
76.882	-0.117\\
76.997	-0.109\\
77.553	-0.111\\
78.001	-0.133\\
79.137	-0.156\\
79.695	-0.143\\
79.810	-0.106\\
80.146	-0.067\\
80.376	-0.039\\
80.481	-0.017\\
80.950	-0.018\\
81.403	-0.035\\
81.518	-0.074\\
82.061	-0.108\\
82.273	-0.127\\
82.494	-0.167\\
82.608	-0.197\\
83.400	-0.130\\
84.622	-0.107\\
85.299	-0.100\\
85.978	-0.063\\
86.088	-0.001\\
86.758	0.011\\
87.319	0.017\\
87.432	0.075\\
88.434	0.105\\
88.545	0.101\\
88.877	0.060\\
91.239	0.089\\
91.348	0.066\\
92.589	0.064\\
92.701	0.065\\
92.813	0.058\\
93.032	0.063\\
93.596	0.075\\
94.053	0.068\\
94.496	0.077\\
95.722	0.095\\
97.171	0.093\\
97.283	0.077\\
97.508	0.092\\
98.744	0.087\\
98.968	0.057\\
99.210	0.030\\
100.226	0.013\\
100.562	-0.012\\
100.672	-0.027\\
101.121	-0.022\\
101.233	-0.002\\
101.458	0.016\\
102.241	0.024\\
102.582	0.032\\
103.030	0.035\\
103.142	0.041\\
103.587	0.026\\
103.923	0.008\\
104.373	-0.011\\
104.597	-0.019\\
104.709	-0.051\\
104.931	-0.040\\
105.042	-0.025\\
106.734	-0.006\\
106.846	0.041\\
107.301	0.062\\
107.642	0.058\\
107.756	0.049\\
107.979	0.038\\
108.321	0.008\\
108.548	0.005\\
109.341	0.004\\
109.455	0.014\\
110.125	0.020\\
110.795	0.012\\
110.904	0.004\\
111.803	-0.003\\
111.913	-0.003\\
112.024	0.006\\
112.916	0.020\\
113.025	0.013\\
113.359	0.009\\
115.040	0.007\\
115.152	0.005\\
115.492	0.032\\
115.824	0.048\\
115.936	0.054\\
116.278	0.046\\
116.837	0.033\\
116.947	-0.007\\
117.063	-0.010\\
117.291	-0.013\\
117.740	-0.012\\
118.746	-0.011\\
118.971	-0.010\\
119.083	-0.014\\
119.225	-0.000\\
119.321	0.012\\
119.885	0.011\\
119.999	0.005\\
120.332	0.013\\
120.667	0.018\\
121.336	0.050\\
122.006	0.076\\
122.235	0.089\\
122.343	0.078\\
122.565	0.055\\
122.677	0.006\\
123.016	0.000\\
123.137	0.008\\
124.581	0.019\\
124.694	0.034\\
124.806	0.015\\
124.918	-0.028\\
125.697	-0.063\\
125.817	-0.096\\
125.920	-0.128\\
126.263	-0.110\\
126.375	-0.096\\
127.156	-0.092\\
127.267	-0.087\\
127.489	-0.077\\
127.823	-0.068\\
127.935	-0.057\\
128.274	-0.040\\
128.384	-0.014\\
128.608	0.002\\
128.942	-0.001\\
129.061	-0.006\\
129.409	-0.012\\
129.523	-0.023\\
129.748	-0.020\\
129.860	-0.022\\
130.304	-0.028\\
131.199	-0.042\\
131.755	-0.057\\
132.202	-0.072\\
132.425	-0.058\\
132.879	-0.035\\
133.442	-0.006\\
133.577	0.026\\
133.786	0.049\\
133.915	0.048\\
134.014	0.044\\
134.341	0.033\\
134.455	0.024\\
134.679	0.013\\
134.794	0.006\\
135.020	-0.004\\
135.580	-0.002\\
135.918	-0.008\\
136.592	-0.007\\
136.703	-0.003\\
137.369	0.004\\
137.927	-0.003\\
138.041	-0.009\\
138.487	-0.016\\
139.963	-0.023\\
140.294	-0.033\\
140.408	-0.030\\
141.310	-0.020\\
141.634	0.021\\
141.747	0.030\\
141.864	0.036\\
142.086	0.034\\
142.416	0.022\\
142.528	-0.030\\
142.761	-0.040\\
142.977	-0.042\\
143.200	-0.036\\
143.311	-0.019\\
143.422	-0.003\\
143.646	0.001\\
143.981	-0.002\\
144.654	0.001\\
145.103	0.000\\
146.003	-0.003\\
146.118	-0.022\\
146.568	-0.029\\
147.693	-0.042\\
148.032	-0.052\\
148.250	-0.052\\
148.599	-0.030\\
148.702	-0.016\\
148.805	-0.007\\
148.919	-0.002\\
149.032	0.002\\
149.383	0.005\\
149.609	0.006\\
149.962	0.009\\
150.061	0.012\\
150.170	0.051\\
150.304	0.058\\
150.737	0.058\\
150.852	0.061\\
151.071	0.053\\
151.189	0.008\\
151.525	-0.004\\
151.636	-0.008\\
152.639	-0.023\\
152.753	-0.030\\
153.087	-0.029\\
153.198	-0.034\\
153.989	-0.032\\
154.098	-0.022\\
154.761	-0.018\\
154.878	-0.017\\
156.010	-0.009\\
156.111	0.003\\
156.226	0.008\\
156.335	0.007\\
156.893	0.001\\
157.339	-0.006\\
157.450	-0.023\\
157.906	-0.036\\
158.026	-0.035\\
158.240	-0.030\\
158.353	-0.032\\
158.703	-0.017\\
158.798	0.002\\
159.247	0.021\\
159.364	0.041\\
159.597	0.044\\
159.708	0.036\\
160.042	0.025\\
161.387	0.007\\
161.503	-0.013\\
161.941	-0.012\\
162.053	-0.008\\
162.279	-0.001\\
162.498	0.010\\
162.612	0.020\\
162.945	0.027\\
163.054	0.026\\
163.167	0.019\\
163.390	0.001\\
163.613	-0.021\\
164.066	-0.023\\
164.180	-0.032\\
165.280	-0.035\\
165.963	-0.025\\
166.308	-0.010\\
166.614	-0.005\\
167.195	-0.005\\
167.531	0.003\\
168.320	0.017\\
169.324	0.040\\
170.002	0.044\\
170.672	0.066\\
170.899	0.083\\
171.803	0.096\\
171.915	0.100\\
172.251	0.111\\
172.367	0.111\\
172.593	0.107\\
172.711	0.100\\
172.824	0.091\\
172.928	0.089\\
173.069	0.091\\
173.152	0.100\\
173.262	0.105\\
174.047	0.109\\
174.158	0.110\\
175.061	0.137\\
175.620	0.122\\
176.071	0.099\\
176.532	0.084\\
176.966	0.071\\
177.301	0.040\\
177.413	0.062\\
177.524	0.094\\
177.860	0.123\\
178.196	0.144\\
178.311	0.154\\
178.752	0.139\\
178.865	0.132\\
179.455	0.119\\
179.563	0.110\\
180.574	0.090\\
180.676	0.084\\
181.689	0.072\\
182.135	0.074\\
182.248	0.084\\
182.472	0.101\\
182.582	0.120\\
183.371	0.121\\
184.040	0.109\\
184.374	0.080\\
184.487	0.081\\
184.819	0.072\\
185.818	0.081\\
186.155	0.088\\
186.269	0.104\\
186.718	0.091\\
186.830	0.081\\
187.164	0.077\\
187.276	0.083\\
187.388	0.086\\
188.289	0.077\\
188.400	0.063\\
188.858	0.045\\
189.080	0.026\\
189.298	0.008\\
189.646	0.007\\
189.897	0.017\\
189.989	0.029\\
190.216	0.038\\
190.322	0.047\\
190.657	0.068\\
190.766	0.084\\
190.879	0.080\\
191.218	0.066\\
191.330	0.051\\
192.117	0.025\\
192.229	0.001\\
192.344	-0.007\\
192.470	-0.006\\
192.574	-0.003\\
193.129	-0.001\\
193.244	0.000\\
193.354	0.008\\
193.464	0.008\\
194.249	0.018\\
194.919	0.029\\
195.814	0.037\\
195.928	0.034\\
196.043	0.021\\
196.266	0.019\\
197.496	0.004\\
198.275	-0.017\\
198.835	-0.042\\
198.946	-0.065\\
199.058	-0.092\\
199.506	-0.118\\
199.630	-0.141\\
199.965	-0.147\\
200.185	-0.150\\
200.631	-0.172\\
201.093	-0.171\\
201.193	-0.159\\
201.415	-0.179\\
201.755	-0.180\\
202.091	-0.178\\
202.214	-0.189\\
202.318	-0.192\\
202.657	-0.176\\
202.892	-0.161\\
203.780	-0.146\\
203.892	-0.147\\
204.021	-0.159\\
204.112	-0.172\\
204.227	-0.183\\
204.339	-0.186\\
204.450	-0.171\\
204.787	-0.163\\
204.899	-0.158\\
205.010	-0.166\\
205.580	-0.181\\
205.686	-0.183\\
205.796	-0.171\\
206.128	-0.150\\
206.572	-0.119\\
207.130	-0.087\\
207.241	-0.075\\
207.352	-0.068\\
207.467	-0.062\\
208.139	-0.062\\
208.365	-0.056\\
208.594	-0.051\\
209.031	-0.057\\
209.720	-0.061\\
209.832	-0.060\\
210.170	-0.062\\
210.291	-0.056\\
210.393	-0.036\\
210.505	-0.024\\
210.958	-0.006\\
211.955	0.018\\
212.068	0.052\\
212.180	0.072\\
212.515	0.060\\
212.627	0.032\\
213.296	0.004\\
213.527	-0.014\\
214.084	-0.020\\
214.417	0.006\\
214.532	0.034\\
215.092	0.052\\
215.200	0.049\\
215.421	0.042\\
215.532	0.038\\
215.982	0.031\\
216.316	0.021\\
216.877	0.011\\
217.325	0.002\\
217.548	0.004\\
217.660	0.005\\
218.330	0.010\\
218.775	0.018\\
219.136	0.025\\
220.030	0.039\\
220.140	0.062\\
220.932	0.103\\
221.064	0.127\\
221.156	0.143\\
221.845	0.155\\
221.949	0.170\\
222.161	0.172\\
222.277	0.185\\
222.390	0.185\\
222.500	0.168\\
222.953	0.144\\
223.178	0.130\\
223.290	0.119\\
223.968	0.119\\
224.304	0.127\\
224.414	0.132\\
224.636	0.133\\
224.748	0.140\\
224.859	0.156\\
224.973	0.156\\
225.299	0.168\\
226.303	0.169\\
226.538	0.155\\
226.870	0.138\\
226.981	0.130\\
227.203	0.123\\
227.318	0.114\\
227.545	0.116\\
227.766	0.138\\
227.878	0.140\\
228.214	0.132\\
228.998	0.149\\
229.111	0.157\\
229.561	0.120\\
230.141	0.119\\
231.497	0.119\\
231.722	0.100\\
232.058	0.082\\
232.173	0.098\\
232.392	0.103\\
233.072	0.080\\
233.180	0.057\\
234.068	0.036\\
234.181	0.007\\
234.296	-0.021\\
234.407	-0.015\\
234.629	-0.010\\
234.742	-0.002\\
235.188	0.016\\
235.298	0.055\\
235.629	0.064\\
235.963	0.037\\
236.075	-0.001\\
236.519	-0.011\\
236.630	-0.049\\
237.867	-0.063\\
237.983	-0.045\\
238.204	0.001\\
238.422	0.021\\
238.868	0.045\\
239.094	0.067\\
239.208	0.075\\
239.323	0.061\\
239.548	0.024\\
239.658	-0.020\\
240.012	-0.070\\
240.235	-0.111\\
240.457	-0.151\\
240.568	-0.173\\
240.793	-0.192\\
240.910	-0.191\\
241.245	-0.181\\
241.357	-0.149\\
241.472	-0.118\\
241.804	-0.081\\
242.168	-0.050\\
242.256	-0.026\\
242.925	-0.027\\
243.036	-0.066\\
243.266	-0.097\\
244.281	-0.111\\
244.623	-0.118\\
245.091	-0.127\\
245.184	-0.109\\
245.629	-0.102\\
245.851	-0.104\\
245.962	-0.113\\
246.419	-0.115\\
246.530	-0.114\\
246.641	-0.128\\
246.752	-0.155\\
247.657	-0.168\\
247.766	-0.178\\
247.990	-0.179\\
248.103	-0.161\\
248.552	-0.147\\
248.681	-0.141\\
249.114	-0.133\\
249.224	-0.152\\
249.562	-0.177\\
249.675	-0.180\\
250.136	-0.188\\
250.249	-0.185\\
250.481	-0.160\\
251.024	-0.128\\
251.135	-0.110\\
251.245	-0.090\\
251.357	-0.087\\
251.472	-0.097\\
251.584	-0.111\\
251.924	-0.125\\
252.151	-0.138\\
252.489	-0.125\\
253.600	-0.107\\
254.609	-0.100\\
254.721	-0.092\\
255.169	-0.079\\
255.277	-0.115\\
255.618	-0.144\\
256.289	-0.151\\
256.402	-0.162\\
256.630	-0.173\\
257.087	-0.131\\
257.528	-0.089\\
258.646	-0.043\\
259.209	-0.031\\
260.003	0.008\\
260.114	0.020\\
262.133	0.029\\
262.245	0.020\\
262.583	0.044\\
262.806	0.030\\
263.481	-0.000\\
263.605	0.020\\
264.269	0.009\\
264.935	-0.003\\
265.384	-0.074\\
265.497	-0.128\\
266.160	-0.237\\
266.271	-0.202\\
266.495	-0.159\\
266.952	-0.063\\
267.055	0.051\\
267.503	0.114\\
267.612	0.059\\
268.288	0.026\\
268.629	-0.005\\
268.965	0.101\\
269.413	0.180\\
269.875	0.262\\
270.329	0.263\\
270.550	0.263\\
271.333	0.110\\
271.669	0.056\\
272.111	0.021\\
272.340	0.014\\
272.467	-0.020\\
272.567	-0.011\\
273.008	-0.025\\
273.568	-0.055\\
273.904	-0.052\\
274.017	-0.017\\
274.356	-0.072\\
274.466	-0.073\\
274.912	-0.050\\
275.024	-0.079\\
275.135	-0.074\\
275.695	-0.011\\
275.805	-0.003\\
276.476	-0.002\\
276.589	0.049\\
276.817	-0.212\\
276.926	-0.228\\
277.260	-0.413\\
277.596	-0.037\\
};
\addplot [color=red,solid,forget plot]
  table[row sep=crcr]{%
nan	0.000\\
1.277	-0.000\\
2.821	-0.000\\
3.840	-0.000\\
4.607	0.000\\
4.861	-0.000\\
5.628	0.000\\
6.655	0.000\\
7.673	0.000\\
8.725	0.000\\
9.491	-0.000\\
9.750	-0.000\\
10.509	0.000\\
11.537	0.000\\
12.803	-0.000\\
14.336	-0.000\\
14.594	-0.000\\
15.355	0.000\\
16.423	0.000\\
16.535	0.000\\
16.762	0.000\\
16.873	-0.000\\
17.207	-0.000\\
17.319	-0.000\\
18.108	-0.000\\
18.571	-0.000\\
18.683	-0.000\\
19.239	0.000\\
19.458	0.000\\
19.913	0.000\\
20.023	0.000\\
20.248	0.000\\
20.359	0.000\\
20.927	0.000\\
21.037	-0.000\\
21.150	-0.000\\
21.376	-0.000\\
21.715	-0.000\\
21.828	0.000\\
21.939	0.000\\
22.507	0.000\\
22.619	0.000\\
22.730	0.000\\
23.067	-0.000\\
23.178	-0.000\\
23.292	-0.000\\
23.627	-0.000\\
23.739	-0.000\\
24.183	0.000\\
24.294	-0.000\\
24.632	-0.000\\
24.968	0.000\\
25.079	0.000\\
25.531	0.000\\
25.643	0.000\\
26.541	-0.000\\
27.330	-0.000\\
27.441	-0.000\\
27.891	0.000\\
28.119	0.000\\
28.232	0.000\\
28.343	-0.000\\
28.688	-0.000\\
28.797	-0.000\\
28.909	-0.000\\
29.246	0.000\\
29.358	0.000\\
29.809	0.000\\
29.922	0.000\\
30.256	0.000\\
30.479	0.000\\
30.821	0.000\\
30.932	0.000\\
31.044	0.000\\
31.382	-0.000\\
31.604	-0.000\\
31.830	-0.000\\
31.942	0.000\\
32.610	0.000\\
32.833	0.000\\
33.054	-0.000\\
33.167	-0.000\\
34.295	0.000\\
34.406	0.000\\
34.517	0.000\\
34.737	0.000\\
35.073	-0.000\\
35.184	-0.000\\
35.636	-0.000\\
36.198	-0.000\\
36.538	-0.000\\
36.757	-0.000\\
37.088	-0.000\\
37.989	0.000\\
38.101	0.000\\
38.326	0.000\\
38.438	0.000\\
38.676	0.000\\
39.013	0.000\\
39.572	0.000\\
39.907	0.000\\
40.132	0.000\\
40.580	-0.000\\
40.910	-0.000\\
41.020	0.000\\
41.132	0.000\\
41.357	0.000\\
41.466	0.000\\
41.581	0.000\\
42.480	-0.000\\
42.592	-0.000\\
43.039	0.000\\
43.154	0.000\\
43.265	0.000\\
43.596	-0.000\\
43.708	-0.000\\
43.820	-0.000\\
43.935	0.000\\
44.344	0.000\\
44.654	-0.000\\
44.859	-0.000\\
44.939	0.000\\
45.471	0.000\\
46.164	0.000\\
46.835	-0.000\\
47.927	-0.000\\
48.869	-0.000\\
49.432	-0.000\\
49.653	0.000\\
50.322	-0.000\\
50.545	-0.000\\
50.768	-0.000\\
50.879	-0.000\\
51.104	-0.000\\
51.555	-0.000\\
51.667	0.000\\
51.888	0.000\\
52.004	-0.000\\
52.451	-0.000\\
52.562	-0.000\\
53.119	-0.000\\
53.455	0.000\\
53.569	0.000\\
53.681	0.000\\
54.014	0.000\\
54.123	0.000\\
54.567	-0.000\\
54.678	-0.000\\
54.900	-0.000\\
55.014	-0.000\\
55.346	-0.000\\
55.456	-0.000\\
55.679	-0.000\\
56.356	-0.000\\
56.692	-0.000\\
56.804	-0.000\\
57.362	0.012\\
57.922	0.044\\
58.034	0.079\\
58.156	0.088\\
58.587	0.082\\
58.699	0.069\\
59.051	0.052\\
59.278	0.031\\
59.389	0.019\\
59.612	0.011\\
59.723	-0.022\\
60.966	-0.072\\
61.193	-0.109\\
61.308	-0.125\\
61.988	-0.137\\
62.095	-0.141\\
62.541	-0.119\\
63.551	-0.094\\
63.775	-0.080\\
64.335	-0.078\\
64.447	-0.074\\
64.559	-0.075\\
64.919	-0.080\\
65.024	-0.084\\
65.130	-0.089\\
65.349	-0.097\\
66.009	-0.106\\
66.120	-0.114\\
67.586	-0.117\\
67.706	-0.100\\
68.148	-0.054\\
68.487	0.008\\
68.949	0.049\\
69.060	0.065\\
69.506	0.063\\
69.617	0.173\\
70.173	0.351\\
71.407	0.424\\
72.086	0.321\\
72.412	0.209\\
72.519	0.078\\
73.521	-0.069\\
73.633	-0.183\\
73.999	-0.221\\
74.082	-0.247\\
74.193	-0.253\\
74.526	-0.249\\
74.641	-0.224\\
75.320	-0.182\\
75.880	-0.136\\
76.320	-0.105\\
76.882	-0.093\\
76.997	-0.097\\
77.553	-0.104\\
78.001	-0.097\\
79.137	-0.074\\
79.695	-0.047\\
79.810	-0.030\\
80.146	-0.022\\
80.376	-0.021\\
80.481	-0.028\\
80.950	-0.044\\
81.403	-0.062\\
81.518	-0.074\\
82.061	-0.080\\
82.273	-0.080\\
82.494	-0.075\\
82.608	-0.068\\
83.400	-0.033\\
84.622	0.015\\
85.299	0.019\\
85.978	-0.021\\
86.088	-0.031\\
86.758	-0.000\\
87.319	0.024\\
87.432	0.030\\
88.434	0.032\\
88.545	0.036\\
88.877	0.039\\
91.239	0.043\\
91.348	0.047\\
92.589	0.048\\
92.701	0.045\\
92.813	0.044\\
93.032	0.052\\
93.596	0.069\\
94.053	0.082\\
94.496	0.092\\
95.722	0.093\\
97.171	0.086\\
97.283	0.076\\
97.508	0.060\\
98.744	0.035\\
98.968	0.007\\
99.210	-0.006\\
100.226	-0.010\\
100.562	-0.004\\
100.672	0.005\\
101.121	0.013\\
101.233	0.018\\
101.458	0.020\\
102.241	0.021\\
102.582	0.017\\
103.030	0.012\\
103.142	0.008\\
103.587	0.005\\
103.923	-0.007\\
104.373	-0.017\\
104.597	-0.018\\
104.709	-0.010\\
104.931	-0.004\\
105.042	0.001\\
106.734	0.007\\
106.846	0.012\\
107.301	0.011\\
107.642	0.001\\
107.756	-0.007\\
107.979	-0.011\\
108.321	-0.011\\
108.548	-0.010\\
109.341	-0.010\\
109.455	-0.011\\
110.125	-0.009\\
110.795	-0.007\\
110.904	-0.006\\
111.803	-0.002\\
111.913	0.000\\
112.024	0.006\\
112.916	0.013\\
113.025	0.014\\
113.359	0.007\\
115.040	0.005\\
115.152	0.006\\
115.492	0.005\\
115.824	0.003\\
115.936	-0.007\\
116.278	-0.019\\
116.837	-0.027\\
116.947	-0.024\\
117.063	-0.022\\
117.291	-0.024\\
117.740	-0.025\\
118.746	-0.020\\
118.971	-0.014\\
119.083	-0.011\\
119.225	-0.006\\
119.321	-0.005\\
119.885	-0.009\\
119.999	-0.014\\
120.332	-0.012\\
120.667	0.000\\
121.336	0.012\\
122.006	0.013\\
122.235	0.006\\
122.343	-0.000\\
122.565	-0.006\\
122.677	-0.008\\
123.016	-0.005\\
123.137	-0.015\\
124.581	-0.036\\
124.694	-0.056\\
124.806	-0.064\\
124.918	-0.078\\
125.697	-0.092\\
125.817	-0.099\\
125.920	-0.097\\
126.263	-0.094\\
126.375	-0.085\\
127.156	-0.076\\
127.267	-0.072\\
127.489	-0.066\\
127.823	-0.052\\
127.935	-0.038\\
128.274	-0.027\\
128.384	-0.021\\
128.608	-0.020\\
128.942	-0.019\\
129.061	-0.017\\
129.409	-0.013\\
129.523	-0.013\\
129.748	-0.013\\
129.860	-0.014\\
130.304	-0.026\\
131.199	-0.046\\
131.755	-0.060\\
132.202	-0.052\\
132.425	-0.029\\
132.879	-0.000\\
133.442	0.019\\
133.577	0.016\\
133.786	0.004\\
133.915	0.001\\
134.014	-0.000\\
134.341	-0.005\\
134.455	-0.006\\
134.679	-0.002\\
134.794	-0.001\\
135.020	0.000\\
135.580	-0.000\\
135.918	0.003\\
136.592	0.013\\
136.703	0.025\\
137.369	0.025\\
137.927	0.017\\
138.041	0.005\\
138.487	-0.007\\
139.963	-0.017\\
140.294	-0.019\\
140.408	-0.015\\
141.310	-0.010\\
141.634	-0.007\\
141.747	-0.009\\
141.864	-0.010\\
142.086	-0.015\\
142.416	-0.021\\
142.528	-0.023\\
142.761	-0.019\\
142.977	-0.012\\
143.200	-0.004\\
143.311	0.001\\
143.422	-0.000\\
143.646	-0.005\\
143.981	-0.006\\
144.654	-0.002\\
145.103	-0.002\\
146.003	-0.010\\
146.118	-0.025\\
146.568	-0.031\\
147.693	-0.028\\
148.032	-0.017\\
148.250	-0.010\\
148.599	-0.005\\
148.702	-0.003\\
148.805	-0.004\\
148.919	-0.002\\
149.032	0.007\\
149.383	0.016\\
149.609	0.022\\
149.962	0.022\\
150.061	0.022\\
150.170	0.022\\
150.304	0.018\\
150.737	0.010\\
150.852	0.005\\
151.071	0.006\\
151.189	0.003\\
151.525	-0.001\\
151.636	-0.023\\
152.639	-0.051\\
152.753	-0.071\\
153.087	-0.069\\
153.198	-0.051\\
153.989	-0.025\\
154.098	0.000\\
154.761	0.013\\
154.878	0.019\\
156.010	0.017\\
156.111	0.021\\
156.226	0.028\\
156.335	0.019\\
156.893	-0.007\\
157.339	-0.029\\
157.450	-0.029\\
157.906	-0.022\\
158.026	-0.023\\
158.240	-0.024\\
158.353	-0.015\\
158.703	-0.000\\
158.798	0.008\\
159.247	0.010\\
159.364	0.005\\
159.597	-0.004\\
159.708	-0.016\\
160.042	-0.023\\
161.387	-0.027\\
161.503	-0.021\\
161.941	-0.014\\
162.053	-0.006\\
162.279	-0.004\\
162.498	0.002\\
162.612	-0.004\\
162.945	-0.015\\
163.054	-0.023\\
163.167	-0.024\\
163.390	-0.033\\
163.613	-0.045\\
164.066	-0.056\\
164.180	-0.048\\
165.280	-0.031\\
165.963	-0.019\\
166.308	-0.023\\
166.614	-0.034\\
167.195	-0.033\\
167.531	-0.014\\
168.320	0.014\\
169.324	0.034\\
170.002	0.049\\
170.672	0.054\\
170.899	0.062\\
171.803	0.076\\
171.915	0.085\\
172.251	0.085\\
172.367	0.084\\
172.593	0.082\\
172.711	0.082\\
172.824	0.087\\
172.928	0.093\\
173.069	0.096\\
173.152	0.097\\
173.262	0.092\\
174.047	0.083\\
174.158	0.076\\
175.061	0.076\\
175.620	0.077\\
176.071	0.063\\
176.532	0.050\\
176.966	0.055\\
177.301	0.077\\
177.413	0.094\\
177.524	0.104\\
177.860	0.108\\
178.196	0.106\\
178.311	0.098\\
178.752	0.092\\
178.865	0.093\\
179.455	0.097\\
179.563	0.088\\
180.574	0.071\\
180.676	0.067\\
181.689	0.082\\
182.135	0.098\\
182.248	0.104\\
182.472	0.106\\
182.582	0.101\\
183.371	0.085\\
184.040	0.065\\
184.374	0.051\\
184.487	0.048\\
184.819	0.062\\
185.818	0.078\\
186.155	0.080\\
186.269	0.073\\
186.718	0.071\\
186.830	0.073\\
187.164	0.076\\
187.276	0.075\\
187.388	0.057\\
188.289	0.030\\
188.400	0.011\\
188.858	0.010\\
189.080	0.015\\
189.298	0.021\\
189.646	0.030\\
189.897	0.039\\
189.989	0.045\\
190.216	0.049\\
190.322	0.051\\
190.657	0.048\\
190.766	0.045\\
190.879	0.038\\
191.218	0.029\\
191.330	0.020\\
192.117	0.012\\
192.229	0.009\\
192.344	0.015\\
192.470	0.022\\
192.574	0.030\\
193.129	0.038\\
193.244	0.043\\
193.354	0.040\\
193.464	0.037\\
194.249	0.029\\
194.919	0.016\\
195.814	0.004\\
195.928	0.002\\
196.043	-0.003\\
196.266	-0.009\\
197.496	-0.016\\
198.275	-0.037\\
198.835	-0.072\\
198.946	-0.099\\
199.058	-0.107\\
199.506	-0.107\\
199.630	-0.108\\
199.965	-0.111\\
200.185	-0.117\\
200.631	-0.130\\
201.093	-0.142\\
201.193	-0.147\\
201.415	-0.152\\
201.755	-0.161\\
202.091	-0.170\\
202.214	-0.176\\
202.318	-0.180\\
202.657	-0.183\\
202.892	-0.190\\
203.780	-0.197\\
203.892	-0.200\\
204.021	-0.199\\
204.112	-0.198\\
204.227	-0.196\\
204.339	-0.194\\
204.450	-0.184\\
204.787	-0.171\\
204.899	-0.160\\
205.010	-0.149\\
205.580	-0.137\\
205.686	-0.128\\
205.796	-0.118\\
206.128	-0.098\\
206.572	-0.075\\
207.130	-0.060\\
207.241	-0.053\\
207.352	-0.051\\
207.467	-0.049\\
208.139	-0.045\\
208.365	-0.040\\
208.594	-0.036\\
209.031	-0.038\\
209.720	-0.040\\
209.832	-0.032\\
210.170	-0.015\\
210.291	-0.000\\
210.393	0.006\\
210.505	0.006\\
210.958	0.012\\
211.955	0.022\\
212.068	0.028\\
212.180	0.020\\
212.515	0.005\\
212.627	-0.015\\
213.296	-0.026\\
213.527	-0.012\\
214.084	0.020\\
214.417	0.043\\
214.532	0.042\\
215.092	0.033\\
215.200	0.027\\
215.421	0.027\\
215.532	0.023\\
215.982	0.013\\
216.316	-0.002\\
216.877	-0.007\\
217.325	-0.005\\
217.548	-0.005\\
217.660	-0.007\\
218.330	-0.007\\
218.775	-0.014\\
219.136	0.001\\
220.030	0.038\\
220.140	0.083\\
220.932	0.104\\
221.064	0.118\\
221.156	0.119\\
221.845	0.120\\
221.949	0.121\\
222.161	0.122\\
222.277	0.123\\
222.390	0.121\\
222.500	0.112\\
222.953	0.098\\
223.178	0.092\\
223.290	0.105\\
223.968	0.124\\
224.304	0.132\\
224.414	0.133\\
224.636	0.136\\
224.748	0.141\\
224.859	0.143\\
224.973	0.141\\
225.299	0.140\\
226.303	0.142\\
226.538	0.147\\
226.870	0.148\\
226.981	0.148\\
227.203	0.150\\
227.318	0.157\\
227.545	0.164\\
227.766	0.169\\
227.878	0.170\\
228.214	0.161\\
228.998	0.144\\
229.111	0.125\\
229.561	0.102\\
230.141	0.084\\
231.497	0.080\\
231.722	0.084\\
232.058	0.084\\
232.173	0.080\\
232.392	0.066\\
233.072	0.048\\
233.180	0.026\\
234.068	0.006\\
234.181	-0.003\\
234.296	-0.001\\
234.407	0.001\\
234.629	-0.001\\
234.742	-0.002\\
235.188	-0.002\\
235.298	-0.006\\
235.629	-0.019\\
235.963	-0.036\\
236.075	-0.052\\
236.519	-0.061\\
236.630	-0.050\\
237.867	-0.025\\
237.983	-0.007\\
238.204	-0.003\\
238.422	-0.002\\
238.868	-0.012\\
239.094	-0.031\\
239.208	-0.048\\
239.323	-0.060\\
239.548	-0.075\\
239.658	-0.099\\
240.012	-0.125\\
240.235	-0.143\\
240.457	-0.148\\
240.568	-0.149\\
240.793	-0.143\\
240.910	-0.123\\
241.245	-0.099\\
241.357	-0.083\\
241.472	-0.068\\
241.804	-0.048\\
242.168	-0.032\\
242.256	-0.033\\
242.925	-0.041\\
243.036	-0.052\\
243.266	-0.083\\
244.281	-0.113\\
244.623	-0.115\\
245.091	-0.095\\
245.184	-0.093\\
245.629	-0.099\\
245.851	-0.104\\
245.962	-0.102\\
246.419	-0.099\\
246.530	-0.099\\
246.641	-0.107\\
246.752	-0.115\\
247.657	-0.114\\
247.766	-0.106\\
247.990	-0.098\\
248.103	-0.096\\
248.552	-0.099\\
248.681	-0.104\\
249.114	-0.107\\
249.224	-0.113\\
249.562	-0.121\\
249.675	-0.120\\
250.136	-0.109\\
250.249	-0.101\\
250.481	-0.095\\
251.024	-0.087\\
251.135	-0.080\\
251.245	-0.082\\
251.357	-0.085\\
251.472	-0.088\\
251.584	-0.089\\
251.924	-0.087\\
252.151	-0.087\\
252.489	-0.087\\
253.600	-0.099\\
254.609	-0.117\\
254.721	-0.137\\
255.169	-0.145\\
255.277	-0.151\\
255.618	-0.149\\
256.289	-0.145\\
256.402	-0.133\\
256.630	-0.115\\
257.087	-0.086\\
257.528	-0.040\\
258.646	0.006\\
259.209	0.026\\
260.003	0.021\\
260.114	0.031\\
262.133	0.053\\
262.245	0.062\\
262.583	0.048\\
262.806	0.015\\
263.481	-0.021\\
263.605	-0.045\\
264.269	-0.041\\
264.935	-0.035\\
265.384	-0.038\\
265.497	-0.021\\
266.160	0.020\\
266.271	0.064\\
266.495	0.101\\
266.952	0.130\\
267.055	0.127\\
267.503	0.105\\
267.612	0.012\\
268.288	-0.154\\
268.629	-0.308\\
268.965	-0.141\\
269.413	0.130\\
269.875	0.215\\
270.329	0.080\\
270.550	0.061\\
271.333	0.193\\
271.669	0.288\\
272.111	0.265\\
272.340	0.169\\
272.467	0.110\\
272.567	0.052\\
273.008	0.012\\
273.568	-0.001\\
273.904	-0.000\\
274.017	-0.000\\
274.356	-0.000\\
274.466	0.000\\
274.912	0.000\\
275.024	0.001\\
275.135	0.001\\
275.695	0.000\\
275.805	-0.000\\
276.476	0.000\\
276.589	0.001\\
276.817	0.001\\
276.926	0.001\\
277.260	0.000\\
nan	-0.000\\
};
\addplot [color=black!30!green,solid,forget plot]
  table[row sep=crcr]{%
1.016	0.000\\
1.277	0.000\\
2.821	0.000\\
3.840	0.000\\
4.607	0.000\\
4.861	0.000\\
5.628	0.000\\
6.655	0.000\\
7.673	0.000\\
8.725	0.000\\
9.491	0.000\\
9.750	0.000\\
10.509	0.000\\
11.537	0.000\\
12.803	0.000\\
14.336	0.000\\
14.594	0.000\\
15.355	0.000\\
16.423	0.000\\
16.535	0.000\\
16.762	0.000\\
16.873	0.000\\
17.207	0.000\\
17.319	0.000\\
18.108	0.000\\
18.571	0.000\\
18.683	0.000\\
19.239	0.000\\
19.458	0.000\\
19.913	0.000\\
20.023	0.000\\
20.248	0.000\\
20.359	0.000\\
20.927	0.000\\
21.037	0.000\\
21.150	0.000\\
21.376	0.000\\
21.715	0.000\\
21.828	0.000\\
21.939	0.000\\
22.507	0.000\\
22.619	0.000\\
22.730	0.000\\
23.067	0.000\\
23.178	0.000\\
23.292	0.000\\
23.627	0.000\\
23.739	0.000\\
24.183	0.000\\
24.294	0.000\\
24.632	0.000\\
24.968	0.000\\
25.079	0.000\\
25.531	0.000\\
25.643	0.000\\
26.541	0.000\\
27.330	0.000\\
27.441	0.000\\
27.891	0.000\\
28.119	0.000\\
28.232	0.000\\
28.343	0.000\\
28.688	0.000\\
28.797	0.000\\
28.909	0.000\\
29.246	0.000\\
29.358	0.000\\
29.809	0.000\\
29.922	0.000\\
30.256	0.000\\
30.479	0.000\\
30.821	0.000\\
30.932	0.000\\
31.044	0.000\\
31.382	0.000\\
31.604	0.000\\
31.830	0.000\\
31.942	0.000\\
32.610	0.000\\
32.833	0.000\\
33.054	0.000\\
33.167	0.000\\
34.295	0.000\\
34.406	0.000\\
34.517	0.000\\
34.737	0.000\\
35.073	0.000\\
35.184	0.000\\
35.636	0.000\\
36.198	0.000\\
36.538	0.000\\
36.757	0.000\\
37.088	0.000\\
37.989	0.000\\
38.101	0.000\\
38.326	0.000\\
38.438	0.000\\
38.676	0.000\\
39.013	0.000\\
39.572	0.000\\
39.907	0.000\\
40.132	0.000\\
40.580	0.000\\
40.910	0.000\\
41.020	0.000\\
41.132	0.000\\
41.357	0.000\\
41.466	0.000\\
41.581	0.000\\
42.480	0.000\\
42.592	0.000\\
43.039	0.000\\
43.154	0.000\\
43.265	0.000\\
43.596	0.000\\
43.708	0.000\\
43.820	0.000\\
43.935	0.000\\
44.344	0.000\\
44.654	0.000\\
44.859	0.000\\
44.939	0.000\\
45.471	0.000\\
46.164	0.000\\
46.835	0.000\\
47.927	0.000\\
48.869	0.000\\
49.432	0.000\\
49.653	0.000\\
50.322	0.000\\
50.545	0.000\\
50.768	0.000\\
50.879	0.000\\
51.104	0.000\\
51.555	0.000\\
51.667	0.000\\
51.888	0.000\\
52.004	0.000\\
52.451	0.000\\
52.562	0.000\\
53.119	0.000\\
53.455	0.000\\
53.569	0.000\\
53.681	0.000\\
54.014	0.000\\
54.123	0.000\\
54.567	0.000\\
54.678	0.000\\
54.900	0.000\\
55.014	0.000\\
55.346	0.000\\
55.456	0.000\\
55.679	0.000\\
56.356	0.000\\
56.692	0.000\\
56.804	0.000\\
57.362	0.000\\
57.922	0.000\\
58.034	0.000\\
58.156	0.000\\
58.587	0.000\\
58.699	0.000\\
59.051	0.000\\
59.278	0.000\\
59.389	0.000\\
59.612	0.000\\
59.723	0.000\\
60.966	0.000\\
61.193	0.000\\
61.308	0.000\\
61.988	0.000\\
62.095	0.000\\
62.541	0.000\\
63.551	0.000\\
63.775	0.000\\
64.335	0.000\\
64.447	0.000\\
64.559	0.000\\
64.919	0.000\\
65.024	0.000\\
65.130	0.000\\
65.349	0.000\\
66.009	0.000\\
66.120	0.000\\
67.586	0.000\\
67.706	0.000\\
68.148	0.000\\
68.487	0.000\\
68.949	0.000\\
69.060	0.000\\
69.506	0.000\\
69.617	0.000\\
70.173	0.000\\
71.407	0.000\\
72.086	0.000\\
72.412	0.000\\
72.519	0.000\\
73.521	0.000\\
73.633	0.000\\
73.999	0.000\\
74.082	0.000\\
74.193	0.000\\
74.526	0.000\\
74.641	0.000\\
75.320	0.000\\
75.880	0.000\\
76.320	0.000\\
76.882	0.000\\
76.997	0.000\\
77.553	0.000\\
78.001	0.000\\
79.137	0.000\\
79.695	0.000\\
79.810	0.000\\
80.146	0.000\\
80.376	0.000\\
80.481	0.000\\
80.950	0.000\\
81.403	0.000\\
81.518	0.000\\
82.061	0.000\\
82.273	0.000\\
82.494	0.000\\
82.608	0.000\\
83.400	0.000\\
84.622	0.000\\
85.299	0.000\\
85.978	0.055\\
86.088	0.069\\
86.758	0.147\\
87.319	0.147\\
87.432	0.147\\
88.434	0.147\\
88.545	0.147\\
88.877	0.147\\
91.239	0.147\\
91.348	0.147\\
92.589	0.147\\
92.701	0.147\\
92.813	0.147\\
93.032	0.147\\
93.596	0.147\\
94.053	0.147\\
94.496	0.147\\
95.722	0.147\\
97.171	0.102\\
97.283	0.087\\
97.508	0.057\\
98.744	0.000\\
98.968	0.000\\
99.210	0.000\\
100.226	0.000\\
100.562	0.000\\
100.672	0.000\\
101.121	0.000\\
101.233	0.000\\
101.458	0.000\\
102.241	0.000\\
102.582	0.000\\
103.030	0.000\\
103.142	0.000\\
103.587	0.000\\
103.923	0.000\\
104.373	0.000\\
104.597	0.000\\
104.709	0.000\\
104.931	0.000\\
105.042	0.000\\
106.734	0.000\\
106.846	0.000\\
107.301	0.000\\
107.642	0.000\\
107.756	0.000\\
107.979	0.000\\
108.321	0.000\\
108.548	0.000\\
109.341	0.000\\
109.455	0.000\\
110.125	0.000\\
110.795	0.000\\
110.904	0.000\\
111.803	0.000\\
111.913	0.000\\
112.024	0.000\\
112.916	0.000\\
113.025	0.000\\
113.359	0.000\\
115.040	0.000\\
115.152	0.000\\
115.492	0.000\\
115.824	0.000\\
115.936	0.000\\
116.278	0.000\\
116.837	0.000\\
116.947	0.000\\
117.063	0.000\\
117.291	0.000\\
117.740	0.000\\
118.746	0.000\\
118.971	0.000\\
119.083	0.000\\
119.225	0.000\\
119.321	0.000\\
119.885	0.000\\
119.999	0.000\\
120.332	0.000\\
120.667	0.000\\
121.336	0.000\\
122.006	0.000\\
122.235	0.000\\
122.343	0.000\\
122.565	0.000\\
122.677	0.000\\
123.016	0.000\\
123.137	0.000\\
124.581	0.000\\
124.694	0.000\\
124.806	0.000\\
124.918	0.000\\
125.697	0.000\\
125.817	0.000\\
125.920	0.000\\
126.263	0.000\\
126.375	0.000\\
127.156	0.000\\
127.267	0.000\\
127.489	0.000\\
127.823	0.000\\
127.935	0.000\\
128.274	0.000\\
128.384	0.000\\
128.608	0.000\\
128.942	0.000\\
129.061	0.000\\
129.409	0.000\\
129.523	0.000\\
129.748	0.000\\
129.860	0.000\\
130.304	0.000\\
131.199	0.000\\
131.755	0.000\\
132.202	0.000\\
132.425	0.000\\
132.879	0.000\\
133.442	0.000\\
133.577	0.000\\
133.786	0.000\\
133.915	0.000\\
134.014	0.000\\
134.341	0.000\\
134.455	0.000\\
134.679	0.000\\
134.794	0.000\\
135.020	0.000\\
135.580	0.000\\
135.918	0.000\\
136.592	0.000\\
136.703	0.000\\
137.369	0.000\\
137.927	0.000\\
138.041	0.000\\
138.487	0.000\\
139.963	0.049\\
140.294	0.049\\
140.408	0.049\\
141.310	0.049\\
141.634	0.049\\
141.747	0.049\\
141.864	0.049\\
142.086	0.049\\
142.416	0.049\\
142.528	0.049\\
142.761	0.049\\
142.977	0.049\\
143.200	0.049\\
143.311	0.049\\
143.422	0.049\\
143.646	0.049\\
143.981	0.049\\
144.654	0.030\\
145.103	0.010\\
146.003	0.000\\
146.118	0.000\\
146.568	0.000\\
147.693	0.000\\
148.032	0.000\\
148.250	0.000\\
148.599	0.000\\
148.702	0.000\\
148.805	0.004\\
148.919	0.009\\
149.032	0.013\\
149.383	0.029\\
149.609	0.038\\
149.962	0.049\\
150.061	0.049\\
150.170	0.049\\
150.304	0.049\\
150.737	0.049\\
150.852	0.049\\
151.071	0.046\\
151.189	0.041\\
151.525	0.026\\
151.636	0.021\\
152.639	0.000\\
152.753	0.000\\
153.087	0.000\\
153.198	0.000\\
153.989	0.000\\
154.098	0.000\\
154.761	0.000\\
154.878	0.000\\
156.010	0.000\\
156.111	0.000\\
156.226	0.000\\
156.335	0.000\\
156.893	0.000\\
157.339	0.000\\
157.450	0.000\\
157.906	0.000\\
158.026	0.000\\
158.240	0.000\\
158.353	0.000\\
158.703	0.000\\
158.798	0.000\\
159.247	0.000\\
159.364	0.000\\
159.597	0.000\\
159.708	0.000\\
160.042	0.000\\
161.387	0.000\\
161.503	0.000\\
161.941	0.000\\
162.053	0.000\\
162.279	0.000\\
162.498	0.000\\
162.612	0.000\\
162.945	0.000\\
163.054	0.000\\
163.167	0.000\\
163.390	0.000\\
163.613	0.000\\
164.066	0.000\\
164.180	0.000\\
165.280	0.000\\
165.963	0.000\\
166.308	0.000\\
166.614	0.000\\
167.195	0.000\\
167.531	0.000\\
168.320	0.029\\
169.324	0.075\\
170.002	0.075\\
170.672	0.075\\
170.899	0.075\\
171.803	0.075\\
171.915	0.075\\
172.251	0.075\\
172.367	0.075\\
172.593	0.075\\
172.711	0.075\\
172.824	0.075\\
172.928	0.075\\
173.069	0.075\\
173.152	0.075\\
173.262	0.075\\
174.047	0.075\\
174.158	0.075\\
175.061	0.075\\
175.620	0.075\\
176.071	0.075\\
176.532	0.075\\
176.966	0.075\\
177.301	0.075\\
177.413	0.075\\
177.524	0.075\\
177.860	0.075\\
178.196	0.075\\
178.311	0.075\\
178.752	0.075\\
178.865	0.075\\
179.455	0.056\\
179.563	0.049\\
180.574	0.000\\
180.676	0.000\\
181.689	0.000\\
182.135	0.000\\
182.248	0.000\\
182.472	0.000\\
182.582	0.000\\
183.371	0.000\\
184.040	0.000\\
184.374	0.000\\
184.487	0.000\\
184.819	0.000\\
185.818	0.000\\
186.155	0.000\\
186.269	0.000\\
186.718	0.000\\
186.830	0.000\\
187.164	0.000\\
187.276	0.000\\
187.388	0.000\\
188.289	0.000\\
188.400	0.000\\
188.858	0.000\\
189.080	0.000\\
189.298	0.000\\
189.646	0.000\\
189.897	0.000\\
189.989	0.000\\
190.216	0.000\\
190.322	0.000\\
190.657	0.000\\
190.766	0.000\\
190.879	0.000\\
191.218	0.000\\
191.330	0.000\\
192.117	0.000\\
192.229	0.000\\
192.344	0.000\\
192.470	0.000\\
192.574	0.000\\
193.129	0.000\\
193.244	0.000\\
193.354	0.000\\
193.464	0.000\\
194.249	0.000\\
194.919	0.000\\
195.814	0.000\\
195.928	0.000\\
196.043	0.000\\
196.266	-0.007\\
197.496	-0.058\\
198.275	-0.091\\
198.835	-0.093\\
198.946	-0.093\\
199.058	-0.093\\
199.506	-0.093\\
199.630	-0.093\\
199.965	-0.093\\
200.185	-0.093\\
200.631	-0.093\\
201.093	-0.093\\
201.193	-0.093\\
201.415	-0.093\\
201.755	-0.093\\
202.091	-0.093\\
202.214	-0.093\\
202.318	-0.093\\
202.657	-0.093\\
202.892	-0.090\\
203.780	-0.018\\
203.892	-0.009\\
204.021	0.000\\
204.112	0.000\\
204.227	0.000\\
204.339	0.000\\
204.450	0.000\\
204.787	0.000\\
204.899	0.000\\
205.010	0.000\\
205.580	0.000\\
205.686	0.000\\
205.796	0.000\\
206.128	0.000\\
206.572	0.000\\
207.130	0.000\\
207.241	0.000\\
207.352	0.000\\
207.467	0.000\\
208.139	0.000\\
208.365	0.000\\
208.594	0.000\\
209.031	0.000\\
209.720	0.000\\
209.832	0.000\\
210.170	0.000\\
210.291	0.000\\
210.393	0.000\\
210.505	0.000\\
210.958	0.000\\
211.955	0.000\\
212.068	0.000\\
212.180	0.000\\
212.515	0.000\\
212.627	0.000\\
213.296	0.000\\
213.527	0.000\\
214.084	0.000\\
214.417	0.000\\
214.532	0.000\\
215.092	0.000\\
215.200	0.000\\
215.421	0.000\\
215.532	0.000\\
215.982	0.000\\
216.316	0.000\\
216.877	0.000\\
217.325	0.000\\
217.548	0.000\\
217.660	0.000\\
218.330	0.000\\
218.775	0.014\\
219.136	0.046\\
220.030	0.101\\
220.140	0.101\\
220.932	0.101\\
221.064	0.101\\
221.156	0.101\\
221.845	0.101\\
221.949	0.101\\
222.161	0.101\\
222.277	0.101\\
222.390	0.101\\
222.500	0.101\\
222.953	0.101\\
223.178	0.101\\
223.290	0.101\\
223.968	0.101\\
224.304	0.101\\
224.414	0.101\\
224.636	0.101\\
224.748	0.101\\
224.859	0.101\\
224.973	0.101\\
225.299	0.101\\
226.303	0.101\\
226.538	0.101\\
226.870	0.101\\
226.981	0.101\\
227.203	0.101\\
227.318	0.101\\
227.545	0.101\\
227.766	0.091\\
227.878	0.081\\
228.214	0.051\\
228.998	0.000\\
229.111	0.000\\
229.561	0.000\\
230.141	0.000\\
231.497	0.000\\
231.722	0.000\\
232.058	0.000\\
232.173	0.000\\
232.392	0.000\\
233.072	0.000\\
233.180	0.000\\
234.068	0.000\\
234.181	0.000\\
234.296	0.000\\
234.407	0.000\\
234.629	0.000\\
234.742	0.000\\
235.188	0.000\\
235.298	0.000\\
235.629	0.000\\
235.963	0.000\\
236.075	0.000\\
236.519	0.000\\
236.630	0.000\\
237.867	-0.005\\
237.983	-0.015\\
238.204	-0.033\\
238.422	-0.051\\
238.868	-0.087\\
239.094	-0.092\\
239.208	-0.092\\
239.323	-0.092\\
239.548	-0.092\\
239.658	-0.092\\
240.012	-0.092\\
240.235	-0.092\\
240.457	-0.092\\
240.568	-0.092\\
240.793	-0.092\\
240.910	-0.092\\
241.245	-0.092\\
241.357	-0.092\\
241.472	-0.092\\
241.804	-0.092\\
242.168	-0.092\\
242.256	-0.092\\
242.925	-0.092\\
243.036	-0.092\\
243.266	-0.092\\
244.281	-0.092\\
244.623	-0.092\\
245.091	-0.092\\
245.184	-0.092\\
245.629	-0.092\\
245.851	-0.092\\
245.962	-0.092\\
246.419	-0.092\\
246.530	-0.092\\
246.641	-0.092\\
246.752	-0.092\\
247.657	-0.092\\
247.766	-0.092\\
247.990	-0.092\\
248.103	-0.092\\
248.552	-0.092\\
248.681	-0.092\\
249.114	-0.092\\
249.224	-0.083\\
249.562	-0.056\\
249.675	-0.047\\
250.136	-0.009\\
250.249	-0.000\\
250.481	0.000\\
251.024	0.000\\
251.135	0.000\\
251.245	0.000\\
251.357	0.000\\
251.472	0.000\\
251.584	0.000\\
251.924	0.000\\
252.151	0.000\\
252.489	0.000\\
253.600	0.000\\
254.609	0.000\\
254.721	0.000\\
255.169	0.000\\
255.277	0.000\\
255.618	0.000\\
256.289	0.000\\
256.402	0.000\\
256.630	0.000\\
257.087	0.000\\
257.528	0.000\\
258.646	0.000\\
259.209	0.000\\
260.003	0.086\\
260.114	0.099\\
262.133	0.133\\
262.245	0.133\\
262.583	0.133\\
262.806	0.116\\
263.481	0.037\\
263.605	0.022\\
264.269	0.000\\
264.935	0.000\\
265.384	0.000\\
265.497	0.000\\
266.160	0.000\\
266.271	0.000\\
266.495	0.000\\
266.952	0.000\\
267.055	0.000\\
267.503	0.000\\
267.612	0.000\\
268.288	0.000\\
268.629	0.000\\
268.965	0.000\\
269.413	0.000\\
269.875	0.000\\
270.329	0.000\\
270.550	0.000\\
271.333	0.000\\
271.669	0.000\\
272.111	0.000\\
272.340	0.000\\
272.467	0.000\\
272.567	0.000\\
273.008	0.000\\
273.568	0.000\\
273.904	0.000\\
274.017	0.000\\
274.356	0.000\\
274.466	0.000\\
274.912	0.000\\
275.024	0.000\\
275.135	0.000\\
275.695	0.000\\
275.805	0.000\\
276.476	0.000\\
276.589	0.000\\
276.817	0.000\\
276.926	0.000\\
277.260	0.000\\
277.596	nan\\
};
\addplot [color=black,dashed,forget plot]
  table[row sep=crcr]{%
0.000	0.000\\
253.600	0.000\\
};
\end{axis}
\end{tikzpicture}%

%% file: images/pocket_velocity_estimate_optitrack_hor.tikz
%
%
\begin{tikzpicture}

\begin{axis}[%
width=0.95092\figurewidth,
height=\figureheight,
at={(0\figurewidth,0\figureheight)},
scale only axis,
unbounded coords=jump,
every outer x axis line/.append style={black},
every x tick label/.append style={font=\color{black}},
xmin=76.384,
xmax=150.344,
xlabel={time [s]},
every outer y axis line/.append style={black},
every y tick label/.append style={font=\color{black}},
ymin=-0.500,
ymax=0.500,
ylabel={velocity [m/s]},
axis x line*=bottom,
axis y line*=left,
legend style={at={(0.37748,0.863878)},anchor=south west,legend cell align=left,align=left,fill=none,draw=none}
]
\addplot [color=blue,solid]
  table[row sep=crcr]{%
13.857	0.000\\
14.345	0.000\\
14.844	0.000\\
15.834	0.000\\
16.834	0.000\\
17.334	0.000\\
17.854	0.000\\
18.354	0.000\\
18.844	0.000\\
19.354	0.000\\
19.835	0.000\\
20.854	0.000\\
21.344	0.000\\
21.874	0.000\\
22.854	0.000\\
23.344	0.000\\
24.344	0.000\\
24.844	0.000\\
25.334	0.000\\
25.834	0.000\\
26.334	0.000\\
26.864	0.000\\
27.334	0.000\\
27.844	0.000\\
28.355	0.000\\
29.345	0.000\\
29.844	0.000\\
30.354	0.000\\
30.834	0.000\\
31.834	0.000\\
32.358	0.000\\
32.845	0.000\\
33.354	-0.024\\
33.844	-0.031\\
34.335	-0.048\\
34.844	-0.090\\
35.345	-0.184\\
35.844	-0.178\\
36.334	-0.184\\
36.854	-0.188\\
37.344	-0.152\\
38.335	-0.116\\
38.844	-0.074\\
40.346	-0.067\\
40.834	-0.062\\
41.357	-0.061\\
42.344	-0.008\\
42.844	-0.037\\
43.344	-0.031\\
43.844	-0.016\\
44.345	-0.011\\
44.844	-0.000\\
45.365	0.013\\
46.344	0.014\\
46.844	0.018\\
47.354	0.014\\
47.844	0.008\\
48.356	-0.004\\
48.844	-0.013\\
49.344	-0.025\\
49.864	-0.019\\
50.365	-0.017\\
52.344	-0.020\\
54.354	-0.039\\
54.835	-0.036\\
55.345	-0.044\\
55.844	-0.047\\
56.244	-0.040\\
56.354	-0.012\\
56.464	-0.011\\
56.584	-0.010\\
56.684	-0.015\\
56.784	-0.033\\
56.894	-0.049\\
57.019	-0.049\\
57.125	-0.032\\
57.224	-0.019\\
57.574	0.001\\
57.664	0.017\\
57.795	0.025\\
57.895	0.017\\
58.014	0.008\\
58.104	-0.022\\
58.204	-0.057\\
58.435	-0.087\\
58.574	-0.118\\
58.764	-0.138\\
59.024	-0.123\\
59.194	-0.094\\
59.314	-0.066\\
59.427	-0.030\\
59.514	0.018\\
59.754	0.050\\
59.855	0.077\\
60.294	0.091\\
60.404	0.093\\
60.534	0.076\\
60.614	0.070\\
60.834	0.066\\
61.054	0.074\\
61.165	0.081\\
61.516	0.077\\
61.705	0.068\\
61.954	0.063\\
62.144	0.059\\
62.255	0.054\\
62.385	0.054\\
62.524	0.054\\
62.585	0.045\\
62.695	0.021\\
62.815	0.007\\
62.914	-0.002\\
63.065	-0.001\\
63.807	0.003\\
63.924	0.023\\
64.024	0.036\\
64.114	0.050\\
64.224	0.053\\
64.444	0.044\\
64.574	0.024\\
64.784	0.001\\
65.034	-0.020\\
65.104	-0.040\\
65.204	-0.048\\
65.345	-0.050\\
65.434	-0.046\\
65.534	-0.041\\
65.655	-0.030\\
65.774	-0.016\\
66.024	-0.005\\
66.194	0.007\\
66.305	0.019\\
66.414	0.025\\
66.544	0.012\\
66.635	0.005\\
66.764	-0.005\\
66.964	-0.016\\
67.064	-0.024\\
67.194	-0.018\\
67.404	-0.017\\
67.514	-0.014\\
67.625	-0.011\\
67.724	-0.010\\
67.834	-0.013\\
67.944	-0.019\\
68.055	-0.020\\
68.174	-0.016\\
68.424	-0.002\\
68.524	0.024\\
68.614	0.050\\
68.734	0.072\\
68.934	0.097\\
69.096	0.086\\
69.167	0.047\\
69.265	0.017\\
69.374	-0.007\\
69.714	-0.022\\
69.804	0.009\\
69.925	0.060\\
70.024	0.096\\
70.134	0.119\\
70.244	0.106\\
70.365	0.087\\
70.464	0.069\\
70.677	0.060\\
70.895	0.057\\
71.024	0.047\\
71.124	0.047\\
71.234	0.053\\
71.444	0.069\\
71.664	0.055\\
71.774	0.077\\
72.024	0.082\\
72.214	0.066\\
72.374	0.038\\
72.435	0.027\\
73.428	0.002\\
73.549	-0.029\\
73.677	-0.023\\
73.854	-0.017\\
73.974	-0.003\\
74.074	0.008\\
74.184	0.024\\
74.305	0.031\\
74.414	0.035\\
74.538	0.040\\
74.624	0.063\\
74.734	0.099\\
74.834	0.092\\
74.965	0.104\\
75.065	0.101\\
75.174	0.088\\
75.285	0.057\\
75.524	0.038\\
75.615	0.015\\
75.714	0.020\\
75.824	0.014\\
76.044	0.005\\
76.264	0.015\\
76.384	0.021\\
76.524	0.019\\
76.584	0.029\\
76.815	0.042\\
77.034	0.043\\
77.134	0.044\\
77.245	0.022\\
77.464	-0.007\\
77.685	-0.031\\
77.804	-0.048\\
77.924	-0.056\\
78.045	-0.035\\
78.356	-0.010\\
78.455	0.002\\
78.564	0.016\\
78.664	0.001\\
78.794	-0.019\\
78.920	-0.046\\
79.036	-0.069\\
79.114	-0.092\\
79.224	-0.101\\
79.434	-0.112\\
79.554	-0.115\\
79.664	-0.087\\
79.864	-0.035\\
80.024	-0.012\\
80.104	-0.001\\
80.324	-0.001\\
80.424	-0.025\\
80.634	-0.070\\
80.754	-0.074\\
80.964	-0.045\\
81.074	-0.031\\
81.314	-0.028\\
81.404	-0.023\\
81.625	-0.037\\
81.744	-0.079\\
81.867	-0.090\\
81.955	-0.105\\
82.174	-0.125\\
82.395	-0.133\\
82.544	-0.127\\
82.614	-0.125\\
82.944	-0.110\\
83.050	-0.087\\
83.264	-0.067\\
83.537	-0.054\\
83.715	-0.032\\
83.814	-0.008\\
84.064	0.011\\
84.164	0.029\\
84.254	0.052\\
84.364	0.061\\
84.474	0.068\\
84.684	0.074\\
84.824	0.090\\
84.924	0.100\\
85.034	0.108\\
85.124	0.116\\
85.234	0.113\\
85.390	0.100\\
85.574	0.088\\
85.694	0.084\\
85.894	0.079\\
86.034	0.082\\
86.224	0.088\\
86.334	0.088\\
86.444	0.100\\
86.654	0.115\\
86.874	0.119\\
87.045	0.120\\
87.094	0.107\\
87.204	0.094\\
87.447	0.063\\
87.644	0.075\\
87.754	0.064\\
87.864	0.054\\
87.984	0.048\\
88.184	0.085\\
88.414	0.105\\
88.534	0.129\\
88.734	0.155\\
88.844	0.146\\
89.084	0.127\\
89.174	0.098\\
89.335	0.088\\
89.424	0.093\\
90.184	0.119\\
90.274	0.109\\
90.384	0.092\\
91.264	0.072\\
91.505	0.065\\
91.614	0.039\\
91.925	0.040\\
92.047	0.048\\
92.134	0.054\\
92.234	0.053\\
92.354	0.066\\
92.455	0.069\\
92.674	0.066\\
92.926	0.071\\
93.036	0.071\\
93.114	0.060\\
93.242	0.073\\
93.464	0.078\\
93.655	0.061\\
93.774	0.034\\
94.034	0.004\\
94.095	-0.053\\
94.345	-0.029\\
94.424	-0.006\\
94.574	0.014\\
94.674	0.020\\
94.864	0.020\\
95.084	-0.052\\
95.184	-0.105\\
95.317	-0.144\\
95.435	-0.154\\
95.534	-0.155\\
95.754	-0.155\\
95.844	-0.149\\
95.955	-0.125\\
96.064	-0.121\\
96.394	-0.119\\
96.547	-0.119\\
96.614	-0.093\\
96.834	-0.099\\
96.945	-0.064\\
97.164	-0.011\\
97.274	0.011\\
97.394	-0.011\\
97.534	-0.021\\
97.594	-0.052\\
97.714	-0.099\\
97.834	-0.113\\
98.044	-0.114\\
98.146	-0.112\\
98.274	-0.111\\
98.364	-0.110\\
98.474	-0.105\\
98.584	-0.099\\
98.705	-0.093\\
99.034	-0.079\\
99.234	-0.058\\
99.454	-0.039\\
99.694	-0.022\\
99.794	-0.006\\
99.944	0.002\\
100.065	0.001\\
100.114	0.000\\
100.224	0.007\\
100.334	0.016\\
100.564	0.027\\
100.704	0.050\\
100.784	0.072\\
100.894	0.082\\
101.034	0.088\\
101.094	0.095\\
101.204	0.093\\
101.324	0.089\\
101.426	0.089\\
101.549	0.094\\
101.644	0.097\\
101.874	0.104\\
101.976	0.106\\
102.094	0.096\\
102.534	0.076\\
102.634	0.084\\
102.734	0.078\\
102.954	0.069\\
103.074	0.090\\
103.174	0.111\\
103.524	0.093\\
103.614	0.097\\
103.724	0.138\\
103.944	0.161\\
104.044	0.202\\
104.274	0.238\\
104.524	0.235\\
104.614	0.226\\
104.824	0.209\\
104.935	0.161\\
105.054	0.120\\
105.167	0.131\\
105.384	0.126\\
105.474	0.119\\
105.714	0.121\\
105.816	0.123\\
105.944	0.104\\
106.047	0.083\\
106.144	0.083\\
106.254	0.088\\
106.354	0.098\\
106.464	0.102\\
106.574	0.102\\
106.686	0.088\\
106.785	0.085\\
106.894	0.101\\
107.124	0.108\\
107.224	0.115\\
107.364	0.125\\
107.554	0.124\\
107.774	0.069\\
107.894	0.029\\
108.094	0.002\\
108.204	-0.021\\
108.314	-0.042\\
108.434	-0.024\\
108.534	-0.004\\
108.644	0.001\\
108.754	0.000\\
108.864	0.002\\
108.986	0.002\\
109.084	-0.007\\
109.314	-0.011\\
109.404	-0.007\\
109.534	-0.001\\
109.747	0.003\\
109.864	0.007\\
109.954	0.003\\
110.074	-0.017\\
110.175	-0.034\\
110.314	-0.049\\
110.427	-0.066\\
110.546	-0.074\\
110.614	-0.072\\
110.724	-0.078\\
110.854	-0.072\\
110.944	-0.048\\
111.064	-0.027\\
111.174	-0.022\\
111.284	-0.014\\
111.536	-0.023\\
111.604	-0.038\\
111.714	-0.059\\
111.834	-0.068\\
111.964	-0.076\\
112.144	-0.077\\
112.364	-0.087\\
112.515	-0.092\\
112.594	-0.093\\
112.704	-0.094\\
112.824	-0.095\\
112.914	-0.095\\
113.144	-0.095\\
113.464	-0.090\\
113.584	-0.077\\
113.815	-0.075\\
113.904	-0.074\\
114.124	-0.066\\
114.244	-0.060\\
114.344	-0.066\\
114.454	-0.067\\
114.674	-0.061\\
115.046	-0.056\\
115.225	-0.058\\
115.325	-0.052\\
115.434	-0.044\\
115.607	-0.042\\
115.674	-0.047\\
115.764	-0.050\\
115.874	-0.055\\
115.984	-0.059\\
116.094	-0.065\\
116.314	-0.071\\
116.435	-0.073\\
116.549	-0.074\\
116.644	-0.069\\
116.744	-0.057\\
116.864	-0.041\\
116.985	-0.026\\
117.086	-0.014\\
117.184	-0.002\\
117.294	0.007\\
117.405	0.014\\
117.544	0.021\\
117.624	0.035\\
117.734	0.058\\
117.844	0.092\\
117.954	0.140\\
118.164	0.196\\
118.384	0.240\\
118.534	0.301\\
118.625	0.338\\
118.714	0.332\\
118.824	0.286\\
119.154	0.216\\
119.374	0.167\\
119.486	0.073\\
119.604	0.040\\
119.704	0.052\\
119.804	0.091\\
120.048	0.079\\
120.134	0.125\\
120.244	0.137\\
120.364	0.141\\
120.474	0.145\\
120.574	0.151\\
120.845	0.147\\
120.924	0.129\\
121.044	0.094\\
121.134	0.108\\
121.226	0.109\\
121.454	0.116\\
121.594	0.131\\
121.780	0.163\\
121.894	0.146\\
122.044	0.146\\
122.214	0.154\\
122.334	0.160\\
122.449	0.160\\
122.544	0.157\\
122.665	0.144\\
122.764	0.062\\
123.005	0.029\\
123.204	0.020\\
123.314	-0.008\\
123.534	-0.035\\
123.634	-0.005\\
123.747	-0.019\\
123.864	-0.061\\
123.964	-0.091\\
124.095	-0.091\\
124.184	-0.078\\
124.549	-0.066\\
124.734	-0.035\\
124.857	0.016\\
124.944	0.021\\
125.054	0.037\\
125.384	0.071\\
125.546	0.092\\
125.604	0.079\\
125.715	0.083\\
125.934	0.062\\
126.064	0.008\\
126.194	-0.053\\
126.264	-0.088\\
126.374	-0.117\\
126.704	-0.130\\
126.824	-0.130\\
126.955	-0.113\\
127.057	-0.087\\
127.144	-0.090\\
127.244	-0.103\\
127.464	-0.109\\
127.584	-0.127\\
127.694	-0.151\\
127.904	-0.152\\
128.044	-0.152\\
128.124	-0.148\\
128.234	-0.141\\
128.347	-0.125\\
128.454	-0.114\\
128.675	-0.106\\
128.774	-0.102\\
129.047	-0.101\\
129.104	-0.112\\
129.334	-0.115\\
129.434	-0.115\\
129.554	-0.112\\
129.654	-0.111\\
129.884	-0.107\\
129.984	-0.102\\
130.194	-0.094\\
130.536	-0.087\\
130.644	-0.076\\
130.754	-0.067\\
130.964	-0.060\\
131.084	-0.053\\
131.194	-0.044\\
131.324	-0.041\\
131.447	-0.042\\
131.544	-0.043\\
131.634	-0.049\\
131.734	-0.056\\
131.844	-0.055\\
132.074	-0.045\\
132.195	-0.034\\
132.404	-0.022\\
132.554	-0.013\\
132.614	-0.004\\
132.844	0.008\\
132.954	0.023\\
133.154	0.036\\
133.264	0.052\\
133.374	0.059\\
133.497	0.075\\
133.924	0.080\\
134.044	0.085\\
134.144	0.091\\
134.464	0.089\\
134.684	0.071\\
135.057	0.077\\
135.234	0.062\\
135.344	0.042\\
135.464	0.037\\
135.564	0.049\\
135.674	0.040\\
136.044	0.066\\
136.324	0.084\\
136.587	0.111\\
136.674	0.105\\
136.764	0.109\\
136.984	0.100\\
137.094	0.114\\
137.224	0.091\\
137.434	0.110\\
137.644	0.090\\
137.754	0.060\\
137.984	0.057\\
138.074	0.073\\
138.194	0.057\\
138.304	0.075\\
138.544	0.083\\
138.734	0.051\\
139.614	0.030\\
139.854	0.009\\
139.964	-0.025\\
140.044	-0.033\\
140.274	-0.047\\
140.384	-0.076\\
140.494	-0.104\\
140.594	-0.118\\
140.824	-0.137\\
141.065	-0.154\\
141.144	-0.160\\
141.254	-0.159\\
141.474	-0.154\\
141.584	-0.170\\
141.824	-0.179\\
142.615	-0.191\\
142.684	-0.200\\
142.784	-0.221\\
143.054	-0.221\\
143.114	-0.215\\
143.484	-0.211\\
143.554	-0.218\\
143.874	-0.217\\
143.994	-0.217\\
144.214	-0.227\\
144.334	-0.232\\
144.554	-0.217\\
144.644	-0.198\\
144.864	-0.176\\
144.974	-0.156\\
145.084	-0.114\\
145.194	-0.096\\
145.854	-0.082\\
145.954	-0.068\\
146.069	-0.055\\
146.184	-0.064\\
146.284	-0.068\\
146.394	-0.071\\
146.567	-0.074\\
146.744	-0.068\\
146.834	-0.060\\
147.064	-0.053\\
147.184	-0.048\\
147.384	-0.045\\
147.494	-0.047\\
147.604	-0.050\\
147.854	-0.047\\
147.944	-0.046\\
148.054	-0.040\\
148.174	-0.034\\
148.364	-0.030\\
148.474	-0.028\\
148.596	-0.018\\
148.714	-0.008\\
148.914	0.004\\
149.045	0.021\\
149.134	0.059\\
149.234	0.085\\
149.574	0.107\\
149.706	0.114\\
149.784	0.123\\
149.894	0.116\\
150.078	0.110\\
150.114	0.089\\
150.224	0.066\\
150.344	0.052\\
150.445	0.025\\
151.644	-0.001\\
151.764	-0.031\\
151.984	-0.026\\
152.084	-0.034\\
152.354	-0.028\\
152.434	-0.005\\
152.544	0.023\\
152.634	0.025\\
152.754	0.030\\
152.854	0.023\\
152.974	0.008\\
153.095	0.007\\
153.194	0.009\\
153.304	0.006\\
153.404	0.007\\
153.553	0.004\\
153.624	0.001\\
153.735	0.000\\
153.844	-0.001\\
153.956	0.001\\
154.054	0.001\\
154.164	0.001\\
154.274	0.001\\
154.384	0.001\\
154.515	0.000\\
154.714	0.000\\
154.844	0.000\\
154.924	-0.002\\
155.064	-0.003\\
155.155	-0.003\\
155.255	-0.003\\
155.694	-0.003\\
155.804	-0.001\\
156.060	0.000\\
156.234	0.000\\
156.364	-0.005\\
156.484	-0.006\\
157.074	-0.007\\
157.114	-0.007\\
157.336	-0.007\\
157.444	-0.002\\
157.687	-0.002\\
157.774	-0.002\\
157.894	-0.003\\
157.984	-0.011\\
158.104	-0.014\\
158.324	-0.015\\
158.445	-0.011\\
158.554	-0.011\\
158.754	-0.003\\
159.204	0.000\\
159.324	0.002\\
159.414	-0.001\\
159.554	-0.001\\
159.634	-0.001\\
159.854	-0.001\\
160.074	-0.001\\
160.559	-0.001\\
160.614	-0.002\\
160.744	-0.003\\
160.944	-0.003\\
161.054	-0.003\\
161.294	-0.003\\
161.504	-0.001\\
161.924	0.000\\
162.054	0.000\\
162.154	0.000\\
162.365	0.000\\
162.474	0.000\\
162.594	-0.001\\
162.714	-0.002\\
162.834	-0.004\\
162.955	-0.004\\
163.045	-0.004\\
163.134	-0.003\\
163.244	-0.002\\
163.358	0.000\\
163.464	0.000\\
163.577	0.000\\
163.699	0.000\\
163.894	0.000\\
164.067	0.000\\
164.124	0.000\\
164.224	0.000\\
};
\addlegendentry{Velocity estimate};

\addplot [color=red,solid]
  table[row sep=crcr]{%
nan	-0.000\\
nan	-0.000\\
nan	-0.000\\
nan	-0.000\\
nan	-0.000\\
17.334	-0.000\\
17.854	-0.000\\
18.354	-0.000\\
18.844	-0.000\\
19.354	-0.000\\
19.835	-0.000\\
20.854	-0.000\\
21.344	-0.000\\
21.874	-0.000\\
22.854	-0.000\\
23.344	-0.000\\
24.344	-0.000\\
24.844	-0.000\\
25.334	-0.000\\
25.834	-0.000\\
26.334	-0.000\\
26.864	-0.000\\
27.334	-0.000\\
27.844	-0.000\\
28.355	-0.000\\
29.345	-0.000\\
29.844	-0.000\\
30.354	-0.000\\
30.834	-0.000\\
31.834	-0.000\\
32.358	-0.000\\
32.845	0.001\\
33.354	0.001\\
33.844	0.013\\
34.335	0.029\\
34.844	0.037\\
35.345	0.024\\
35.844	0.003\\
36.334	-0.013\\
36.854	-0.021\\
37.344	-0.021\\
38.335	-0.044\\
38.844	-0.067\\
40.346	-0.087\\
40.834	-0.142\\
41.357	-0.180\\
42.344	-0.134\\
42.844	-0.048\\
43.344	-0.017\\
43.844	0.014\\
44.345	0.065\\
44.844	0.140\\
45.365	0.174\\
46.344	0.154\\
46.844	0.101\\
47.354	0.062\\
47.844	0.022\\
48.356	0.001\\
48.844	-0.003\\
49.344	-0.000\\
49.864	0.030\\
50.365	0.036\\
52.344	0.016\\
54.354	-0.010\\
54.835	-0.006\\
55.345	-0.008\\
55.844	-0.007\\
56.244	-0.003\\
56.354	-0.001\\
56.464	-0.000\\
56.584	-0.000\\
56.684	-0.000\\
56.784	-0.000\\
56.894	-0.000\\
57.019	-0.000\\
57.125	-0.002\\
57.224	-0.003\\
57.574	-0.003\\
57.664	-0.002\\
57.795	-0.002\\
57.895	-0.002\\
58.014	-0.002\\
58.104	-0.002\\
58.204	-0.003\\
58.435	-0.003\\
58.574	-0.004\\
58.764	-0.004\\
59.024	-0.003\\
59.194	-0.002\\
59.314	-0.000\\
59.427	0.004\\
59.514	0.007\\
59.754	0.012\\
59.855	0.014\\
60.294	0.013\\
60.404	0.008\\
60.534	0.010\\
60.614	0.014\\
60.834	0.014\\
61.054	0.016\\
61.165	0.018\\
61.516	0.019\\
61.705	0.016\\
61.954	0.014\\
62.144	0.010\\
62.255	0.007\\
62.385	0.004\\
62.524	0.000\\
62.585	-0.001\\
62.695	-0.001\\
62.815	-0.001\\
62.914	-0.002\\
63.065	-0.002\\
63.807	-0.002\\
63.924	-0.001\\
64.024	-0.001\\
64.114	-0.001\\
64.224	-0.001\\
64.444	-0.003\\
64.574	-0.010\\
64.784	-0.013\\
65.034	-0.010\\
65.104	-0.008\\
65.204	-0.009\\
65.345	-0.008\\
65.434	-0.007\\
65.534	-0.008\\
65.655	-0.010\\
65.774	-0.012\\
66.024	-0.010\\
66.194	-0.006\\
66.305	-0.004\\
66.414	-0.004\\
66.544	-0.004\\
66.635	-0.005\\
66.764	-0.005\\
66.964	-0.004\\
67.064	-0.004\\
67.194	-0.005\\
67.404	-0.004\\
67.514	-0.003\\
67.625	-0.000\\
67.724	0.003\\
67.834	0.004\\
67.944	0.004\\
68.055	0.006\\
68.174	0.015\\
68.424	0.027\\
68.524	0.036\\
68.614	0.048\\
68.734	0.059\\
68.934	0.052\\
69.096	0.036\\
69.167	0.029\\
69.265	0.051\\
69.374	0.059\\
69.714	0.050\\
69.804	0.031\\
69.925	0.031\\
70.024	0.031\\
70.134	0.032\\
70.244	0.032\\
70.365	0.039\\
70.464	0.051\\
70.677	0.054\\
70.895	0.041\\
71.024	0.031\\
71.124	0.039\\
71.234	0.064\\
71.444	0.075\\
71.664	0.082\\
71.774	0.088\\
72.024	0.092\\
72.214	0.066\\
72.374	0.171\\
72.435	0.228\\
73.428	0.178\\
73.549	0.061\\
73.677	0.058\\
73.854	0.048\\
73.974	0.040\\
74.074	0.041\\
74.184	0.042\\
74.305	0.043\\
74.414	0.040\\
74.538	0.038\\
74.624	0.037\\
74.734	0.041\\
74.834	0.042\\
74.965	0.041\\
75.065	0.034\\
75.174	0.038\\
75.285	0.037\\
75.524	0.031\\
75.615	0.023\\
75.714	0.027\\
75.824	0.027\\
76.044	0.020\\
76.264	0.013\\
76.384	0.009\\
76.524	-0.000\\
76.584	-0.021\\
76.815	-0.032\\
77.034	-0.029\\
77.134	-0.029\\
77.245	-0.040\\
77.464	-0.040\\
77.685	-0.031\\
77.804	-0.025\\
77.924	-0.036\\
78.045	-0.040\\
78.356	-0.034\\
78.455	-0.023\\
78.564	-0.025\\
78.664	-0.027\\
78.794	-0.049\\
78.920	-0.068\\
79.036	-0.083\\
79.114	-0.110\\
79.224	-0.134\\
79.434	-0.124\\
79.554	-0.122\\
79.664	-0.144\\
79.864	-0.132\\
80.024	-0.128\\
80.104	-0.130\\
80.324	-0.132\\
80.424	-0.104\\
80.634	-0.109\\
80.754	-0.099\\
80.964	-0.111\\
81.074	-0.100\\
81.314	-0.106\\
81.404	-0.096\\
81.625	-0.092\\
81.744	-0.066\\
81.867	-0.059\\
81.955	-0.058\\
82.174	-0.057\\
82.395	-0.042\\
82.544	-0.047\\
82.614	-0.054\\
82.944	-0.056\\
83.050	-0.042\\
83.264	-0.021\\
83.537	0.008\\
83.715	0.024\\
83.814	0.024\\
84.064	0.021\\
84.164	0.014\\
84.254	0.015\\
84.364	0.020\\
84.474	0.024\\
84.684	0.022\\
84.824	0.016\\
84.924	0.014\\
85.034	0.014\\
85.124	0.016\\
85.234	0.021\\
85.390	0.023\\
85.574	0.030\\
85.694	0.037\\
85.894	0.043\\
86.034	0.038\\
86.224	0.034\\
86.334	0.044\\
86.444	0.068\\
86.654	0.079\\
86.874	0.058\\
87.045	0.039\\
87.094	0.050\\
87.204	0.072\\
87.447	0.069\\
87.644	0.049\\
87.754	0.044\\
87.864	0.065\\
87.984	0.095\\
88.184	0.099\\
88.414	0.089\\
88.534	0.075\\
88.734	0.083\\
88.844	0.075\\
89.084	0.072\\
89.174	0.057\\
89.335	0.089\\
89.424	0.091\\
90.184	0.064\\
90.274	0.074\\
90.384	0.109\\
91.264	0.089\\
91.505	0.038\\
91.614	0.028\\
91.925	0.019\\
92.047	0.010\\
92.134	0.010\\
92.234	0.011\\
92.354	0.014\\
92.455	0.020\\
92.674	0.020\\
92.926	0.014\\
93.036	0.010\\
93.114	0.014\\
93.242	0.018\\
93.464	0.017\\
93.655	0.002\\
93.774	-0.016\\
94.034	-0.047\\
94.095	-0.059\\
94.345	-0.060\\
94.424	-0.047\\
94.574	-0.073\\
94.674	-0.112\\
94.864	-0.123\\
95.084	-0.101\\
95.184	-0.085\\
95.317	-0.084\\
95.435	-0.100\\
95.534	-0.107\\
95.754	-0.097\\
95.844	-0.077\\
95.955	-0.136\\
96.064	-0.179\\
96.394	-0.154\\
96.547	-0.121\\
96.614	-0.129\\
96.834	-0.140\\
96.945	-0.112\\
97.164	-0.098\\
97.274	-0.083\\
97.394	-0.075\\
97.534	-0.065\\
97.594	-0.056\\
97.714	-0.065\\
97.834	-0.062\\
98.044	-0.056\\
98.146	-0.044\\
98.274	-0.042\\
98.364	-0.041\\
98.474	-0.045\\
98.584	-0.069\\
98.705	-0.085\\
99.034	-0.067\\
99.234	-0.030\\
99.454	-0.016\\
99.694	-0.013\\
99.794	-0.003\\
99.944	0.005\\
100.065	0.011\\
100.114	0.012\\
100.224	0.018\\
100.334	0.022\\
100.564	0.023\\
100.704	0.022\\
100.784	0.027\\
100.894	0.026\\
101.034	0.024\\
101.094	0.024\\
101.204	0.027\\
101.324	0.027\\
101.426	0.026\\
101.549	0.034\\
101.644	0.037\\
101.874	0.034\\
101.976	0.065\\
102.094	0.088\\
102.534	0.074\\
102.634	0.052\\
102.734	0.064\\
102.954	0.061\\
103.074	0.079\\
103.174	0.091\\
103.524	0.076\\
103.614	0.060\\
103.724	0.068\\
103.944	0.075\\
104.044	0.083\\
104.274	0.078\\
104.524	0.059\\
104.614	0.046\\
104.824	0.044\\
104.935	0.036\\
105.054	0.045\\
105.167	0.047\\
105.384	0.052\\
105.474	0.049\\
105.714	0.046\\
105.816	0.035\\
105.944	0.034\\
106.047	0.032\\
106.144	0.032\\
106.254	0.033\\
106.354	0.034\\
106.464	0.035\\
106.574	0.034\\
106.686	0.034\\
106.785	0.045\\
106.894	0.051\\
107.124	0.048\\
107.224	0.047\\
107.364	0.060\\
107.554	0.060\\
107.774	0.057\\
107.894	0.047\\
108.094	0.029\\
108.204	0.004\\
108.314	-0.004\\
108.434	-0.004\\
108.534	-0.004\\
108.644	-0.004\\
108.754	-0.005\\
108.864	-0.006\\
108.986	-0.009\\
109.084	-0.010\\
109.314	-0.009\\
109.404	-0.015\\
109.534	-0.028\\
109.747	-0.035\\
109.864	-0.036\\
109.954	-0.036\\
110.074	-0.039\\
110.175	-0.041\\
110.314	-0.041\\
110.427	-0.035\\
110.546	-0.032\\
110.614	-0.036\\
110.724	-0.039\\
110.854	-0.039\\
110.944	-0.039\\
111.064	-0.055\\
111.174	-0.109\\
111.284	-0.130\\
111.536	-0.110\\
111.604	-0.084\\
111.714	-0.099\\
111.834	-0.108\\
111.964	-0.124\\
112.144	-0.128\\
112.364	-0.103\\
112.515	-0.075\\
112.594	-0.072\\
112.704	-0.075\\
112.824	-0.096\\
112.914	-0.148\\
113.144	-0.155\\
113.464	-0.119\\
113.584	-0.076\\
113.815	-0.079\\
113.904	-0.069\\
114.124	-0.058\\
114.244	-0.034\\
114.344	-0.038\\
114.454	-0.058\\
114.674	-0.059\\
115.046	-0.037\\
115.225	-0.015\\
115.325	-0.016\\
115.434	-0.016\\
115.607	-0.013\\
115.674	-0.011\\
115.764	-0.013\\
115.874	-0.014\\
115.984	-0.018\\
116.094	-0.020\\
116.314	-0.010\\
116.435	0.005\\
116.549	0.012\\
116.644	0.013\\
116.744	0.014\\
116.864	0.014\\
116.985	0.013\\
117.086	0.013\\
117.184	0.013\\
117.294	0.014\\
117.405	0.014\\
117.544	0.012\\
117.624	0.012\\
117.734	0.014\\
117.844	0.042\\
117.954	0.083\\
118.164	0.099\\
118.384	0.076\\
118.534	0.052\\
118.625	0.046\\
118.714	0.079\\
118.824	0.110\\
119.154	0.102\\
119.374	0.066\\
119.486	0.050\\
119.604	0.047\\
119.704	0.063\\
119.804	0.070\\
120.048	0.061\\
120.134	0.044\\
120.244	0.046\\
120.364	0.043\\
120.474	0.059\\
120.574	0.065\\
120.845	0.057\\
120.924	0.039\\
121.044	0.039\\
121.134	0.051\\
121.226	0.064\\
121.454	0.069\\
121.594	0.059\\
121.780	0.056\\
121.894	0.055\\
122.044	0.057\\
122.214	0.051\\
122.334	0.039\\
122.449	0.030\\
122.544	0.024\\
122.665	0.033\\
122.764	0.041\\
123.005	0.034\\
123.204	0.017\\
123.314	0.007\\
123.534	0.006\\
123.634	0.005\\
123.747	0.005\\
123.864	0.004\\
123.964	0.002\\
124.095	0.002\\
124.184	0.002\\
124.549	0.002\\
124.734	0.001\\
124.857	0.001\\
124.944	0.002\\
125.054	-0.001\\
125.384	-0.005\\
125.546	-0.008\\
125.604	-0.012\\
125.715	-0.015\\
125.934	-0.014\\
126.064	-0.020\\
126.194	-0.039\\
126.264	-0.095\\
126.374	-0.125\\
126.704	-0.118\\
126.824	-0.086\\
126.955	-0.080\\
127.057	-0.075\\
127.144	-0.105\\
127.244	-0.127\\
127.464	-0.118\\
127.584	-0.116\\
127.694	-0.133\\
127.904	-0.119\\
128.044	-0.089\\
128.124	-0.086\\
128.234	-0.093\\
128.347	-0.119\\
128.454	-0.129\\
128.675	-0.154\\
128.774	-0.134\\
129.047	-0.135\\
129.104	-0.107\\
129.334	-0.103\\
129.434	-0.076\\
129.554	-0.099\\
129.654	-0.102\\
129.884	-0.088\\
129.984	-0.077\\
130.194	-0.078\\
130.536	-0.055\\
130.644	-0.036\\
130.754	-0.037\\
130.964	-0.033\\
131.084	-0.027\\
131.194	-0.028\\
131.324	-0.023\\
131.447	-0.015\\
131.544	-0.010\\
131.634	-0.010\\
131.734	-0.015\\
131.844	-0.017\\
132.074	-0.000\\
132.195	0.023\\
132.404	0.028\\
132.554	0.027\\
132.614	0.029\\
132.844	0.033\\
132.954	0.029\\
133.154	0.026\\
133.264	0.021\\
133.374	0.065\\
133.497	0.099\\
133.924	0.092\\
134.044	0.090\\
134.144	0.120\\
134.464	0.149\\
134.684	0.136\\
135.057	0.105\\
135.234	0.064\\
135.344	0.054\\
135.464	0.053\\
135.564	0.088\\
135.674	0.131\\
136.044	0.147\\
136.324	0.106\\
136.587	0.068\\
136.674	0.062\\
136.764	0.075\\
136.984	0.073\\
137.094	0.073\\
137.224	0.082\\
137.434	0.068\\
137.644	0.054\\
137.754	0.049\\
137.984	0.043\\
138.074	0.025\\
138.194	0.015\\
138.304	0.009\\
138.544	0.018\\
138.734	0.012\\
139.614	-0.006\\
139.854	-0.020\\
139.964	-0.024\\
140.044	-0.028\\
140.274	-0.028\\
140.384	-0.022\\
140.494	-0.030\\
140.594	-0.044\\
140.824	-0.044\\
141.065	-0.037\\
141.144	-0.056\\
141.254	-0.077\\
141.474	-0.092\\
141.584	-0.189\\
141.824	-0.225\\
142.615	-0.155\\
142.684	-0.091\\
142.784	-0.108\\
143.054	-0.141\\
143.114	-0.133\\
143.484	-0.156\\
143.554	-0.133\\
143.874	-0.142\\
143.994	-0.100\\
144.214	-0.086\\
144.334	-0.060\\
144.554	-0.063\\
144.644	-0.058\\
144.864	-0.054\\
144.974	-0.042\\
145.084	-0.100\\
145.194	-0.128\\
145.854	-0.099\\
145.954	-0.040\\
146.069	-0.040\\
146.184	-0.039\\
146.284	-0.029\\
146.394	-0.009\\
146.567	0.008\\
146.744	0.014\\
146.834	0.013\\
147.064	0.014\\
147.184	0.012\\
147.384	0.010\\
147.494	0.010\\
147.604	0.011\\
147.854	0.009\\
147.944	0.007\\
148.054	0.008\\
148.174	0.009\\
148.364	0.007\\
148.474	0.005\\
148.596	0.004\\
148.714	0.001\\
148.914	-0.000\\
149.045	-0.000\\
149.134	0.009\\
149.234	0.024\\
149.574	0.030\\
149.706	0.027\\
149.784	0.032\\
149.894	0.032\\
150.078	0.026\\
150.114	0.024\\
150.224	0.029\\
150.344	0.110\\
150.445	0.144\\
151.644	0.089\\
151.764	-0.006\\
151.984	-0.014\\
152.084	-0.013\\
152.354	-0.011\\
152.434	-0.007\\
152.544	-0.008\\
152.634	-0.008\\
152.754	-0.008\\
152.854	-0.008\\
152.974	-0.008\\
153.095	-0.008\\
153.194	-0.008\\
153.304	-0.009\\
153.404	-0.008\\
153.553	-0.008\\
153.624	-0.007\\
153.735	-0.008\\
153.844	-0.008\\
153.956	-0.008\\
154.054	-0.008\\
154.164	-0.008\\
154.274	-0.009\\
154.384	-0.010\\
154.515	-0.009\\
154.714	-0.004\\
154.844	-0.000\\
154.924	-0.000\\
155.064	-0.000\\
155.155	-0.000\\
155.255	-0.000\\
155.694	-0.000\\
155.804	-0.000\\
156.060	-0.000\\
156.234	-0.000\\
156.364	-0.000\\
156.484	-0.000\\
157.074	-0.000\\
157.114	-0.000\\
157.336	-0.000\\
157.444	-0.000\\
157.687	-0.000\\
157.774	-0.000\\
157.894	-0.000\\
157.984	-0.000\\
158.104	-0.000\\
158.324	-0.000\\
158.445	-0.000\\
158.554	-0.000\\
158.754	-0.000\\
159.204	-0.000\\
159.324	-0.000\\
159.414	-0.000\\
159.554	-0.000\\
159.634	-0.000\\
159.854	-0.000\\
160.074	-0.000\\
160.559	-0.000\\
160.614	-0.000\\
160.744	-0.000\\
160.944	-0.000\\
161.054	-0.000\\
161.294	-0.000\\
161.504	-0.000\\
161.924	-0.000\\
162.054	-0.000\\
162.154	-0.000\\
162.365	-0.000\\
162.474	-0.000\\
162.594	-0.000\\
162.714	-0.000\\
162.834	-0.000\\
162.955	-0.000\\
163.045	-0.000\\
163.134	-0.000\\
163.244	-0.000\\
163.358	-0.000\\
163.464	-0.000\\
163.577	-0.000\\
nan	-0.000\\
nan	-0.000\\
nan	-0.000\\
nan	-0.000\\
nan	-0.000\\
};
\addlegendentry{Opti-track};

\addplot [color=black,dashed,forget plot]
  table[row sep=crcr]{%
0.000	0.000\\
150.344	0.000\\
};
\end{axis}
\end{tikzpicture}%

%% file: images/pocket_velocity_estimate_optitrack_ver.tikz
%
%
\begin{tikzpicture}

\begin{axis}[%
width=0.95092\figurewidth,
height=\figureheight,
at={(0\figurewidth,0\figureheight)},
scale only axis,
unbounded coords=jump,
every outer x axis line/.append style={black},
every x tick label/.append style={font=\color{black}},
xmin=76.384,
xmax=150.344,
xlabel={time [s]},
every outer y axis line/.append style={black},
every y tick label/.append style={font=\color{black}},
ymin=-0.500,
ymax=0.500,
ylabel={velocity [m/s]},
axis x line*=bottom,
axis y line*=left
]
\addplot [color=blue,solid,forget plot]
  table[row sep=crcr]{%
13.857	0.000\\
14.345	0.000\\
14.844	0.000\\
15.834	0.000\\
16.834	0.000\\
17.334	0.000\\
17.854	0.000\\
18.354	0.000\\
18.844	0.000\\
19.354	0.000\\
19.835	0.000\\
20.854	0.000\\
21.344	0.000\\
21.874	0.000\\
22.854	0.000\\
23.344	0.000\\
24.344	0.000\\
24.844	0.000\\
25.334	0.000\\
25.834	0.000\\
26.334	0.000\\
26.864	0.000\\
27.334	0.000\\
27.844	0.000\\
28.355	0.000\\
29.345	0.000\\
29.844	0.000\\
30.354	0.000\\
30.834	0.000\\
31.834	0.000\\
32.358	0.000\\
32.845	0.000\\
33.354	0.026\\
33.844	0.025\\
34.335	0.019\\
34.844	0.016\\
35.345	0.058\\
35.844	0.019\\
36.334	0.016\\
36.854	0.003\\
37.344	0.015\\
38.335	-0.109\\
38.844	-0.087\\
40.346	-0.082\\
40.834	-0.067\\
41.357	-0.073\\
42.344	0.015\\
42.844	0.011\\
43.344	0.011\\
43.844	0.011\\
44.345	-0.003\\
44.844	-0.014\\
45.365	-0.028\\
46.344	-0.030\\
46.844	-0.020\\
47.354	-0.015\\
47.844	-0.013\\
48.356	0.006\\
48.844	0.015\\
49.344	0.022\\
49.864	0.028\\
50.365	0.031\\
52.344	0.021\\
54.354	0.022\\
54.835	0.020\\
55.345	0.031\\
55.844	0.039\\
56.244	0.053\\
56.354	0.058\\
56.464	0.063\\
56.584	0.063\\
56.684	0.066\\
56.784	0.082\\
56.894	0.091\\
57.019	0.081\\
57.125	0.069\\
57.224	0.059\\
57.574	0.032\\
57.664	0.011\\
57.795	0.007\\
57.895	0.015\\
58.014	0.029\\
58.104	0.026\\
58.204	0.031\\
58.435	0.049\\
58.574	0.075\\
58.764	0.099\\
59.024	0.130\\
59.194	0.151\\
59.314	0.145\\
59.427	0.116\\
59.514	0.080\\
59.754	0.059\\
59.855	0.041\\
60.294	0.036\\
60.404	0.038\\
60.534	0.035\\
60.614	0.017\\
60.834	-0.002\\
61.054	-0.033\\
61.165	-0.059\\
61.516	-0.082\\
61.705	-0.098\\
61.954	-0.111\\
62.144	-0.115\\
62.255	-0.118\\
62.385	-0.119\\
62.524	-0.119\\
62.585	-0.122\\
62.695	-0.116\\
62.815	-0.108\\
62.914	-0.107\\
63.065	-0.095\\
63.807	-0.073\\
63.924	-0.052\\
64.024	-0.030\\
64.114	0.016\\
64.224	0.055\\
64.444	0.084\\
64.574	0.100\\
64.784	0.113\\
65.034	0.106\\
65.104	0.095\\
65.204	0.081\\
65.345	0.074\\
65.434	0.061\\
65.534	0.050\\
65.655	0.036\\
65.774	0.028\\
66.024	0.027\\
66.194	0.032\\
66.305	0.040\\
66.414	0.048\\
66.544	0.059\\
66.635	0.070\\
66.764	0.078\\
66.964	0.080\\
67.064	0.080\\
67.194	0.072\\
67.404	0.056\\
67.514	0.039\\
67.625	0.025\\
67.724	0.015\\
67.834	0.006\\
67.944	-0.003\\
68.055	-0.020\\
68.174	-0.034\\
68.424	-0.045\\
68.524	-0.054\\
68.614	-0.053\\
68.734	-0.045\\
68.934	-0.043\\
69.096	-0.044\\
69.167	-0.048\\
69.265	-0.071\\
69.374	-0.091\\
69.714	-0.100\\
69.804	-0.107\\
69.925	-0.106\\
70.024	-0.096\\
70.134	-0.085\\
70.244	-0.088\\
70.365	-0.106\\
70.464	-0.078\\
70.677	-0.069\\
70.895	-0.071\\
71.024	-0.078\\
71.124	-0.074\\
71.234	-0.100\\
71.444	-0.117\\
71.664	-0.142\\
71.774	-0.134\\
72.024	-0.141\\
72.214	-0.155\\
72.374	-0.155\\
72.435	-0.150\\
73.428	-0.153\\
73.549	-0.138\\
73.677	-0.133\\
73.854	-0.136\\
73.974	-0.120\\
74.074	-0.112\\
74.184	-0.104\\
74.305	-0.096\\
74.414	-0.079\\
74.538	-0.070\\
74.624	-0.068\\
74.734	-0.072\\
74.834	-0.077\\
74.965	-0.081\\
75.065	-0.080\\
75.174	-0.078\\
75.285	-0.069\\
75.524	-0.059\\
75.615	-0.058\\
75.714	-0.063\\
75.824	-0.051\\
76.044	-0.035\\
76.264	-0.020\\
76.384	0.004\\
76.524	0.029\\
76.584	0.037\\
76.815	0.040\\
77.034	0.047\\
77.134	0.053\\
77.245	0.064\\
77.464	0.071\\
77.685	0.079\\
77.804	0.080\\
77.924	0.081\\
78.045	0.078\\
78.356	0.081\\
78.455	0.081\\
78.564	0.078\\
78.664	0.079\\
78.794	0.082\\
78.920	0.086\\
79.036	0.078\\
79.114	0.083\\
79.224	0.084\\
79.434	0.070\\
79.554	0.068\\
79.664	0.084\\
79.864	0.091\\
80.024	0.096\\
80.104	0.114\\
80.324	0.120\\
80.424	0.122\\
80.634	0.123\\
80.754	0.120\\
80.964	0.091\\
81.074	0.068\\
81.314	0.062\\
81.404	0.061\\
81.625	0.064\\
81.744	0.100\\
81.867	0.119\\
81.955	0.129\\
82.174	0.133\\
82.395	0.133\\
82.544	0.112\\
82.614	0.099\\
82.944	0.072\\
83.050	0.044\\
83.264	0.012\\
83.537	-0.017\\
83.715	-0.055\\
83.814	-0.081\\
84.064	-0.108\\
84.164	-0.125\\
84.254	-0.125\\
84.364	-0.130\\
84.474	-0.146\\
84.684	-0.152\\
84.824	-0.152\\
84.924	-0.172\\
85.034	-0.151\\
85.124	-0.106\\
85.234	-0.073\\
85.390	-0.054\\
85.574	-0.037\\
85.694	-0.055\\
85.894	-0.095\\
86.034	-0.112\\
86.224	-0.111\\
86.334	-0.079\\
86.444	-0.027\\
86.654	0.023\\
86.874	0.035\\
87.045	0.036\\
87.094	-0.014\\
87.204	-0.094\\
87.447	-0.123\\
87.644	-0.150\\
87.754	-0.172\\
87.864	-0.147\\
87.984	-0.122\\
88.184	-0.136\\
88.414	-0.142\\
88.534	-0.158\\
88.734	-0.132\\
88.844	-0.118\\
89.084	-0.106\\
89.174	-0.091\\
89.335	-0.091\\
89.424	-0.133\\
90.184	-0.144\\
90.274	-0.139\\
90.384	-0.122\\
91.264	-0.085\\
91.505	-0.059\\
91.614	-0.032\\
91.925	-0.024\\
92.047	-0.021\\
92.134	-0.020\\
92.234	-0.020\\
92.354	-0.015\\
92.455	-0.009\\
92.674	-0.002\\
92.926	0.006\\
93.036	0.017\\
93.114	0.023\\
93.242	0.027\\
93.464	0.034\\
93.655	0.053\\
93.774	0.072\\
94.034	0.093\\
94.095	0.114\\
94.345	0.129\\
94.424	0.133\\
94.574	0.119\\
94.674	0.111\\
94.864	0.114\\
95.084	0.120\\
95.184	0.106\\
95.317	0.117\\
95.435	0.109\\
95.534	0.102\\
95.754	0.098\\
95.844	0.115\\
95.955	0.126\\
96.064	0.137\\
96.394	0.138\\
96.547	0.143\\
96.614	0.146\\
96.834	0.130\\
96.945	0.106\\
97.164	0.093\\
97.274	0.072\\
97.394	0.050\\
97.534	0.046\\
97.594	0.064\\
97.714	0.064\\
97.834	0.057\\
98.044	0.057\\
98.146	0.055\\
98.274	0.050\\
98.364	0.049\\
98.474	0.056\\
98.584	0.047\\
98.705	0.032\\
99.034	0.015\\
99.234	-0.004\\
99.454	-0.025\\
99.694	-0.038\\
99.794	-0.046\\
99.944	-0.061\\
100.065	-0.080\\
100.114	-0.095\\
100.224	-0.107\\
100.334	-0.125\\
100.564	-0.114\\
100.704	-0.091\\
100.784	-0.087\\
100.894	-0.089\\
101.034	-0.082\\
101.094	-0.091\\
101.204	-0.109\\
101.324	-0.095\\
101.426	-0.071\\
101.549	-0.026\\
101.644	0.016\\
101.874	0.034\\
101.976	0.033\\
102.094	0.033\\
102.534	0.019\\
102.634	-0.009\\
102.734	-0.030\\
102.954	-0.077\\
103.074	-0.130\\
103.174	-0.150\\
103.524	-0.182\\
103.614	-0.203\\
103.724	-0.219\\
103.944	-0.188\\
104.044	-0.214\\
104.274	-0.220\\
104.524	-0.195\\
104.614	-0.171\\
104.824	-0.223\\
104.935	-0.273\\
105.054	-0.334\\
105.167	-0.355\\
105.384	-0.356\\
105.474	-0.323\\
105.714	-0.268\\
105.816	-0.190\\
105.944	-0.173\\
106.047	-0.158\\
106.144	-0.150\\
106.254	-0.140\\
106.354	-0.134\\
106.464	-0.126\\
106.574	-0.095\\
106.686	-0.070\\
106.785	-0.054\\
106.894	-0.043\\
107.124	-0.033\\
107.224	-0.035\\
107.364	-0.027\\
107.554	-0.021\\
107.774	-0.011\\
107.894	0.001\\
108.094	0.008\\
108.204	0.013\\
108.314	0.020\\
108.434	0.025\\
108.534	0.027\\
108.644	0.028\\
108.754	0.029\\
108.864	0.031\\
108.986	0.035\\
109.084	0.042\\
109.314	0.049\\
109.404	0.057\\
109.534	0.062\\
109.747	0.064\\
109.864	0.061\\
109.954	0.060\\
110.074	0.063\\
110.175	0.064\\
110.314	0.070\\
110.427	0.077\\
110.546	0.084\\
110.614	0.087\\
110.724	0.092\\
110.854	0.089\\
110.944	0.075\\
111.064	0.073\\
111.174	0.062\\
111.284	0.055\\
111.536	0.057\\
111.604	0.067\\
111.714	0.059\\
111.834	0.069\\
111.964	0.072\\
112.144	0.068\\
112.364	0.066\\
112.515	0.065\\
112.594	0.060\\
112.704	0.063\\
112.824	0.067\\
112.914	0.065\\
113.144	0.076\\
113.464	0.083\\
113.584	0.085\\
113.815	0.087\\
113.904	0.083\\
114.124	0.072\\
114.244	0.060\\
114.344	0.049\\
114.454	0.034\\
114.674	0.036\\
115.046	0.033\\
115.225	0.026\\
115.325	0.020\\
115.434	0.016\\
115.607	-0.001\\
115.674	-0.010\\
115.764	-0.030\\
115.874	-0.044\\
115.984	-0.054\\
116.094	-0.056\\
116.314	-0.062\\
116.435	-0.053\\
116.549	-0.051\\
116.644	-0.054\\
116.744	-0.055\\
116.864	-0.055\\
116.985	-0.055\\
117.086	-0.056\\
117.184	-0.056\\
117.294	-0.057\\
117.405	-0.058\\
117.544	-0.059\\
117.624	-0.063\\
117.734	-0.070\\
117.844	-0.086\\
117.954	-0.089\\
118.164	-0.109\\
118.384	-0.119\\
118.534	-0.153\\
118.625	-0.172\\
118.714	-0.215\\
118.824	-0.246\\
119.154	-0.209\\
119.374	-0.200\\
119.486	-0.198\\
119.604	-0.182\\
119.704	-0.149\\
119.804	-0.170\\
120.048	-0.145\\
120.134	-0.142\\
120.244	-0.129\\
120.364	-0.143\\
120.474	-0.204\\
120.574	-0.232\\
120.845	-0.254\\
120.924	-0.262\\
121.044	-0.192\\
121.134	-0.116\\
121.226	-0.031\\
121.454	-0.020\\
121.594	-0.054\\
121.780	-0.156\\
121.894	-0.196\\
122.044	-0.277\\
122.214	-0.281\\
122.334	-0.266\\
122.449	-0.242\\
122.544	-0.253\\
122.665	-0.244\\
122.764	-0.213\\
123.005	-0.182\\
123.204	-0.151\\
123.314	-0.121\\
123.534	-0.094\\
123.634	-0.072\\
123.747	-0.049\\
123.864	-0.027\\
123.964	-0.005\\
124.095	0.030\\
124.184	0.063\\
124.549	0.084\\
124.734	0.092\\
124.857	0.087\\
124.944	0.051\\
125.054	0.012\\
125.384	-0.017\\
125.546	-0.035\\
125.604	-0.036\\
125.715	-0.024\\
125.934	-0.007\\
126.064	0.018\\
126.194	0.049\\
126.264	0.067\\
126.374	0.081\\
126.704	0.090\\
126.824	0.088\\
126.955	0.079\\
127.057	0.083\\
127.144	0.098\\
127.244	0.116\\
127.464	0.125\\
127.584	0.143\\
127.694	0.148\\
127.904	0.140\\
128.044	0.138\\
128.124	0.136\\
128.234	0.120\\
128.347	0.107\\
128.454	0.093\\
128.675	0.077\\
128.774	0.061\\
129.047	0.054\\
129.104	0.050\\
129.334	0.050\\
129.434	0.040\\
129.554	0.058\\
129.654	0.072\\
129.884	0.080\\
129.984	0.087\\
130.194	0.093\\
130.536	0.080\\
130.644	0.064\\
130.754	0.051\\
130.964	0.041\\
131.084	0.036\\
131.194	0.027\\
131.324	0.017\\
131.447	0.006\\
131.544	-0.008\\
131.634	-0.022\\
131.734	-0.039\\
131.844	-0.050\\
132.074	-0.060\\
132.195	-0.068\\
132.404	-0.078\\
132.554	-0.079\\
132.614	-0.083\\
132.844	-0.105\\
132.954	-0.111\\
133.154	-0.129\\
133.264	-0.148\\
133.374	-0.129\\
133.497	-0.110\\
133.924	-0.087\\
134.044	-0.081\\
134.144	-0.074\\
134.464	-0.094\\
134.684	-0.079\\
135.057	-0.101\\
135.234	-0.070\\
135.344	-0.044\\
135.464	-0.025\\
135.564	-0.049\\
135.674	-0.032\\
136.044	-0.045\\
136.324	-0.068\\
136.587	-0.089\\
136.674	-0.107\\
136.764	-0.120\\
136.984	-0.099\\
137.094	-0.078\\
137.224	-0.044\\
137.434	-0.048\\
137.644	-0.070\\
137.754	-0.114\\
137.984	-0.135\\
138.074	-0.174\\
138.194	-0.145\\
138.304	-0.128\\
138.544	-0.082\\
138.734	-0.041\\
139.614	-0.000\\
139.854	0.035\\
139.964	0.074\\
140.044	0.084\\
140.274	0.100\\
140.384	0.111\\
140.494	0.120\\
140.594	0.133\\
140.824	0.141\\
141.065	0.155\\
141.144	0.178\\
141.254	0.203\\
141.474	0.222\\
141.584	0.236\\
141.824	0.213\\
142.615	0.186\\
142.684	0.168\\
142.784	0.144\\
143.054	0.150\\
143.114	0.179\\
143.484	0.184\\
143.554	0.189\\
143.874	0.209\\
143.994	0.206\\
144.214	0.175\\
144.334	0.155\\
144.554	0.152\\
144.644	0.138\\
144.864	0.120\\
144.974	0.102\\
145.084	0.097\\
145.194	0.063\\
145.854	0.030\\
145.954	0.002\\
146.069	0.000\\
146.184	-0.009\\
146.284	-0.016\\
146.394	-0.016\\
146.567	-0.016\\
146.744	-0.013\\
146.834	-0.004\\
147.064	0.005\\
147.184	0.007\\
147.384	0.007\\
147.494	-0.005\\
147.604	-0.017\\
147.854	-0.031\\
147.944	-0.042\\
148.054	-0.055\\
148.174	-0.058\\
148.364	-0.061\\
148.474	-0.061\\
148.596	-0.064\\
148.714	-0.063\\
148.914	-0.065\\
149.045	-0.058\\
149.134	-0.055\\
149.234	-0.076\\
149.574	-0.069\\
149.706	-0.054\\
149.784	-0.046\\
149.894	-0.048\\
150.078	-0.008\\
150.114	0.003\\
150.224	-0.012\\
150.344	-0.011\\
150.445	0.005\\
151.644	0.006\\
151.764	0.011\\
151.984	0.026\\
152.084	0.038\\
152.354	0.040\\
152.434	0.035\\
152.544	-0.010\\
152.634	-0.019\\
152.754	-0.043\\
152.854	-0.054\\
152.974	-0.063\\
153.095	-0.035\\
153.194	-0.029\\
153.304	-0.023\\
153.404	-0.020\\
153.553	-0.012\\
153.624	-0.006\\
153.735	-0.003\\
153.844	0.000\\
153.956	0.002\\
154.054	0.001\\
154.164	0.001\\
154.274	0.001\\
154.384	0.001\\
154.515	0.000\\
154.714	0.000\\
154.844	0.000\\
154.924	0.000\\
155.064	0.000\\
155.155	-0.003\\
155.255	-0.003\\
155.694	-0.003\\
155.804	-0.003\\
156.060	-0.003\\
156.234	0.000\\
156.364	0.000\\
156.484	0.000\\
157.074	0.002\\
157.114	0.003\\
157.336	0.003\\
157.444	0.010\\
157.687	0.017\\
157.774	0.016\\
157.894	0.020\\
157.984	0.018\\
158.104	0.010\\
158.324	0.003\\
158.445	0.002\\
158.554	0.003\\
158.754	0.006\\
159.204	0.007\\
159.324	0.007\\
159.414	0.007\\
159.554	0.001\\
159.634	0.000\\
159.854	0.000\\
160.074	0.000\\
160.559	0.000\\
160.614	0.000\\
160.744	0.000\\
160.944	0.000\\
161.054	0.000\\
161.294	0.000\\
161.504	0.000\\
161.924	0.000\\
162.054	0.000\\
162.154	0.000\\
162.365	0.000\\
162.474	0.000\\
162.594	0.002\\
162.714	0.002\\
162.834	0.002\\
162.955	0.002\\
163.045	0.002\\
163.134	0.000\\
163.244	0.000\\
163.358	0.000\\
163.464	0.000\\
163.577	0.000\\
163.699	0.000\\
163.894	0.000\\
164.067	0.000\\
164.124	0.000\\
164.224	0.000\\
};
\addplot [color=red,solid,forget plot]
  table[row sep=crcr]{%
nan	-0.000\\
nan	-0.000\\
nan	-0.000\\
nan	-0.000\\
nan	-0.000\\
17.334	-0.000\\
17.854	-0.000\\
18.354	-0.000\\
18.844	-0.000\\
19.354	-0.000\\
19.835	-0.000\\
20.854	-0.000\\
21.344	-0.000\\
21.874	-0.000\\
22.854	-0.000\\
23.344	-0.000\\
24.344	-0.000\\
24.844	-0.000\\
25.334	-0.000\\
25.834	-0.000\\
26.334	-0.000\\
26.864	-0.000\\
27.334	-0.000\\
27.844	-0.000\\
28.355	-0.000\\
29.345	-0.000\\
29.844	-0.000\\
30.354	-0.000\\
30.834	-0.000\\
31.834	-0.000\\
32.358	-0.000\\
32.845	-0.002\\
33.354	-0.011\\
33.844	-0.027\\
34.335	-0.044\\
34.844	-0.054\\
35.345	-0.043\\
35.844	-0.020\\
36.334	-0.004\\
36.854	-0.005\\
37.344	-0.009\\
38.335	-0.041\\
38.844	-0.078\\
40.346	-0.075\\
40.834	0.049\\
41.357	0.149\\
42.344	0.148\\
42.844	0.078\\
43.344	0.058\\
43.844	0.044\\
44.345	0.005\\
44.844	-0.061\\
45.365	-0.098\\
46.344	-0.095\\
46.844	-0.084\\
47.354	-0.069\\
47.844	-0.029\\
48.356	0.006\\
48.844	0.011\\
49.344	0.011\\
49.864	0.020\\
50.365	0.037\\
52.344	0.040\\
54.354	0.028\\
54.835	0.024\\
55.345	0.032\\
55.844	0.027\\
56.244	0.012\\
56.354	0.003\\
56.464	0.002\\
56.584	0.002\\
56.684	0.002\\
56.784	0.002\\
56.894	0.002\\
57.019	0.003\\
57.125	0.008\\
57.224	0.011\\
57.574	0.010\\
57.664	0.006\\
57.795	0.006\\
57.895	0.006\\
58.014	0.006\\
58.104	0.007\\
58.204	0.009\\
58.435	0.010\\
58.574	0.011\\
58.764	0.012\\
59.024	0.010\\
59.194	0.007\\
59.314	0.006\\
59.427	0.009\\
59.514	0.010\\
59.754	0.010\\
59.855	0.004\\
60.294	-0.003\\
60.404	-0.008\\
60.534	-0.010\\
60.614	-0.013\\
60.834	-0.014\\
61.054	-0.015\\
61.165	-0.017\\
61.516	-0.018\\
61.705	-0.016\\
61.954	-0.013\\
62.144	-0.010\\
62.255	-0.007\\
62.385	-0.003\\
62.524	-0.000\\
62.585	0.001\\
62.695	0.001\\
62.815	0.001\\
62.914	0.002\\
63.065	0.003\\
63.807	0.002\\
63.924	0.001\\
64.024	0.001\\
64.114	0.001\\
64.224	0.001\\
64.444	0.005\\
64.574	0.016\\
64.784	0.021\\
65.034	0.017\\
65.104	0.014\\
65.204	0.015\\
65.345	0.012\\
65.434	0.009\\
65.534	0.009\\
65.655	0.012\\
65.774	0.015\\
66.024	0.014\\
66.194	0.010\\
66.305	0.009\\
66.414	0.009\\
66.544	0.009\\
66.635	0.011\\
66.764	0.012\\
66.964	0.011\\
67.064	0.012\\
67.194	0.013\\
67.404	0.011\\
67.514	0.007\\
67.625	0.003\\
67.724	-0.002\\
67.834	-0.004\\
67.944	-0.004\\
68.055	-0.005\\
68.174	-0.010\\
68.424	-0.015\\
68.524	-0.018\\
68.614	-0.024\\
68.734	-0.029\\
68.934	-0.025\\
69.096	-0.024\\
69.167	-0.033\\
69.265	-0.069\\
69.374	-0.081\\
69.714	-0.068\\
69.804	-0.043\\
69.925	-0.044\\
70.024	-0.043\\
70.134	-0.046\\
70.244	-0.045\\
70.365	-0.056\\
70.464	-0.074\\
70.677	-0.078\\
70.895	-0.060\\
71.024	-0.046\\
71.124	-0.056\\
71.234	-0.080\\
71.444	-0.083\\
71.664	-0.084\\
71.774	-0.088\\
72.024	-0.084\\
72.214	-0.053\\
72.374	-0.124\\
72.435	-0.166\\
73.428	-0.130\\
73.549	-0.041\\
73.677	-0.033\\
73.854	-0.024\\
73.974	-0.019\\
74.074	-0.020\\
74.184	-0.020\\
74.305	-0.021\\
74.414	-0.019\\
74.538	-0.018\\
74.624	-0.018\\
74.734	-0.020\\
74.834	-0.020\\
74.965	-0.019\\
75.065	-0.016\\
75.174	-0.015\\
75.285	-0.014\\
75.524	-0.011\\
75.615	-0.008\\
75.714	-0.006\\
75.824	-0.000\\
76.044	0.006\\
76.264	0.007\\
76.384	0.005\\
76.524	0.013\\
76.584	0.031\\
76.815	0.039\\
77.034	0.033\\
77.134	0.033\\
77.245	0.046\\
77.464	0.047\\
77.685	0.037\\
77.804	0.030\\
77.924	0.043\\
78.045	0.048\\
78.356	0.039\\
78.455	0.024\\
78.564	0.026\\
78.664	0.027\\
78.794	0.041\\
78.920	0.052\\
79.036	0.059\\
79.114	0.077\\
79.224	0.091\\
79.434	0.082\\
79.554	0.077\\
79.664	0.089\\
79.864	0.080\\
80.024	0.076\\
80.104	0.077\\
80.324	0.080\\
80.424	0.067\\
80.634	0.072\\
80.754	0.067\\
80.964	0.076\\
81.074	0.070\\
81.314	0.076\\
81.404	0.070\\
81.625	0.069\\
81.744	0.051\\
81.867	0.041\\
81.955	0.032\\
82.174	0.029\\
82.395	0.022\\
82.544	0.025\\
82.614	0.029\\
82.944	0.032\\
83.050	0.019\\
83.264	-0.006\\
83.537	-0.030\\
83.715	-0.045\\
83.814	-0.045\\
84.064	-0.038\\
84.164	-0.027\\
84.254	-0.029\\
84.364	-0.038\\
84.474	-0.044\\
84.684	-0.041\\
84.824	-0.032\\
84.924	-0.028\\
85.034	-0.028\\
85.124	-0.033\\
85.234	-0.042\\
85.390	-0.044\\
85.574	-0.047\\
85.694	-0.049\\
85.894	-0.054\\
86.034	-0.048\\
86.224	-0.042\\
86.334	-0.048\\
86.444	-0.070\\
86.654	-0.080\\
86.874	-0.060\\
87.045	-0.041\\
87.094	-0.053\\
87.204	-0.078\\
87.447	-0.076\\
87.644	-0.056\\
87.754	-0.050\\
87.864	-0.071\\
87.984	-0.103\\
88.184	-0.109\\
88.414	-0.100\\
88.534	-0.085\\
88.734	-0.093\\
88.844	-0.083\\
89.084	-0.080\\
89.174	-0.062\\
89.335	-0.089\\
89.424	-0.086\\
90.184	-0.060\\
90.274	-0.066\\
90.384	-0.097\\
91.264	-0.078\\
91.505	-0.023\\
91.614	-0.003\\
91.925	0.005\\
92.047	0.006\\
92.134	0.006\\
92.234	0.006\\
92.354	0.008\\
92.455	0.011\\
92.674	0.012\\
92.926	0.008\\
93.036	0.006\\
93.114	0.009\\
93.242	0.012\\
93.464	0.012\\
93.655	0.019\\
93.774	0.025\\
94.034	0.040\\
94.095	0.041\\
94.345	0.041\\
94.424	0.031\\
94.574	0.051\\
94.674	0.081\\
94.864	0.089\\
95.084	0.071\\
95.184	0.057\\
95.317	0.055\\
95.435	0.065\\
95.534	0.068\\
95.754	0.060\\
95.844	0.046\\
95.955	0.086\\
96.064	0.115\\
96.394	0.100\\
96.547	0.077\\
96.614	0.082\\
96.834	0.088\\
96.945	0.070\\
97.164	0.061\\
97.274	0.052\\
97.394	0.048\\
97.534	0.040\\
97.594	0.028\\
97.714	0.024\\
97.834	0.018\\
98.044	0.017\\
98.146	0.014\\
98.274	0.014\\
98.364	0.014\\
98.474	0.016\\
98.584	0.026\\
98.705	0.030\\
99.034	0.015\\
99.234	-0.011\\
99.454	-0.019\\
99.694	-0.015\\
99.794	-0.018\\
99.944	-0.021\\
100.065	-0.022\\
100.114	-0.024\\
100.224	-0.037\\
100.334	-0.044\\
100.564	-0.039\\
100.704	-0.029\\
100.784	-0.031\\
100.894	-0.031\\
101.034	-0.028\\
101.094	-0.028\\
101.204	-0.032\\
101.324	-0.032\\
101.426	-0.031\\
101.549	-0.041\\
101.644	-0.045\\
101.874	-0.041\\
101.976	-0.080\\
102.094	-0.109\\
102.534	-0.093\\
102.634	-0.067\\
102.734	-0.079\\
102.954	-0.073\\
103.074	-0.092\\
103.174	-0.106\\
103.524	-0.089\\
103.614	-0.071\\
103.724	-0.081\\
103.944	-0.091\\
104.044	-0.097\\
104.274	-0.086\\
104.524	-0.053\\
104.614	-0.035\\
104.824	-0.034\\
104.935	-0.028\\
105.054	-0.035\\
105.167	-0.037\\
105.384	-0.042\\
105.474	-0.039\\
105.714	-0.037\\
105.816	-0.028\\
105.944	-0.026\\
106.047	-0.025\\
106.144	-0.025\\
106.254	-0.025\\
106.354	-0.026\\
106.464	-0.026\\
106.574	-0.025\\
106.686	-0.025\\
106.785	-0.033\\
106.894	-0.036\\
107.124	-0.034\\
107.224	-0.032\\
107.364	-0.040\\
107.554	-0.040\\
107.774	-0.037\\
107.894	-0.029\\
108.094	-0.015\\
108.204	0.003\\
108.314	0.009\\
108.434	0.009\\
108.534	0.009\\
108.644	0.009\\
108.754	0.011\\
108.864	0.014\\
108.986	0.023\\
109.084	0.026\\
109.314	0.024\\
109.404	0.027\\
109.534	0.037\\
109.747	0.037\\
109.864	0.033\\
109.954	0.033\\
110.074	0.036\\
110.175	0.037\\
110.314	0.037\\
110.427	0.031\\
110.546	0.029\\
110.614	0.031\\
110.724	0.033\\
110.854	0.031\\
110.944	0.030\\
111.064	0.039\\
111.174	0.070\\
111.284	0.081\\
111.536	0.067\\
111.604	0.049\\
111.714	0.057\\
111.834	0.063\\
111.964	0.074\\
112.144	0.076\\
112.364	0.059\\
112.515	0.042\\
112.594	0.039\\
112.704	0.041\\
112.824	0.053\\
112.914	0.083\\
113.144	0.093\\
113.464	0.085\\
113.584	0.066\\
113.815	0.071\\
113.904	0.064\\
114.124	0.050\\
114.244	0.024\\
114.344	0.023\\
114.454	0.033\\
114.674	0.026\\
115.046	0.007\\
115.225	-0.006\\
115.325	-0.006\\
115.434	-0.006\\
115.607	-0.005\\
115.674	-0.004\\
115.764	-0.005\\
115.874	-0.005\\
115.984	-0.006\\
116.094	-0.007\\
116.314	-0.010\\
116.435	-0.013\\
116.549	-0.016\\
116.644	-0.016\\
116.744	-0.018\\
116.864	-0.018\\
116.985	-0.016\\
117.086	-0.016\\
117.184	-0.017\\
117.294	-0.018\\
117.405	-0.018\\
117.544	-0.016\\
117.624	-0.016\\
117.734	-0.018\\
117.844	-0.044\\
117.954	-0.083\\
118.164	-0.100\\
118.384	-0.079\\
118.534	-0.056\\
118.625	-0.051\\
118.714	-0.090\\
118.824	-0.128\\
119.154	-0.121\\
119.374	-0.080\\
119.486	-0.062\\
119.604	-0.059\\
119.704	-0.079\\
119.804	-0.089\\
120.048	-0.078\\
120.134	-0.059\\
120.244	-0.063\\
120.364	-0.060\\
120.474	-0.083\\
120.574	-0.091\\
120.845	-0.080\\
120.924	-0.055\\
121.044	-0.054\\
121.134	-0.071\\
121.226	-0.090\\
121.454	-0.098\\
121.594	-0.084\\
121.780	-0.080\\
121.894	-0.080\\
122.044	-0.083\\
122.214	-0.073\\
122.334	-0.052\\
122.449	-0.037\\
122.544	-0.027\\
122.665	-0.037\\
122.764	-0.045\\
123.005	-0.038\\
123.204	-0.018\\
123.314	-0.008\\
123.534	-0.007\\
123.634	-0.006\\
123.747	-0.005\\
123.864	-0.000\\
123.964	0.005\\
124.095	0.011\\
124.184	0.015\\
124.549	0.014\\
124.734	0.008\\
124.857	0.007\\
124.944	0.010\\
125.054	0.015\\
125.384	0.015\\
125.546	0.012\\
125.604	0.016\\
125.715	0.020\\
125.934	0.020\\
126.064	0.027\\
126.194	0.049\\
126.264	0.116\\
126.374	0.146\\
126.704	0.122\\
126.824	0.070\\
126.955	0.057\\
127.057	0.047\\
127.144	0.061\\
127.244	0.072\\
127.464	0.066\\
127.584	0.062\\
127.694	0.069\\
127.904	0.060\\
128.044	0.043\\
128.124	0.041\\
128.234	0.044\\
128.347	0.057\\
128.454	0.062\\
128.675	0.075\\
128.774	0.067\\
129.047	0.076\\
129.104	0.072\\
129.334	0.075\\
129.434	0.060\\
129.554	0.080\\
129.654	0.083\\
129.884	0.067\\
129.984	0.054\\
130.194	0.055\\
130.536	0.039\\
130.644	0.020\\
130.754	0.015\\
130.964	0.012\\
131.084	0.011\\
131.194	0.011\\
131.324	0.004\\
131.447	-0.007\\
131.544	-0.015\\
131.634	-0.015\\
131.734	-0.022\\
131.844	-0.025\\
132.074	-0.037\\
132.195	-0.046\\
132.404	-0.044\\
132.554	-0.042\\
132.614	-0.046\\
132.844	-0.053\\
132.954	-0.047\\
133.154	-0.043\\
133.264	-0.036\\
133.374	-0.092\\
133.497	-0.134\\
133.924	-0.124\\
134.044	-0.123\\
134.144	-0.165\\
134.464	-0.207\\
134.684	-0.191\\
135.057	-0.151\\
135.234	-0.094\\
135.344	-0.079\\
135.464	-0.077\\
135.564	-0.129\\
135.674	-0.191\\
136.044	-0.213\\
136.324	-0.154\\
136.587	-0.098\\
136.674	-0.089\\
136.764	-0.107\\
136.984	-0.104\\
137.094	-0.104\\
137.224	-0.112\\
137.434	-0.088\\
137.644	-0.065\\
137.754	-0.058\\
137.984	-0.051\\
138.074	-0.028\\
138.194	-0.016\\
138.304	-0.007\\
138.544	-0.014\\
138.734	0.000\\
139.614	0.024\\
139.854	0.038\\
139.964	0.043\\
140.044	0.050\\
140.274	0.046\\
140.384	0.033\\
140.494	0.042\\
140.594	0.058\\
140.824	0.055\\
141.065	0.040\\
141.144	0.049\\
141.254	0.063\\
141.474	0.073\\
141.584	0.143\\
141.824	0.168\\
142.615	0.115\\
142.684	0.073\\
142.784	0.087\\
143.054	0.112\\
143.114	0.103\\
143.484	0.118\\
143.554	0.097\\
143.874	0.099\\
143.994	0.063\\
144.214	0.042\\
144.334	0.020\\
144.554	0.022\\
144.644	0.021\\
144.864	0.020\\
144.974	0.016\\
145.084	0.041\\
145.194	0.054\\
145.854	0.043\\
145.954	0.020\\
146.069	0.020\\
146.184	0.020\\
146.284	0.002\\
146.394	-0.034\\
146.567	-0.055\\
146.744	-0.066\\
146.834	-0.068\\
147.064	-0.075\\
147.184	-0.065\\
147.384	-0.059\\
147.494	-0.065\\
147.604	-0.071\\
147.854	-0.062\\
147.944	-0.047\\
148.054	-0.060\\
148.174	-0.064\\
148.364	-0.060\\
148.474	-0.048\\
148.596	-0.032\\
148.714	-0.011\\
148.914	-0.000\\
149.045	-0.000\\
149.134	-0.004\\
149.234	-0.010\\
149.574	-0.013\\
149.706	-0.013\\
149.784	-0.015\\
149.894	-0.015\\
150.078	-0.013\\
150.114	-0.012\\
150.224	-0.015\\
150.344	-0.057\\
150.445	-0.076\\
151.644	-0.048\\
151.764	0.003\\
151.984	0.008\\
152.084	0.007\\
152.354	0.006\\
152.434	0.003\\
152.544	0.003\\
152.634	0.003\\
152.754	0.003\\
152.854	0.003\\
152.974	0.003\\
153.095	0.003\\
153.194	0.002\\
153.304	0.003\\
153.404	0.003\\
153.553	0.002\\
153.624	0.002\\
153.735	0.002\\
153.844	0.002\\
153.956	0.002\\
154.054	0.002\\
154.164	0.002\\
154.274	0.003\\
154.384	0.003\\
154.515	0.003\\
154.714	0.001\\
154.844	-0.000\\
154.924	-0.000\\
155.064	-0.000\\
155.155	-0.000\\
155.255	-0.000\\
155.694	-0.000\\
155.804	-0.000\\
156.060	-0.000\\
156.234	-0.000\\
156.364	-0.000\\
156.484	-0.000\\
157.074	-0.000\\
157.114	-0.000\\
157.336	-0.000\\
157.444	-0.000\\
157.687	-0.000\\
157.774	-0.000\\
157.894	-0.000\\
157.984	-0.000\\
158.104	-0.000\\
158.324	-0.000\\
158.445	-0.000\\
158.554	-0.000\\
158.754	-0.000\\
159.204	-0.000\\
159.324	-0.000\\
159.414	-0.000\\
159.554	-0.000\\
159.634	-0.000\\
159.854	-0.000\\
160.074	-0.000\\
160.559	-0.000\\
160.614	-0.000\\
160.744	-0.000\\
160.944	-0.000\\
161.054	-0.000\\
161.294	-0.000\\
161.504	-0.000\\
161.924	-0.000\\
162.054	-0.000\\
162.154	-0.000\\
162.365	-0.000\\
162.474	-0.000\\
162.594	-0.000\\
162.714	-0.000\\
162.834	-0.000\\
162.955	-0.000\\
163.045	-0.000\\
163.134	-0.000\\
163.244	-0.000\\
163.358	-0.000\\
163.464	-0.000\\
163.577	-0.000\\
nan	-0.000\\
nan	-0.000\\
nan	-0.000\\
nan	-0.000\\
nan	-0.000\\
};
\addplot [color=black,dashed,forget plot]
  table[row sep=crcr]{%
0.000	0.000\\
150.344	0.000\\
};
\end{axis}
\end{tikzpicture}%

%% file: images/pocket_velocity_estimate_control_ref_hor.tikz
%
%
\begin{tikzpicture}

\begin{axis}[%
width=0.95092\figurewidth,
height=\figureheight,
at={(0\figurewidth,0\figureheight)},
scale only axis,
unbounded coords=jump,
every outer x axis line/.append style={black},
every x tick label/.append style={font=\color{black}},
xmin=160.000,
xmax=280.000,
xlabel={time [s]},
every outer y axis line/.append style={black},
every y tick label/.append style={font=\color{black}},
ymin=-0.500,
ymax=0.500,
ylabel={velocity [m/s]},
axis x line*=bottom,
axis y line*=left,
legend style={at={(0.375248,0.812597)},anchor=south west,legend cell align=left,align=left,fill=none,draw=none}
]
\addplot [color=blue,solid]
  table[row sep=crcr]{%
159.920	-0.314\\
160.030	-0.308\\
160.090	-0.306\\
160.170	-0.302\\
160.312	-0.304\\
160.390	-0.322\\
160.470	-0.328\\
160.591	-0.320\\
160.750	-0.308\\
161.062	-0.290\\
161.120	-0.272\\
161.220	-0.262\\
161.420	-0.260\\
161.540	-0.260\\
161.771	-0.260\\
161.891	-0.260\\
162.032	-0.260\\
162.100	-0.254\\
162.210	-0.248\\
162.280	-0.242\\
162.420	-0.232\\
162.541	-0.220\\
162.650	-0.214\\
162.740	-0.208\\
162.840	-0.202\\
162.970	-0.200\\
163.010	-0.198\\
163.140	-0.176\\
163.390	-0.150\\
163.600	-0.122\\
163.690	-0.092\\
163.770	-0.062\\
163.850	-0.050\\
163.990	-0.042\\
164.020	-0.036\\
164.160	-0.030\\
164.210	-0.018\\
164.340	-0.010\\
164.620	0.004\\
164.660	0.018\\
164.800	0.038\\
164.881	0.062\\
165.110	0.092\\
165.220	0.116\\
165.370	0.144\\
165.470	0.164\\
165.670	0.174\\
165.780	0.180\\
165.880	0.188\\
165.950	0.194\\
166.080	0.200\\
166.170	0.204\\
166.270	0.204\\
166.310	0.202\\
166.510	0.200\\
166.670	0.196\\
166.730	0.196\\
166.841	0.200\\
167.060	0.214\\
167.180	0.220\\
167.300	0.230\\
167.420	0.232\\
167.490	0.236\\
167.750	0.238\\
167.850	0.248\\
167.930	0.252\\
168.041	0.258\\
168.080	0.260\\
168.220	0.254\\
168.340	0.252\\
168.400	0.268\\
168.490	0.288\\
168.540	0.336\\
168.710	0.358\\
168.751	0.346\\
168.890	0.320\\
169.010	0.292\\
169.080	0.236\\
169.230	0.206\\
169.300	0.204\\
169.410	0.212\\
169.480	0.216\\
169.811	0.218\\
169.910	0.218\\
170.000	0.224\\
170.030	0.228\\
170.310	0.236\\
170.490	0.248\\
170.590	0.262\\
170.671	0.270\\
170.820	0.270\\
170.881	0.270\\
170.940	0.270\\
171.010	0.266\\
171.180	0.266\\
171.380	0.266\\
171.660	0.268\\
171.710	0.270\\
171.930	0.276\\
172.000	0.272\\
172.120	0.268\\
172.200	0.256\\
172.331	0.246\\
172.500	0.236\\
172.572	0.232\\
172.690	0.228\\
172.964	0.230\\
173.020	0.230\\
173.130	0.230\\
173.220	0.230\\
173.440	0.230\\
173.731	0.232\\
173.820	0.216\\
173.990	0.200\\
174.501	0.184\\
174.530	0.168\\
174.612	0.132\\
174.712	0.112\\
175.040	0.092\\
175.120	0.070\\
175.240	0.046\\
175.300	0.042\\
175.440	0.038\\
175.470	0.030\\
175.680	0.024\\
175.760	0.012\\
175.870	-0.012\\
176.040	-0.048\\
176.170	-0.090\\
176.290	-0.132\\
176.520	-0.166\\
176.601	-0.192\\
176.720	-0.204\\
176.870	-0.206\\
177.032	-0.208\\
177.250	-0.210\\
177.340	-0.214\\
177.420	-0.218\\
177.540	-0.224\\
177.701	-0.236\\
177.820	-0.244\\
177.850	-0.244\\
177.910	-0.244\\
178.040	-0.240\\
178.090	-0.234\\
178.220	-0.232\\
178.410	-0.234\\
178.520	-0.254\\
178.580	-0.272\\
178.660	-0.282\\
178.740	-0.290\\
178.900	-0.298\\
179.050	-0.288\\
179.170	-0.282\\
179.390	-0.280\\
179.470	-0.280\\
179.650	-0.280\\
179.720	-0.280\\
179.850	-0.280\\
179.911	-0.280\\
180.020	-0.268\\
180.120	-0.256\\
180.160	-0.244\\
180.220	-0.232\\
180.290	-0.220\\
180.360	-0.220\\
180.440	-0.220\\
180.662	-0.220\\
180.770	-0.226\\
180.840	-0.234\\
181.180	-0.244\\
181.280	-0.264\\
181.310	-0.284\\
181.550	-0.304\\
181.590	-0.320\\
181.730	-0.326\\
181.830	-0.324\\
181.920	-0.322\\
182.010	-0.316\\
182.100	-0.312\\
182.220	-0.344\\
182.301	-0.350\\
182.530	-0.352\\
182.770	-0.344\\
182.851	-0.336\\
183.020	-0.300\\
183.050	-0.276\\
183.260	-0.252\\
183.450	-0.236\\
183.550	-0.220\\
183.630	-0.210\\
183.760	-0.202\\
183.890	-0.202\\
183.960	-0.208\\
184.030	-0.214\\
184.180	-0.210\\
184.230	-0.222\\
184.381	-0.230\\
184.490	-0.238\\
184.520	-0.240\\
184.620	-0.248\\
184.790	-0.256\\
184.840	-0.266\\
184.980	-0.280\\
185.060	-0.308\\
185.390	-0.336\\
185.510	-0.340\\
185.590	-0.346\\
185.850	-0.326\\
185.990	-0.304\\
186.070	-0.266\\
186.110	-0.246\\
186.200	-0.232\\
186.420	-0.220\\
186.520	-0.202\\
186.600	-0.214\\
186.700	-0.232\\
186.770	-0.224\\
186.902	-0.214\\
186.960	-0.206\\
187.000	-0.188\\
187.090	-0.182\\
187.130	-0.202\\
187.210	-0.242\\
187.300	-0.272\\
187.500	-0.264\\
187.540	-0.234\\
187.660	-0.192\\
187.740	-0.138\\
187.840	-0.080\\
187.990	-0.054\\
188.100	-0.032\\
188.180	-0.012\\
188.260	0.000\\
188.321	0.006\\
188.361	0.012\\
188.540	0.018\\
188.620	0.026\\
188.710	0.036\\
188.790	0.044\\
188.870	0.056\\
189.030	0.072\\
189.140	0.084\\
189.180	0.110\\
189.321	0.124\\
189.360	0.128\\
189.520	0.136\\
189.620	0.166\\
189.700	0.172\\
189.861	0.176\\
190.030	0.204\\
190.211	0.208\\
190.380	0.194\\
190.460	0.196\\
190.710	0.202\\
190.900	0.192\\
191.060	0.200\\
191.181	0.210\\
191.222	0.212\\
191.370	0.218\\
191.470	0.222\\
191.590	0.232\\
191.670	0.236\\
191.880	0.238\\
191.960	0.240\\
192.090	0.242\\
192.170	0.238\\
192.280	0.234\\
192.460	0.236\\
192.509	0.252\\
192.670	0.274\\
192.810	0.288\\
193.171	0.304\\
193.310	0.316\\
193.340	0.316\\
193.440	0.322\\
193.550	0.342\\
193.660	0.360\\
193.700	0.366\\
193.759	0.368\\
193.830	0.358\\
194.030	0.340\\
194.152	0.324\\
194.300	0.320\\
194.490	0.320\\
194.540	0.320\\
194.700	0.320\\
194.800	0.320\\
194.949	0.320\\
194.990	0.312\\
195.320	0.304\\
195.479	0.296\\
195.580	0.288\\
195.660	0.272\\
195.820	0.258\\
195.940	0.246\\
196.040	0.228\\
196.130	0.200\\
196.313	0.174\\
196.450	0.154\\
196.520	0.134\\
196.590	0.118\\
196.780	0.106\\
196.820	0.098\\
196.930	0.082\\
196.960	0.064\\
197.140	0.048\\
197.232	0.038\\
197.320	0.030\\
197.350	0.026\\
197.480	0.016\\
197.589	0.002\\
197.693	-0.012\\
197.800	-0.026\\
197.969	-0.046\\
198.050	-0.074\\
198.160	-0.098\\
198.529	-0.122\\
198.620	-0.146\\
198.710	-0.160\\
198.860	-0.162\\
198.940	-0.164\\
199.030	-0.166\\
199.070	-0.168\\
199.150	-0.170\\
199.250	-0.170\\
199.330	-0.170\\
199.419	-0.170\\
199.500	-0.172\\
199.580	-0.194\\
199.701	-0.212\\
199.989	-0.238\\
200.070	-0.266\\
200.231	-0.296\\
200.270	-0.294\\
200.390	-0.290\\
200.489	-0.278\\
200.610	-0.258\\
200.650	-0.234\\
200.840	-0.222\\
200.950	-0.222\\
201.020	-0.240\\
201.110	-0.268\\
201.231	-0.296\\
201.350	-0.324\\
201.449	-0.346\\
201.499	-0.362\\
201.700	-0.376\\
201.960	-0.398\\
202.020	-0.410\\
202.109	-0.420\\
202.350	-0.410\\
202.530	-0.372\\
202.659	-0.326\\
202.889	-0.294\\
203.000	-0.274\\
203.101	-0.252\\
203.400	-0.262\\
203.500	-0.272\\
203.590	-0.272\\
203.720	-0.262\\
203.839	-0.264\\
203.929	-0.266\\
204.010	-0.268\\
204.150	-0.270\\
204.299	-0.266\\
204.380	-0.256\\
204.529	-0.238\\
204.610	-0.220\\
204.670	-0.208\\
204.790	-0.202\\
205.070	-0.200\\
205.129	-0.212\\
205.209	-0.224\\
205.330	-0.236\\
205.400	-0.248\\
205.489	-0.260\\
205.620	-0.260\\
205.660	-0.260\\
205.800	-0.260\\
205.879	-0.260\\
206.210	-0.260\\
206.289	-0.260\\
206.369	-0.260\\
206.520	-0.260\\
206.560	-0.266\\
206.709	-0.266\\
206.820	-0.266\\
206.881	-0.266\\
206.961	-0.266\\
206.990	-0.260\\
207.170	-0.260\\
207.290	-0.214\\
207.440	-0.166\\
207.710	-0.118\\
207.779	-0.070\\
207.899	-0.020\\
207.970	-0.014\\
208.090	-0.010\\
208.270	0.002\\
208.320	0.012\\
208.439	0.032\\
208.530	0.046\\
208.600	0.058\\
208.730	0.068\\
208.789	0.074\\
208.840	0.086\\
208.880	0.102\\
209.070	0.120\\
209.330	0.132\\
209.449	0.150\\
209.489	0.150\\
209.589	0.150\\
209.830	0.150\\
209.860	0.150\\
209.960	0.142\\
210.040	0.134\\
210.100	0.130\\
210.160	0.130\\
210.230	0.130\\
210.440	0.130\\
210.520	0.138\\
210.640	0.144\\
211.030	0.146\\
211.200	0.148\\
211.300	0.158\\
211.380	0.160\\
211.520	0.160\\
211.579	0.160\\
211.619	0.172\\
212.040	0.182\\
212.159	0.192\\
212.230	0.202\\
212.431	0.212\\
212.579	0.210\\
212.630	0.210\\
212.730	0.210\\
212.880	0.216\\
213.081	0.222\\
213.150	0.236\\
213.190	0.248\\
213.309	0.260\\
213.434	0.266\\
213.490	0.272\\
213.530	0.270\\
213.580	0.270\\
213.661	0.270\\
213.751	0.270\\
213.811	0.270\\
213.970	0.266\\
214.060	0.262\\
214.189	0.250\\
214.289	0.238\\
214.381	0.242\\
214.500	0.248\\
214.610	0.252\\
214.690	0.264\\
214.810	0.222\\
215.059	0.208\\
215.160	0.196\\
215.380	0.186\\
215.441	0.182\\
215.590	0.236\\
215.651	0.252\\
215.760	0.266\\
215.831	0.286\\
216.090	0.290\\
216.230	0.290\\
216.330	0.286\\
216.419	0.286\\
216.560	0.288\\
216.791	0.302\\
216.901	0.312\\
217.020	0.308\\
217.080	0.306\\
217.269	0.300\\
217.360	0.280\\
217.451	0.262\\
217.730	0.254\\
217.809	0.242\\
217.870	0.226\\
217.949	0.224\\
218.029	0.220\\
218.182	0.218\\
218.322	0.216\\
218.481	0.214\\
218.599	0.200\\
218.679	0.190\\
218.851	0.172\\
218.860	0.154\\
218.980	0.126\\
219.010	0.106\\
219.290	0.076\\
219.399	0.054\\
219.584	0.028\\
219.730	0.004\\
219.850	-0.014\\
220.050	-0.038\\
220.159	-0.056\\
220.280	-0.070\\
220.379	-0.072\\
220.480	-0.086\\
220.564	-0.096\\
220.730	-0.108\\
220.880	-0.122\\
220.939	-0.142\\
221.009	-0.164\\
221.129	-0.174\\
221.379	-0.188\\
221.491	-0.200\\
221.569	-0.210\\
221.649	-0.210\\
221.789	-0.216\\
221.880	-0.220\\
221.989	-0.224\\
222.060	-0.240\\
222.200	-0.256\\
222.439	-0.274\\
222.540	-0.302\\
222.590	-0.338\\
222.683	-0.380\\
222.841	-0.418\\
222.950	-0.446\\
223.160	-0.458\\
223.230	-0.454\\
223.379	-0.418\\
223.461	-0.382\\
223.599	-0.336\\
223.799	-0.304\\
223.920	-0.278\\
224.009	-0.266\\
224.120	-0.268\\
224.179	-0.280\\
224.371	-0.280\\
224.431	-0.282\\
224.791	-0.284\\
224.889	-0.276\\
224.970	-0.270\\
225.179	-0.270\\
225.270	-0.270\\
225.370	-0.270\\
225.489	-0.270\\
225.549	-0.270\\
225.760	-0.274\\
225.829	-0.276\\
225.939	-0.274\\
225.969	-0.274\\
226.029	-0.260\\
226.210	-0.248\\
226.249	-0.246\\
226.350	-0.246\\
226.559	-0.244\\
226.659	-0.256\\
226.699	-0.274\\
226.839	-0.284\\
226.970	-0.290\\
227.079	-0.292\\
227.160	-0.284\\
227.469	-0.264\\
227.670	-0.244\\
227.749	-0.230\\
227.830	-0.220\\
227.919	-0.220\\
228.019	-0.222\\
228.179	-0.232\\
228.259	-0.242\\
228.380	-0.250\\
228.430	-0.250\\
228.590	-0.248\\
228.819	-0.238\\
228.880	-0.228\\
229.040	-0.204\\
229.200	-0.176\\
229.280	-0.138\\
229.409	-0.094\\
229.549	-0.046\\
229.680	-0.014\\
229.929	0.006\\
230.009	0.016\\
230.129	0.020\\
230.229	0.024\\
230.341	0.030\\
230.391	0.052\\
230.490	0.068\\
230.920	0.072\\
231.030	0.090\\
231.219	0.112\\
231.380	0.132\\
231.539	0.156\\
231.730	0.190\\
231.811	0.208\\
231.859	0.220\\
232.151	0.218\\
232.219	0.218\\
232.349	0.220\\
232.470	0.220\\
232.569	0.220\\
232.730	0.218\\
232.829	0.216\\
233.009	0.214\\
233.070	0.212\\
233.330	0.212\\
233.429	0.214\\
233.611	0.216\\
233.749	0.218\\
233.819	0.220\\
233.919	0.220\\
234.009	0.222\\
234.039	0.224\\
234.100	0.226\\
234.249	0.228\\
234.411	0.230\\
234.489	0.230\\
234.570	0.230\\
234.630	0.230\\
234.679	0.238\\
234.760	0.260\\
234.840	0.286\\
234.949	0.318\\
235.101	0.352\\
235.319	0.372\\
235.409	0.368\\
235.480	0.360\\
235.719	0.340\\
235.890	0.318\\
236.019	0.290\\
236.179	0.270\\
236.409	0.248\\
236.489	0.234\\
236.579	0.222\\
236.729	0.222\\
236.809	0.224\\
236.959	0.228\\
237.039	0.230\\
237.129	0.230\\
237.209	0.230\\
237.250	0.230\\
237.389	0.230\\
237.560	0.230\\
237.749	0.230\\
237.930	0.230\\
238.060	0.230\\
238.120	0.230\\
238.209	0.230\\
238.320	0.230\\
238.449	0.230\\
238.499	0.230\\
238.539	0.230\\
238.659	0.230\\
238.719	0.230\\
238.800	0.232\\
238.989	0.236\\
239.080	0.240\\
239.199	0.244\\
239.272	0.242\\
239.369	0.238\\
239.489	0.222\\
239.540	0.206\\
239.799	0.190\\
239.859	0.180\\
239.990	0.162\\
240.090	0.154\\
240.159	0.134\\
240.350	0.118\\
240.549	0.096\\
240.650	0.078\\
240.709	0.058\\
240.849	0.050\\
240.909	0.038\\
241.049	0.032\\
241.130	-0.002\\
241.269	-0.036\\
241.629	-0.072\\
241.730	-0.108\\
241.809	-0.144\\
241.930	-0.148\\
241.970	-0.150\\
242.150	-0.150\\
242.219	-0.150\\
242.259	-0.150\\
242.379	-0.150\\
242.460	-0.162\\
242.599	-0.170\\
242.679	-0.178\\
242.971	-0.186\\
243.029	-0.194\\
243.209	-0.190\\
243.340	-0.190\\
243.509	-0.190\\
243.570	-0.198\\
243.670	-0.212\\
243.809	-0.224\\
243.910	-0.244\\
243.959	-0.268\\
244.039	-0.286\\
244.211	-0.300\\
244.299	-0.316\\
244.400	-0.324\\
244.530	-0.328\\
244.580	-0.328\\
244.719	-0.328\\
244.799	-0.328\\
244.909	-0.328\\
244.999	-0.328\\
245.090	-0.330\\
245.249	-0.330\\
245.320	-0.330\\
245.409	-0.330\\
245.479	-0.330\\
245.539	-0.330\\
245.680	-0.330\\
245.879	-0.330\\
245.999	-0.330\\
246.069	-0.330\\
246.129	-0.330\\
246.250	-0.330\\
246.350	-0.330\\
246.409	-0.330\\
246.450	-0.330\\
246.619	-0.330\\
246.699	-0.332\\
246.859	-0.336\\
246.919	-0.342\\
246.999	-0.348\\
247.159	-0.352\\
247.369	-0.354\\
247.459	-0.354\\
247.490	-0.350\\
247.669	-0.346\\
247.870	-0.344\\
247.951	-0.340\\
248.029	-0.336\\
248.152	-0.334\\
248.189	-0.332\\
248.369	-0.330\\
248.489	-0.330\\
248.549	-0.318\\
248.609	-0.306\\
248.660	-0.294\\
249.019	-0.282\\
249.089	-0.264\\
249.189	-0.254\\
249.269	-0.238\\
249.349	-0.218\\
249.509	-0.198\\
249.589	-0.176\\
249.789	-0.158\\
249.950	-0.138\\
250.009	-0.120\\
250.089	-0.096\\
250.139	-0.076\\
250.200	-0.054\\
250.319	-0.038\\
250.499	-0.018\\
250.579	-0.002\\
250.679	0.010\\
250.759	0.020\\
250.830	0.028\\
250.960	0.034\\
251.009	0.042\\
251.119	0.052\\
251.219	0.058\\
251.330	0.066\\
251.660	0.072\\
251.739	0.074\\
251.929	0.074\\
252.149	0.078\\
252.229	0.080\\
252.349	0.080\\
252.409	0.082\\
252.609	0.084\\
252.729	0.088\\
252.801	0.098\\
252.920	0.112\\
253.089	0.120\\
253.139	0.128\\
253.460	0.134\\
253.599	0.140\\
253.669	0.146\\
253.800	0.150\\
253.869	0.168\\
253.909	0.192\\
254.030	0.210\\
254.309	0.224\\
254.369	0.244\\
254.490	0.250\\
254.579	0.250\\
254.699	0.250\\
254.760	0.250\\
254.820	0.250\\
254.899	0.250\\
255.239	0.250\\
255.319	0.250\\
255.409	0.250\\
255.739	0.250\\
256.000	0.250\\
256.079	0.250\\
256.152	0.250\\
256.299	0.250\\
256.369	0.250\\
256.439	0.256\\
256.609	0.262\\
256.769	0.270\\
256.892	0.278\\
257.039	0.286\\
257.340	0.288\\
257.439	0.290\\
257.479	0.290\\
257.599	0.290\\
257.749	0.290\\
257.840	0.288\\
257.949	0.286\\
258.069	0.278\\
258.149	0.270\\
258.309	0.254\\
258.389	0.242\\
258.509	0.226\\
258.599	0.212\\
258.871	0.204\\
258.969	0.210\\
259.059	0.208\\
259.239	0.216\\
259.379	0.222\\
259.500	0.220\\
259.731	0.210\\
259.800	0.204\\
259.909	0.182\\
260.019	0.166\\
260.089	0.152\\
260.209	0.140\\
260.282	0.130\\
260.429	0.130\\
260.520	0.116\\
260.650	0.102\\
260.759	0.092\\
260.940	0.080\\
261.020	0.066\\
261.109	0.062\\
261.189	0.056\\
261.229	0.040\\
261.349	0.024\\
261.519	0.010\\
261.610	0.000\\
261.719	-0.008\\
261.759	-0.010\\
261.859	-0.024\\
262.030	-0.044\\
262.289	-0.064\\
262.490	-0.084\\
262.569	-0.112\\
262.690	-0.130\\
262.751	-0.142\\
262.899	-0.154\\
263.079	-0.172\\
263.139	-0.178\\
263.189	-0.208\\
263.289	-0.238\\
263.459	-0.270\\
263.539	-0.288\\
263.672	-0.304\\
263.751	-0.288\\
263.800	-0.278\\
263.850	-0.260\\
263.962	-0.258\\
264.119	-0.254\\
264.209	-0.250\\
264.319	-0.240\\
264.409	-0.236\\
264.560	-0.224\\
264.659	-0.220\\
264.699	-0.220\\
264.769	-0.220\\
264.869	-0.220\\
264.919	-0.220\\
265.109	-0.220\\
265.229	-0.220\\
265.299	-0.220\\
265.349	-0.220\\
265.429	-0.220\\
265.549	-0.220\\
265.659	-0.220\\
265.699	-0.220\\
265.749	-0.220\\
265.849	-0.224\\
266.049	-0.238\\
266.119	-0.256\\
266.249	-0.276\\
266.479	-0.304\\
266.519	-0.328\\
266.659	-0.348\\
266.699	-0.370\\
266.849	-0.386\\
266.989	-0.386\\
267.069	-0.392\\
267.251	-0.392\\
267.329	-0.376\\
267.379	-0.356\\
267.569	-0.346\\
267.729	-0.322\\
267.919	-0.298\\
268.019	-0.284\\
268.089	-0.290\\
268.240	-0.296\\
268.349	-0.296\\
268.469	-0.294\\
268.850	-0.292\\
268.969	-0.278\\
269.069	-0.262\\
269.289	-0.256\\
269.400	-0.256\\
269.499	-0.254\\
269.579	-0.248\\
269.639	-0.240\\
269.699	-0.236\\
269.879	-0.222\\
269.999	-0.210\\
270.209	-0.202\\
270.309	-0.196\\
270.421	-0.192\\
270.509	-0.194\\
270.629	-0.196\\
270.749	-0.200\\
270.909	-0.202\\
270.980	-0.226\\
271.079	-0.252\\
271.370	-0.280\\
271.429	-0.308\\
271.510	-0.338\\
271.899	-0.344\\
271.940	-0.348\\
272.079	-0.350\\
272.119	-0.350\\
272.239	-0.350\\
272.299	-0.350\\
272.329	-0.336\\
272.530	-0.316\\
272.629	-0.290\\
272.769	-0.264\\
272.929	-0.224\\
273.049	-0.196\\
273.189	-0.172\\
273.230	-0.160\\
273.369	-0.138\\
273.500	-0.122\\
273.529	-0.104\\
273.640	-0.086\\
273.769	-0.060\\
273.819	-0.042\\
273.949	-0.032\\
274.009	-0.026\\
274.089	-0.018\\
274.249	-0.010\\
274.520	-0.004\\
274.609	0.000\\
274.699	0.010\\
274.829	0.022\\
275.029	0.036\\
275.110	0.058\\
275.150	0.082\\
275.289	0.100\\
275.350	0.112\\
275.419	0.120\\
275.579	0.120\\
275.639	0.120\\
275.729	0.108\\
275.919	0.104\\
276.079	0.104\\
276.219	0.100\\
276.299	0.100\\
276.379	0.112\\
276.471	0.116\\
276.589	0.116\\
276.709	0.120\\
276.799	0.114\\
276.889	0.114\\
276.929	0.114\\
277.089	0.114\\
277.209	0.122\\
277.329	0.144\\
277.399	0.166\\
277.519	0.186\\
277.719	0.206\\
277.779	0.224\\
277.859	0.240\\
278.019	0.252\\
278.219	0.264\\
278.269	0.276\\
278.469	0.286\\
278.569	0.292\\
278.689	0.290\\
278.729	0.304\\
278.869	0.304\\
279.019	0.310\\
279.199	0.314\\
279.299	0.330\\
279.379	0.328\\
279.509	0.344\\
279.699	0.342\\
279.789	0.334\\
279.849	0.306\\
279.969	0.284\\
280.109	0.266\\
};
\addlegendentry{Velocity estimate};

\addplot [color=red,solid]
  table[row sep=crcr]{%
159.920	-0.300\\
160.030	-0.306\\
160.090	-0.308\\
160.170	-0.308\\
160.312	-0.314\\
160.390	-0.316\\
160.470	-0.318\\
160.591	-0.322\\
160.750	-0.322\\
161.062	-0.316\\
161.120	-0.310\\
161.220	-0.302\\
161.420	-0.296\\
161.540	-0.294\\
161.771	-0.290\\
161.891	-0.286\\
162.032	-0.284\\
162.100	-0.282\\
162.210	-0.280\\
162.280	-0.280\\
162.420	-0.280\\
162.541	-0.276\\
162.650	-0.268\\
162.740	-0.260\\
162.840	-0.252\\
162.970	-0.234\\
163.010	-0.220\\
163.140	-0.188\\
163.390	-0.156\\
163.600	-0.116\\
163.690	-0.086\\
163.770	-0.056\\
163.850	-0.048\\
163.990	-0.034\\
164.020	-0.028\\
164.160	-0.020\\
164.210	-0.006\\
164.340	0.012\\
164.620	0.024\\
164.660	0.044\\
164.800	0.070\\
164.881	0.090\\
165.110	0.110\\
165.220	0.136\\
165.370	0.158\\
165.470	0.172\\
165.670	0.186\\
165.780	0.196\\
165.880	0.204\\
165.950	0.208\\
166.080	0.210\\
166.170	0.212\\
166.270	0.218\\
166.310	0.220\\
166.510	0.224\\
166.670	0.230\\
166.730	0.238\\
166.841	0.242\\
167.060	0.246\\
167.180	0.248\\
167.300	0.250\\
167.420	0.260\\
167.490	0.270\\
167.750	0.280\\
167.850	0.292\\
167.930	0.304\\
168.041	0.306\\
168.080	0.308\\
168.220	0.310\\
168.340	0.310\\
168.400	0.310\\
168.490	0.310\\
168.540	0.310\\
168.710	0.310\\
168.751	0.304\\
168.890	0.298\\
169.010	0.292\\
169.080	0.286\\
169.230	0.280\\
169.300	0.280\\
169.410	0.278\\
169.480	0.276\\
169.811	0.274\\
169.910	0.272\\
170.000	0.272\\
170.030	0.280\\
170.310	0.288\\
170.490	0.296\\
170.590	0.304\\
170.671	0.310\\
170.820	0.308\\
170.881	0.306\\
170.940	0.304\\
171.010	0.302\\
171.180	0.298\\
171.380	0.296\\
171.660	0.294\\
171.710	0.292\\
171.930	0.288\\
172.000	0.286\\
172.120	0.286\\
172.200	0.286\\
172.331	0.286\\
172.500	0.288\\
172.572	0.294\\
172.690	0.298\\
172.964	0.300\\
173.020	0.302\\
173.130	0.304\\
173.220	0.304\\
173.440	0.304\\
173.731	0.304\\
173.820	0.290\\
173.990	0.276\\
174.501	0.260\\
174.530	0.238\\
174.612	0.202\\
174.712	0.180\\
175.040	0.150\\
175.120	0.120\\
175.240	0.096\\
175.300	0.088\\
175.440	0.070\\
175.470	0.054\\
175.680	0.038\\
175.760	0.018\\
175.870	-0.012\\
176.040	-0.040\\
176.170	-0.066\\
176.290	-0.100\\
176.520	-0.130\\
176.601	-0.150\\
176.720	-0.162\\
176.870	-0.178\\
177.032	-0.186\\
177.250	-0.204\\
177.340	-0.222\\
177.420	-0.246\\
177.540	-0.262\\
177.701	-0.278\\
177.820	-0.284\\
177.850	-0.290\\
177.910	-0.290\\
178.040	-0.292\\
178.090	-0.294\\
178.220	-0.296\\
178.410	-0.298\\
178.520	-0.300\\
178.580	-0.300\\
178.660	-0.298\\
178.740	-0.296\\
178.900	-0.294\\
179.050	-0.286\\
179.170	-0.278\\
179.390	-0.272\\
179.470	-0.266\\
179.650	-0.258\\
179.720	-0.256\\
179.850	-0.254\\
179.911	-0.252\\
180.020	-0.250\\
180.120	-0.250\\
180.160	-0.250\\
180.220	-0.252\\
180.290	-0.254\\
180.360	-0.256\\
180.440	-0.258\\
180.662	-0.260\\
180.770	-0.258\\
180.840	-0.258\\
181.180	-0.258\\
181.280	-0.258\\
181.310	-0.258\\
181.550	-0.262\\
181.590	-0.264\\
181.730	-0.262\\
181.830	-0.260\\
181.920	-0.258\\
182.010	-0.254\\
182.100	-0.248\\
182.220	-0.246\\
182.301	-0.244\\
182.530	-0.242\\
182.770	-0.236\\
182.851	-0.228\\
183.020	-0.220\\
183.050	-0.210\\
183.260	-0.200\\
183.450	-0.192\\
183.550	-0.188\\
183.630	-0.184\\
183.760	-0.182\\
183.890	-0.180\\
183.960	-0.178\\
184.030	-0.176\\
184.180	-0.176\\
184.230	-0.176\\
184.381	-0.174\\
184.490	-0.174\\
184.520	-0.174\\
184.620	-0.172\\
184.790	-0.170\\
184.840	-0.170\\
184.980	-0.170\\
185.060	-0.170\\
185.390	-0.168\\
185.510	-0.160\\
185.590	-0.152\\
185.850	-0.144\\
185.990	-0.136\\
186.070	-0.130\\
186.110	-0.128\\
186.200	-0.124\\
186.420	-0.120\\
186.520	-0.116\\
186.600	-0.112\\
186.700	-0.110\\
186.770	-0.112\\
186.902	-0.114\\
186.960	-0.116\\
187.000	-0.118\\
187.090	-0.120\\
187.130	-0.120\\
187.210	-0.120\\
187.300	-0.118\\
187.500	-0.116\\
187.540	-0.114\\
187.660	-0.112\\
187.740	-0.098\\
187.840	-0.086\\
187.990	-0.068\\
188.100	-0.050\\
188.180	-0.028\\
188.260	-0.018\\
188.321	-0.004\\
188.361	0.004\\
188.540	0.014\\
188.620	0.020\\
188.710	0.030\\
188.790	0.036\\
188.870	0.044\\
189.030	0.050\\
189.140	0.060\\
189.180	0.066\\
189.321	0.072\\
189.360	0.078\\
189.520	0.084\\
189.620	0.090\\
189.700	0.098\\
189.861	0.110\\
190.030	0.120\\
190.211	0.130\\
190.380	0.136\\
190.460	0.154\\
190.710	0.170\\
190.900	0.186\\
191.060	0.202\\
191.181	0.224\\
191.222	0.232\\
191.370	0.238\\
191.470	0.244\\
191.590	0.254\\
191.670	0.262\\
191.880	0.270\\
191.960	0.278\\
192.090	0.286\\
192.170	0.290\\
192.280	0.292\\
192.460	0.294\\
192.509	0.298\\
192.670	0.312\\
192.810	0.324\\
193.171	0.334\\
193.310	0.348\\
193.340	0.360\\
193.440	0.362\\
193.550	0.366\\
193.660	0.370\\
193.700	0.374\\
193.759	0.378\\
193.830	0.384\\
194.030	0.390\\
194.152	0.392\\
194.300	0.390\\
194.490	0.394\\
194.540	0.396\\
194.700	0.398\\
194.800	0.404\\
194.949	0.406\\
194.990	0.402\\
195.320	0.388\\
195.479	0.374\\
195.580	0.360\\
195.660	0.342\\
195.820	0.324\\
195.940	0.310\\
196.040	0.296\\
196.130	0.282\\
196.313	0.276\\
196.450	0.238\\
196.520	0.206\\
196.590	0.174\\
196.780	0.142\\
196.820	0.102\\
196.930	0.086\\
196.960	0.070\\
197.140	0.054\\
197.232	0.032\\
197.320	0.018\\
197.350	0.010\\
197.480	0.002\\
197.589	-0.012\\
197.693	-0.026\\
197.800	-0.040\\
197.969	-0.052\\
198.050	-0.080\\
198.160	-0.106\\
198.529	-0.126\\
198.620	-0.146\\
198.710	-0.166\\
198.860	-0.170\\
198.940	-0.170\\
199.030	-0.170\\
199.070	-0.170\\
199.150	-0.170\\
199.250	-0.182\\
199.330	-0.194\\
199.419	-0.210\\
199.500	-0.226\\
199.580	-0.244\\
199.701	-0.254\\
199.989	-0.266\\
200.070	-0.274\\
200.231	-0.284\\
200.270	-0.294\\
200.390	-0.300\\
200.489	-0.302\\
200.610	-0.306\\
200.650	-0.308\\
200.840	-0.308\\
200.950	-0.308\\
201.020	-0.310\\
201.110	-0.312\\
201.231	-0.314\\
201.350	-0.316\\
201.449	-0.320\\
201.499	-0.324\\
201.700	-0.326\\
201.960	-0.324\\
202.020	-0.322\\
202.109	-0.318\\
202.350	-0.314\\
202.530	-0.306\\
202.659	-0.302\\
202.889	-0.294\\
203.000	-0.286\\
203.101	-0.278\\
203.400	-0.274\\
203.500	-0.270\\
203.590	-0.270\\
203.720	-0.270\\
203.839	-0.270\\
203.929	-0.270\\
204.010	-0.270\\
204.150	-0.268\\
204.299	-0.266\\
204.380	-0.260\\
204.529	-0.254\\
204.610	-0.246\\
204.670	-0.236\\
204.790	-0.226\\
205.070	-0.222\\
205.129	-0.218\\
205.209	-0.216\\
205.330	-0.218\\
205.400	-0.224\\
205.489	-0.228\\
205.620	-0.236\\
205.660	-0.244\\
205.800	-0.252\\
205.879	-0.256\\
206.210	-0.260\\
206.289	-0.262\\
206.369	-0.264\\
206.520	-0.266\\
206.560	-0.268\\
206.709	-0.270\\
206.820	-0.266\\
206.881	-0.262\\
206.961	-0.254\\
206.990	-0.246\\
207.170	-0.228\\
207.290	-0.196\\
207.440	-0.164\\
207.710	-0.136\\
207.779	-0.108\\
207.899	-0.090\\
207.970	-0.072\\
208.090	-0.054\\
208.270	-0.032\\
208.320	-0.010\\
208.439	0.016\\
208.530	0.024\\
208.600	0.036\\
208.730	0.044\\
208.789	0.052\\
208.840	0.062\\
208.880	0.080\\
209.070	0.094\\
209.330	0.108\\
209.449	0.124\\
209.489	0.136\\
209.589	0.140\\
209.830	0.144\\
209.860	0.148\\
209.960	0.150\\
210.040	0.150\\
210.100	0.152\\
210.160	0.156\\
210.230	0.160\\
210.440	0.166\\
210.520	0.176\\
210.640	0.188\\
211.030	0.198\\
211.200	0.208\\
211.300	0.216\\
211.380	0.220\\
211.520	0.222\\
211.579	0.226\\
211.619	0.230\\
212.040	0.234\\
212.159	0.238\\
212.230	0.242\\
212.431	0.244\\
212.579	0.246\\
212.630	0.248\\
212.730	0.252\\
212.880	0.254\\
213.081	0.256\\
213.150	0.258\\
213.190	0.266\\
213.309	0.272\\
213.434	0.278\\
213.490	0.284\\
213.530	0.288\\
213.580	0.286\\
213.661	0.286\\
213.751	0.286\\
213.811	0.288\\
213.970	0.292\\
214.060	0.296\\
214.189	0.298\\
214.289	0.300\\
214.381	0.300\\
214.500	0.300\\
214.610	0.300\\
214.690	0.302\\
214.810	0.308\\
215.059	0.314\\
215.160	0.318\\
215.380	0.322\\
215.441	0.324\\
215.590	0.322\\
215.651	0.320\\
215.760	0.320\\
215.831	0.320\\
216.090	0.320\\
216.230	0.320\\
216.330	0.320\\
216.419	0.316\\
216.560	0.310\\
216.791	0.304\\
216.901	0.298\\
217.020	0.294\\
217.080	0.294\\
217.269	0.296\\
217.360	0.296\\
217.451	0.296\\
217.730	0.294\\
217.809	0.292\\
217.870	0.284\\
217.949	0.278\\
218.029	0.268\\
218.182	0.258\\
218.322	0.238\\
218.481	0.224\\
218.599	0.190\\
218.679	0.160\\
218.851	0.130\\
218.860	0.110\\
218.980	0.086\\
219.010	0.082\\
219.290	0.064\\
219.399	0.046\\
219.584	0.024\\
219.730	0.006\\
219.850	-0.012\\
220.050	-0.026\\
220.159	-0.040\\
220.280	-0.050\\
220.379	-0.060\\
220.480	-0.076\\
220.564	-0.088\\
220.730	-0.100\\
220.880	-0.112\\
220.939	-0.124\\
221.009	-0.140\\
221.129	-0.150\\
221.379	-0.164\\
221.491	-0.178\\
221.569	-0.192\\
221.649	-0.200\\
221.789	-0.208\\
221.880	-0.214\\
221.989	-0.224\\
222.060	-0.240\\
222.200	-0.252\\
222.439	-0.266\\
222.540	-0.278\\
222.590	-0.286\\
222.683	-0.288\\
222.841	-0.300\\
222.950	-0.310\\
223.160	-0.324\\
223.230	-0.338\\
223.379	-0.352\\
223.461	-0.352\\
223.599	-0.350\\
223.799	-0.344\\
223.920	-0.338\\
224.009	-0.330\\
224.120	-0.326\\
224.179	-0.324\\
224.371	-0.322\\
224.431	-0.320\\
224.791	-0.318\\
224.889	-0.320\\
224.970	-0.322\\
225.179	-0.324\\
225.270	-0.320\\
225.370	-0.318\\
225.489	-0.318\\
225.549	-0.316\\
225.760	-0.314\\
225.829	-0.318\\
225.939	-0.324\\
225.969	-0.324\\
226.029	-0.326\\
226.210	-0.328\\
226.249	-0.330\\
226.350	-0.330\\
226.559	-0.330\\
226.659	-0.330\\
226.699	-0.330\\
226.839	-0.326\\
226.970	-0.322\\
227.079	-0.314\\
227.160	-0.304\\
227.469	-0.294\\
227.670	-0.288\\
227.749	-0.282\\
227.830	-0.280\\
227.919	-0.276\\
228.019	-0.268\\
228.179	-0.260\\
228.259	-0.252\\
228.380	-0.244\\
228.430	-0.232\\
228.590	-0.224\\
228.819	-0.206\\
228.880	-0.188\\
229.040	-0.160\\
229.200	-0.140\\
229.280	-0.112\\
229.409	-0.086\\
229.549	-0.054\\
229.680	-0.032\\
229.929	-0.010\\
230.009	0.004\\
230.129	0.014\\
230.229	0.018\\
230.341	0.026\\
230.391	0.046\\
230.490	0.066\\
230.920	0.086\\
231.030	0.110\\
231.219	0.130\\
231.380	0.152\\
231.539	0.174\\
231.730	0.192\\
231.811	0.206\\
231.859	0.220\\
232.151	0.220\\
232.219	0.220\\
232.349	0.220\\
232.470	0.220\\
232.569	0.220\\
232.730	0.222\\
232.829	0.224\\
233.009	0.226\\
233.070	0.228\\
233.330	0.236\\
233.429	0.242\\
233.611	0.248\\
233.749	0.254\\
233.819	0.260\\
233.919	0.260\\
234.009	0.260\\
234.039	0.260\\
234.100	0.260\\
234.249	0.260\\
234.411	0.260\\
234.489	0.260\\
234.570	0.260\\
234.630	0.260\\
234.679	0.260\\
234.760	0.262\\
234.840	0.264\\
234.949	0.268\\
235.101	0.272\\
235.319	0.276\\
235.409	0.278\\
235.480	0.278\\
235.719	0.276\\
235.890	0.274\\
236.019	0.270\\
236.179	0.266\\
236.409	0.264\\
236.489	0.262\\
236.579	0.260\\
236.729	0.262\\
236.809	0.264\\
236.959	0.266\\
237.039	0.268\\
237.129	0.270\\
237.209	0.270\\
237.250	0.272\\
237.389	0.276\\
237.560	0.282\\
237.749	0.288\\
237.930	0.294\\
238.060	0.298\\
238.120	0.304\\
238.209	0.308\\
238.320	0.310\\
238.449	0.312\\
238.499	0.314\\
238.539	0.316\\
238.659	0.318\\
238.719	0.320\\
238.800	0.322\\
238.989	0.324\\
239.080	0.322\\
239.199	0.320\\
239.272	0.310\\
239.369	0.300\\
239.489	0.278\\
239.540	0.256\\
239.799	0.228\\
239.859	0.210\\
239.990	0.192\\
240.090	0.182\\
240.159	0.166\\
240.350	0.156\\
240.549	0.136\\
240.650	0.116\\
240.709	0.100\\
240.849	0.090\\
240.909	0.066\\
241.049	0.052\\
241.130	0.018\\
241.269	-0.016\\
241.629	-0.050\\
241.730	-0.084\\
241.809	-0.118\\
241.930	-0.132\\
241.970	-0.146\\
242.150	-0.166\\
242.219	-0.172\\
242.259	-0.184\\
242.379	-0.196\\
242.460	-0.208\\
242.599	-0.218\\
242.679	-0.228\\
242.971	-0.232\\
243.029	-0.240\\
243.209	-0.248\\
243.340	-0.252\\
243.509	-0.256\\
243.570	-0.264\\
243.670	-0.268\\
243.809	-0.272\\
243.910	-0.280\\
243.959	-0.288\\
244.039	-0.294\\
244.211	-0.300\\
244.299	-0.306\\
244.400	-0.308\\
244.530	-0.310\\
244.580	-0.310\\
244.719	-0.316\\
244.799	-0.322\\
244.909	-0.334\\
244.999	-0.346\\
245.090	-0.358\\
245.249	-0.362\\
245.320	-0.366\\
245.409	-0.364\\
245.479	-0.364\\
245.539	-0.362\\
245.680	-0.362\\
245.879	-0.362\\
245.999	-0.362\\
246.069	-0.360\\
246.129	-0.362\\
246.250	-0.364\\
246.350	-0.366\\
246.409	-0.370\\
246.450	-0.374\\
246.619	-0.378\\
246.699	-0.382\\
246.859	-0.386\\
246.919	-0.390\\
246.999	-0.394\\
247.159	-0.396\\
247.369	-0.398\\
247.459	-0.408\\
247.490	-0.416\\
247.669	-0.418\\
247.870	-0.420\\
247.951	-0.422\\
248.029	-0.416\\
248.152	-0.410\\
248.189	-0.406\\
248.369	-0.402\\
248.489	-0.398\\
248.549	-0.394\\
248.609	-0.378\\
248.660	-0.366\\
249.019	-0.344\\
249.089	-0.322\\
249.189	-0.300\\
249.269	-0.282\\
249.349	-0.264\\
249.509	-0.244\\
249.589	-0.216\\
249.789	-0.188\\
249.950	-0.158\\
250.009	-0.128\\
250.089	-0.110\\
250.139	-0.092\\
250.200	-0.066\\
250.319	-0.050\\
250.499	-0.032\\
250.579	-0.014\\
250.679	-0.004\\
250.759	0.000\\
250.830	0.004\\
250.960	0.012\\
251.009	0.020\\
251.119	0.028\\
251.219	0.044\\
251.330	0.062\\
251.660	0.074\\
251.739	0.098\\
251.929	0.122\\
252.149	0.138\\
252.229	0.152\\
252.349	0.170\\
252.409	0.180\\
252.609	0.190\\
252.729	0.204\\
252.801	0.218\\
252.920	0.230\\
253.089	0.244\\
253.139	0.254\\
253.460	0.258\\
253.599	0.262\\
253.669	0.264\\
253.800	0.260\\
253.869	0.260\\
253.909	0.270\\
254.030	0.280\\
254.309	0.294\\
254.369	0.308\\
254.490	0.322\\
254.579	0.328\\
254.699	0.334\\
254.760	0.336\\
254.820	0.340\\
254.899	0.344\\
255.239	0.346\\
255.319	0.348\\
255.409	0.352\\
255.739	0.354\\
256.000	0.356\\
256.079	0.358\\
256.152	0.360\\
256.299	0.360\\
256.369	0.360\\
256.439	0.358\\
256.609	0.358\\
256.769	0.358\\
256.892	0.354\\
257.039	0.350\\
257.340	0.348\\
257.439	0.342\\
257.479	0.340\\
257.599	0.342\\
257.749	0.338\\
257.840	0.334\\
257.949	0.336\\
258.069	0.334\\
258.149	0.328\\
258.309	0.328\\
258.389	0.328\\
258.509	0.322\\
258.599	0.316\\
258.871	0.314\\
258.969	0.312\\
259.059	0.310\\
259.239	0.310\\
259.379	0.304\\
259.500	0.298\\
259.731	0.292\\
259.800	0.286\\
259.909	0.280\\
260.019	0.274\\
260.089	0.268\\
260.209	0.256\\
260.282	0.236\\
260.429	0.216\\
260.520	0.194\\
260.650	0.172\\
260.759	0.156\\
260.940	0.148\\
261.020	0.140\\
261.109	0.126\\
261.189	0.106\\
261.229	0.078\\
261.349	0.050\\
261.519	0.022\\
261.610	0.008\\
261.719	0.000\\
261.759	-0.006\\
261.859	-0.018\\
262.030	-0.046\\
262.289	-0.074\\
262.490	-0.102\\
262.569	-0.124\\
262.690	-0.140\\
262.751	-0.152\\
262.899	-0.164\\
263.079	-0.176\\
263.139	-0.188\\
263.189	-0.216\\
263.289	-0.232\\
263.459	-0.244\\
263.539	-0.256\\
263.672	-0.268\\
263.751	-0.264\\
263.800	-0.260\\
263.850	-0.260\\
263.962	-0.258\\
264.119	-0.256\\
264.209	-0.254\\
264.319	-0.250\\
264.409	-0.246\\
264.560	-0.244\\
264.659	-0.242\\
264.699	-0.240\\
264.769	-0.242\\
264.869	-0.244\\
264.919	-0.248\\
265.109	-0.252\\
265.229	-0.256\\
265.299	-0.258\\
265.349	-0.262\\
265.429	-0.264\\
265.549	-0.266\\
265.659	-0.270\\
265.699	-0.274\\
265.749	-0.276\\
265.849	-0.278\\
266.049	-0.286\\
266.119	-0.294\\
266.249	-0.302\\
266.479	-0.310\\
266.519	-0.320\\
266.659	-0.324\\
266.699	-0.328\\
266.849	-0.332\\
266.989	-0.336\\
267.069	-0.338\\
267.251	-0.338\\
267.329	-0.334\\
267.379	-0.330\\
267.569	-0.326\\
267.729	-0.316\\
267.919	-0.308\\
268.019	-0.302\\
268.089	-0.292\\
268.240	-0.282\\
268.349	-0.280\\
268.469	-0.274\\
268.850	-0.268\\
268.969	-0.266\\
269.069	-0.268\\
269.289	-0.268\\
269.400	-0.272\\
269.499	-0.276\\
269.579	-0.280\\
269.639	-0.272\\
269.699	-0.264\\
269.879	-0.260\\
269.999	-0.256\\
270.209	-0.252\\
270.309	-0.256\\
270.421	-0.260\\
270.509	-0.262\\
270.629	-0.264\\
270.749	-0.266\\
270.909	-0.268\\
270.980	-0.276\\
271.079	-0.282\\
271.370	-0.292\\
271.429	-0.306\\
271.510	-0.320\\
271.899	-0.328\\
271.940	-0.336\\
272.079	-0.340\\
272.119	-0.340\\
272.239	-0.334\\
272.299	-0.328\\
272.329	-0.320\\
272.530	-0.302\\
272.629	-0.284\\
272.769	-0.266\\
272.929	-0.230\\
273.049	-0.196\\
273.189	-0.172\\
273.230	-0.148\\
273.369	-0.122\\
273.500	-0.114\\
273.529	-0.106\\
273.640	-0.098\\
273.769	-0.090\\
273.819	-0.078\\
273.949	-0.066\\
274.009	-0.050\\
274.089	-0.026\\
274.249	-0.002\\
274.520	0.010\\
274.609	0.022\\
274.699	0.036\\
274.829	0.042\\
275.029	0.052\\
275.110	0.066\\
275.150	0.080\\
275.289	0.092\\
275.350	0.104\\
275.419	0.112\\
275.579	0.116\\
275.639	0.122\\
275.729	0.124\\
275.919	0.130\\
276.079	0.136\\
276.219	0.142\\
276.299	0.146\\
276.379	0.150\\
276.471	0.154\\
276.589	0.158\\
276.709	0.166\\
276.799	0.174\\
276.889	0.184\\
276.929	0.194\\
277.089	0.204\\
277.209	0.210\\
277.329	0.228\\
277.399	0.248\\
277.519	0.264\\
277.719	0.280\\
277.779	0.300\\
277.859	0.312\\
278.019	0.320\\
278.219	0.332\\
278.269	0.344\\
278.469	0.352\\
278.569	0.356\\
278.689	0.362\\
278.729	0.364\\
278.869	0.368\\
279.019	0.372\\
279.199	0.376\\
279.299	0.376\\
279.379	0.376\\
279.509	0.374\\
279.699	0.372\\
279.789	0.370\\
279.849	0.368\\
279.969	0.366\\
280.109	0.366\\
};
\addlegendentry{Opti-Track};

\addplot [color=black!30!green,solid]
  table[row sep=crcr]{%
159.960	-0.257\\
160.650	-0.255\\
160.831	-0.255\\
160.981	-0.255\\
161.242	-0.255\\
161.500	-0.255\\
161.600	-0.255\\
161.650	-0.245\\
162.140	0.300\\
163.300	0.300\\
163.730	0.300\\
164.060	0.300\\
164.300	0.300\\
164.420	0.300\\
164.560	0.300\\
164.700	0.300\\
164.910	0.300\\
165.270	0.300\\
165.430	0.300\\
165.720	0.300\\
166.430	0.300\\
166.770	0.300\\
166.950	0.300\\
167.220	0.300\\
167.890	0.300\\
168.260	0.300\\
168.440	0.300\\
169.050	0.300\\
169.340	0.300\\
169.580	0.300\\
169.850	0.300\\
170.090	0.300\\
170.620	0.300\\
171.040	0.300\\
171.090	0.300\\
171.220	0.300\\
171.420	0.300\\
171.530	0.300\\
171.600	0.300\\
172.091	0.300\\
172.400	0.300\\
172.770	0.300\\
173.290	0.071\\
173.482	-0.214\\
173.590	-0.255\\
173.780	-0.255\\
173.940	-0.255\\
174.051	-0.255\\
174.390	-0.255\\
174.670	-0.255\\
174.850	-0.255\\
175.210	-0.255\\
175.601	-0.255\\
176.640	-0.255\\
177.279	-0.255\\
177.620	-0.257\\
178.330	-0.262\\
178.819	-0.257\\
178.940	-0.257\\
179.250	-0.257\\
179.770	-0.257\\
179.940	-0.255\\
180.700	-0.255\\
181.019	-0.255\\
181.360	-0.255\\
181.480	-0.255\\
182.180	-0.255\\
182.400	-0.255\\
182.620	-0.257\\
182.890	-0.260\\
183.690	-0.257\\
184.090	-0.257\\
184.429	-0.257\\
184.919	-0.257\\
185.129	-0.257\\
185.311	-0.257\\
185.920	-0.255\\
186.159	-0.255\\
186.280	-0.255\\
186.639	0.133\\
187.250	0.300\\
187.430	0.300\\
187.779	0.300\\
188.580	0.300\\
189.770	0.300\\
189.810	0.300\\
189.929	0.300\\
190.090	0.300\\
190.429	0.300\\
190.590	0.300\\
190.800	0.300\\
191.410	0.300\\
191.722	0.300\\
192.221	0.300\\
192.539	0.300\\
192.749	0.300\\
192.880	0.300\\
192.930	0.300\\
193.019	0.300\\
193.480	0.300\\
193.940	0.300\\
194.070	0.300\\
194.660	0.205\\
195.070	-0.250\\
195.619	-0.255\\
195.880	-0.255\\
196.270	-0.255\\
196.622	-0.255\\
197.039	-0.255\\
197.080	-0.255\\
197.870	-0.260\\
198.340	-0.260\\
198.559	-0.257\\
198.900	-0.257\\
199.540	-0.257\\
199.719	-0.257\\
199.950	-0.257\\
200.151	-0.257\\
200.729	-0.257\\
201.139	-0.257\\
201.280	-0.257\\
201.569	-0.257\\
202.141	-0.257\\
202.269	-0.257\\
202.380	-0.257\\
202.579	-0.257\\
203.149	-0.257\\
203.329	-0.257\\
203.779	-0.257\\
204.200	-0.257\\
204.719	-0.257\\
204.909	-0.257\\
205.749	-0.276\\
205.919	-0.271\\
206.150	0.036\\
206.760	0.300\\
207.049	0.300\\
207.209	0.300\\
207.539	0.300\\
207.840	0.300\\
208.049	0.300\\
208.399	0.300\\
208.959	0.300\\
209.039	0.300\\
209.369	0.300\\
209.549	0.300\\
209.660	0.300\\
210.339	0.300\\
211.419	0.300\\
211.699	0.300\\
211.840	0.300\\
211.961	0.300\\
212.199	0.300\\
212.770	0.300\\
212.969	0.300\\
214.229	0.300\\
214.750	0.300\\
215.199	0.300\\
215.519	0.300\\
215.940	0.300\\
216.129	0.300\\
216.849	0.300\\
217.319	0.057\\
217.669	-0.255\\
218.159	-0.255\\
218.250	-0.255\\
218.359	-0.255\\
219.059	-0.255\\
219.499	-0.255\\
219.989	-0.255\\
221.060	-0.255\\
221.529	-0.255\\
221.949	-0.260\\
222.129	-0.271\\
222.389	-0.276\\
223.349	-0.255\\
223.539	-0.255\\
223.639	-0.257\\
224.039	-0.255\\
224.249	-0.257\\
224.470	-0.255\\
224.569	-0.255\\
224.709	-0.257\\
225.099	-0.257\\
225.410	-0.257\\
225.589	-0.257\\
226.479	-0.257\\
227.270	-0.148\\
227.400	0.000\\
227.609	0.300\\
227.869	0.300\\
228.299	0.300\\
228.559	0.300\\
228.789	0.300\\
229.120	0.300\\
229.439	0.300\\
229.869	0.300\\
230.169	0.300\\
230.449	0.300\\
230.609	0.300\\
230.719	0.300\\
230.839	0.300\\
231.129	0.300\\
231.259	0.300\\
231.439	0.300\\
231.929	0.300\\
232.290	0.300\\
232.609	0.300\\
232.889	0.300\\
233.110	0.195\\
233.251	0.114\\
233.549	0.129\\
233.859	0.171\\
234.349	0.283\\
234.889	0.283\\
235.030	0.300\\
235.559	0.300\\
235.639	0.300\\
235.799	0.300\\
235.949	0.300\\
236.059	0.300\\
236.151	0.300\\
236.249	0.300\\
236.999	0.300\\
237.439	0.300\\
237.589	0.300\\
237.649	0.300\\
237.779	0.300\\
239.599	-0.255\\
240.019	-0.255\\
240.239	-0.255\\
240.289	-0.255\\
240.429	-0.255\\
240.779	-0.255\\
241.100	-0.255\\
241.209	-0.255\\
241.369	-0.255\\
241.769	-0.255\\
242.089	-0.271\\
242.719	-0.271\\
242.889	-0.271\\
243.269	-0.271\\
243.389	-0.271\\
243.730	-0.271\\
243.869	-0.271\\
244.240	-0.257\\
244.359	-0.257\\
244.469	-0.255\\
245.939	-0.255\\
246.779	-0.255\\
247.189	-0.255\\
247.529	-0.255\\
247.622	-0.255\\
247.729	-0.255\\
248.249	0.226\\
248.729	0.300\\
249.129	0.300\\
249.669	0.300\\
249.729	0.300\\
249.869	0.300\\
250.239	0.300\\
250.359	0.300\\
250.539	0.300\\
250.660	0.300\\
251.549	0.300\\
251.699	0.300\\
252.269	0.300\\
252.489	0.300\\
252.869	0.300\\
252.999	0.300\\
253.260	0.300\\
253.709	0.300\\
254.440	0.300\\
254.999	0.300\\
255.271	0.300\\
255.589	0.300\\
256.039	0.300\\
256.509	0.300\\
256.649	0.300\\
256.930	0.300\\
257.009	0.300\\
257.301	0.300\\
258.669	0.300\\
259.149	0.000\\
259.279	-0.255\\
259.459	-0.255\\
259.559	-0.255\\
259.949	-0.255\\
261.049	-0.255\\
261.569	-0.255\\
261.819	-0.255\\
261.979	-0.255\\
262.109	-0.255\\
262.189	-0.255\\
262.599	-0.255\\
262.819	-0.255\\
263.019	-0.255\\
263.319	-0.290\\
263.620	-0.290\\
264.469	-0.260\\
265.189	-0.260\\
265.469	-0.260\\
265.949	-0.260\\
266.279	-0.260\\
266.419	-0.260\\
266.789	-0.260\\
267.179	-0.260\\
267.459	-0.260\\
267.799	-0.260\\
268.319	-0.260\\
268.429	-0.260\\
268.629	-0.260\\
268.769	-0.260\\
268.929	-0.260\\
269.129	-0.260\\
269.439	-0.260\\
269.549	-0.260\\
269.939	-0.260\\
270.379	-0.260\\
271.259	-0.164\\
271.549	0.000\\
271.669	0.000\\
271.990	0.274\\
272.459	0.300\\
272.569	0.300\\
273.309	0.300\\
274.199	0.300\\
274.379	0.300\\
274.749	0.300\\
274.899	0.300\\
275.459	0.300\\
275.839	0.300\\
275.969	0.300\\
277.369	0.300\\
277.639	0.300\\
277.979	0.300\\
278.129	0.300\\
278.429	0.300\\
278.779	0.300\\
278.901	0.300\\
278.979	0.300\\
279.099	0.300\\
279.569	0.300\\
279.929	0.300\\
280.290	0.300\\
};
\addlegendentry{Speed Reference};

\addplot [color=black,dashed,forget plot]
  table[row sep=crcr]{%
148.000	0.000\\
nan	nan\\
292.000	0.000\\
};
\end{axis}
\end{tikzpicture}%

%% file: images/pocket_velocity_estimate_control_ref_ver.tikz
%
%
\begin{tikzpicture}

\begin{axis}[%
width=0.95092\figurewidth,
height=\figureheight,
at={(0\figurewidth,0\figureheight)},
scale only axis,
unbounded coords=jump,
every outer x axis line/.append style={black},
every x tick label/.append style={font=\color{black}},
xmin=160.000,
xmax=280.000,
xlabel={time [s]},
every outer y axis line/.append style={black},
every y tick label/.append style={font=\color{black}},
ymin=-0.500,
ymax=0.500,
ylabel={velocity [m/s]},
axis x line*=bottom,
axis y line*=left
]
\addplot [color=blue,solid,forget plot]
  table[row sep=crcr]{%
159.920	-0.030\\
160.030	-0.030\\
160.090	-0.030\\
160.170	-0.030\\
160.312	-0.026\\
160.390	-0.016\\
160.470	-0.002\\
160.591	0.010\\
160.750	0.024\\
161.062	0.032\\
161.120	0.034\\
161.220	0.034\\
161.420	0.032\\
161.540	0.028\\
161.771	0.026\\
161.891	0.022\\
162.032	0.014\\
162.100	0.010\\
162.210	0.006\\
162.280	0.004\\
162.420	0.004\\
162.541	0.000\\
162.650	-0.006\\
162.740	-0.012\\
162.840	-0.020\\
162.970	-0.028\\
163.010	-0.030\\
163.140	-0.030\\
163.390	-0.030\\
163.600	-0.030\\
163.690	-0.030\\
163.770	-0.030\\
163.850	-0.028\\
163.990	-0.022\\
164.020	-0.016\\
164.160	-0.010\\
164.210	-0.002\\
164.340	0.004\\
164.620	0.006\\
164.660	0.008\\
164.800	0.010\\
164.881	0.010\\
165.110	0.010\\
165.220	0.014\\
165.370	0.018\\
165.470	0.022\\
165.670	0.026\\
165.780	0.030\\
165.880	0.030\\
165.950	0.030\\
166.080	0.030\\
166.170	0.030\\
166.270	0.028\\
166.310	0.022\\
166.510	0.016\\
166.670	0.006\\
166.730	-0.004\\
166.841	-0.014\\
167.060	-0.022\\
167.180	-0.028\\
167.300	-0.030\\
167.420	-0.032\\
167.490	-0.032\\
167.750	-0.032\\
167.850	-0.032\\
167.930	-0.032\\
168.041	-0.032\\
168.080	-0.032\\
168.220	-0.030\\
168.340	-0.030\\
168.400	-0.030\\
168.490	-0.030\\
168.540	-0.030\\
168.710	-0.030\\
168.751	-0.028\\
168.890	-0.026\\
169.010	-0.024\\
169.080	-0.018\\
169.230	-0.012\\
169.300	-0.008\\
169.410	-0.004\\
169.480	0.000\\
169.811	0.000\\
169.910	0.000\\
170.000	0.000\\
170.030	0.000\\
170.310	0.000\\
170.490	0.000\\
170.590	0.000\\
170.671	0.000\\
170.820	0.000\\
170.881	0.000\\
170.940	0.000\\
171.010	0.000\\
171.180	0.000\\
171.380	0.000\\
171.660	0.000\\
171.710	0.002\\
171.930	0.004\\
172.000	0.006\\
172.120	0.008\\
172.200	0.010\\
172.331	0.010\\
172.500	0.010\\
172.572	0.010\\
172.690	0.012\\
172.964	0.014\\
173.020	0.018\\
173.130	0.022\\
173.220	0.026\\
173.440	0.028\\
173.731	0.030\\
173.820	0.030\\
173.990	0.030\\
174.501	0.030\\
174.530	0.030\\
174.612	0.030\\
174.712	0.030\\
175.040	0.030\\
175.120	0.030\\
175.240	0.030\\
175.300	0.030\\
175.440	0.030\\
175.470	0.030\\
175.680	0.030\\
175.760	0.030\\
175.870	0.030\\
176.040	0.030\\
176.170	0.030\\
176.290	0.032\\
176.520	0.032\\
176.601	0.034\\
176.720	0.024\\
176.870	0.016\\
177.032	0.010\\
177.250	0.010\\
177.340	0.012\\
177.420	0.038\\
177.540	0.056\\
177.701	0.066\\
177.820	0.070\\
177.850	0.078\\
177.910	0.078\\
178.040	0.078\\
178.090	0.076\\
178.220	0.078\\
178.410	0.070\\
178.520	0.060\\
178.580	0.050\\
178.660	0.046\\
178.740	0.040\\
178.900	0.032\\
179.050	0.020\\
179.170	0.014\\
179.390	0.008\\
179.470	0.002\\
179.650	0.000\\
179.720	0.000\\
179.850	0.000\\
179.911	0.000\\
180.020	0.000\\
180.120	0.000\\
180.160	0.000\\
180.220	0.000\\
180.290	-0.004\\
180.360	-0.010\\
180.440	-0.016\\
180.662	-0.022\\
180.770	-0.028\\
180.840	-0.030\\
181.180	-0.030\\
181.280	-0.034\\
181.310	-0.036\\
181.550	-0.036\\
181.590	-0.036\\
181.730	-0.038\\
181.830	-0.034\\
181.920	-0.034\\
182.010	-0.042\\
182.100	-0.048\\
182.220	-0.052\\
182.301	-0.052\\
182.530	-0.050\\
182.770	-0.048\\
182.851	-0.044\\
183.020	-0.038\\
183.050	-0.040\\
183.260	-0.040\\
183.450	-0.034\\
183.550	-0.034\\
183.630	-0.034\\
183.760	-0.032\\
183.890	-0.032\\
183.960	-0.032\\
184.030	-0.030\\
184.180	-0.030\\
184.230	-0.034\\
184.381	-0.038\\
184.490	-0.038\\
184.520	-0.042\\
184.620	-0.044\\
184.790	-0.040\\
184.840	-0.036\\
184.980	-0.040\\
185.060	-0.044\\
185.390	-0.042\\
185.510	-0.042\\
185.590	-0.042\\
185.850	-0.038\\
185.990	-0.030\\
186.070	-0.030\\
186.110	-0.030\\
186.200	-0.032\\
186.420	-0.036\\
186.520	-0.040\\
186.600	-0.042\\
186.700	-0.042\\
186.770	-0.040\\
186.902	-0.036\\
186.960	-0.032\\
187.000	-0.032\\
187.090	-0.032\\
187.130	-0.032\\
187.210	-0.032\\
187.300	-0.036\\
187.500	-0.034\\
187.540	-0.038\\
187.660	-0.042\\
187.740	-0.048\\
187.840	-0.048\\
187.990	-0.052\\
188.100	-0.048\\
188.180	-0.048\\
188.260	-0.042\\
188.321	-0.038\\
188.361	-0.034\\
188.540	-0.036\\
188.620	-0.036\\
188.710	-0.038\\
188.790	-0.038\\
188.870	-0.038\\
189.030	-0.036\\
189.140	-0.032\\
189.180	-0.028\\
189.321	-0.022\\
189.360	-0.014\\
189.520	-0.006\\
189.620	0.004\\
189.700	0.010\\
189.861	0.016\\
190.030	0.020\\
190.211	0.022\\
190.380	0.026\\
190.460	0.030\\
190.710	0.030\\
190.900	0.030\\
191.060	0.032\\
191.181	0.030\\
191.222	0.030\\
191.370	0.030\\
191.470	0.030\\
191.590	0.030\\
191.670	0.030\\
191.880	0.030\\
191.960	0.030\\
192.090	0.030\\
192.170	0.030\\
192.280	0.030\\
192.460	0.030\\
192.509	0.030\\
192.670	0.030\\
192.810	0.030\\
193.171	0.030\\
193.310	0.030\\
193.340	0.030\\
193.440	0.030\\
193.550	0.030\\
193.660	0.030\\
193.700	0.030\\
193.759	0.030\\
193.830	0.030\\
194.030	0.030\\
194.152	0.028\\
194.300	0.026\\
194.490	0.024\\
194.540	0.022\\
194.700	0.014\\
194.800	0.008\\
194.949	0.002\\
194.990	-0.004\\
195.320	-0.010\\
195.479	-0.010\\
195.580	-0.010\\
195.660	-0.010\\
195.820	-0.010\\
195.940	-0.010\\
196.040	-0.010\\
196.130	-0.010\\
196.313	-0.010\\
196.450	-0.010\\
196.520	-0.010\\
196.590	-0.010\\
196.780	-0.010\\
196.820	-0.010\\
196.930	-0.010\\
196.960	-0.010\\
197.140	-0.010\\
197.232	-0.010\\
197.320	-0.010\\
197.350	-0.010\\
197.480	-0.010\\
197.589	-0.012\\
197.693	-0.016\\
197.800	-0.020\\
197.969	-0.024\\
198.050	-0.026\\
198.160	-0.026\\
198.529	-0.024\\
198.620	-0.022\\
198.710	-0.020\\
198.860	-0.020\\
198.940	-0.018\\
199.030	-0.012\\
199.070	-0.004\\
199.150	0.004\\
199.250	0.012\\
199.330	0.020\\
199.419	0.024\\
199.500	0.026\\
199.580	0.028\\
199.701	0.030\\
199.989	0.030\\
200.070	0.030\\
200.231	0.030\\
200.270	0.028\\
200.390	0.024\\
200.489	0.018\\
200.610	0.006\\
200.650	-0.006\\
200.840	-0.016\\
200.950	-0.024\\
201.020	-0.030\\
201.110	-0.032\\
201.231	-0.032\\
201.350	-0.032\\
201.449	-0.032\\
201.499	-0.034\\
201.700	-0.036\\
201.960	-0.038\\
202.020	-0.044\\
202.109	-0.046\\
202.350	-0.048\\
202.530	-0.044\\
202.659	-0.042\\
202.889	-0.036\\
203.000	-0.034\\
203.101	-0.030\\
203.400	-0.030\\
203.500	-0.030\\
203.590	-0.030\\
203.720	-0.030\\
203.839	-0.030\\
203.929	-0.030\\
204.010	-0.026\\
204.150	-0.022\\
204.299	-0.018\\
204.380	-0.014\\
204.529	-0.010\\
204.610	-0.010\\
204.670	-0.008\\
204.790	-0.008\\
205.070	-0.010\\
205.129	-0.012\\
205.209	-0.014\\
205.330	-0.018\\
205.400	-0.020\\
205.489	-0.020\\
205.620	-0.026\\
205.660	-0.030\\
205.800	-0.032\\
205.879	-0.034\\
206.210	-0.034\\
206.289	-0.022\\
206.369	-0.012\\
206.520	-0.004\\
206.560	0.004\\
206.709	0.010\\
206.820	0.010\\
206.881	0.010\\
206.961	0.010\\
206.990	0.010\\
207.170	0.010\\
207.290	0.014\\
207.440	0.018\\
207.710	0.022\\
207.779	0.026\\
207.899	0.030\\
207.970	0.026\\
208.090	0.022\\
208.270	0.018\\
208.320	0.012\\
208.439	0.006\\
208.530	0.004\\
208.600	-0.002\\
208.730	-0.010\\
208.789	-0.016\\
208.840	-0.022\\
208.880	-0.028\\
209.070	-0.032\\
209.330	-0.032\\
209.449	-0.032\\
209.489	-0.032\\
209.589	-0.032\\
209.830	-0.030\\
209.860	-0.030\\
209.960	-0.030\\
210.040	-0.030\\
210.100	-0.030\\
210.160	-0.030\\
210.230	-0.026\\
210.440	-0.020\\
210.520	-0.012\\
210.640	0.002\\
211.030	0.014\\
211.200	0.022\\
211.300	0.028\\
211.380	0.032\\
211.520	0.030\\
211.579	0.030\\
211.619	0.030\\
212.040	0.028\\
212.159	0.026\\
212.230	0.024\\
212.431	0.022\\
212.579	0.020\\
212.630	0.020\\
212.730	0.020\\
212.880	0.020\\
213.081	0.020\\
213.150	0.020\\
213.190	0.020\\
213.309	0.020\\
213.434	0.020\\
213.490	0.020\\
213.530	0.022\\
213.580	0.026\\
213.661	0.028\\
213.751	0.030\\
213.811	0.032\\
213.970	0.032\\
214.060	0.030\\
214.189	0.030\\
214.289	0.032\\
214.381	0.032\\
214.500	0.032\\
214.610	0.032\\
214.690	0.032\\
214.810	0.032\\
215.059	0.032\\
215.160	0.032\\
215.380	0.032\\
215.441	0.032\\
215.590	0.030\\
215.651	0.030\\
215.760	0.032\\
215.831	0.036\\
216.090	0.036\\
216.230	0.038\\
216.330	0.040\\
216.419	0.040\\
216.560	0.040\\
216.791	0.042\\
216.901	0.044\\
217.020	0.040\\
217.080	0.036\\
217.269	0.030\\
217.360	0.028\\
217.451	0.024\\
217.730	0.024\\
217.809	0.024\\
217.870	0.024\\
217.949	0.022\\
218.029	0.020\\
218.182	0.020\\
218.322	0.020\\
218.481	0.020\\
218.599	0.020\\
218.679	0.020\\
218.851	0.020\\
218.860	0.020\\
218.980	0.020\\
219.010	0.020\\
219.290	0.016\\
219.399	0.012\\
219.584	0.004\\
219.730	-0.006\\
219.850	-0.016\\
220.050	-0.022\\
220.159	-0.028\\
220.280	-0.032\\
220.379	-0.036\\
220.480	-0.038\\
220.564	-0.038\\
220.730	-0.038\\
220.880	-0.036\\
220.939	-0.032\\
221.009	-0.030\\
221.129	-0.030\\
221.379	-0.030\\
221.491	-0.030\\
221.569	-0.030\\
221.649	-0.030\\
221.789	-0.030\\
221.880	-0.030\\
221.989	-0.030\\
222.060	-0.030\\
222.200	-0.030\\
222.439	-0.030\\
222.540	-0.030\\
222.590	-0.028\\
222.683	-0.014\\
222.841	0.004\\
222.950	0.024\\
223.160	0.042\\
223.230	0.060\\
223.379	0.066\\
223.461	0.064\\
223.599	0.066\\
223.799	0.066\\
223.920	0.066\\
224.009	0.064\\
224.120	0.064\\
224.179	0.060\\
224.371	0.054\\
224.431	0.048\\
224.791	0.044\\
224.889	0.040\\
224.970	0.034\\
225.179	0.034\\
225.270	0.032\\
225.370	0.030\\
225.489	0.034\\
225.549	0.034\\
225.760	0.036\\
225.829	0.038\\
225.939	0.038\\
225.969	0.034\\
226.029	0.034\\
226.210	0.032\\
226.249	0.030\\
226.350	0.032\\
226.559	0.032\\
226.659	0.032\\
226.699	0.032\\
226.839	0.032\\
226.970	0.030\\
227.079	0.026\\
227.160	0.020\\
227.469	0.014\\
227.670	0.008\\
227.749	0.002\\
227.830	0.000\\
227.919	-0.006\\
228.019	-0.012\\
228.179	-0.018\\
228.259	-0.024\\
228.380	-0.030\\
228.430	-0.030\\
228.590	-0.030\\
228.819	-0.030\\
228.880	-0.030\\
229.040	-0.030\\
229.200	-0.030\\
229.280	-0.030\\
229.409	-0.036\\
229.549	-0.036\\
229.680	-0.036\\
229.929	-0.036\\
230.009	-0.036\\
230.129	-0.030\\
230.229	-0.030\\
230.341	-0.030\\
230.391	-0.030\\
230.490	-0.032\\
230.920	-0.032\\
231.030	-0.042\\
231.219	-0.054\\
231.380	-0.054\\
231.539	-0.052\\
231.730	-0.052\\
231.811	-0.028\\
231.859	-0.004\\
232.151	0.010\\
232.219	0.022\\
232.349	0.036\\
232.470	0.034\\
232.569	0.034\\
232.730	0.034\\
232.829	0.034\\
233.009	0.034\\
233.070	0.038\\
233.330	0.038\\
233.429	0.036\\
233.611	0.036\\
233.749	0.034\\
233.819	0.030\\
233.919	0.030\\
234.009	0.030\\
234.039	0.030\\
234.100	0.028\\
234.249	0.028\\
234.411	0.028\\
234.489	0.028\\
234.570	0.028\\
234.630	0.030\\
234.679	0.030\\
234.760	0.030\\
234.840	0.030\\
234.949	0.030\\
235.101	0.030\\
235.319	0.030\\
235.409	0.030\\
235.480	0.028\\
235.719	0.028\\
235.890	0.028\\
236.019	0.028\\
236.179	0.028\\
236.409	0.030\\
236.489	0.030\\
236.579	0.030\\
236.729	0.036\\
236.809	0.040\\
236.959	0.046\\
237.039	0.054\\
237.129	0.064\\
237.209	0.068\\
237.250	0.068\\
237.389	0.072\\
237.560	0.070\\
237.749	0.064\\
237.930	0.054\\
238.060	0.050\\
238.120	0.040\\
238.209	0.034\\
238.320	0.030\\
238.449	0.030\\
238.499	0.030\\
238.539	0.030\\
238.659	0.030\\
238.719	0.034\\
238.800	0.040\\
238.989	0.044\\
239.080	0.044\\
239.199	0.046\\
239.272	0.042\\
239.369	0.036\\
239.489	0.032\\
239.540	0.032\\
239.799	0.030\\
239.859	0.030\\
239.990	0.030\\
240.090	0.030\\
240.159	0.030\\
240.350	0.028\\
240.549	0.026\\
240.650	0.026\\
240.709	0.026\\
240.849	0.026\\
240.909	0.028\\
241.049	0.032\\
241.130	0.032\\
241.269	0.032\\
241.629	0.030\\
241.730	0.030\\
241.809	0.032\\
241.930	0.034\\
241.970	0.036\\
242.150	0.038\\
242.219	0.056\\
242.259	0.066\\
242.379	0.070\\
242.460	0.072\\
242.599	0.074\\
242.679	0.056\\
242.971	0.044\\
243.029	0.038\\
243.209	0.034\\
243.340	0.032\\
243.509	0.032\\
243.570	0.030\\
243.670	0.030\\
243.809	0.030\\
243.910	0.030\\
243.959	0.030\\
244.039	0.030\\
244.211	0.024\\
244.299	0.014\\
244.400	0.004\\
244.530	-0.008\\
244.580	-0.020\\
244.719	-0.026\\
244.799	-0.028\\
244.909	-0.030\\
244.999	-0.030\\
245.090	-0.030\\
245.249	-0.030\\
245.320	-0.030\\
245.409	-0.030\\
245.479	-0.020\\
245.539	-0.010\\
245.680	0.000\\
245.879	0.010\\
245.999	0.020\\
246.069	0.020\\
246.129	0.022\\
246.250	0.024\\
246.350	0.028\\
246.409	0.030\\
246.450	0.032\\
246.619	0.032\\
246.699	0.032\\
246.859	0.030\\
246.919	0.034\\
246.999	0.034\\
247.159	0.034\\
247.369	0.040\\
247.459	0.042\\
247.490	0.036\\
247.669	0.034\\
247.870	0.032\\
247.951	0.024\\
248.029	0.020\\
248.152	0.020\\
248.189	0.020\\
248.369	0.020\\
248.489	0.020\\
248.549	0.020\\
248.609	0.020\\
248.660	0.020\\
249.019	0.020\\
249.089	0.020\\
249.189	0.020\\
249.269	0.024\\
249.349	0.028\\
249.509	0.028\\
249.589	0.028\\
249.789	0.028\\
249.950	0.024\\
250.009	0.020\\
250.089	0.020\\
250.139	0.020\\
250.200	0.020\\
250.319	0.016\\
250.499	0.010\\
250.579	0.004\\
250.679	-0.002\\
250.759	-0.008\\
250.830	-0.010\\
250.960	-0.010\\
251.009	-0.010\\
251.119	-0.010\\
251.219	-0.008\\
251.330	-0.006\\
251.660	-0.004\\
251.739	0.000\\
251.929	0.004\\
252.149	0.006\\
252.229	0.008\\
252.349	0.010\\
252.409	0.010\\
252.609	0.010\\
252.729	0.010\\
252.801	0.010\\
252.920	0.010\\
253.089	0.010\\
253.139	0.010\\
253.460	0.010\\
253.599	0.010\\
253.669	0.010\\
253.800	0.010\\
253.869	0.010\\
253.909	0.014\\
254.030	0.018\\
254.309	0.022\\
254.369	0.026\\
254.490	0.030\\
254.579	0.030\\
254.699	0.030\\
254.760	0.030\\
254.820	0.030\\
254.899	0.030\\
255.239	0.030\\
255.319	0.030\\
255.409	0.026\\
255.739	0.020\\
256.000	0.014\\
256.079	0.004\\
256.152	-0.006\\
256.299	-0.012\\
256.369	-0.018\\
256.439	-0.028\\
256.609	-0.032\\
256.769	-0.034\\
256.892	-0.034\\
257.039	-0.032\\
257.340	-0.026\\
257.439	-0.022\\
257.479	-0.020\\
257.599	-0.020\\
257.749	-0.020\\
257.840	-0.022\\
257.949	-0.024\\
258.069	-0.026\\
258.149	-0.028\\
258.309	-0.030\\
258.389	-0.030\\
258.509	-0.030\\
258.599	-0.030\\
258.871	-0.030\\
258.969	-0.036\\
259.059	-0.038\\
259.239	-0.038\\
259.379	-0.040\\
259.500	-0.040\\
259.731	-0.034\\
259.800	-0.032\\
259.909	-0.032\\
260.019	-0.030\\
260.089	-0.030\\
260.209	-0.036\\
260.282	-0.048\\
260.429	-0.056\\
260.520	-0.056\\
260.650	-0.056\\
260.759	-0.050\\
260.940	-0.038\\
261.020	-0.030\\
261.109	-0.030\\
261.189	-0.030\\
261.229	-0.030\\
261.349	-0.030\\
261.519	-0.030\\
261.610	-0.030\\
261.719	-0.030\\
261.759	-0.030\\
261.859	-0.032\\
262.030	-0.030\\
262.289	-0.028\\
262.490	-0.026\\
262.569	-0.024\\
262.690	-0.020\\
262.751	-0.016\\
262.899	-0.010\\
263.079	-0.004\\
263.139	0.002\\
263.189	0.012\\
263.289	0.018\\
263.459	0.022\\
263.539	0.026\\
263.672	0.032\\
263.751	0.032\\
263.800	0.032\\
263.850	0.032\\
263.962	0.032\\
264.119	0.028\\
264.209	0.026\\
264.319	0.022\\
264.409	0.018\\
264.560	0.014\\
264.659	0.012\\
264.699	0.010\\
264.769	0.010\\
264.869	0.010\\
264.919	0.010\\
265.109	0.010\\
265.229	0.010\\
265.299	0.010\\
265.349	0.010\\
265.429	0.010\\
265.549	0.008\\
265.659	0.004\\
265.699	0.000\\
265.749	-0.008\\
265.849	-0.018\\
266.049	-0.024\\
266.119	-0.032\\
266.249	-0.038\\
266.479	-0.042\\
266.519	-0.042\\
266.659	-0.044\\
266.699	-0.048\\
266.849	-0.052\\
266.989	-0.054\\
267.069	-0.056\\
267.251	-0.060\\
267.329	-0.056\\
267.379	-0.050\\
267.569	-0.044\\
267.729	-0.042\\
267.919	-0.036\\
268.019	-0.032\\
268.089	-0.032\\
268.240	-0.032\\
268.349	-0.030\\
268.469	-0.032\\
268.850	-0.032\\
268.969	-0.032\\
269.069	-0.032\\
269.289	-0.034\\
269.400	-0.030\\
269.499	-0.028\\
269.579	-0.028\\
269.639	-0.028\\
269.699	-0.026\\
269.879	-0.044\\
269.999	-0.066\\
270.209	-0.084\\
270.309	-0.096\\
270.421	-0.102\\
270.509	-0.102\\
270.629	-0.094\\
270.749	-0.084\\
270.909	-0.076\\
270.980	-0.080\\
271.079	-0.074\\
271.370	-0.068\\
271.429	-0.060\\
271.510	-0.056\\
271.899	-0.046\\
271.940	-0.036\\
272.079	-0.030\\
272.119	-0.030\\
272.239	-0.030\\
272.299	-0.030\\
272.329	-0.030\\
272.530	-0.030\\
272.629	-0.030\\
272.769	-0.030\\
272.929	-0.030\\
273.049	-0.030\\
273.189	-0.030\\
273.230	-0.030\\
273.369	-0.030\\
273.500	-0.030\\
273.529	-0.030\\
273.640	-0.030\\
273.769	-0.030\\
273.819	-0.030\\
273.949	-0.030\\
274.009	-0.030\\
274.089	-0.030\\
274.249	-0.032\\
274.520	-0.032\\
274.609	-0.032\\
274.699	-0.032\\
274.829	-0.032\\
275.029	-0.030\\
275.110	-0.032\\
275.150	-0.032\\
275.289	-0.032\\
275.350	-0.032\\
275.419	-0.034\\
275.579	-0.032\\
275.639	-0.032\\
275.729	-0.032\\
275.919	-0.032\\
276.079	-0.030\\
276.219	-0.030\\
276.299	-0.030\\
276.379	-0.030\\
276.471	-0.030\\
276.589	-0.030\\
276.709	-0.030\\
276.799	-0.030\\
276.889	-0.030\\
276.929	-0.030\\
277.089	-0.030\\
277.209	-0.034\\
277.329	-0.034\\
277.399	-0.034\\
277.519	-0.034\\
277.719	-0.034\\
277.779	-0.030\\
277.859	-0.032\\
278.019	-0.032\\
278.219	-0.032\\
278.269	-0.032\\
278.469	-0.034\\
278.569	-0.032\\
278.689	-0.032\\
278.729	-0.032\\
278.869	-0.032\\
279.019	-0.030\\
279.199	-0.030\\
279.299	-0.034\\
279.379	-0.040\\
279.509	-0.044\\
279.699	-0.046\\
279.789	-0.046\\
279.849	-0.046\\
279.969	-0.042\\
280.109	-0.042\\
};
\addplot [color=red,solid,forget plot]
  table[row sep=crcr]{%
159.920	-0.058\\
160.030	-0.056\\
160.090	-0.052\\
160.170	-0.048\\
160.312	-0.038\\
160.390	-0.030\\
160.470	-0.022\\
160.591	-0.014\\
160.750	-0.006\\
161.062	-0.004\\
161.120	-0.002\\
161.220	0.000\\
161.420	0.000\\
161.540	-0.002\\
161.771	-0.008\\
161.891	-0.014\\
162.032	-0.020\\
162.100	-0.026\\
162.210	-0.030\\
162.280	-0.030\\
162.420	-0.030\\
162.541	-0.028\\
162.650	-0.028\\
162.740	-0.028\\
162.840	-0.028\\
162.970	-0.026\\
163.010	-0.026\\
163.140	-0.022\\
163.390	-0.018\\
163.600	-0.012\\
163.690	-0.008\\
163.770	-0.004\\
163.850	-0.002\\
163.990	0.000\\
164.020	0.000\\
164.160	0.004\\
164.210	0.008\\
164.340	0.010\\
164.620	0.012\\
164.660	0.016\\
164.800	0.018\\
164.881	0.020\\
165.110	0.024\\
165.220	0.030\\
165.370	0.032\\
165.470	0.030\\
165.670	0.028\\
165.780	0.026\\
165.880	0.022\\
165.950	0.020\\
166.080	0.020\\
166.170	0.020\\
166.270	0.018\\
166.310	0.016\\
166.510	0.012\\
166.670	0.008\\
166.730	0.004\\
166.841	0.002\\
167.060	0.000\\
167.180	0.000\\
167.300	0.000\\
167.420	0.000\\
167.490	0.000\\
167.750	0.000\\
167.850	0.000\\
167.930	0.000\\
168.041	0.000\\
168.080	0.000\\
168.220	0.000\\
168.340	0.000\\
168.400	0.000\\
168.490	0.000\\
168.540	0.000\\
168.710	0.000\\
168.751	0.008\\
168.890	0.016\\
169.010	0.024\\
169.080	0.032\\
169.230	0.040\\
169.300	0.040\\
169.410	0.040\\
169.480	0.040\\
169.811	0.040\\
169.910	0.040\\
170.000	0.044\\
170.030	0.046\\
170.310	0.048\\
170.490	0.050\\
170.590	0.052\\
170.671	0.050\\
170.820	0.052\\
170.881	0.054\\
170.940	0.056\\
171.010	0.058\\
171.180	0.060\\
171.380	0.060\\
171.660	0.060\\
171.710	0.060\\
171.930	0.062\\
172.000	0.064\\
172.120	0.064\\
172.200	0.064\\
172.331	0.064\\
172.500	0.062\\
172.572	0.062\\
172.690	0.064\\
172.964	0.066\\
173.020	0.070\\
173.130	0.072\\
173.220	0.074\\
173.440	0.076\\
173.731	0.074\\
173.820	0.072\\
173.990	0.072\\
174.501	0.070\\
174.530	0.066\\
174.612	0.060\\
174.712	0.052\\
175.040	0.042\\
175.120	0.032\\
175.240	0.024\\
175.300	0.022\\
175.440	0.020\\
175.470	0.020\\
175.680	0.020\\
175.760	0.018\\
175.870	0.018\\
176.040	0.018\\
176.170	0.016\\
176.290	0.014\\
176.520	0.014\\
176.601	0.012\\
176.720	0.010\\
176.870	0.010\\
177.032	0.010\\
177.250	0.014\\
177.340	0.018\\
177.420	0.026\\
177.540	0.034\\
177.701	0.042\\
177.820	0.046\\
177.850	0.050\\
177.910	0.050\\
178.040	0.050\\
178.090	0.048\\
178.220	0.042\\
178.410	0.036\\
178.520	0.030\\
178.580	0.020\\
178.660	0.010\\
178.740	0.004\\
178.900	-0.002\\
179.050	-0.016\\
179.170	-0.026\\
179.390	-0.034\\
179.470	-0.042\\
179.650	-0.050\\
179.720	-0.050\\
179.850	-0.050\\
179.911	-0.050\\
180.020	-0.052\\
180.120	-0.054\\
180.160	-0.056\\
180.220	-0.060\\
180.290	-0.064\\
180.360	-0.068\\
180.440	-0.072\\
180.662	-0.076\\
180.770	-0.076\\
180.840	-0.076\\
181.180	-0.074\\
181.280	-0.072\\
181.310	-0.070\\
181.550	-0.072\\
181.590	-0.074\\
181.730	-0.078\\
181.830	-0.082\\
181.920	-0.086\\
182.010	-0.088\\
182.100	-0.090\\
182.220	-0.090\\
182.301	-0.090\\
182.530	-0.090\\
182.770	-0.088\\
182.851	-0.084\\
183.020	-0.080\\
183.050	-0.074\\
183.260	-0.068\\
183.450	-0.064\\
183.550	-0.062\\
183.630	-0.060\\
183.760	-0.056\\
183.890	-0.052\\
183.960	-0.050\\
184.030	-0.048\\
184.180	-0.048\\
184.230	-0.052\\
184.381	-0.056\\
184.490	-0.058\\
184.520	-0.064\\
184.620	-0.068\\
184.790	-0.072\\
184.840	-0.076\\
184.980	-0.076\\
185.060	-0.072\\
185.390	-0.070\\
185.510	-0.066\\
185.590	-0.062\\
185.850	-0.062\\
185.990	-0.062\\
186.070	-0.062\\
186.110	-0.062\\
186.200	-0.062\\
186.420	-0.062\\
186.520	-0.062\\
186.600	-0.060\\
186.700	-0.060\\
186.770	-0.058\\
186.902	-0.056\\
186.960	-0.054\\
187.000	-0.052\\
187.090	-0.050\\
187.130	-0.050\\
187.210	-0.050\\
187.300	-0.052\\
187.500	-0.054\\
187.540	-0.056\\
187.660	-0.058\\
187.740	-0.058\\
187.840	-0.056\\
187.990	-0.052\\
188.100	-0.048\\
188.180	-0.042\\
188.260	-0.038\\
188.321	-0.028\\
188.361	-0.020\\
188.540	-0.014\\
188.620	-0.010\\
188.710	-0.004\\
188.790	-0.004\\
188.870	-0.002\\
189.030	0.002\\
189.140	0.008\\
189.180	0.012\\
189.321	0.016\\
189.360	0.024\\
189.520	0.032\\
189.620	0.040\\
189.700	0.048\\
189.861	0.058\\
190.030	0.062\\
190.211	0.066\\
190.380	0.068\\
190.460	0.070\\
190.710	0.072\\
190.900	0.074\\
191.060	0.076\\
191.181	0.074\\
191.222	0.072\\
191.370	0.068\\
191.470	0.064\\
191.590	0.062\\
191.670	0.066\\
191.880	0.070\\
191.960	0.074\\
192.090	0.078\\
192.170	0.080\\
192.280	0.080\\
192.460	0.080\\
192.509	0.080\\
192.670	0.080\\
192.810	0.078\\
193.171	0.076\\
193.310	0.076\\
193.340	0.076\\
193.440	0.076\\
193.550	0.078\\
193.660	0.080\\
193.700	0.080\\
193.759	0.080\\
193.830	0.076\\
194.030	0.072\\
194.152	0.068\\
194.300	0.064\\
194.490	0.060\\
194.540	0.060\\
194.700	0.060\\
194.800	0.060\\
194.949	0.058\\
194.990	0.056\\
195.320	0.054\\
195.479	0.052\\
195.580	0.050\\
195.660	0.052\\
195.820	0.054\\
195.940	0.052\\
196.040	0.050\\
196.130	0.048\\
196.313	0.044\\
196.450	0.034\\
196.520	0.026\\
196.590	0.018\\
196.780	0.012\\
196.820	0.004\\
196.930	0.000\\
196.960	-0.002\\
197.140	-0.004\\
197.232	-0.008\\
197.320	-0.010\\
197.350	-0.014\\
197.480	-0.018\\
197.589	-0.024\\
197.693	-0.028\\
197.800	-0.032\\
197.969	-0.032\\
198.050	-0.028\\
198.160	-0.022\\
198.529	-0.018\\
198.620	-0.014\\
198.710	-0.010\\
198.860	-0.010\\
198.940	-0.010\\
199.030	-0.010\\
199.070	-0.010\\
199.150	-0.010\\
199.250	-0.006\\
199.330	-0.002\\
199.419	0.000\\
199.500	0.002\\
199.580	0.002\\
199.701	-0.002\\
199.989	-0.010\\
200.070	-0.016\\
200.231	-0.024\\
200.270	-0.032\\
200.390	-0.040\\
200.489	-0.046\\
200.610	-0.054\\
200.650	-0.060\\
200.840	-0.064\\
200.950	-0.070\\
201.020	-0.074\\
201.110	-0.074\\
201.231	-0.074\\
201.350	-0.074\\
201.449	-0.074\\
201.499	-0.074\\
201.700	-0.076\\
201.960	-0.082\\
202.020	-0.088\\
202.109	-0.092\\
202.350	-0.096\\
202.530	-0.096\\
202.659	-0.092\\
202.889	-0.086\\
203.000	-0.082\\
203.101	-0.078\\
203.400	-0.078\\
203.500	-0.076\\
203.590	-0.076\\
203.720	-0.074\\
203.839	-0.072\\
203.929	-0.070\\
204.010	-0.068\\
204.150	-0.068\\
204.299	-0.068\\
204.380	-0.068\\
204.529	-0.068\\
204.610	-0.070\\
204.670	-0.068\\
204.790	-0.066\\
205.070	-0.068\\
205.129	-0.070\\
205.209	-0.072\\
205.330	-0.076\\
205.400	-0.080\\
205.489	-0.080\\
205.620	-0.078\\
205.660	-0.076\\
205.800	-0.068\\
205.879	-0.060\\
206.210	-0.052\\
206.289	-0.042\\
206.369	-0.032\\
206.520	-0.028\\
206.560	-0.024\\
206.709	-0.020\\
206.820	-0.020\\
206.881	-0.020\\
206.961	-0.018\\
206.990	-0.016\\
207.170	-0.012\\
207.290	-0.006\\
207.440	0.000\\
207.710	0.004\\
207.779	0.008\\
207.899	0.010\\
207.970	0.008\\
208.090	0.006\\
208.270	0.004\\
208.320	0.002\\
208.439	0.000\\
208.530	0.000\\
208.600	-0.002\\
208.730	-0.004\\
208.789	-0.006\\
208.840	-0.008\\
208.880	-0.008\\
209.070	-0.006\\
209.330	-0.004\\
209.449	-0.004\\
209.489	-0.002\\
209.589	-0.002\\
209.830	-0.002\\
209.860	-0.002\\
209.960	0.000\\
210.040	0.000\\
210.100	0.000\\
210.160	0.004\\
210.230	0.008\\
210.440	0.014\\
210.520	0.020\\
210.640	0.028\\
211.030	0.032\\
211.200	0.036\\
211.300	0.038\\
211.380	0.040\\
211.520	0.038\\
211.579	0.036\\
211.619	0.034\\
212.040	0.032\\
212.159	0.030\\
212.230	0.034\\
212.431	0.038\\
212.579	0.042\\
212.630	0.046\\
212.730	0.050\\
212.880	0.050\\
213.081	0.050\\
213.150	0.050\\
213.190	0.052\\
213.309	0.054\\
213.434	0.056\\
213.490	0.058\\
213.530	0.064\\
213.580	0.068\\
213.661	0.074\\
213.751	0.080\\
213.811	0.086\\
213.970	0.088\\
214.060	0.092\\
214.189	0.094\\
214.289	0.096\\
214.381	0.098\\
214.500	0.100\\
214.610	0.100\\
214.690	0.100\\
214.810	0.100\\
215.059	0.100\\
215.160	0.098\\
215.380	0.096\\
215.441	0.094\\
215.590	0.092\\
215.651	0.090\\
215.760	0.094\\
215.831	0.098\\
216.090	0.102\\
216.230	0.106\\
216.330	0.110\\
216.419	0.118\\
216.560	0.124\\
216.791	0.130\\
216.901	0.136\\
217.020	0.140\\
217.080	0.136\\
217.269	0.134\\
217.360	0.130\\
217.451	0.126\\
217.730	0.124\\
217.809	0.122\\
217.870	0.116\\
217.949	0.112\\
218.029	0.106\\
218.182	0.100\\
218.322	0.088\\
218.481	0.080\\
218.599	0.066\\
218.679	0.054\\
218.851	0.042\\
218.860	0.036\\
218.980	0.028\\
219.010	0.026\\
219.290	0.020\\
219.399	0.014\\
219.584	0.004\\
219.730	-0.004\\
219.850	-0.012\\
220.050	-0.018\\
220.159	-0.024\\
220.280	-0.032\\
220.379	-0.040\\
220.480	-0.048\\
220.564	-0.052\\
220.730	-0.056\\
220.880	-0.054\\
220.939	-0.052\\
221.009	-0.048\\
221.129	-0.046\\
221.379	-0.044\\
221.491	-0.042\\
221.569	-0.040\\
221.649	-0.042\\
221.789	-0.044\\
221.880	-0.044\\
221.989	-0.042\\
222.060	-0.042\\
222.200	-0.040\\
222.439	-0.034\\
222.540	-0.030\\
222.590	-0.028\\
222.683	-0.024\\
222.841	-0.010\\
222.950	0.000\\
223.160	0.008\\
223.230	0.016\\
223.379	0.024\\
223.461	0.024\\
223.599	0.024\\
223.799	0.028\\
223.920	0.032\\
224.009	0.034\\
224.120	0.034\\
224.179	0.034\\
224.371	0.026\\
224.431	0.018\\
224.791	0.012\\
224.889	0.006\\
224.970	0.000\\
225.179	0.000\\
225.270	0.000\\
225.370	0.000\\
225.489	0.000\\
225.549	0.004\\
225.760	0.008\\
225.829	0.012\\
225.939	0.012\\
225.969	0.012\\
226.029	0.008\\
226.210	0.004\\
226.249	0.000\\
226.350	0.000\\
226.559	0.000\\
226.659	0.000\\
226.699	0.000\\
226.839	-0.002\\
226.970	-0.004\\
227.079	-0.014\\
227.160	-0.024\\
227.469	-0.034\\
227.670	-0.042\\
227.749	-0.050\\
227.830	-0.050\\
227.919	-0.054\\
228.019	-0.056\\
228.179	-0.058\\
228.259	-0.060\\
228.380	-0.062\\
228.430	-0.056\\
228.590	-0.052\\
228.819	-0.046\\
228.880	-0.040\\
229.040	-0.034\\
229.200	-0.032\\
229.280	-0.030\\
229.409	-0.030\\
229.549	-0.026\\
229.680	-0.022\\
229.929	-0.018\\
230.009	-0.014\\
230.129	-0.010\\
230.229	-0.010\\
230.341	-0.012\\
230.391	-0.012\\
230.490	-0.012\\
230.920	-0.014\\
231.030	-0.020\\
231.219	-0.024\\
231.380	-0.022\\
231.539	-0.020\\
231.730	-0.016\\
231.811	0.002\\
231.859	0.020\\
232.151	0.030\\
232.219	0.040\\
232.349	0.050\\
232.470	0.050\\
232.569	0.050\\
232.730	0.058\\
232.829	0.066\\
233.009	0.076\\
233.070	0.086\\
233.330	0.096\\
233.429	0.098\\
233.611	0.102\\
233.749	0.104\\
233.819	0.106\\
233.919	0.108\\
234.009	0.110\\
234.039	0.110\\
234.100	0.108\\
234.249	0.106\\
234.411	0.104\\
234.489	0.102\\
234.570	0.100\\
234.630	0.100\\
234.679	0.100\\
234.760	0.102\\
234.840	0.104\\
234.949	0.104\\
235.101	0.104\\
235.319	0.104\\
235.409	0.102\\
235.480	0.102\\
235.719	0.104\\
235.890	0.106\\
236.019	0.106\\
236.179	0.108\\
236.409	0.108\\
236.489	0.108\\
236.579	0.108\\
236.729	0.112\\
236.809	0.114\\
236.959	0.116\\
237.039	0.118\\
237.129	0.120\\
237.209	0.120\\
237.250	0.122\\
237.389	0.122\\
237.560	0.122\\
237.749	0.122\\
237.930	0.122\\
238.060	0.120\\
238.120	0.118\\
238.209	0.116\\
238.320	0.112\\
238.449	0.108\\
238.499	0.104\\
238.539	0.104\\
238.659	0.106\\
238.719	0.108\\
238.800	0.110\\
238.989	0.112\\
239.080	0.112\\
239.199	0.110\\
239.272	0.110\\
239.369	0.110\\
239.489	0.108\\
239.540	0.106\\
239.799	0.100\\
239.859	0.094\\
239.990	0.088\\
240.090	0.082\\
240.159	0.076\\
240.350	0.074\\
240.549	0.072\\
240.650	0.070\\
240.709	0.070\\
240.849	0.070\\
240.909	0.064\\
241.049	0.058\\
241.130	0.054\\
241.269	0.050\\
241.629	0.046\\
241.730	0.048\\
241.809	0.050\\
241.930	0.050\\
241.970	0.050\\
242.150	0.052\\
242.219	0.054\\
242.259	0.052\\
242.379	0.050\\
242.460	0.048\\
242.599	0.040\\
242.679	0.032\\
242.971	0.028\\
243.029	0.020\\
243.209	0.012\\
243.340	0.008\\
243.509	0.004\\
243.570	-0.002\\
243.670	-0.010\\
243.809	-0.018\\
243.910	-0.030\\
243.959	-0.042\\
244.039	-0.054\\
244.211	-0.060\\
244.299	-0.066\\
244.400	-0.068\\
244.530	-0.072\\
244.580	-0.074\\
244.719	-0.072\\
244.799	-0.070\\
244.909	-0.068\\
244.999	-0.064\\
245.090	-0.060\\
245.249	-0.058\\
245.320	-0.056\\
245.409	-0.054\\
245.479	-0.050\\
245.539	-0.042\\
245.680	-0.036\\
245.879	-0.030\\
245.999	-0.024\\
246.069	-0.020\\
246.129	-0.016\\
246.250	-0.012\\
246.350	-0.008\\
246.409	-0.004\\
246.450	0.000\\
246.619	0.000\\
246.699	0.000\\
246.859	0.000\\
246.919	0.000\\
246.999	0.000\\
247.159	0.000\\
247.369	0.000\\
247.459	-0.002\\
247.490	-0.006\\
247.669	-0.012\\
247.870	-0.018\\
247.951	-0.024\\
248.029	-0.028\\
248.152	-0.032\\
248.189	-0.034\\
248.369	-0.038\\
248.489	-0.042\\
248.549	-0.046\\
248.609	-0.046\\
248.660	-0.046\\
249.019	-0.044\\
249.089	-0.042\\
249.189	-0.040\\
249.269	-0.038\\
249.349	-0.036\\
249.509	-0.032\\
249.589	-0.026\\
249.789	-0.020\\
249.950	-0.016\\
250.009	-0.012\\
250.089	-0.010\\
250.139	-0.008\\
250.200	-0.006\\
250.319	-0.004\\
250.499	-0.002\\
250.579	0.000\\
250.679	0.000\\
250.759	0.002\\
250.830	0.004\\
250.960	0.008\\
251.009	0.012\\
251.119	0.016\\
251.219	0.022\\
251.330	0.026\\
251.660	0.028\\
251.739	0.030\\
251.929	0.032\\
252.149	0.030\\
252.229	0.030\\
252.349	0.028\\
252.409	0.028\\
252.609	0.028\\
252.729	0.028\\
252.801	0.028\\
252.920	0.032\\
253.089	0.034\\
253.139	0.038\\
253.460	0.042\\
253.599	0.046\\
253.669	0.048\\
253.800	0.050\\
253.869	0.050\\
253.909	0.056\\
254.030	0.062\\
254.309	0.072\\
254.369	0.082\\
254.490	0.092\\
254.579	0.094\\
254.699	0.096\\
254.760	0.094\\
254.820	0.088\\
254.899	0.082\\
255.239	0.078\\
255.319	0.072\\
255.409	0.062\\
255.739	0.056\\
256.000	0.050\\
256.079	0.044\\
256.152	0.036\\
256.299	0.032\\
256.369	0.028\\
256.439	0.022\\
256.609	0.016\\
256.769	0.014\\
256.892	0.018\\
257.039	0.022\\
257.340	0.028\\
257.439	0.032\\
257.479	0.034\\
257.599	0.030\\
257.749	0.022\\
257.840	0.014\\
257.949	0.008\\
258.069	0.004\\
258.149	0.000\\
258.309	0.000\\
258.389	-0.002\\
258.509	-0.004\\
258.599	-0.006\\
258.871	-0.008\\
258.969	-0.010\\
259.059	-0.012\\
259.239	-0.014\\
259.379	-0.016\\
259.500	-0.018\\
259.731	-0.020\\
259.800	-0.020\\
259.909	-0.020\\
260.019	-0.022\\
260.089	-0.024\\
260.209	-0.026\\
260.282	-0.030\\
260.429	-0.032\\
260.520	-0.034\\
260.650	-0.036\\
260.759	-0.038\\
260.940	-0.038\\
261.020	-0.040\\
261.109	-0.038\\
261.189	-0.034\\
261.229	-0.032\\
261.349	-0.030\\
261.519	-0.028\\
261.610	-0.028\\
261.719	-0.030\\
261.759	-0.032\\
261.859	-0.034\\
262.030	-0.032\\
262.289	-0.030\\
262.490	-0.028\\
262.569	-0.024\\
262.690	-0.020\\
262.751	-0.018\\
262.899	-0.016\\
263.079	-0.014\\
263.139	-0.012\\
263.189	-0.010\\
263.289	-0.010\\
263.459	-0.012\\
263.539	-0.014\\
263.672	-0.016\\
263.751	-0.018\\
263.800	-0.020\\
263.850	-0.020\\
263.962	-0.026\\
264.119	-0.032\\
264.209	-0.038\\
264.319	-0.048\\
264.409	-0.058\\
264.560	-0.062\\
264.659	-0.066\\
264.699	-0.070\\
264.769	-0.070\\
264.869	-0.070\\
264.919	-0.070\\
265.109	-0.070\\
265.229	-0.070\\
265.299	-0.070\\
265.349	-0.074\\
265.429	-0.078\\
265.549	-0.082\\
265.659	-0.084\\
265.699	-0.086\\
265.749	-0.084\\
265.849	-0.082\\
266.049	-0.086\\
266.119	-0.092\\
266.249	-0.098\\
266.479	-0.104\\
266.519	-0.110\\
266.659	-0.110\\
266.699	-0.112\\
266.849	-0.116\\
266.989	-0.120\\
267.069	-0.124\\
267.251	-0.124\\
267.329	-0.120\\
267.379	-0.114\\
267.569	-0.108\\
267.729	-0.098\\
267.919	-0.092\\
268.019	-0.088\\
268.089	-0.080\\
268.240	-0.072\\
268.349	-0.070\\
268.469	-0.070\\
268.850	-0.070\\
268.969	-0.074\\
269.069	-0.080\\
269.289	-0.086\\
269.400	-0.090\\
269.499	-0.094\\
269.579	-0.098\\
269.639	-0.102\\
269.699	-0.104\\
269.879	-0.110\\
269.999	-0.116\\
270.209	-0.122\\
270.309	-0.126\\
270.421	-0.130\\
270.509	-0.130\\
270.629	-0.130\\
270.749	-0.130\\
270.909	-0.130\\
270.980	-0.132\\
271.079	-0.134\\
271.370	-0.136\\
271.429	-0.132\\
271.510	-0.128\\
271.899	-0.122\\
271.940	-0.116\\
272.079	-0.110\\
272.119	-0.110\\
272.239	-0.104\\
272.299	-0.098\\
272.329	-0.092\\
272.530	-0.082\\
272.629	-0.072\\
272.769	-0.066\\
272.929	-0.056\\
273.049	-0.046\\
273.189	-0.040\\
273.230	-0.034\\
273.369	-0.030\\
273.500	-0.030\\
273.529	-0.030\\
273.640	-0.030\\
273.769	-0.030\\
273.819	-0.026\\
273.949	-0.022\\
274.009	-0.020\\
274.089	-0.014\\
274.249	-0.008\\
274.520	-0.006\\
274.609	-0.004\\
274.699	-0.002\\
274.829	-0.004\\
275.029	-0.006\\
275.110	-0.010\\
275.150	-0.014\\
275.289	-0.016\\
275.350	-0.018\\
275.419	-0.020\\
275.579	-0.020\\
275.639	-0.020\\
275.729	-0.020\\
275.919	-0.016\\
276.079	-0.012\\
276.219	-0.008\\
276.299	-0.004\\
276.379	0.000\\
276.471	-0.002\\
276.589	-0.004\\
276.709	-0.008\\
276.799	-0.012\\
276.889	-0.014\\
276.929	-0.016\\
277.089	-0.018\\
277.209	-0.018\\
277.329	-0.018\\
277.399	-0.020\\
277.519	-0.020\\
277.719	-0.020\\
277.779	-0.020\\
277.859	-0.018\\
278.019	-0.016\\
278.219	-0.014\\
278.269	-0.012\\
278.469	-0.010\\
278.569	-0.010\\
278.689	-0.012\\
278.729	-0.014\\
278.869	-0.016\\
279.019	-0.018\\
279.199	-0.016\\
279.299	-0.012\\
279.379	-0.008\\
279.509	-0.004\\
279.699	0.000\\
279.789	0.000\\
279.849	0.000\\
279.969	0.000\\
280.109	0.002\\
};
\addplot [color=black!30!green,solid,forget plot]
  table[row sep=crcr]{%
159.960	-0.000\\
160.650	-0.000\\
160.831	-0.000\\
160.981	-0.000\\
161.242	-0.000\\
161.500	-0.000\\
161.600	-0.000\\
161.650	-0.000\\
162.140	-0.000\\
163.300	-0.000\\
163.730	-0.000\\
164.060	-0.000\\
164.300	-0.000\\
164.420	-0.000\\
164.560	-0.000\\
164.700	-0.000\\
164.910	-0.000\\
165.270	-0.000\\
165.430	-0.000\\
165.720	-0.000\\
166.430	-0.000\\
166.770	-0.000\\
166.950	-0.000\\
167.220	-0.000\\
167.890	-0.000\\
168.260	0.149\\
168.440	0.154\\
169.050	-0.000\\
169.340	-0.000\\
169.580	-0.000\\
169.850	-0.000\\
170.090	-0.000\\
170.620	-0.000\\
171.040	-0.000\\
171.090	-0.000\\
171.220	-0.000\\
171.420	-0.000\\
171.530	-0.000\\
171.600	-0.000\\
172.091	-0.000\\
172.400	-0.000\\
172.770	-0.000\\
173.290	-0.000\\
173.482	-0.037\\
173.590	-0.037\\
173.780	-0.000\\
173.940	-0.000\\
174.051	-0.000\\
174.390	-0.000\\
174.670	-0.000\\
174.850	-0.000\\
175.210	-0.000\\
175.601	-0.000\\
176.640	-0.000\\
177.279	-0.024\\
177.620	-0.085\\
178.330	-0.141\\
178.819	-0.000\\
178.940	-0.005\\
179.250	-0.000\\
179.770	-0.000\\
179.940	-0.000\\
180.700	-0.000\\
181.019	-0.000\\
181.360	-0.000\\
181.480	-0.000\\
182.180	-0.000\\
182.400	-0.000\\
182.620	0.049\\
182.890	0.100\\
183.690	-0.000\\
184.090	-0.000\\
184.429	-0.000\\
184.919	-0.000\\
185.129	-0.000\\
185.311	-0.000\\
185.920	-0.000\\
186.159	-0.000\\
186.280	-0.000\\
186.639	-0.000\\
187.250	0.076\\
187.430	0.076\\
187.779	0.076\\
188.580	0.202\\
189.770	0.037\\
189.810	0.037\\
189.929	0.037\\
190.090	0.037\\
190.429	0.017\\
190.590	0.017\\
190.800	0.017\\
191.410	0.015\\
191.722	0.015\\
192.221	0.015\\
192.539	0.015\\
192.749	0.015\\
192.880	0.015\\
192.930	0.015\\
193.019	0.015\\
193.480	0.015\\
193.940	0.015\\
194.070	0.015\\
194.660	-0.000\\
195.070	-0.000\\
195.619	-0.000\\
195.880	-0.000\\
196.270	-0.000\\
196.622	-0.000\\
197.039	-0.000\\
197.080	-0.000\\
197.870	0.090\\
198.340	0.078\\
198.559	-0.000\\
198.900	-0.000\\
199.540	-0.000\\
199.719	-0.000\\
199.950	-0.000\\
200.151	-0.000\\
200.729	-0.000\\
201.139	-0.000\\
201.280	-0.000\\
201.569	-0.000\\
202.141	-0.000\\
202.269	-0.000\\
202.380	-0.000\\
202.579	-0.000\\
203.149	-0.000\\
203.329	-0.000\\
203.779	-0.000\\
204.200	-0.000\\
204.719	-0.000\\
204.909	0.005\\
205.749	0.222\\
205.919	0.200\\
206.150	0.010\\
206.760	-0.000\\
207.049	-0.000\\
207.209	-0.000\\
207.539	-0.000\\
207.840	-0.000\\
208.049	-0.000\\
208.399	-0.000\\
208.959	0.122\\
209.039	0.141\\
209.369	0.149\\
209.549	0.151\\
209.660	0.154\\
210.339	0.061\\
211.419	0.278\\
211.699	0.154\\
211.840	0.068\\
211.961	0.029\\
212.199	0.029\\
212.770	0.037\\
212.969	0.046\\
214.229	0.066\\
214.750	0.068\\
215.199	0.068\\
215.519	0.305\\
215.940	0.305\\
216.129	0.305\\
216.849	0.056\\
217.319	0.056\\
217.669	-0.000\\
218.159	-0.000\\
218.250	-0.000\\
218.359	-0.000\\
219.059	-0.000\\
219.499	-0.000\\
219.989	-0.000\\
221.060	-0.000\\
221.529	-0.000\\
221.949	0.088\\
222.129	0.202\\
222.389	0.232\\
223.349	0.005\\
223.539	0.007\\
223.639	0.007\\
224.039	0.005\\
224.249	-0.000\\
224.470	-0.000\\
224.569	0.005\\
224.709	0.010\\
225.099	0.010\\
225.410	0.010\\
225.589	0.010\\
226.479	0.010\\
227.270	-0.000\\
227.400	-0.000\\
227.609	-0.000\\
227.869	-0.000\\
228.299	-0.000\\
228.559	-0.000\\
228.789	-0.000\\
229.120	-0.000\\
229.439	-0.000\\
229.869	-0.000\\
230.169	-0.000\\
230.449	-0.000\\
230.609	-0.000\\
230.719	-0.000\\
230.839	-0.000\\
231.129	-0.000\\
231.259	-0.000\\
231.439	-0.000\\
231.929	-0.000\\
232.290	-0.000\\
232.609	-0.000\\
232.889	-0.000\\
233.110	-0.000\\
233.251	-0.000\\
233.549	-0.000\\
233.859	-0.000\\
234.349	-0.000\\
234.889	0.010\\
235.030	0.017\\
235.559	0.027\\
235.639	0.027\\
235.799	0.027\\
235.949	0.027\\
236.059	0.027\\
236.151	0.027\\
236.249	0.027\\
236.999	0.027\\
237.439	0.029\\
237.589	0.029\\
237.649	0.029\\
237.779	0.029\\
239.599	-0.000\\
240.019	-0.000\\
240.239	-0.000\\
240.289	-0.000\\
240.429	-0.000\\
240.779	-0.000\\
241.100	-0.000\\
241.209	-0.000\\
241.369	-0.000\\
241.769	-0.000\\
242.089	-0.220\\
242.719	-0.212\\
242.889	-0.212\\
243.269	-0.212\\
243.389	-0.212\\
243.730	-0.212\\
243.869	-0.212\\
244.240	-0.024\\
244.359	-0.005\\
244.469	-0.000\\
245.939	-0.000\\
246.779	-0.000\\
247.189	-0.000\\
247.529	-0.000\\
247.622	-0.000\\
247.729	-0.000\\
248.249	-0.000\\
248.729	-0.000\\
249.129	-0.000\\
249.669	-0.000\\
249.729	-0.000\\
249.869	-0.000\\
250.239	-0.015\\
250.359	-0.015\\
250.539	-0.015\\
250.660	-0.015\\
251.549	-0.110\\
251.699	-0.083\\
252.269	-0.046\\
252.489	-0.156\\
252.869	-0.073\\
252.999	-0.032\\
253.260	-0.007\\
253.709	-0.005\\
254.440	-0.005\\
254.999	-0.005\\
255.271	-0.156\\
255.589	-0.295\\
256.039	-0.156\\
256.509	-0.037\\
256.649	-0.032\\
256.930	-0.032\\
257.009	-0.032\\
257.301	-0.300\\
258.669	-0.012\\
259.149	-0.000\\
259.279	-0.000\\
259.459	-0.000\\
259.559	-0.000\\
259.949	-0.000\\
261.049	-0.000\\
261.569	-0.000\\
261.819	-0.000\\
261.979	-0.000\\
262.109	-0.000\\
262.189	-0.000\\
262.599	-0.000\\
262.819	-0.000\\
263.019	-0.054\\
263.319	-0.300\\
263.620	-0.283\\
264.469	-0.088\\
265.189	-0.090\\
265.469	-0.090\\
265.949	-0.085\\
266.279	-0.085\\
266.419	-0.088\\
266.789	-0.085\\
267.179	-0.085\\
267.459	-0.085\\
267.799	-0.085\\
268.319	-0.088\\
268.429	-0.088\\
268.629	-0.090\\
268.769	-0.095\\
268.929	-0.100\\
269.129	-0.100\\
269.439	-0.098\\
269.549	-0.090\\
269.939	-0.090\\
270.379	-0.090\\
271.259	-0.000\\
271.549	-0.000\\
271.669	-0.000\\
271.990	-0.000\\
272.459	-0.000\\
272.569	-0.000\\
273.309	-0.000\\
274.199	-0.063\\
274.379	-0.059\\
274.749	-0.059\\
274.899	-0.063\\
275.459	-0.059\\
275.839	-0.063\\
275.969	-0.063\\
277.369	-0.056\\
277.639	-0.000\\
277.979	0.027\\
278.129	0.027\\
278.429	0.090\\
278.779	0.117\\
278.901	0.127\\
278.979	0.127\\
279.099	0.127\\
279.569	0.110\\
279.929	0.110\\
280.290	0.185\\
};
\addplot [color=black,dashed,forget plot]
  table[row sep=crcr]{%
148.000	0.000\\
nan	nan\\
292.000	0.000\\
};
\end{axis}
\end{tikzpicture}%